\documentclass[sigconf]{acmart}

\newcommand{\myparatight}[1]{\smallskip\noindent{\bf {#1}:}~}
\usepackage{amsthm}
\usepackage{amsmath}

\usepackage{float}
\usepackage{graphics}
\usepackage{graphicx}
\usepackage[skip=0pt]{caption}
\usepackage{algorithm}
\usepackage{algorithmic}
\usepackage{bm}
\usepackage{color}
\usepackage{multirow}
\usepackage{makecell}
\usepackage{pdfx}
\usepackage{subfig}

\usepackage{amssymb}
\allowdisplaybreaks

\newtheorem{thm}{Theorem}
\newtheorem{definition}{Definition}
\renewcommand{\algorithmicrequire}{\textbf{Input:}}
\renewcommand{\algorithmicensure}{\textbf{Output:}}

\DeclareMathOperator*{\argmin}{arg\,min}

\newcommand{\alg}{WEvade~}
\newcommand{\algns}{WEvade}

\copyrightyear{2023}
\acmYear{2023}
\setcopyright{acmlicensed}
\acmConference[CCS '23] {Proceedings of the 2023 ACM SIGSAC Conference on Computer and Communications Security}{November 26--30, 2023}{Copenhagen, Denmark.}
\acmBooktitle{Proceedings of the 2023 ACM SIGSAC Conference on Computer and Communications Security (CCS '23), November 26--30, 2023, Copenhagen, Denmark}
\acmPrice{15.00}
\acmISBN{979-8-4007-0050-7/23/11}
\acmDOI{10.1145/3576915.3623189}

\begin{document}

\title{{Evading Watermark based Detection of AI-Generated Content}}

\author{Zhengyuan Jiang}
\authornotemark[1]
\affiliation{%
  \institution{Duke University}
  \country{}
}
\email{zhengyuan.jiang@duke.edu}

\author{Jinghuai Zhang}
\authornote{Equal contribution.}
\affiliation{%
  \institution{Duke University}
  \country{}
}
\email{jinghuai.zhang@duke.edu}

\author{Neil Zhenqiang Gong}
\affiliation{%
  \institution{Duke University}
  \country{}
}
\email{neil.gong@duke.edu}

\begin{abstract}
  A generative AI model can generate extremely realistic-looking content, posing growing challenges to the authenticity of information. To address the challenges, watermark has been leveraged to detect AI-generated content. Specifically, a watermark is embedded into an AI-generated content before it is released. A content is  detected as AI-generated if a similar watermark can be decoded from it. In this work, we perform a systematic study on the robustness of such watermark-based AI-generated content detection. Our work shows that an attacker can post-process a watermarked image via adding a small, human-imperceptible perturbation to it, such that the post-processed image evades detection while maintaining its visual quality. We show the effectiveness of our attack both theoretically and empirically.   Moreover, to evade detection,  our adversarial post-processing method adds much smaller perturbations to  AI-generated images and thus better maintain their visual quality than existing popular post-processing methods such as JPEG compression, Gaussian blur, and Brightness/Contrast. Our work shows the insufficiency of existing watermark-based detection of AI-generated content, highlighting the urgent needs of new methods. Our code is publicly available: \url{https://github.com/zhengyuan-jiang/WEvade}.

\end{abstract}

\ccsdesc[500]{Security and privacy~Security services}

\keywords{AI-generated content detection; Watermarking; Robustness}

\maketitle
\section{Introduction}

\begin{figure*}[!t]
\centering
{\includegraphics[width=0.12\textwidth]{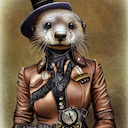}}\hspace{0.2mm}
{\includegraphics[width=0.12\textwidth]{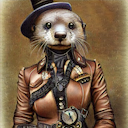}}\hspace{0.2mm}
{\includegraphics[width=0.12\textwidth]{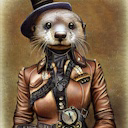}}\hspace{0.2mm}
{\includegraphics[width=0.12\textwidth]{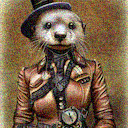}}\hspace{0.2mm}
{\includegraphics[width=0.12\textwidth]{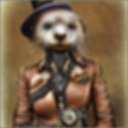}}\hspace{0.2mm}
{\includegraphics[width=0.12\textwidth]{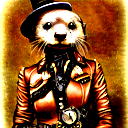}}\hspace{0.2mm}
{\includegraphics[width=0.12\textwidth]{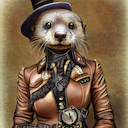}}\hspace{0.2mm}
{\includegraphics[width=0.12\textwidth]{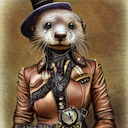}}

\subfloat[Original]{\includegraphics[width=0.12\textwidth]{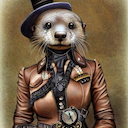}}
\hspace{0.2mm}
\subfloat[Watermarked]{\includegraphics[width=0.12\textwidth]{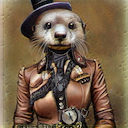}}
\hspace{0.2mm}
\subfloat[JPEG]{\includegraphics[width=0.12\textwidth]{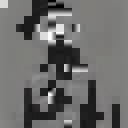}}
\hspace{0.2mm}
\subfloat[GN]{\includegraphics[width=0.12\textwidth]{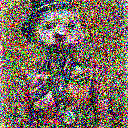}}
\hspace{0.2mm}
\subfloat[GB]{\includegraphics[width=0.12\textwidth]{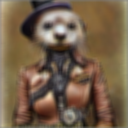}}
\hspace{0.2mm}
\subfloat[B/C]{\includegraphics[width=0.12\textwidth]{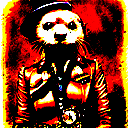}}
\hspace{0.2mm}
\subfloat[\algns-W-II]{\includegraphics[width=0.12\textwidth]{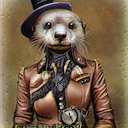}}
\hspace{0.2mm}
\subfloat[\algns-B-Q]{\includegraphics[width=0.12\textwidth]{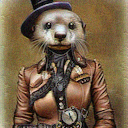}}

\caption{Illustration of original image, watermarked image, and watermarked images post-processed by existing and our methods (last two columns) to evade detection. The watermarking method is HiDDeN. GN: Gaussian noise. GB: Gaussian blur. B/C: Brightness/Contrast. The encoder/decoder are trained via standard training (\emph{first row}) or adversarial training (\emph{second row}).}
\label{examples-intro}
\vspace{-3mm}
\end{figure*}

Given a prompt, generative AI--such as DALL-E, Stable Diffusion, and ChatGPT--can generate extremely realistic looking content including image and text. Like any advanced technology, generative AI is also a double-edged sword. On one hand, generative AI can assist human to enhance effectiveness and efficiency in various domains such as searching, art image creation, and character design in online games. The market for generative AI was predicted to increase to 50 billion by 2028~\cite{generative-ai-market}.   
On the other hand, generative AI also raises many ethical concerns. For instance, their generated realistic looking content can be used to aid disinformation campaigns on social media; they are disruptive for learning and education as students can use them to complete/aid homework and exams; 
and people can use them to generate content and claim its ownership/copyright, though not allowed by US Copyright Office~\cite{us-copyright-office}.    

Watermark-based detection~\cite{fernandez2023stable,yu2021artificial,yang2021faceguard,abdelnabi2021adversarial,kirchenbauer2023watermark} of AI-generated content is a key technology to address these ethical concerns. Multiple AI companies--such as OpenAI, Google, and Meta--have made voluntary commitments to watermark AI-generated content~\cite{developing-AI-responsibility}.
In particular, a watermark is embedded into an AI-generated content when it is generated. The watermark enables proactive detection of AI-generated content in the future: a content is AI-generated if a similar watermark can be extracted from it. In this work, we focus on AI-generated images.  For instance, DALL-E embeds a \emph{visible} watermark at the bottom right corner of its generated images (Figure~\ref{dall-e} in Appendix shows an example); Stable Diffusion uses a \emph{non-learning-based} watermarking method~\cite{invisible-watermark} to embed an invisible watermark into generated images; and Meta~\cite{fernandez2023stable} proposed to use \emph{learning-based} watermarking methods.

A watermarking method~\cite{pereira2000robust, bi2007robust, invisible-watermark, zhu2018hidden, zhang2020udh, luo2020distortion, tancik2020stegastamp, zhang2019steganogan, wen2019romark} consists of three key components, i.e., \emph{watermark} (we represent it as a bitstring), \emph{encoder}, and \emph{decoder}. Given an image and a watermark, an encoder embeds the watermark into the image to produce a \emph{watermarked image}; and a decoder decodes a watermark from an image (a watermarked image or an \emph{original image} without watermark). We note that some watermarking methods~\cite{fernandez2023stable, yu2021artificial, wen2023tree} embed the encoder into a generative AI model, so the watermark is already embedded into its generated images at generation. An image is predicted as AI-generated if the \emph{bitwise accuracy} of the decoded watermark is larger than a threshold $\tau$, where bitwise accuracy is the fraction of matched bits in the decoded watermark and the ground-truth one. The threshold $\tau$ should be larger than 0.5 since the bitwise accuracy of original images without watermarks would be around 0.5. In a non-learning-based watermarking method~\cite{pereira2000robust, bi2007robust, invisible-watermark}, which has been studied for  decades, both encoder and decoder are designed based on heuristics, while they are neural networks and  automatically learnt using a set of images in learning-based watermarking methods~\cite{zhu2018hidden,zhang2020udh, luo2020distortion, tancik2020stegastamp, zhang2019steganogan, wen2019romark}, an emerging category of watermarking methods.

Robustness  against \emph{post-processing}, which post-processes an AI-generated image, is crucial for a watermark-based detector. Unfortunately, the visible watermark adopted by DALL-E can be easily removed without sacrificing the image quality at all~\cite{remove-dall-e}. Non-learning-based watermarks (e.g., the one used by Stable Diffusion) can be removed by popular image post-processing methods (e.g., JPEG compression)~\cite{zhu2018hidden,fernandez2023stable}, which we also confirm in our experiments in Section~\ref{stablediffusion}. Learning-based watermarking methods were believed to be robust against post-processing~\cite{luo2020distortion,zhu2018hidden,zhang2020udh,fernandez2023stable}. In particular, the encoder and decoder can be trained using \emph{adversarial training}~\cite{goodfellow2014explaining} to enhance robustness against post-processing. In adversarial training, a post-processing layer is added between the encoder and decoder; it post-processes a watermarked image outputted by the encoder before feeding it into the decoder; and the encoder and decoder are adversarially trained such that the watermark decoded from a post-processed watermarked image is still similar to the ground-truth one.  However, existing studies only evaluated the robustness of learning-based watermarking methods against popular image post-processing methods such as JPEG compression, Gaussian blur, and Brightness/Contrast, leaving their robustness against adversarial post-processing unexplored.  

\myparatight{Our work} We aim to bridge this gap in this work. We propose \algns, an adversarial post-processing method to evade watermark-based detection of AI-generated images. \alg adds a small, human-imperceptible perturbation to a watermarked image such that the perturbed image is falsely detected as non-AI-generated. \alg can be viewed as adversarial examples~\cite{szegedy2013intriguing} to watermarking methods. However, as we discuss below, simply extending standard adversarial examples to watermarking is insufficient. \alg considers the unique characteristics of watermarking to construct adversarial examples.  

{\bf White-box setting.} In this threat model, we assume the attacker has access to the decoder used by  detectors, but no access to the ground-truth watermark and encoder. Given a watermarked image generated by an AI model, an attacker aims to post-process it via adding a small perturbation to it, such that detectors with any threshold $\tau>0.5$ would falsely detect the post-processed watermarked image as non-AI-generated. One way (denoted as \algns-W-I) to achieve the goal is to simply extend the standard adversarial examples to the decoder. In particular, an attacker finds the perturbation such that each bit of the decoded watermark flips, leading to a very small bitwise accuracy and thus evasion. However, we show that such attack can be mitigated by a \emph{double-tail detector}, which we propose to detect an image as AI-generated if the decoded watermark has either too small or too large bitwise accuracy. 

To address the challenge, we propose \algns-W-II, which adds perturbation to a watermarked image such that the decoded watermark has a bitwise accuracy close to 0.5, making the post-processed image indistinguishable with original images without watermarks. However, since the attacker does not know the ground-truth watermark, it is challenging to measure the bitwise accuracy of the decoded watermark. Our key observation to address the challenge is that a watermark selected uniformly at random would have a bitwise accuracy close to 0.5, no matter what the ground-truth watermark is. Based on this observation, we find the perturbation with which the decoded watermark is close to a random watermark.  We formulate finding such perturbation as an optimization problem and propose a solution to solve it.  

{\bf Black-box setting.} In this threat model, we  assume the attacker can only query the detector API, which returns a binary result ("AI-generated" or "non-AI-generated") for any image. One way (called \algns-B-S) to evade detection is that the attacker trains a surrogate encoder and decoder using a watermarking algorithm. Then, given a watermarked image, the attacker finds the perturbation  based on the surrogate decoder using the white-box attack \algns-W-II. However, such attack achieves limited evasion rates because the surrogate decoder and target decoder output dissimilar watermarks for an image. 

To address the challenge, we propose \algns-B-Q, which extends state-of-the-art hard-label based adversarial example technique called HopSkipJump~\cite{chen2020hopskipjumpattack} to watermark-based detector. Given a watermarked image, HopSkipJump can iteratively find a post-processed version to evade detection via just querying the detector API. Specifically, starting from a \emph{random} initial image that is predicted as non-AI-generated by the detector, HopSkipJump iteratively moves the image closer to the given watermarked image to reduce the added perturbation while always guaranteeing that the image  evades detection. Essentially, in each iteration, HopSkipJump returns 1) a perturbation to update the image and 2) the number of queries to the detector API used to find such perturbation. The iterative process stops when HopSkipJump uses a given \emph{query budget}. However, simply applying HopSkipJump to watermarking may end up with a large perturbation. The reasons include 1) the random initial image may be far away from the given  watermarked  image, and 2) the iterative process does not always reduce the perturbation, and thus an improper setting of query budget may actually enlarge the perturbation. To address the challenges, our \algns-B-Q constructs the initial image using the  watermarked image post-processed by popular methods such as JPEG compression, which results in an initial image closer to the watermarked image. Moreover, \algns-B-Q stops the iterative process when the added perturbation starts to increase, which reduces both perturbation and number of queries to the detector API.  

{\bf Theoretical and empirical evaluation.} Theoretically, we derive the evasion rates of different variants of \algns. For instance, \algns-W-I achieves evasion rate of 1 against the standard \emph{single-tail detector}, but its evasion rate reduces to 0 when our proposed double-tail detector is used. We also derive a lower bound of the evasion rate of \algns-W-II using triangle inequality. Moreover, we derive the evasion rate of \algns-B-S based on a formal similarity quantification between the watermarks outputted by the surrogate decoder and target decoder. We also show that \algns-B-Q achieves evasion rate of 1. 

Empirically, we evaluate our attacks  using multiple datasets and multiple watermarking methods, including two learning-based ones (HiDDeN~\cite{zhu2018hidden} and UDH~\cite{zhang2020udh}) and the non-learning-based one adopted by Stable Diffusion~\cite{invisible-watermark}. Our results show that our method is effective and outperforms existing post-processing methods. In particular, existing post-processing methods need to add much larger perturbations in order to achieve evasion rates comparable to our method. We find that adversarial training can enhance robustness of watermarking, i.e., a post-processing method needs to add larger perturbation to evade detection. However, the perturbation added by our method is still small and maintains image quality, indicating the insufficiency of adversarial training.  Figure~\ref{examples-intro} shows an original image, its watermarked version, and the watermarked versions post-processed by different methods such that the decoded watermarks achieve bitwise accuracy close to 0.55 (indistinguishable with original images without watermarks). The results show that existing post-processing methods substantially sacrifice image quality to evade a watermark-based detector based on adversarial training, while our methods still maintain image quality.  

To summarize, our key contributions are as follows:
\begin{itemize}
\item We propose \alg, which adds small, human-imperceptible perturbations to AI-generated images to evade watermark-based detectors. 
\item We theoretically analyze the evasion rates of  \alg in both white-box and black-box settings.
\item We empirically evaluate \alg on multiple watermarking methods and datasets in various scenarios.  
\end{itemize}

\section{Related Work}

\subsection{Detecting AI-generated Content}
Generative AI models could be GANs~\cite{goodfellow2020generative}, diffusion models (e.g., DALL-E~\cite{ramesh2021zero}, Stable Diffusion~\cite{rombach2022high}), or language models (e.g., ChatGPT~\cite{openai-chatgpt}).  AI-generated content could be image (our focus in this work) or text. 
 AI-generated content detection include \emph{passive detection}~\cite{sha2022fake, wang2019detecting, yu2019attributing, frank2020leveraging, zhao2021multi,mitchell2023detectgpt} and \emph{proactive detection}~\cite{pereira2000robust, bi2007robust, invisible-watermark, zhu2018hidden, zhang2020udh, luo2020distortion, tancik2020stegastamp, zhang2019steganogan, wen2019romark}. Passive detection aims to leverage statistical artifacts in AI-generated content to distinguish them with non-AI-generated content, while proactive detection aims to proactively embed a watermark into AI-generated content when it is generated, which enables detection in the future.
 Several studies~\cite{cao2022understanding,corvi2022detection,sadasivan2023can} showed that passive detectors are not robust to evasion attacks, i.e., an attacker can slightly perturb an AI-generated content to remove the statistical artifacts exploited by a passive detector and thus evade detection. However, the robustness of proactive detectors against evasion attacks is much less explored. For instance, recent studies~\cite{fernandez2023stable} suggested that proactive detectors are more robust than passive ones. Our work focuses on proactive detectors and shows that they are not as robust as previously thought.

\begin{figure}[!t]
\centering
\vspace{-4mm}
{\includegraphics[width=0.46\textwidth]{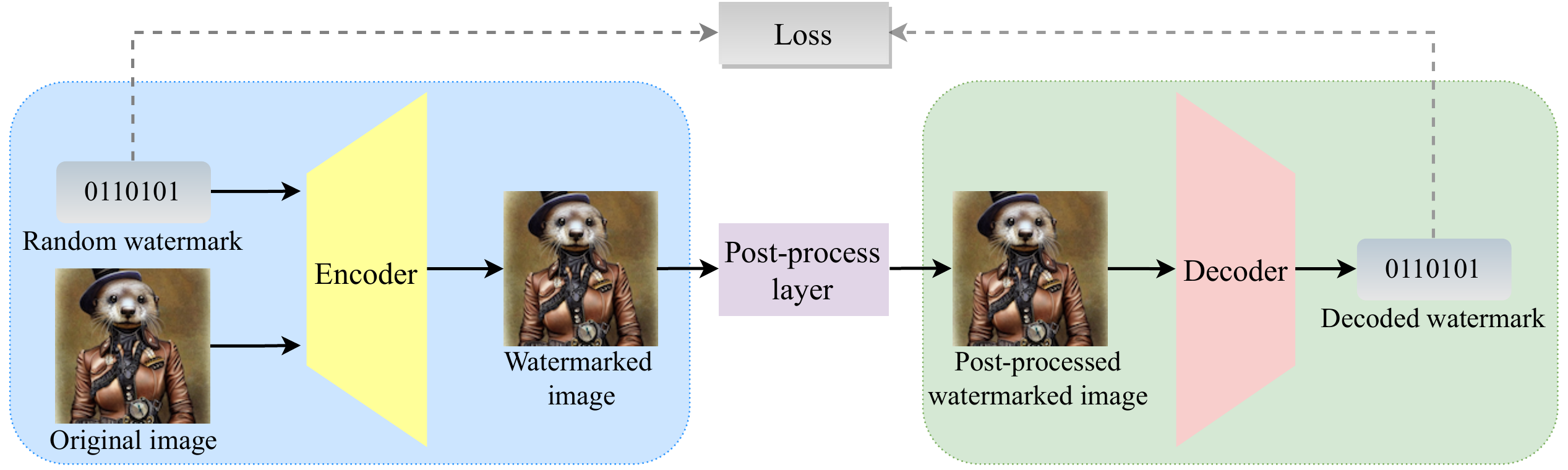}}
\caption{Illustration of training encoder and decoder in learning-based watermarking methods.}
\label{learning-based-method}
\vspace{-3mm}
\end{figure}

\subsection{Watermarking Methods} 
\label{sec:watermarkingmethod}
Since we focus on AI-generated images, we review image watermarking methods.  A watermarking method has three key components: \emph{watermark}, \emph{encoder}, and \emph{decoder}. We consider a watermark $w$ as a $n$-bit bitstring, e.g., $w=0110101$. An encoder takes an image $I$ and a watermark $w$ as input and produces a watermarked image $I_w$. Formally, we have $I_w=E(I,w)$, where $E$ stands for encoder. A decoder takes an image as input and outputs a watermark. Formally, we have $w_I=D(I)$. Note that, given any image (e.g., an original image without watermark or a watermarked image) as input, the decoder can output a watermark.  Watermarking methods can be categorized into two groups depending on how the encoder and decoder are designed, i.e., \emph{non-learning-based} and \emph{learning-based}.

\myparatight{Non-learning-based methods} In these methods~\cite{pereira2000robust, bi2007robust, invisible-watermark}, the encoder and decoder are hand-crafted based on heuristics. Non-learning-based methods have been studied for around three decades. \emph{Invisible-watermark}~\cite{invisible-watermark} is a representative non-learning-based method, which is adopted by Stable Diffusion. Roughly speaking, this method uses Discrete Wavelet Transform (DWT) to decompose an image into several  frequency sub-bands, applies Discrete Cosine Transform (DCT) to each block of some carefully selected sub-bands, and alters certain frequency coefficients of each block via adding a bit of the watermark. The watermark is embedded in selected frequency sub-bands of the image, and the watermarked image is obtained via  inverse transform.

\myparatight{Learning-based methods} In these methods~\cite{zhu2018hidden, zhang2020udh, luo2020distortion, tancik2020stegastamp, zhang2019steganogan, wen2019romark}, the encoder and decoder are neural networks and automatically learnt via deep learning techniques. Roughly speaking, the second-to-last layer of the decoder outputs a vector of real-value numbers, each entry of which indicates the likelihood that the corresponding bit of the watermark is 1. Formally, we denote by $F(I)$ such vector for an image $I$, where $F(I)_i$ is the likelihood that the $i$th bit of the decoded watermark is 1; and the decoded watermark $w_I$ is obtained by thresholding $F(I)$, i.e., the $i$th bit of $w_I$ is 1 if and only if $F(I)_i>0.5$. HiDDeN~\cite{zhu2018hidden} and UDH~\cite{zhang2020udh} are two representative learning-based methods. In HiDDeN, the encoder concatenates a watermark and an image to produce a watermarked image.  In UDH, the  encoder transforms the watermark into a QR code, maps the QR code to a secret image which has the same size as an original image, and pixel-wisely adds the secret image to an original image  as a watermarked image. Figure~\ref{learning-based-method} illustrates how the encoder and decoder are trained, which we discuss next.  

{\bf Standard training.} The encoder and decoder are iteratively trained using a set of images and  the standard Stochastic Gradient Descent (SGD) algorithm. In each iteration, a mini-batch of images are  used to update the encoder and decoder. Specifically, for each image $I$ in the mini-batch, a random watermark $w_I$ is sampled. The encoder $E$ produces a watermarked image $E(I,w_I)$ for each image $I$ and the corresponding random watermark $w_I$. The decoder $D$ takes each watermarked image $E(I,w_I)$ as input and outputs a watermark $D(E(I,w_I))$. The encoder and decoder are learnt such that the decoded watermark $D(E(I,w_I))$ is close to $w_I$. In particular, they are updated via SGD to minimize a loss function $\sum_{I}loss(D(E(I,w_I)), w_I)$.  

{\bf Adversarial training.} A key advantage of learning-based methods is that they can leverage adversarial training~\cite{goodfellow2014explaining,madry2017towards} to enhance their robustness against post-processing~\cite{luo2020distortion,wen2019romark}. Specifically, as illustrated in Figure~\ref{learning-based-method}, a post-processing layer is added between the encoder and decoder, which post-processes each watermarked image before feeding it to the decoder during training. For each image in a mini-batch during training, a post-processing method is randomly selected from  a given set of ones, e.g., JPEG compression, Gaussian noise, Gaussian blur, Brightness/Contrast, and our \algns. The encoder and decoder are updated via SGD to minimize a loss function $\sum_{I}loss(D(E(I,w_I) +  \delta_I), w_I)$, where $\delta_I$ is the perturbation introduced by the post-processing method to the watermarked image $E(I,w_I)$. As shown by previous works~\cite{luo2020distortion,wen2019romark} and confirmed by our experiments, adversarial training makes learning-based watermarking  robust against popular post-processing methods. However, it is still vulnerable to our adversarial post-processing method.   

We note that some watermarking methods~\cite{fernandez2023stable, yu2021artificial, wen2023tree} embed the encoder into a generative AI model, so its generated images are already embedded with the watermark, but they still rely on the decoder  for detection. For instance, Fernandez et al.~\cite{fernandez2023stable} trains encoder/decoder using HiDDeN, embeds the encoder into image generator via fine-tuning it, and uses the decoder for detection. Our attacks are also applicable to such watermarking methods since they are agnostic to how a watermark is embedded into an AI-generated image.

\section{Watermark-based Detectors} 
\label{sec:detectors}
We formally define the detection setup and the standard \emph{single-tail detector}.  Moreover, we propose a \emph{double-tail detector}, which can defend against the evasion attack (discussed in Section~\ref{whiteboxattack}) that simply extends standard adversarial examples to watermarking.

\begin{figure}[!t]
\centering
\vspace{-5mm}
\subfloat[Single-tail detector]{\includegraphics[width=0.23\textwidth]{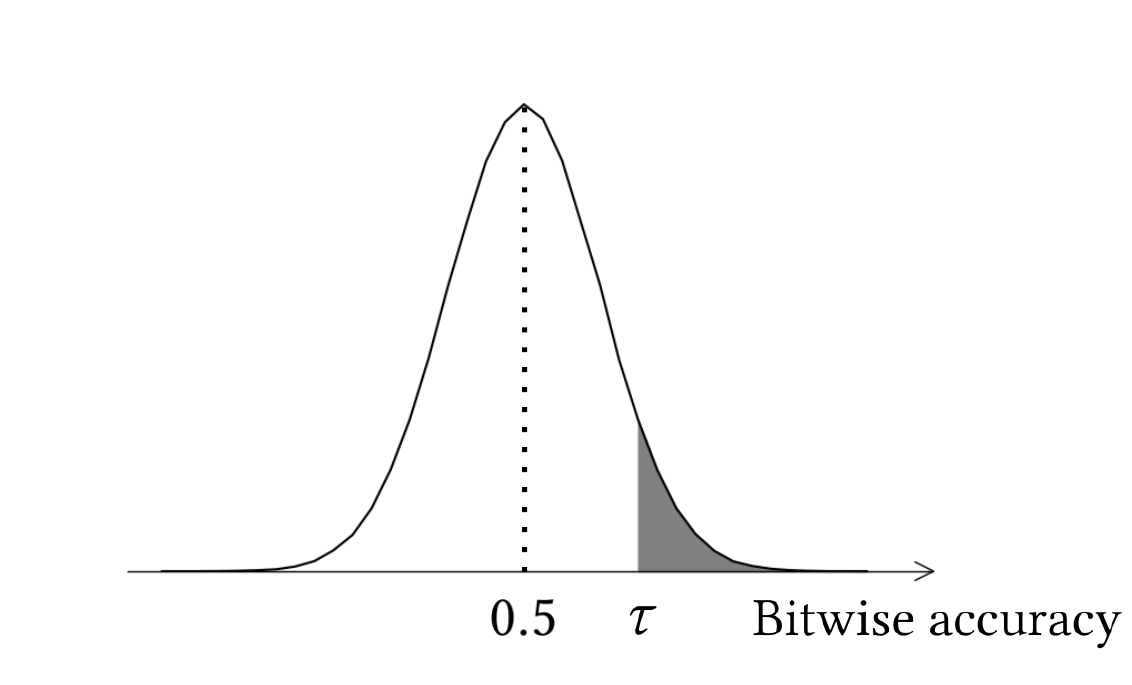}}
\subfloat[Double-tail detector]{\includegraphics[width=0.23\textwidth]{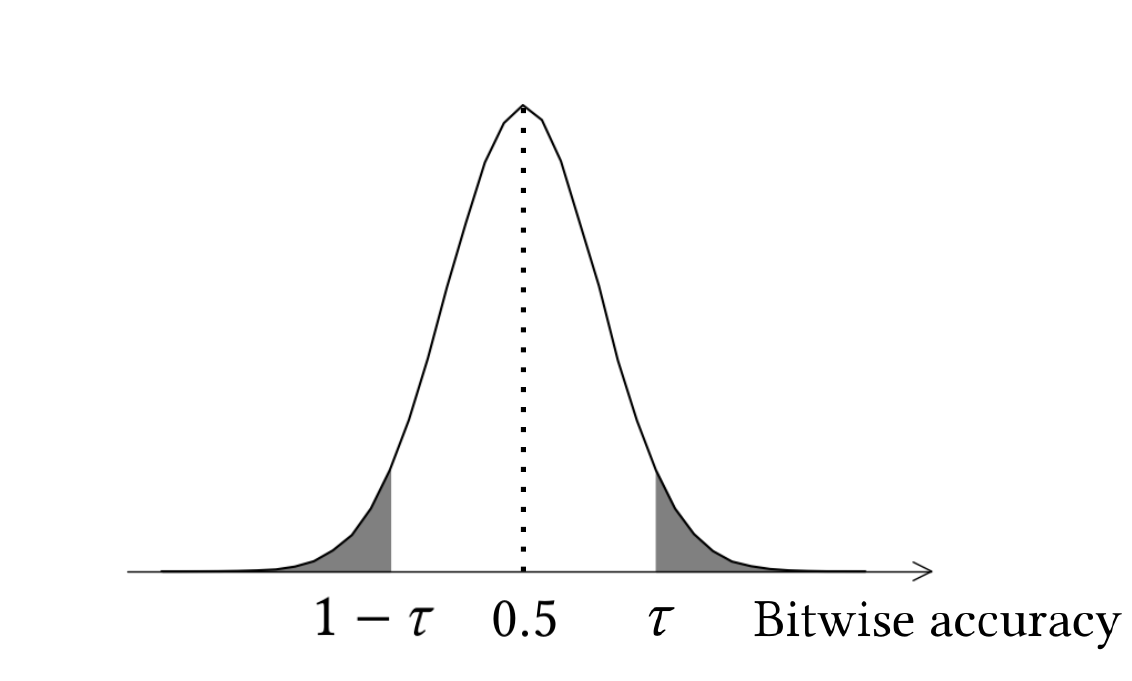}}
\caption{Illustration of (a) single-tail detector  and (b) double-tail detector  with threshold $\tau$.  The bitwise accuracy of an original image $I_o$ follows a binomial distribution divided by $n$, i.e., $BA(D(I_o), w) \sim B(n,0.5)/n$. The area of the shaded region(s) is the \emph{false positive rate (FPR)} of a detector.}
\label{illustration-detector}
\vspace{-5mm}
\end{figure}

\myparatight{Detection setup} We use $I$ to denote an \emph{image}, $I_o$ to denote an \emph{original image} without watermark, $I_w$ to denote a \emph{watermarked image}, and $I_{pw}$ to denote a \emph{post-processed watermarked image}.  Note that, in our notations, $I$ could be an $I_o$,  $I_w$, or  $I_{pw}$. We use $BA(w_1,w_2)$ to denote the \emph{bitwise accuracy} of watermark $w_1$ compared to watermark $w_2$, i.e., $BA(w_1,w_2)$ is the fraction of bits that match in $w_1$ and $w_2$. 
Suppose a service provider (e.g., OpenAI) deploys a generative AI model (e.g., a text-to-image generative model)  as a cloud service and has a ground-truth watermark $w$. Given a  user query (known as \emph{prompt}), the cloud service uses the AI model to generate an image, embeds its watermark $w$ into it using the encoder (or the generated image already has watermark $w$~\cite{fernandez2023stable, yu2021artificial, wen2023tree}), and returns the watermarked image to the user. In such cloud service, detecting AI-generated images reduces to detecting watermarked images. Specifically, given an image $I$, we can decode a watermark $D(I)$ using the decoder. Then, we calculate the bitwise accuracy $BA(D(I),w)$ of the watermark $D(I)$ with respect to the ground-truth watermark $w$. A watermark-based detector (shown in Figure~\ref{illustration-detector}) leverages the bitwise accuracy to detect watermarked images, which we discuss below.

\myparatight{Single-tail detector} In the standard single-tail detector~\cite{fernandez2023stable,yu2021artificial}, an image $I$ is predicted as AI-generated if the bitwise accuracy of its decoded watermark is larger than a threshold $\tau$, i.e., $BA(D(I),w)>\tau$, where $w$ is the ground-truth watermark. A key challenge is how to set the threshold $\tau$ such that the \emph{false positive rate (FPR)}, i.e., the probability that an original image is falsely detected as AI-generated, is bounded by a small value $\eta$, e.g., $\eta=10^{-4}$. This challenge can be addressed by formally analyzing the relationship between the threshold $\tau$ and the FPR of the single-tail detector~\cite{fernandez2023stable,yu2021artificial}.  

Suppose $BA(D(I_o),w)=\frac{m}{n}$ for an original image $I_o$, where $n$ is the length (i.e., number of bits) of the watermark and $m$ is the number of matched bits between $D(I_o)$ and $w$. The key idea is that the service provider should pick the ground-truth watermark $w$ uniformly at random. Thus, the decoded watermark $D(I_o)$ is not related to the randomly picked $w$, and each bit of  $D(I_o)$ matches with the corresponding bit of $w$ with probability 0.5. As a result, $m$ is a random variable and follows a \emph{binomial distribution $B(n,0.5)$}. Therefore, the FPR (denoted as  $FPR_s(\tau)$) of the single-tail detector with threshold $\tau$ can be calculated as follows~\cite{fernandez2023stable,yu2021artificial}:
\begin{align}
    FPR_s(\tau) &= \text{Pr}(BA(D(I_o),w)>\tau) \nonumber \\
    \label{fpr-standard}
    &= \text{Pr}(m>n\tau) =\sum_{k=\lceil n\tau \rceil}^n {n \choose k} \frac{1}{2^n}, 
\end{align}
where $FPR_s(\tau)$ is defined for any original image and the randomness in calculating the probability stems from picking the ground-truth watermark $w$ uniformly at random.
Thus, to make $FPR_s(\tau) < \eta$, $\tau$ should be at least $\tau^*=\argmin_{\tau}$ $ \sum_{k=\lceil n\tau \rceil}^n {n \choose k} \frac{1}{2^n} < \eta$. For instance, when $n=256$ and $\eta=10^{-4}$, we have $\tau\geq \tau^* \approx 0.613$.

\myparatight{Double-tail detector} The single-tail detector can be easily evaded by simply extending standard adversarial examples to watermarking. In particular, a standard adversarial example based evasion attack  adds perturbation to a watermarked image such that the decoded watermark  has a very small bitwise accuracy, e.g., close to 0. However, we propose a double-tail detector to detect such perturbed images. Our key observation is that the watermarks decoded from original images have bitwise accuracy close to 0.5, while those decoded from watermarked images have large bitwise accuracy, e.g., close to 1. Thus, if the bitwise accuracy of the watermark decoded from an image is significantly smaller than 0.5, it is likely to be an adversarially perturbed image. Based on this observation, we propose a double-tail detector that detects an image $I$ as AI-generated if its decoded watermark has a bitwise accuracy larger than $\tau$ or smaller than $1-\tau$, i.e., $BA(D(I),w)>\tau$ or $BA(D(I),w)<1-\tau$. We  can calculate the FPR (denoted as $FPR_d(\tau)$) of the double-tail detector with threshold $\tau$ as follows:
\begin{align}
    FPR_d(\tau) &= \text{Pr}(BA(D(I_o),w)>\tau \text{ or } BA(D(I_o),w)<1-\tau) \nonumber \\
    \label{fpr-adaptive}
    &= \text{Pr}(m>n\tau \text{ or } m<n-n\tau) =2\sum_{k=\lceil n\tau \rceil}^n {n \choose k} \frac{1}{2^n}, 
\end{align}
where $FPR_d(\tau)$ is defined for any original image and the randomness stems from picking the ground-truth watermark $w$ uniformly at random. Therefore, to make $FPR_d(\tau) < \eta$, $\tau$ should be at least $\tau^*=\argmin_{\tau}$ $ 2\sum_{k=\lceil n\tau \rceil}^n {n \choose k} \frac{1}{2^n} < \eta$. For instance, 
when $n=256$ and $\eta=10^{-4}$, we have $\tau\geq \tau^* \approx 0.621$.

\myparatight{Deployment scenarios}  Watermark-based detection of AI-generated content is an emerging topic, and  how  watermark-based detectors will be deployed in the real-world is still an open question. Nevertheless, we envision  the following four deployment scenarios:

     {\bf Detection-as-a-service.} In this scenario, the service provider, who provides the cloud service to generate images, also provides detection-as-a-service to detect its generated images. A user can upload an image to the detection-as-a-service, which returns  a binary answer "AI-generated" or "non-AI-generated". In this scenario, the service provider is a computation/communication bottleneck.

         {\bf End-user detection.} In this scenario, the detector is deployed as an  end-user application (e.g., a mobile app, a browser plugin), which runs on end-user devices (e.g., smartphone, laptop).

    {\bf Public detection.} In this scenario, the service provider makes its decoder and ground-truth watermark $w$ public so everyone can locally detect images generated by the service provider's AI model. Note that  individuals may select their own personalized detection thresholds $\tau$ in public detection. 

    {\bf Third-party detection.} In this scenario, the service provider  shares its decoder and watermark $w$ with selected  third parties, so they can locally detect images generated by the service provider's AI model. For instance, OpenAI may share its decoder and watermark with Twitter, so the latter can detect images generated by OpenAI's models that are propagated on Twitter. Note that  third parties may select their preferred thresholds $\tau$ in third-party detection.

\section{Threat Model}
\vspace{-2mm}
\myparatight{Attacker's goal} Suppose an attacker uses the aforementioned cloud service to generate a watermarked image $I_w$.  The attacker aims to \emph{post-process} the watermarked image to evade watermark-based detection while maintaining its visual quality. The attacker may desire to achieve such goals in various scenarios. For instance, the attacker may use the generated image to spread disinformation on the Internet; and the attacker may claim ownership of the AI-generated image. Formally, the attacker aims to turn the watermarked image $I_w$ into a post-processed one $I_{pw}$ via adding  a small, human-imperceptible perturbation to it such that a detector falsely predicts $I_{pw}$ as non-AI-generated.

\myparatight{Attacker's background knowledge} Recall that a watermarking method has a ground-truth watermark $w$, an encoder, and a decoder. A watermark-based detector requires $w$, the decoder, and  a detection threshold $\tau$. Since detection does not involve the encoder, whether it is available to the attacker is not relevant. Nevertheless, we assume the attacker does not have access to the encoder. Since our attack is encoder-agnostic, it is applicable to watermarking methods~\cite{fernandez2023stable, yu2021artificial, wen2023tree} that embed watermarks to images at generation. Moreover, we assume the attacker does not have access to $w$. Depending on what information (decoder and/or $\tau$) of the detectors the attacker has access to, we consider the following two scenarios:

     {\bf White-box.} In this threat model, we assume the attacker has white-box access to the decoder of the detectors. This scenario arises in various circumstances: 1) an attacker can directly access the decoder when the service provider makes it public in public detection, e.g., the decoder used by Stable Diffusion is public~\cite{stable-diffusion-watermark-decoder}; 2) an attacker can reverse engineer the end-user application to obtain the decoder when the detector is deployed as an end-user application in end-user detection; 3) a third-party may leak the decoder in third-party detection; and 4) an insider may leak the decoder or an attacker can exploit the computer system vulnerabilities to perform a data leakage attack in detection-as-a-service. A recent example of third-party leakage (not watermarking model, though) is that Meta shared its LLaMA model with verified third parties, one of which leaked it to the public~\cite{Meta-leaked}.

   Note that, given a decoder, different detectors may use different $\tau$. For instance, in public detection (or third-party detection), different individuals (or third-parties) can choose their own $\tau$. Therefore, instead of evading a particular detector with a specific $\tau$, an attacker aims to post-process a watermarked image that can evade detectors with any detection threshold $\tau>0.5$ in the white-box setting.

     {\bf Black-box.} In this threat model, we assume the attacker has black-box access to a particular detector with a decoder and a $\tau$ (called \emph{target detector}), and the attacker aims to evade this target detector. Specifically, the attacker only has access to the binary detection result ("AI-generated" or "non-AI-generated") for any image. This threat model may arise in detection-as-a-service, end-user detection, or third-party detection. For instance, in detection-as-a-service or end-user detection, the attacker can query the target detector to obtain the detection result for any image. In third-party detection, the attacker can also obtain the detection result for any image from a particular third party, e.g., the attacker can upload an image to Twitter and obtain the detection result depending on whether the image is blocked by Twitter or not.

\myparatight{Attacker's capability} In the white-box setting, an attacker can post-process a watermarked image via analyzing the decoder. In the black-box setting, the attacker can query the target detector to obtain the detection result for any image. Moreover, we assume the attacker can query the target detector multiple times. For instance,  the attacker can easily send multiple query images to  detection-as-a-service or end-user detection and obtain detection results. 
We acknowledge that it may take a longer time for the attacker to query a target detector in third-party detection. For instance, when the third-party is Twitter, the attacker uploads a query image to Twitter and may have to wait for some time  before obtaining the detection result, i.e., Twitter  blocks or does not block the query image. However, as our experiments will show, an attacker only needs dozens of queries to evade a target detector while adding a small perturbation to a watermarked image.

\section{Our \algns}

\subsection{White-box Setting} 
\label{whiteboxattack}
Suppose we are given a watermarked image $I_w$ and a decoder $D$. An attacker's goal is to add a small, human-imperceptible perturbation $\delta$ to $I_w$ such that the post-processed watermarked image $I_{pw}=I_w+\delta$ {evades detectors with any $\tau>0.5$.} 
We first extend standard adversarial examples to watermarking to find the perturbation $\delta$, which, however, can be defended by the double-tail detector. Then, to address the limitation, we propose a new optimization problem to formulate finding the perturbation $\delta$ to  evade detection and design an algorithm to solve the optimization problem.

\subsubsection{Extending Standard Adversarial Examples to Watermarking (\algns-W-I)} We denote this variant as \algns-W-I, where W indicates the white-box threat model. 
The decoder  $D$ outputs a watermark, each bit of which can be viewed as a binary class. Therefore, given a watermarked image $I_w$, one way is to add perturbation $\delta$ to it such that $D$ outputs a different binary value for each bit of the watermark. Formally, inspired by the standard adversarial examples~\cite{szegedy2013intriguing}, we formulate the following optimization problem:
\begin{align}
    \label{opt-problem-adv}
     \min_\delta  & \ ||\delta||_\infty \nonumber \\
     s.t.\ & D(I_{w} + \delta) =\neg D(I_{w}),
\end{align}
where $||\delta||_\infty$ is the $\ell_\infty$-norm of the perturbation $\delta$ and $\neg$ means flipping each bit of the watermark $D(I_{w})$. This optimization problem is hard to solve due to the highly nonlinear constraint. To address the challenge, we reformulate the optimization problem as follows:
\begin{align}
    \label{opt-problem-adv-re}
     \min_\delta  & \ l(D(I_{w} + \delta), \neg D(I_{w})) \\
    s.t.\ &||\delta||_\infty \leq r, \nonumber \\
    & D(I_{w} + \delta) =\neg D(I_{w}), \label{adv-re-constraint}
\end{align}
where $l$ is a loss function to measure the distance between two watermarks and $r$ is a perturbation bound.
We discuss more details on solving this reformulated optimization problem in Section~\ref{solvinproblem}. 

 The loss function should be small when $D(I_{w} + \delta)$ is close to $\neg D(I_{w})$. For instance, the loss function could be $\ell_2$ distance, $\ell_1$ distance,  negative cosine similarity, or average cross-entropy loss. In defining the loss function, we treat $\neg D(I_{w})$ as desired "labels". Formally, for $\ell_2$ distance, we have $l(D(I_{w} + \delta), \neg D(I_{w}))=\sum_i (F(I_{w} + \delta)_i-\neg D(I_{w})_{i})^2$, where $F(I_{w} + \delta)$ is the second-to-last layer outputs of the decoder neural network $D$ and the subscript $i$ is the index in a vector/bitstring;  for $\ell_1$ distance, we have $l(D(I_{w} + \delta), \neg D(I_{w}))=\sum_i |F(I_{w} + \delta)_i-\neg D(I_{w})_{i}|$; and for negative cosine similarity, we have $l(D(I_{w} + \delta), \neg D(I_{w}))=1 - cos(F(I_{w} + \delta), \neg D(I_{w}))$, where we treat $F(I_{w} + \delta)$ and $w_t$ as vectors and $cos$ is the cosine similarity between them.  For cross-entropy loss, we can treat  $F(I_{w} + \delta)_i$ as the possibility that the $i$th bit is predicted as 1. Then we have $l(D(I_{w} + \delta),\neg D(I_{w}))=-\sum_i (\neg D(I_{w})_{i} \log F(I_{w} + \delta)_{i} + (1-\neg D(I_{w})_{i}) \log (1-F(I_{w} + \delta)_{i}))$. We use the second-to-last layer continuous-value outputs instead of the final binary outputs, because the binary outputs are obtained by thresholding the continuous-value outputs (see details in Section~\ref{sec:watermarkingmethod}) and thus contain no useful gradient information for updating the perturbation $\delta$.

 \subsubsection{Formulating a New Optimization Problem (\algns-W-II)} Given a watermarked image $I_w$, the perturbation $\delta$ found by solving the above optimization problem can evade the single-tail detectors with any  threshold $\tau > 0.5$. However, our double-tail detector can still detect such post-processed watermarked images because their watermarks have too small bitwise accuracy, as we formally show in our theoretical analysis in Section~\ref{theoryanalysis}.  To address the limitation, we propose a new optimization problem to formulate finding the perturbation $\delta$. Specifically, we aim to find a small perturbation $\delta$ such that  the decoded watermark $D(I_w+\delta)$ has a bitwise accuracy close to 0.5, compared to the ground-truth watermark $w$. As a result, the post-processed watermarked image is indistinguishable with original images with respect to bitwise accuracy, evading both single-tail and double-tail detectors. Formally, we formulate finding the  perturbation $\delta$ as the following optimization problem:
\begin{align}
    \label{opt-problem}
     \min_\delta  & \ ||\delta||_\infty \\
     \label{bitaccconstraint}
     s.t.\ & |BA(D(I_{w} + \delta), w) - 0.5| \leq \epsilon,
\end{align}
where  $BA(D(I_{w} + \delta), w)$ measures the bitwise accuracy of the watermark $D(I_{w} + \delta)$ compared to the ground-truth one $w$,  $\epsilon$ is a small value characterizing the difference between $BA(D(I_{w} + \delta), w)$ and 0.5, and we call the constraint of the optimization problem \emph{bitwise-accuracy constraint}. However, solving the above optimization problem faces two challenges:  1)  the attacker does not have access to the ground-truth watermark $w$, and 2) the constraint is highly nonlinear, making standard optimization method like \emph{gradient descent (GD)} hard to apply. Next, we discuss how to address the two challenges.

\myparatight{Addressing the first challenge} One way to address the first challenge is to replace the ground-truth watermark $w$ as the watermark $D(I_w)$ decoded from the watermarked image $I_w$ in the optimization problem. However, when the decoded watermark $D(I_w)$ is quite different from $w$,  even if the found perturbation $\delta$ satisfies $|BA(D(I_{w} + \delta), D(I_w)) - 0.5| \leq \epsilon$, there is no formal guarantee that the bitwise-accuracy constraint in Equation~\ref{bitaccconstraint} is satisfied. To address the challenge, we replace the ground-truth watermark $w$ as a watermark $w_t$ picked uniformly at random, where we call $w_t$ \emph{target watermark}. Moreover, we reformulate the optimization problem such that when the watermark $D(I_{w} + \delta)$ decoded from the post-processed watermarked image is very close to $w_t$, it is guaranteed to satisfy the bitwise-accuracy constraint in Equation~\ref{bitaccconstraint} with high probability. Intuitively, since $w_t$ is picked uniformly at random, it has a bitwise accuracy close to 0.5 compared to any ground-truth watermark $w$. Therefore,  when $D(I_{w} + \delta)$ is close to $w_t$, it is likely to have a bitwise accuracy close to 0.5 as well. 

\myparatight{Addressing the second challenge} Due to the bitwise-accuracy constraint, it is hard to apply an iterative method like GD. This is because it is hard to find the gradient of $\delta$, moving $\delta$ along which  can make the bitwise-accuracy constraint more likely to be satisfied. To address this challenge, we reformulate the optimization problem such that it is easier to find a gradient along which  $\delta$ should be moved. Combining our strategies to address the two challenges, we reformulate the optimization problem as follows:  
\begin{align}
    \label{re-opt-problem}
     \min_\delta  & \ l(D(I_{w} + \delta), w_t)\\
     s.t.\ &||\delta||_\infty \leq r, \nonumber \\
     \label{bitconstraint}
     & BA(D(I_{w} + \delta), w_t) \geq 1-\epsilon,
\end{align}
where  $l$ is a loss function to measure the distance between $D(I_{w} + \delta)$ and $w_t$,   $r$ is a perturbation bound, and $\epsilon$ is a small number. 
Our reformulated optimization problem means that we aim to find a perturbation bounded by $r$ to minimize the loss between $D(I_{w} + \delta)$ and $w_t$ such that the bitwise accuracy $BA(D(I_{w} + \delta), w_t)$ is close to 1. Note that a small $r$ may not be able to generate a perturbation $\delta$ that satisfies the constraint in Equation~\ref{bitconstraint}. Therefore, as detailed in our method to solve the optimization problem, we  perform a binary search to find the smallest $r$ such that the found perturbation $\delta$ satisfies the constraint in Equation~\ref{bitconstraint}.

\subsubsection{Solving the Optimization Problems} 
\label{solvinproblem}
We propose a unified framework to solve the reformulated optimization problems in \algns-W-I and \algns-W-II. Our key idea of solving the reformulated optimization problems is that we use the popular~\emph{projected gradient descent (PGD)}~\cite{madry2017towards} to iteratively find the perturbation $\delta$ that satisfies the constraints (if possible) for a given $r$. Then, we perform binary search over $r$ to find the smallest perturbation $\delta$ that satisfies the constraints. Specifically, the binary search interval $[r_a, r_b]$ is initialized such that $r_a=0$ and $r_b$ is a large value (e.g., 2). Then, we pick $r=(r_a+r_b)/2$ and solve a reformulated optimization problem for the given $r$. If the found perturbation $\delta$ satisfies the constraint in the reformulated optimization problem, then we update $r_b=r$, otherwise we update $r_a=r$. We repeat the process until the binary search interval size is smaller than a threshold, e.g., $r_b-r_a\leq 0.001$ in our experiments. Algorithm~\ref{WEvadeWhite} in Appendix shows our binary search process, where the target watermark $w_t=\neg D(I_{w})$ in \algns-W-I and $w_t$ is a randomly picked watermark in \algns-W-II. The function FindPerturbation solves a reformulated optimization problem to find $\delta$ for a given $r$.

Next, we discuss  the function FindPerturbation, which is illustrated in Algorithm~\ref{PGD} in Appendix.  We solve the  optimization problem for a given $r$ using PGD. The perturbation $\delta$ is initialized to be 0. In each iteration, we compute the gradient of the loss function $l(D(I_w +\delta), w_t)$ with respect to $\delta$ and move $\delta$ towards the inverse of the gradient by a small step $\alpha$, which is known as \emph{learning rate}.  
If the $\ell_\infty$-norm of  $\delta$ is larger than the perturbation bound $r$, we project it so its  $\ell_\infty$-norm is $r$. We repeat the process for \emph{max\_iter} iterations and stop the iterative process early if the constraint in the reformulated optimization problem (i.e., Equation~\ref{adv-re-constraint} in \algns-W-I or Equation~\ref{bitconstraint} in \algns-W-II) is already satisfied.

\subsection{Black-box Setting}

\vspace{-2mm}
\myparatight{Surrogate-model-based (\algns-B-S)} The first direction is that the attacker trains a surrogate encoder/decoder, and then performs white-box attacks based on its surrogate decoder. The key hypothesis of such method is that the surrogate detector outputs a similar watermark with the target decoder for a post-processed watermarked image, and thus the post-processed watermarked image constructed to evade the surrogate decoder based detector may also evade the target detector. Specifically, the attacker collects some images and trains an encoder/decoder using the watermarking algorithm on its own images. The attacker's images and the service provider's images used to train encoders/decoders  may be from different distributions. After training a surrogate encoder and decoder, the attacker can turn a watermarked image $I_w$ into a post-processed one $I_{pw}$ using the surrogate decoder and the white-box attack, e.g., \algns-W-II in our experiments. Note that \algns-B-S does not rely on information of the target detector (e.g., target decoder and $\tau$), and thus the same $I_{pw}$ could be used for all detectors.

\myparatight{Query-based (\algns-B-Q)} \algns-B-S does not directly take information about the \emph{target detector} into consideration. As a result, the surrogate decoder may be quite different from the target decoder, leading to low evasion rates as shown in our experiments. To address the challenge, \algns-B-Q finds the post-processed watermarked image $I_{pw}$ by directly querying the target detector. Note that in this setting, we post-process a watermarked image to evade a target detector with a particular threshold $\tau$, unlike the white-box setting where we aim to evade detectors with any threshold $\tau>0.5$. 
Finding $I_{pw}$ in such scenario can be viewed as finding adversarial example to the target detector (i.e., a binary classifier) which returns a hard label for a query image. Therefore, we extend state-of-the-art hard-label query-based adversarial example technique called HopSkipJump~\cite{chen2020hopskipjumpattack} to find $I_{pw}$ in our problem. 

Specifically, HopSkipJump first generates a random initial $I_{pw}$ that evades the target detector by blending the given watermarked image $I_w$ with uniform random noise. Then, HopSkipJump  iteratively moves $I_{pw}$ towards  $I_w$ to reduce perturbation while always guaranteeing that $I_{pw}$  evades detection. In each iteration, HopSkipJump returns a new $I_{pw}$ and the number of queries to the target detector API used to find such $I_{pw}$. HopSkipJump stops the iterative process when reaching a given \emph{query budget}. We found that simply applying HopSkipJump to watermark-based detector leads to large perturbations. This is because 1) the random initial $I_{pw}$ may be far away from $I_w$, and 2) the perturbation may increase after some iterations before reaching the query budget.

Our \algns-B-S extends HopSkipJump by addressing the two limitations. First, instead of using a random initial $I_{pw}$, \algns-B-S uses a post-processed version of $I_{w}$ as the initial  $I_{pw}$. For instance, we can use JPEG compression to post-process $I_w$ as the  initial $I_{pw}$. In particular, we decrease the quality factor $Q$ of JPEG in the list [99, 90, 70, 50, 30, 10, 1] until finding a post-processed version of $I_{w}$ that evades detection, which is our initial $I_{pw}$. When none of the quality factor can generate a post-processed version of $I_w$ that evades the target detector, we revert to the random initial $I_{pw}$ adopted by HopSkipJump. Second, we early stop the iterative process when the perturbation in $I_{pw}$ increases in multiple (denoted as $ES$) consecutive iterations. Algorithm~\ref{blackboxquery} in Appendix shows our \algns-B-S, where the function HopSkipJump($I_{pw}$) returns a new $I_{pw}$ and the number of queries to the API used to find it.

\section{Theoretical Analysis} 
\label{theoryanalysis}
Given a watermarked image $I_{w}$, our attack turns it into a post-processed watermarked image $I_{pw}$. We define \emph{evasion rate} of $I_{pw}$ as  the probability that it is falsely detected as non-AI-generated, where the randomness (if any) in calculating the probability  stems from our attack, e.g., the randomness in picking the target watermark $w_t$ in \algns-W-II. We formally analyze the evasion rate of \alg against both single-tail detector and double-tail detector in the white-box and black-box settings. All the proofs are shown in Appendix.

\subsection{White-box Setting}

\myparatight{\algns-W-I} Suppose a watermarked image $I_w$ can be correctly detected by a (single-tail or double-tail) detector with threshold $\tau > 0.5$.  The following theorem shows that the post-processed watermarked image $I_{pw}$ found by \algns-W-I is guaranteed to evade the single-tail detector with evasion rate 1, while it is guaranteed to be detected by the double-tail detector (i.e., evasion rate is 0). 
\begin{thm}
\label{theorem1}
Given a watermarked image $I_w$ that can be detected by a single-tail or double-tail detector with a threshold $\tau>0.5$. Suppose $I_{pw}$ is found by our \algns-W-I. $I_{pw}$ is guaranteed to evade the single-tail detector, but is guaranteed to be detected by the double-tail detector. Formally, we have the following:
\begin{align}
     \text{Single-tail detector: } &BA(D(I_{pw}), w) < \tau,  \\
     \text{Double-tail detector: } &BA(D(I_{pw}), w) < 1-\tau \nonumber \\
     &\text{ or } BA(D(I_{pw}), w) > \tau,
\end{align}
where $w$ is any unknown ground-truth watermark.
\end{thm}

\myparatight{\algns-W-II} The following theorems  show the evasion rates of \algns-W-II  against single-tail  and double-tail detectors.

\begin{thm}
\label{theorem2}
Given a watermarked image $I_w$ and a single-tail detector with any threshold $\tau > 0.5$. Suppose $I_{pw}$ is found by our \algns-W-II. For any ground-truth watermark $w$, the probability (i.e., evasion rate) that $I_{pw}$ successfully evades the single-tail detector can be lower bounded as follows:
\begin{align}
    & \text{Pr}(BA(D(I_{pw}), w) \leq \tau) \geq P(\lfloor (\tau-\epsilon)n \rfloor),
\end{align}
where $n$ is the watermark length and $P(t)=\text{Pr}(m \leq t)$ is the cumulative distribution function of the binomial distribution $m \sim B(n,0.5)$.
\end{thm}

\begin{thm}
\label{theorem3}
Given a watermarked image $I_w$ and a double-tail detector with any threshold $\tau > 0.5$. Suppose $I_{pw}$ is found by our \algns-W-II. For any ground-truth watermark $w$, the probability (i.e., evasion rate) that  $I_{pw}$  successfully evades the double-tail detector can be lower bounded as follows:
\begin{align}
    & \text{Pr}(1-\tau \leq BA(D(I_{pw}), w) \leq \tau) \geq 2P(\lfloor (\tau-\epsilon)n \rfloor)-1,
\end{align}
where $n$ is the watermark length and $P(t)=\text{Pr}(m \leq t)$ is the cumulative distribution function of the binomial distribution $m \sim B(n,0.5)$.
\end{thm}

Theorem~\ref{theorem2} and~\ref{theorem3} indicate that the evasion rate lower bound of a post-processed watermarked image $I_{pw}$ constructed by \algns-W-II depends on $\tau$ used by the detector, $\epsilon$ adopted by the attacker in  \algns-W-II, and the watermark length $n$. For instance, for a detector with a larger $\tau$, the evasion rate is larger.

\subsection{Black-box Setting}

\myparatight{\algns-B-S} The evasion rate of \algns-B-S relies on the "similarity" between the surrogate decoder $D'$ and target decoder $D$. Based on a formal definition of similarity between the watermarks decoded by the surrogate decoder $D'$ and target decoder $D$ for any image, we can derive the evasion rate of \algns-B-S. First, we formally define the similarity between $D'$ and $D$ as follows:
\begin{definition}[$(\beta,\gamma)$-similar]
\label{definition4}
Suppose we are given a surrogate decoder $D'$ and target decoder $D$. We say $D'$ and $D$ are $(\beta,\gamma)$-similar if their outputted watermarks have bitwise accuracy at least $\beta$ with probability at least $\gamma$ for an image $I$ picked from the watermarked-image space uniformly at random. Formally, we have:
\begin{align}
    Pr(BA(D'(I),D(I)) \geq \beta) \geq \gamma. 
\end{align}
\end{definition}

Then, given that $D'$ and $D$ are $(\beta,\gamma)$-similar,  the following theorem shows lower bounds of the evasion rates of \algns-B-S against single-tail detector and double-tail detector. 
\begin{thm}
\label{theorem-black}
     Suppose \algns-B-S finds an $I_{pw}$ based on a surrogate decoder $D'$; and $D'$ and the target decoder $D$ are $(\beta,\gamma)$-similar. Then, the evasion rates of $I_{pw}$ for a single-tail detector or double-tail detector with threshold $\tau>0.5$ are lower bounded as follows:
    \begin{align}
        &\text{Single-tail detector: } \nonumber \\
        &Pr(BA(D(I_{pw}),w) \leq \tau) \geq \gamma P(\lfloor (\tau+\beta-\epsilon-1)n \rfloor) \\
        &\text{Double-tail detector: } \nonumber \\
        &Pr(1-\tau \leq BA(D(I_{pw}),w) \leq \tau) \geq 2\gamma P(\lfloor (\tau+\beta-\epsilon-1)n \rfloor)-1, 
    \end{align}
    where $w$ is the unknown ground-truth watermark. 
\end{thm}

\myparatight{\algns-B-Q} \algns-B-Q starts from an initial $I_{pw}$ that evades the target detector. During the iterative process to reduce the perturbation, \algns-B-Q always guarantees that $I_{pw}$ evades detection. Therefore, the evasion rate of \algns-B-Q  is 1. Note that the evasion rate is only for  the target detector.
\section{Evaluation}

\subsection{Experimental Setup}
\label{sec:setup}

\myparatight{Datasets} We use three benchmark datasets, including COCO~\cite{lin2014microsoft}, ImageNet~\cite{deng2009imagenet}, and Conceptual Caption (CC)~\cite{sharma2018conceptual}. Following  HiDDeN~\cite{zhu2018hidden} and UDH~\cite{zhang2020udh}, we randomly sample 10,000 training images from each dataset to train watermarking encoder and decoder. For evaluation, we randomly sample 100 images from the testing set and embed a watermark into each image.  
For each image in all datasets, we re-scale its size to 128 $\times$ 128.

\myparatight{Post-processing methods} We compare with the following existing post-processing methods, which are widely used to measure robustness of watermarking methods.  Each of these post-processing methods has some parameter, which controls the amount of perturbation added to a watermarked image and thus evasion rate. 
    
     {\bf JPEG.} JPEG~\cite{zhang2020towards} is a popular image compression method. It has a parameter called  \emph{quality factor $Q$}. A smaller quality factor compresses an image more, is more likely to evade detection, and also adds larger perturbation. 
    
     {\bf Gaussian noise.} This method adds a random Gaussian noise to each pixel of a watermarked image. The Gaussian distribution has a mean 0 and standard deviation $\sigma$. The  $\sigma$ controls the perturbation and thus evasion rate. 
    
     {\bf Gaussian blur.} This method blurs a watermarked image. It has a parameter called kernel size $s$ and  standard deviation $\sigma$. We did not observe much impact of the kernel size once it is small enough, and thus we set $s=5$. However, we will vary $\sigma$ to control the perturbation added to watermarked images and thus evasion rate. 
    
     {\bf Brightness/Contrast.} This method adjusts the brightness and contrast of an image. Formally, the method has two parameters $a$ and $b$, where each pixel value $x$ is converted to $ax+b$. $b$ has a smaller impact. We set $b=0.2$ and vary $a$ to control the perturbation added to watermarked images.  

\myparatight{Watermarking methods} We consider two representative learning-based methods HiDDeN~\cite{zhu2018hidden} and UDH~\cite{zhang2020udh}, whose implementations are publicly available. To consider watermarks with different lengths, we use  30-bit watermarks in HiDDeN and  256-bit watermarks in UDH.  We use the default parameter settings of HiDDeN and UDH in their publicly available code. HiDDeN normalizes the pixel value range [0, 255] to be [-1, 1], while UDH normalizes to [0, 1]. We consider both standard training and adversarial training as described in Section~\ref{sec:watermarkingmethod}.  but  the encoders/decoders are trained using standard training unless otherwise mentioned. 
In adversarial training,  we randomly sample a post-processing method from no post-processing, the existing ones, and ours with a random parameter to post-process each watermarked image in a mini-batch. We use \algns-W-II  with the parameter $\epsilon=0.01$ if our adversarial post-processing method is sampled. For the existing methods, we consider the following range of parameters during adversarial training: $Q \in$ [10, 99] for JPEG, $\sigma \in$ [0, 0.1] for Gaussian noise, $\sigma \in$[0, 1.0] for Gaussian blur, and $a \in$ [1, 5] for Brightness/Contrast. We consider these parameter ranges  because parameters out of the ranges  impact the images' visual quality.

\myparatight{Evaluation metrics} We consider \emph{bitwise accuracy}, \emph{evasion rate}, and \emph{average perturbation}.  Bitwise accuracy of an image is the fraction of the bits of its watermark that match with the ground-truth one. Evasion rate is the fraction of post-processed watermarked images that evade detection. Perturbation added to a watermarked image is measured by its $\ell_\infty$-norm. For each dataset, we report bitwise accuracy, evasion rate, and perturbation averaged over 100 original/watermarked/post-processed testing images. Note that HiDDeN normalizes the pixel value range [0, 255] to be [-1, 1]. Therefore, we divide the  perturbation in HiDDeN by 2, so the perturbation represents the fraction of the pixel value range [0, 255] in both HiDDeN and UDH. For instance, a perturbation of 0.02 means changing each pixel value by at most $0.02 * 255=5$ of an image.

\myparatight{Parameter settings}  We set $max\_iter=5,000$, $\alpha=0.1$ for HiDDeN and $\alpha=1$ for UDH in \algns-W-I and \algns-W-II. We use a larger $\alpha$ for UDH because its watermark length is larger. We set $\epsilon=0.01$ in \algns-W-II. For \algns-B-Q, unless otherwise mentioned, we set the query budget  to be 2,000 and the early stopping threshold $ES=5$.  By default,  we use the $\ell_2$-distance as the loss function.  Unless otherwise mentioned, we show results when the dataset is COCO, watermarking method is HiDDeN, and detector is the double-tail detector.

\begin{figure}[!t]
\centering
\vspace{-3mm}
\subfloat[FPR]{\includegraphics[width=0.23\textwidth]{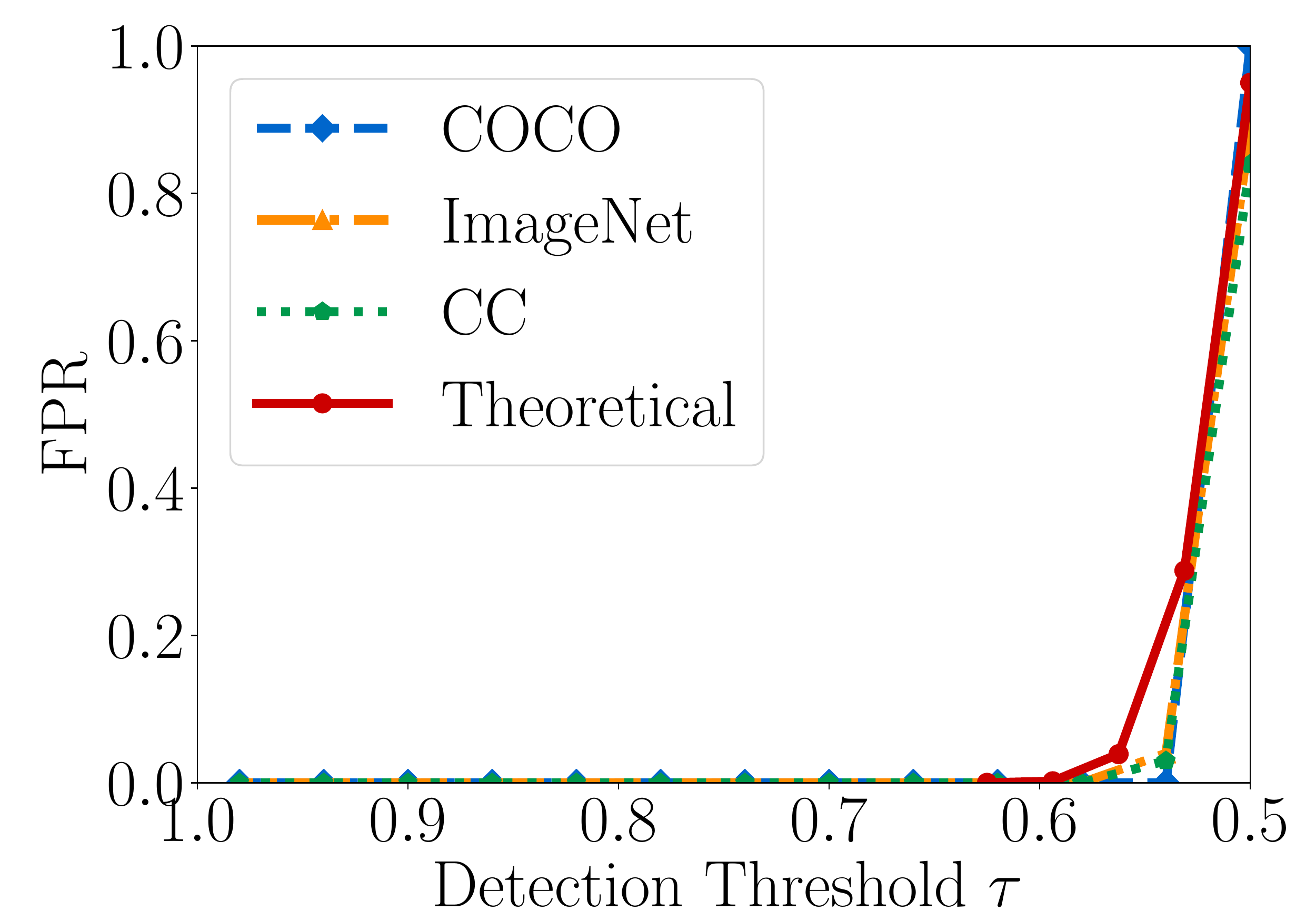}}
\subfloat[FNR]{\includegraphics[width=0.23\textwidth]{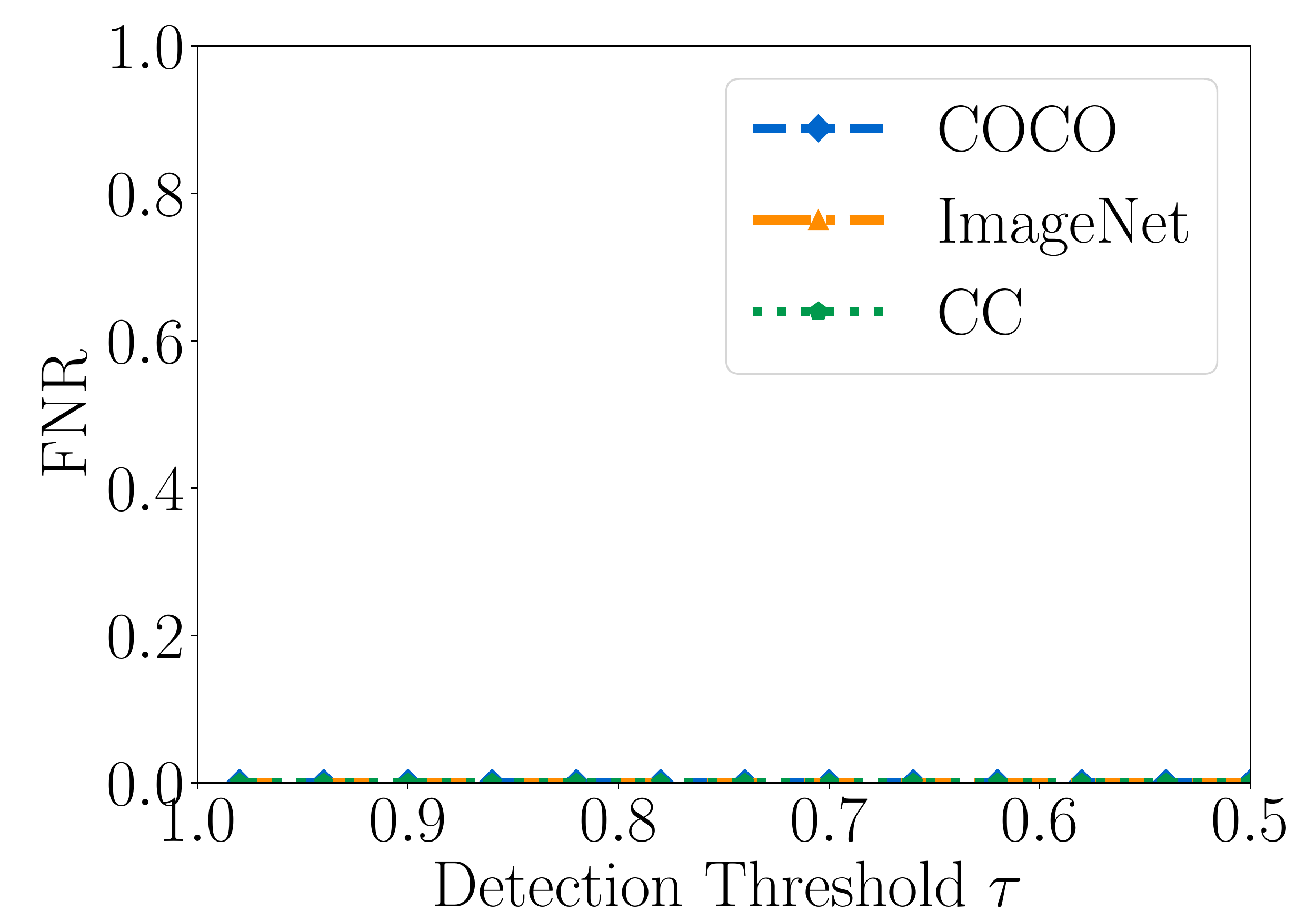}}
\caption{False positive rate (FPR) and false negative rate (FNR) of the double-tail detector based on UDH as the threshold $\tau$ varies  when there are no attacks.}
\label{no-attack-detector}
\vspace{-3mm}
\end{figure}

\begin{figure}[!t]
\centering
{
\includegraphics[width=0.3\textwidth]{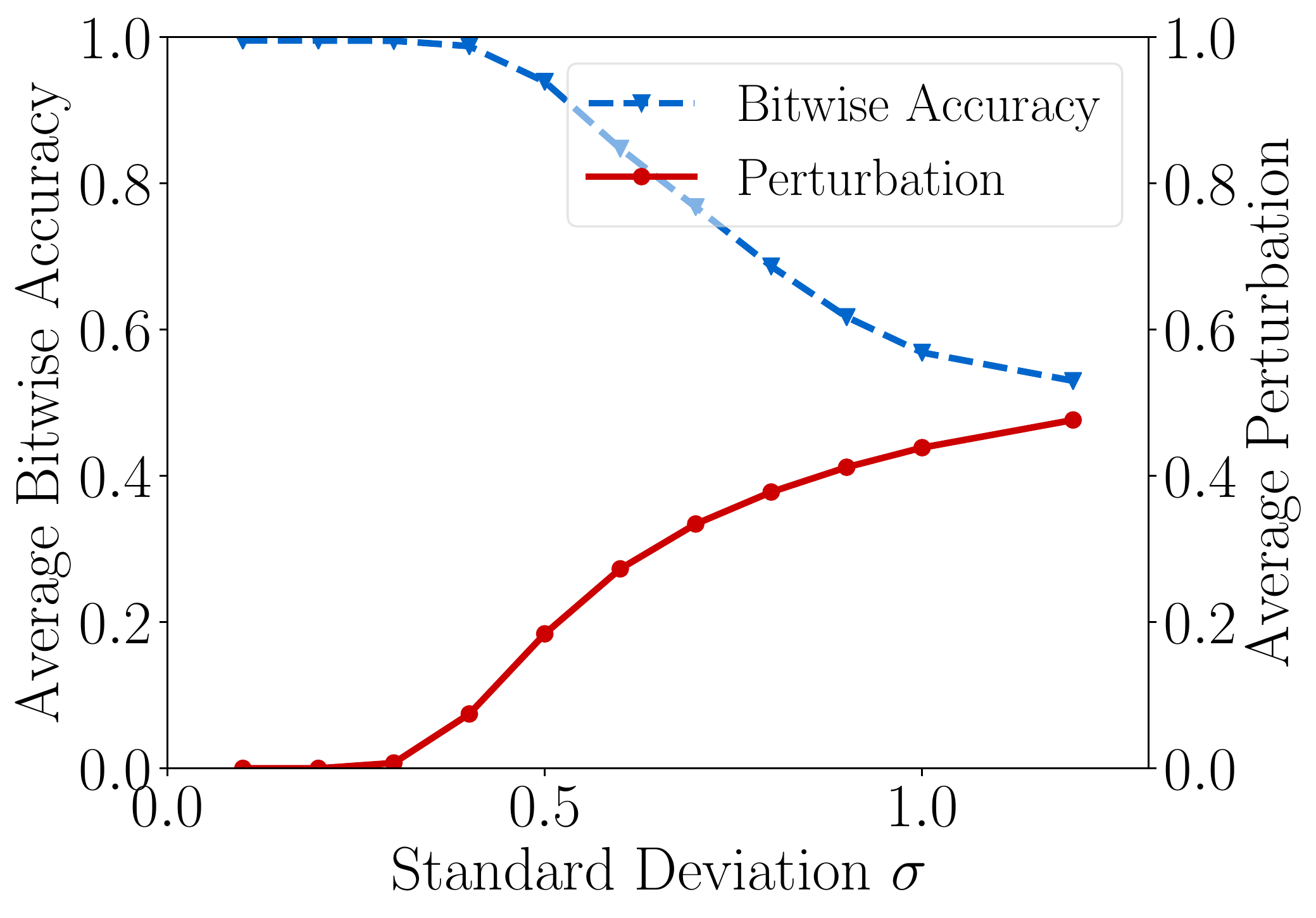}
}%
\caption{Average bitwise accuracy and average  perturbation of the post-processed watermarked images  when Gaussian blur uses different standard deviations. }
\label{hidden-coco-standard-parameter}
\end{figure}

\begin{figure}[!t]
\centering
\vspace{-3mm}
\subfloat[HiDDeN]{\includegraphics[width=0.23\textwidth]{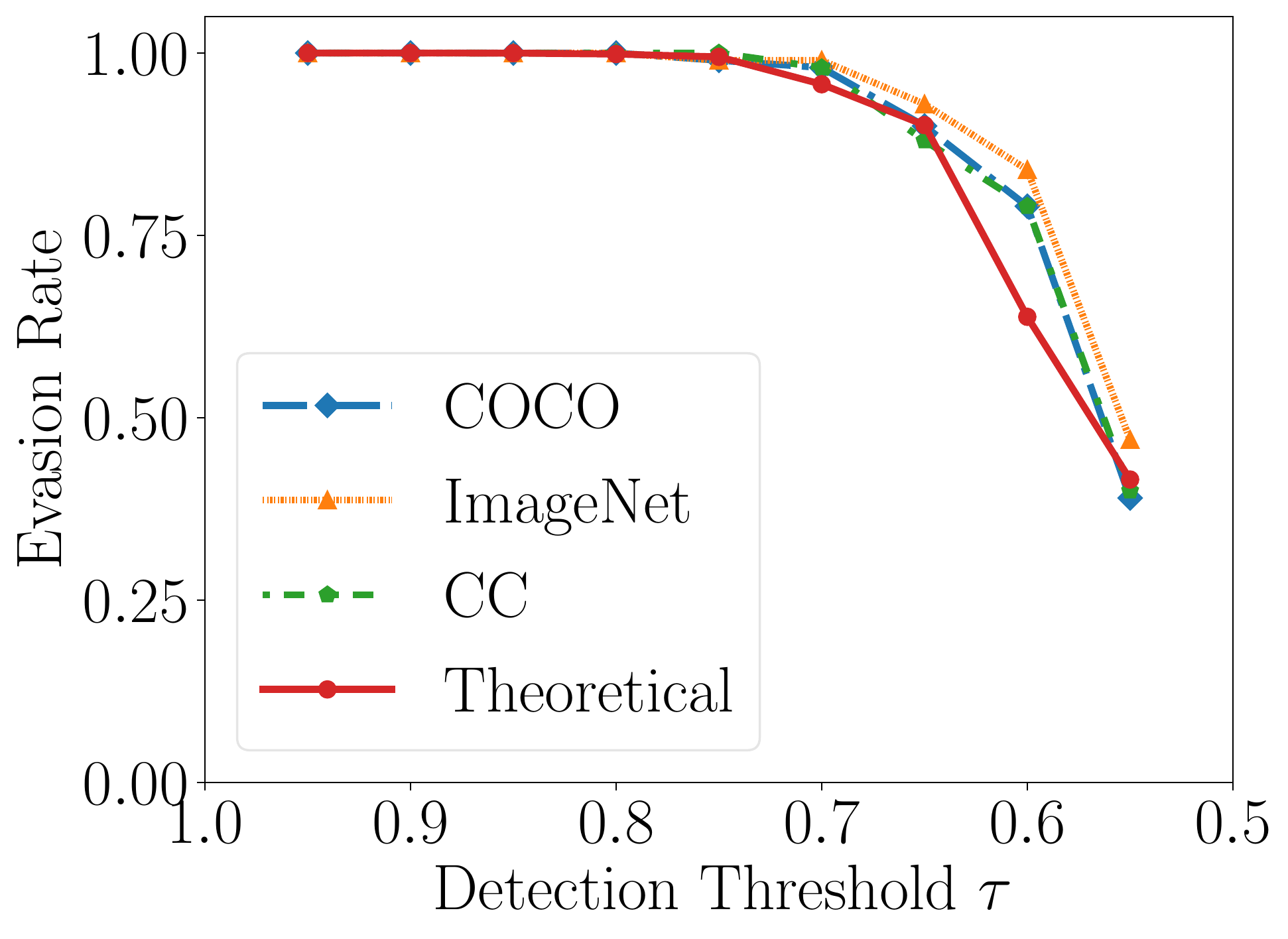}}
\subfloat[UDH]{\includegraphics[width=0.23\textwidth]{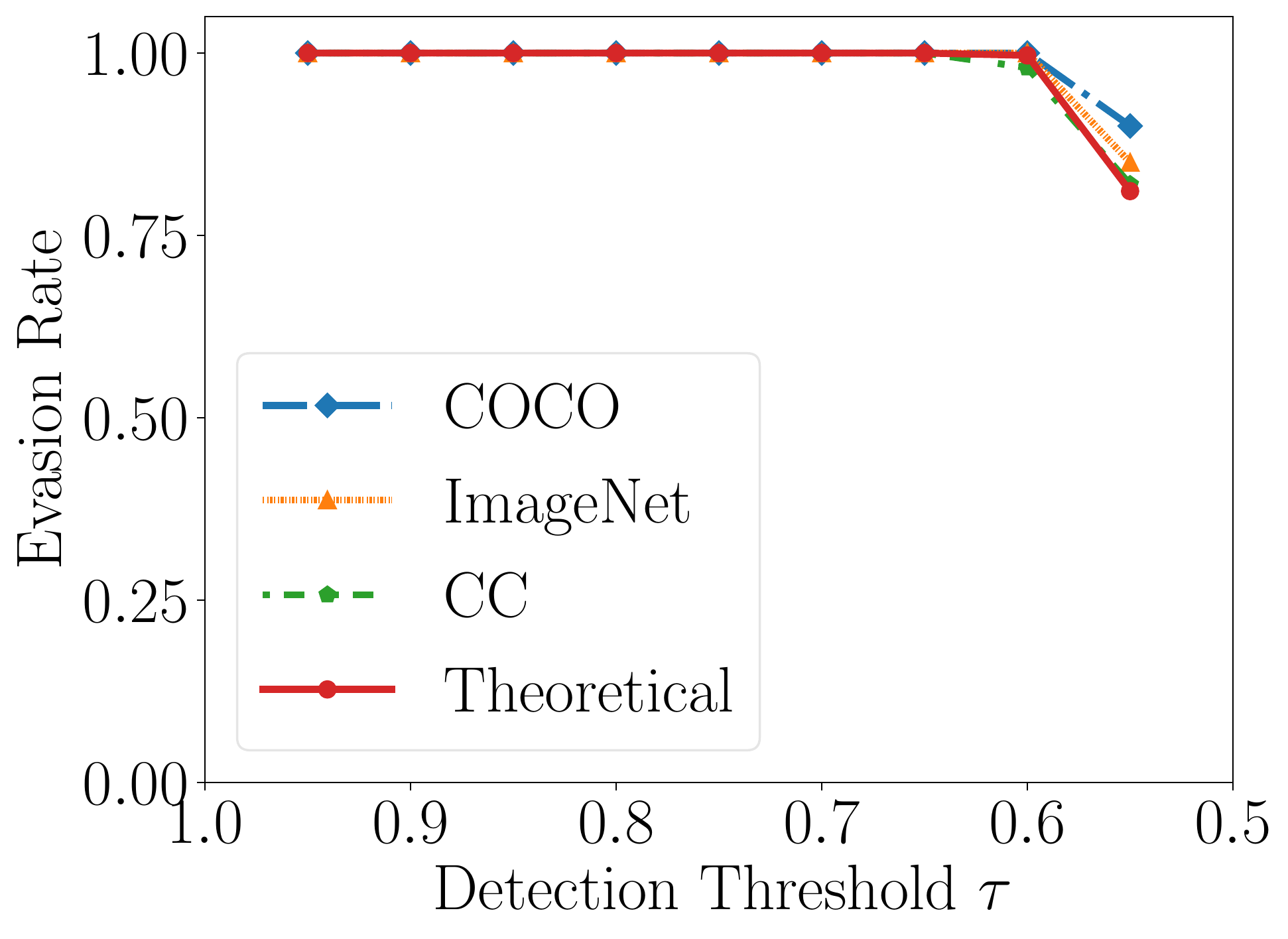}}

\caption{Evasion rates of \algns-W-II against the double-tail detector with different $\tau$ for the three datasets when the watermarking method is (a) HiDDeN and (b) UDH.}
\label{white-evasion}
\vspace{-3mm}
\end{figure}

\begin{figure*}[!t]
\centering
\vspace{-4mm}
\subfloat[COCO]{\includegraphics[width=0.33\textwidth]{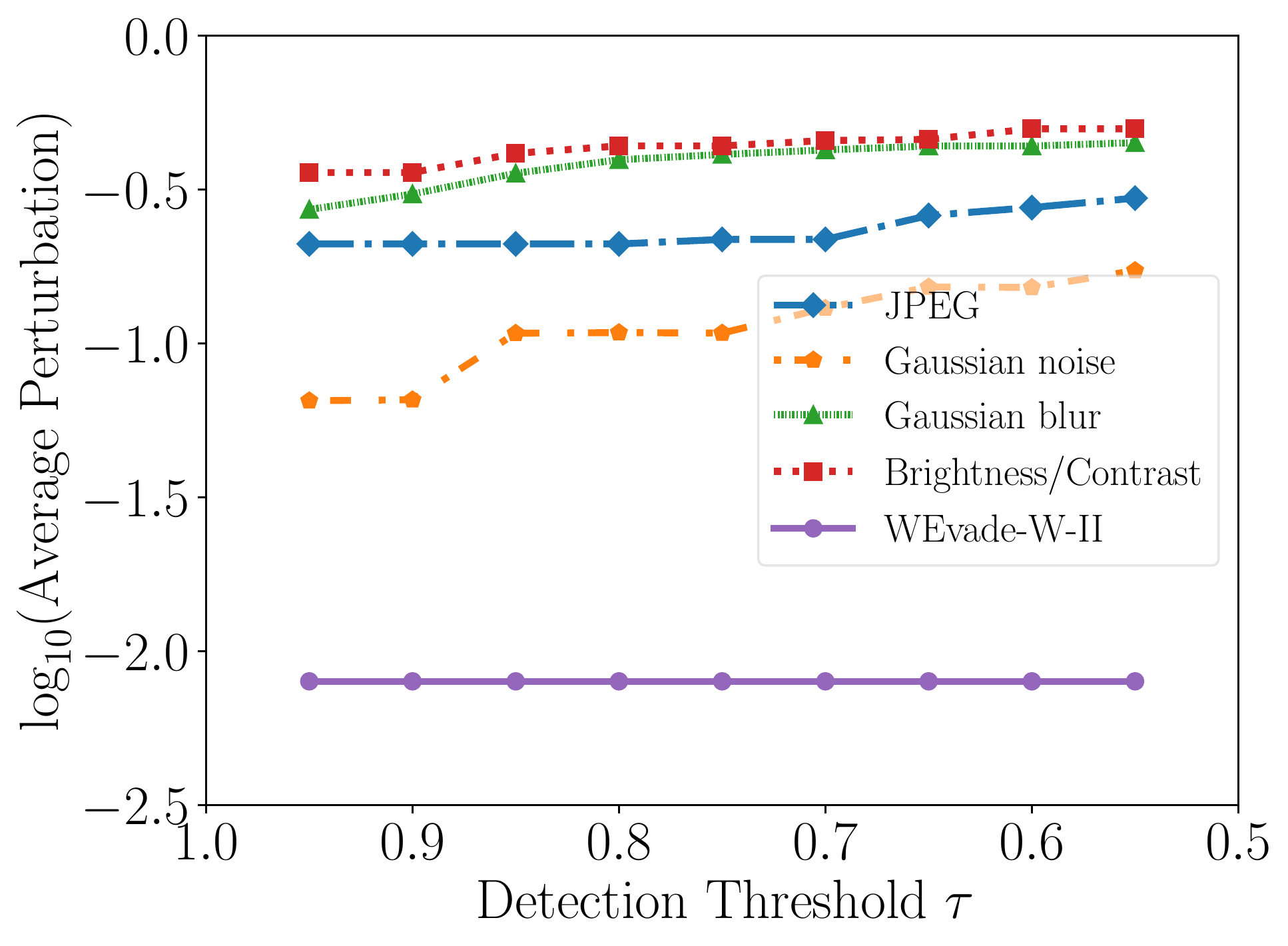}}
\subfloat[ImageNet]{\includegraphics[width=0.33\textwidth]{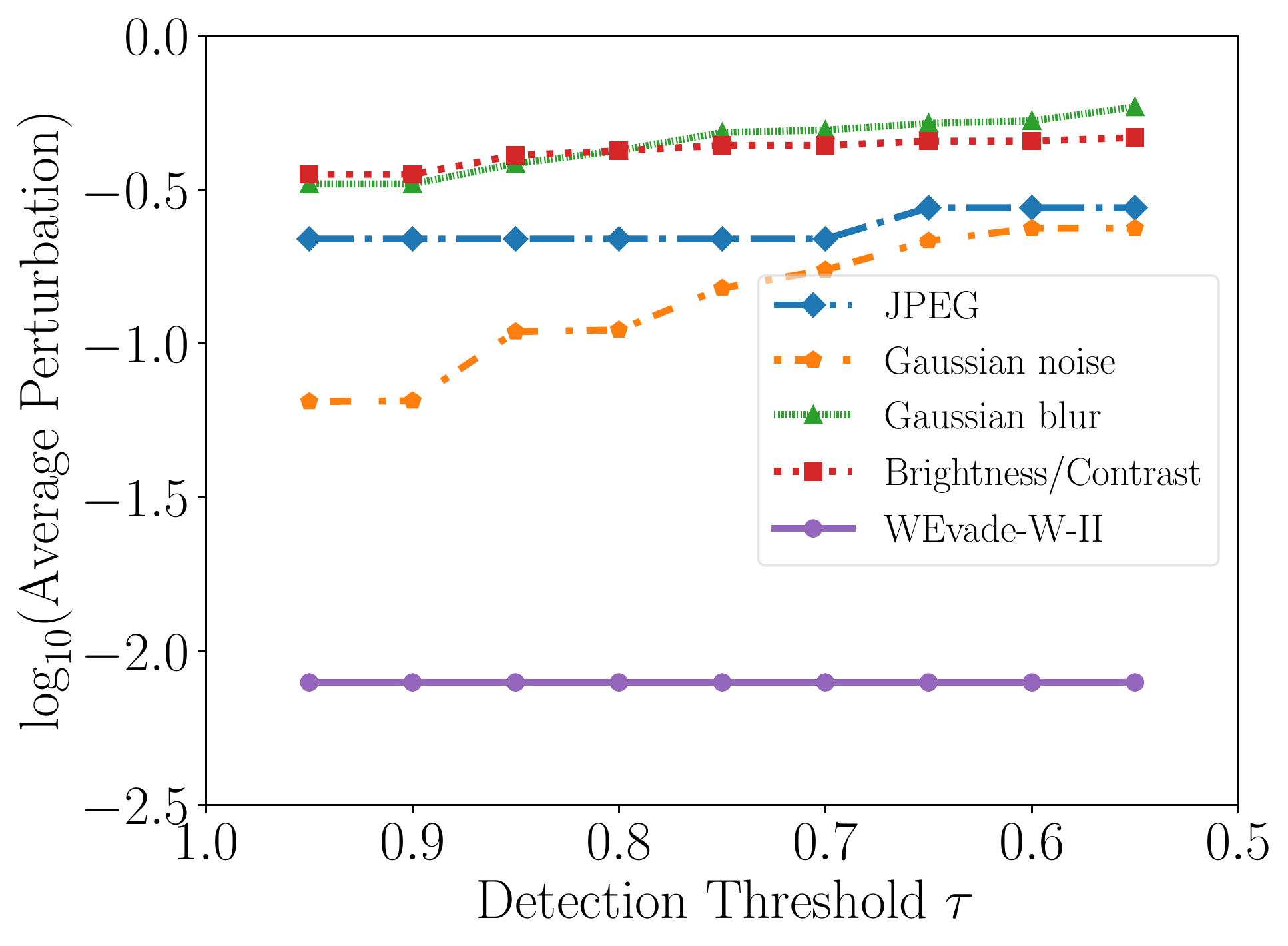}}
\subfloat[CC]{\includegraphics[width=0.33\textwidth]{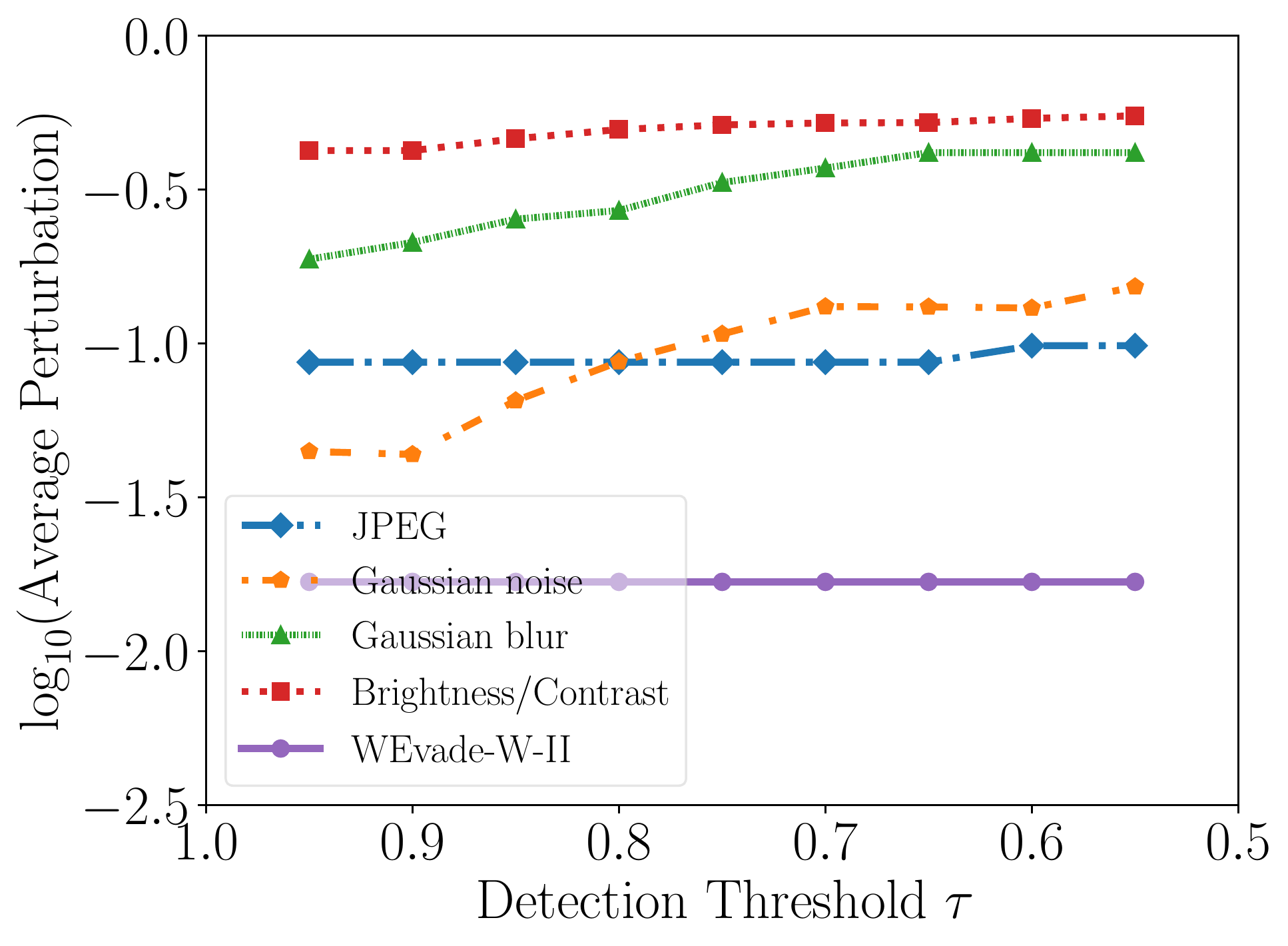}}

\caption{Average perturbation added by each post-processing method to evade the double-tail detector with different threshold $\tau$ in the white-box setting. We set the parameters of existing post-processing methods such that they achieve the same evasion rate as our \algns-W-II. The watermarking method is HiDDeN and the results for UDH are shown in Figure~\ref{white-standard-perturbation-udh} in Appendix.}
\label{white-standard-perturbation}
\vspace{-3mm}
\end{figure*}

\subsection{Detection Results without Attacks}
We first show detection results when there are no attacks to post-process watermarked images.  Figure~\ref{no-attack-detector} shows the false positive rate (FPR) and false negative rate (FNR) of the double-tail detector based on UDH when the threshold $\tau$ varies from 0.99 to 0.50, where FPR is the fraction of original testing images that are falsely detected as watermarked and FNR is the fraction of watermarked testing images that are falsely detected as original. The results for the double-tail detector based on HiDDeN and single-tail detector are shown in Figure~\ref{no-attack-detector-hidden} and Figure~\ref{no-attack-standard-detector} in Appendix, respectively. The "Theoretical" curves are the theoretical FPRs of the  detectors, i.e., $FPR_s(\tau)$ in  Equation~\ref{fpr-standard} and $FPR_d(\tau)$ in Equation~\ref{fpr-adaptive}.  There is no theoretical analytical form for FNR, and thus there are no curves corresponding to "Theoretical" in the FNR graphs. Note that $FPR_s(\tau)$ or $FPR_d(\tau)$ is the theoretical FPR for any original image when the ground-truth watermark is picked uniformly at random. More specifically, given any original image, if we pick 100 ground-truth watermarks uniformly at random, the theoretical FPR is roughly the fraction of the 100 trials in which the original image is falsely detected as watermarked.  The empirical FPR shown in  Figure~\ref{no-attack-detector} can be viewed as estimating the theoretical FPR of each original testing image  using one randomly picked ground-truth watermark and then averaging the estimated theoretical FPRs among the  original testing images.

We have three observations. First, under no attacks, both single-tail and double-tail detectors are accurate when the threshold $\tau$ is set properly. In particular, for HiDDeN (or UDH), both FPR and FNR of both detectors are consistently close to 0 on the three datasets when $\tau$ varies from 0.7 to 0.95 (or from 0.6 to 0.99). The range of such $\tau$ is wider for UDH than for HiDDeN, i.e., [0.6, 0.99] vs. [0.7, 0.95]. This is because UDH uses a longer watermark than HiDDeN, i.e., 256 vs. 30 bits. 
Second, the theoretical FPR is close to the empirical FPRs, i.e., the "Theoretical" curve is close to the other three FPR curves in a graph. They do not exactly match because the empirical FPRs are estimated using only one randomly picked ground-truth watermark. Third, given the same threshold $\tau$, the double-tail detector has a higher FPR than the single-tail detector, which is more noticeable when $\tau$ is small (e.g., 0.55). This is because the double-tail detector considers both the left and right tails of the bitwise-accuracy distribution (see illustration in Figure~\ref{illustration-detector}).

\subsection{Attack Results in the White-box Setting}
\myparatight{\alg outperforms existing post-processing methods} Each existing post-processing method has a parameter (discussed in Section~\ref{sec:setup}), which controls how much perturbation is added to a watermarked image. Figure~\ref{hidden-coco-standard-parameter} shows the average bitwise accuracy and average perturbation of the watermarked images post-processed by Gaussian blur with different parameter values, where HiDDeN and COCO dataset are used. Figure~\ref{hidden-imagenet-cc-parameter} and Figure~\ref{udh-standard-parameter} in Appendix show the results on other post-processing methods and datasets. Based on these results, we compare \alg with existing post-processing methods with respect to evasion rate and average perturbation added to the watermarked images. Note that there exists a trade-off between evasion rate and average perturbation. Therefore, for a given  threshold $\tau$, we tune the parameters of the existing methods such that they achieve  similar evasion rates (within 1\% difference) with \alg and we compare the average perturbation. 

Figure~\ref{white-evasion} shows the evasion rates of \algns-W-II when the double-tail detector uses different threshold $\tau$, while  Figure~\ref{white-standard-perturbation} shows the average perturbations  that each method requires to achieve such evasion rates. The "Theoretical" curves in Figure~\ref{white-evasion} correspond to the theoretical lower bounds of evasion rates of \algns-W-II in Theorem~\ref{theorem3}, i.e., $2P(\lfloor (\tau-\epsilon)n \rfloor)-1$. Specifically,   $\epsilon=0.01$ and $n=30$ in our experiments and we  use $2P(\lfloor (\tau-\epsilon)n \rfloor)-1$ to  calculate the lower bound of evasion rate for any $\tau$. The average perturbation of \algns-W-II is a straight line in Figure~\ref{white-standard-perturbation} because the perturbation added by \algns-W-II does not depend on $\tau$. Note that, in our experiments, we give advantages to existing post-processing methods, i.e., we assume they can tune their parameters for a given threshold $\tau$, while our \algns-W-II does not assume the knowledge of $\tau$.

First,  the empirical evasion rates are close to the "Theoretical" lower bounds in  Figure~\ref{white-evasion}, which validates our theoretical analysis. The empirical evasion rates are sometimes slightly lower than the theoretical lower bounds because the empirical evasion rates are calculated using a small number (100 in our experiments) of watermarked images. Second, our results show that \algns-W-II substantially outperforms existing post-processing methods. In particular, \algns-W-II requires much smaller perturbations to achieve high evasion rates. We also found that when existing  methods use parameter values to achieve average perturbations  no more than  \algns-W-II, their evasion rates are all 0. 

\begin{figure}[!t]
\centering
{
\includegraphics[width=0.227\textwidth]{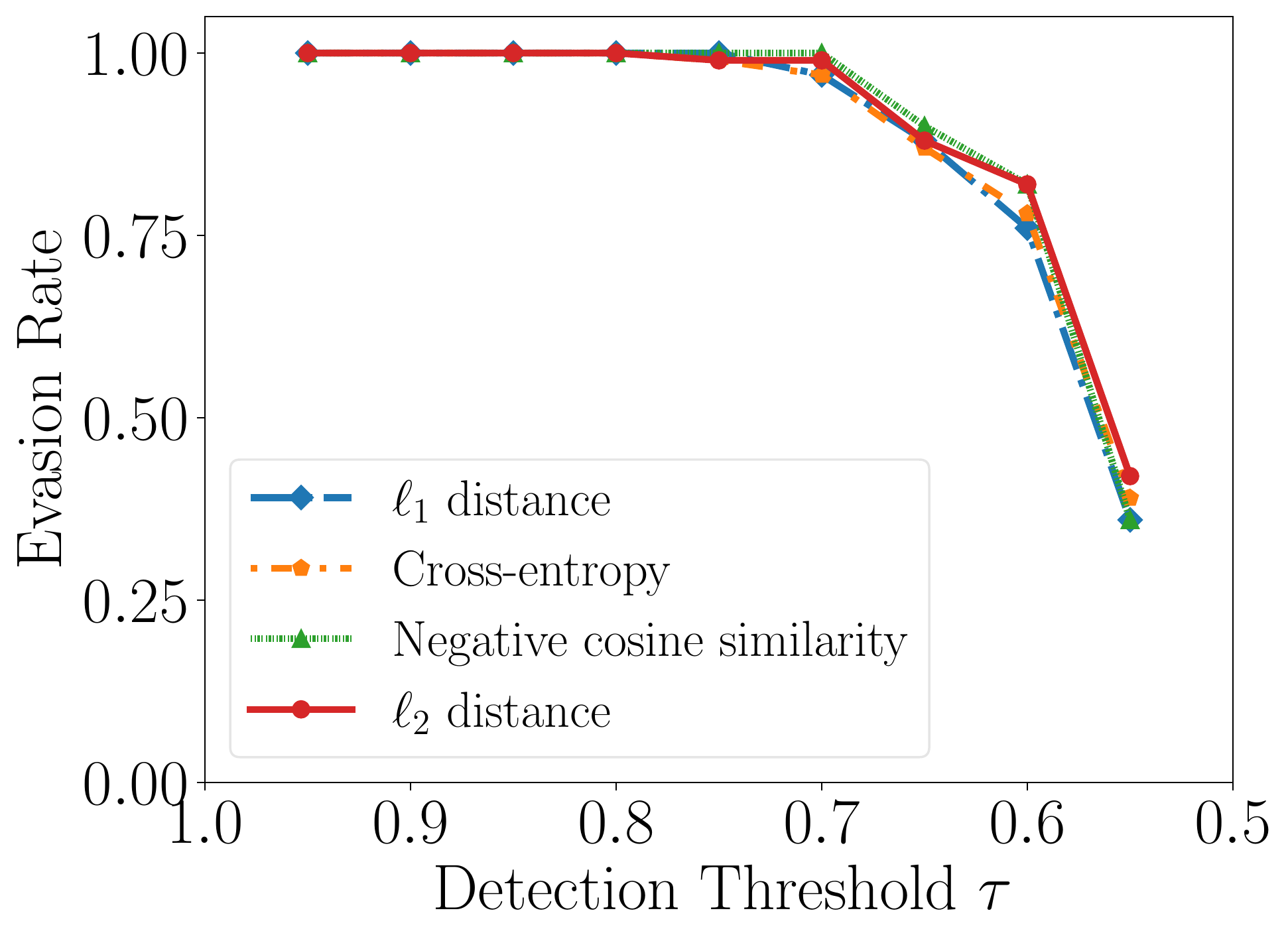}
}%
{
\includegraphics[width=0.233\textwidth]{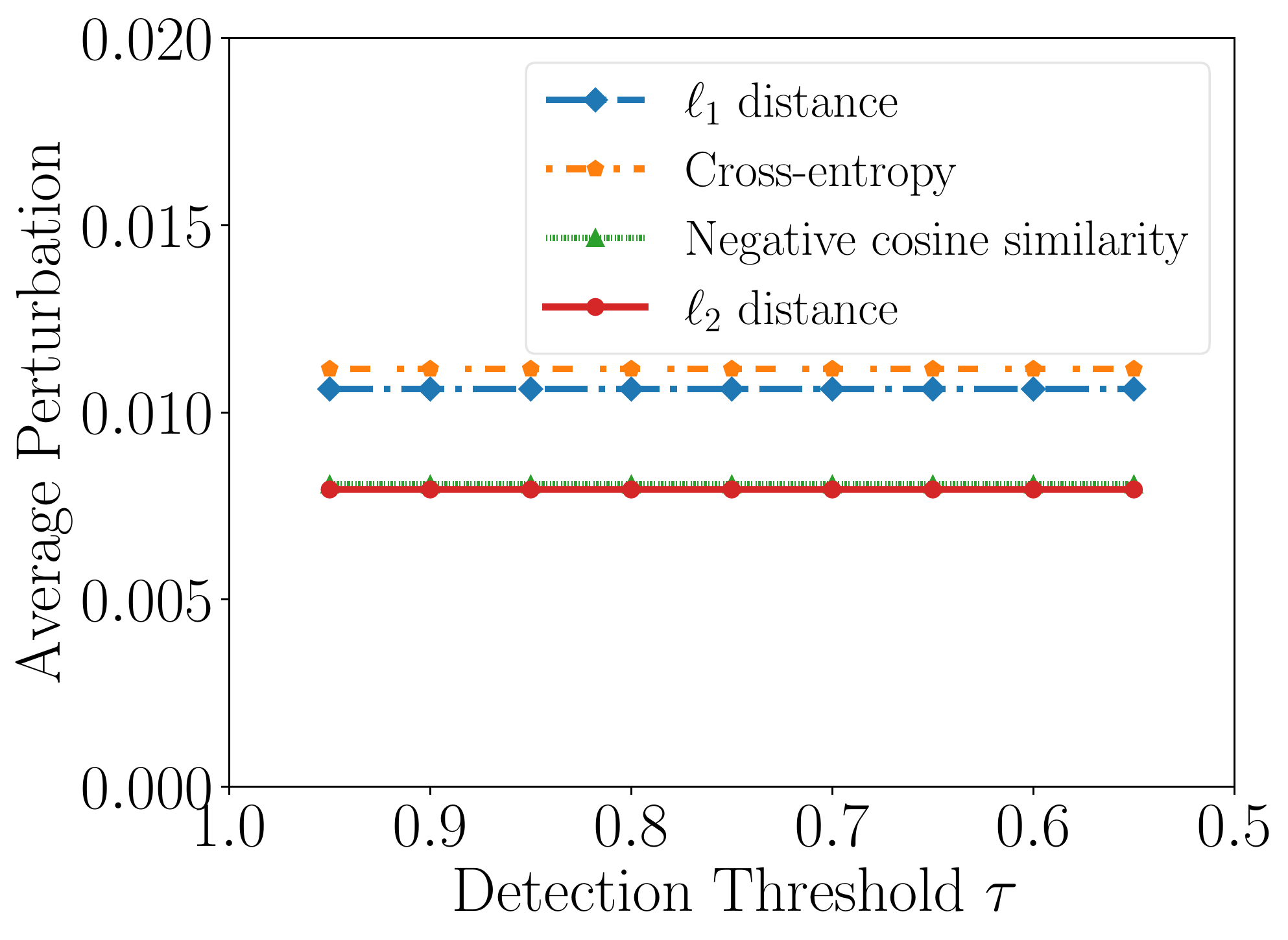}}
\caption{Comparing different loss functions.}
\label{loss-function-comparison}
\vspace{-3mm}
\end{figure}

\myparatight{Comparing \algns-W-I and \algns-W-II} Figure~\ref{variant-comparison} in Appendix shows the evasion rates and average perturbations of \algns-W-I and \algns-W-II as the single-tail detector or double-tail detector uses different threshold $\tau$, where the dataset is COCO and watermarking method is HiDDeN. First, we observe that  \algns-W-I achieves evasion rate of 1 for the single-tail detector while 0 for the double-tail detector, which is consistent with our Theorem~\ref{theorem1}.  Second, for the single-tail detector, \algns-W-I achieves higher evasion rates than \algns-W-II when $\tau$ is small (e.g., 0.6) but incurs larger average perturbation than \algns-W-II. This is because \algns-W-I adds (larger) perturbation to flip each bit of the watermark of the watermarked image. However, we stress that their average perturbations are both very small.  Third, for the double-tail  detector, \algns-W-II achieves higher evasion rates and incurs smaller average perturbations than \algns-W-I.  Note that the perturbations added by both \algns-W-I and \algns-W-II do not depend on the detector, and thus the average-perturbation curves for \algns-W-I (or \algns-W-II) are the same for the single-tail detector and double-tail detector in Figure~\ref{variant-comparison}.

\begin{figure}[!t]
\centering
{
\includegraphics[width=0.227\textwidth]{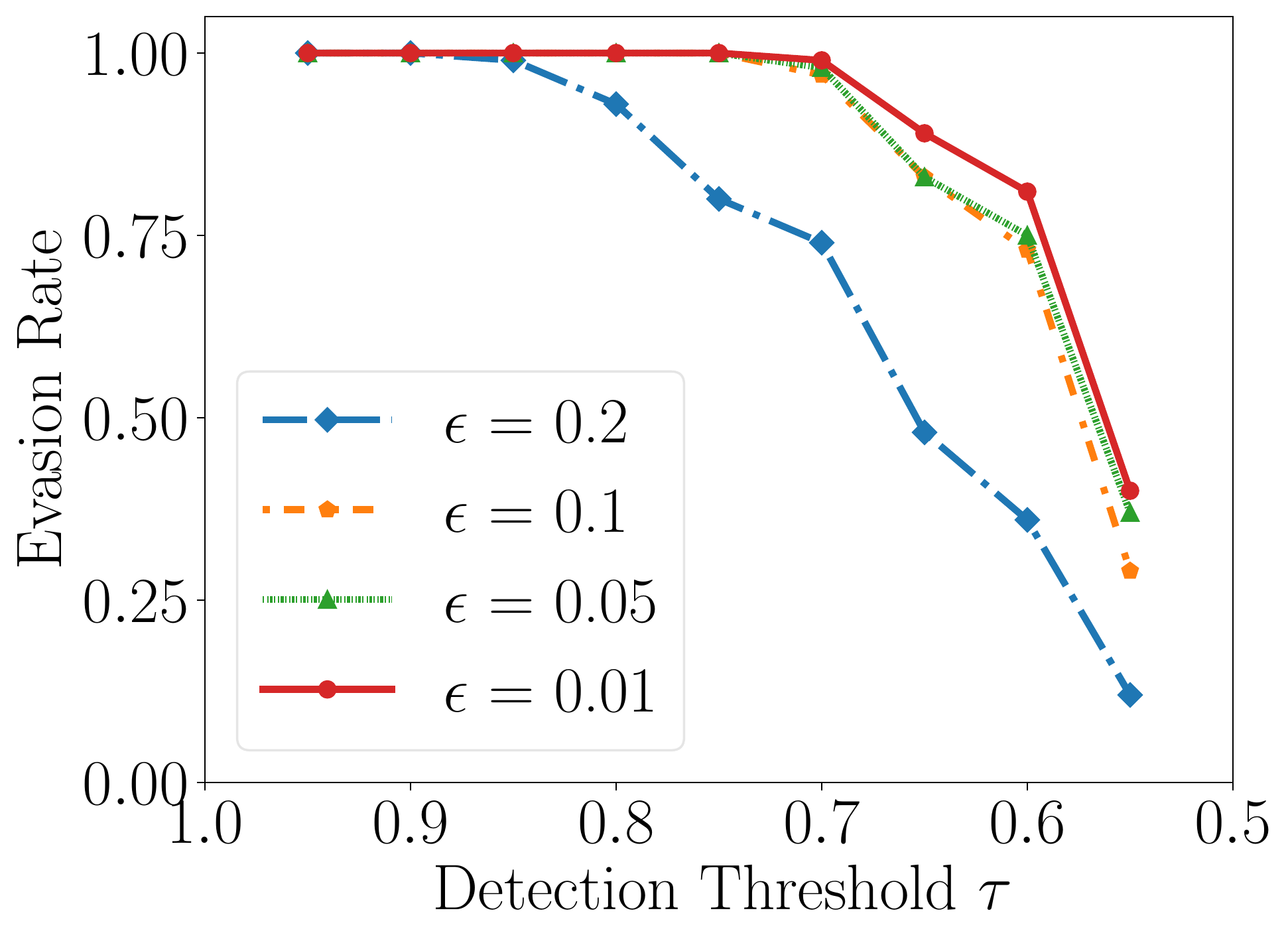}
}%
{
\includegraphics[width=0.233\textwidth]{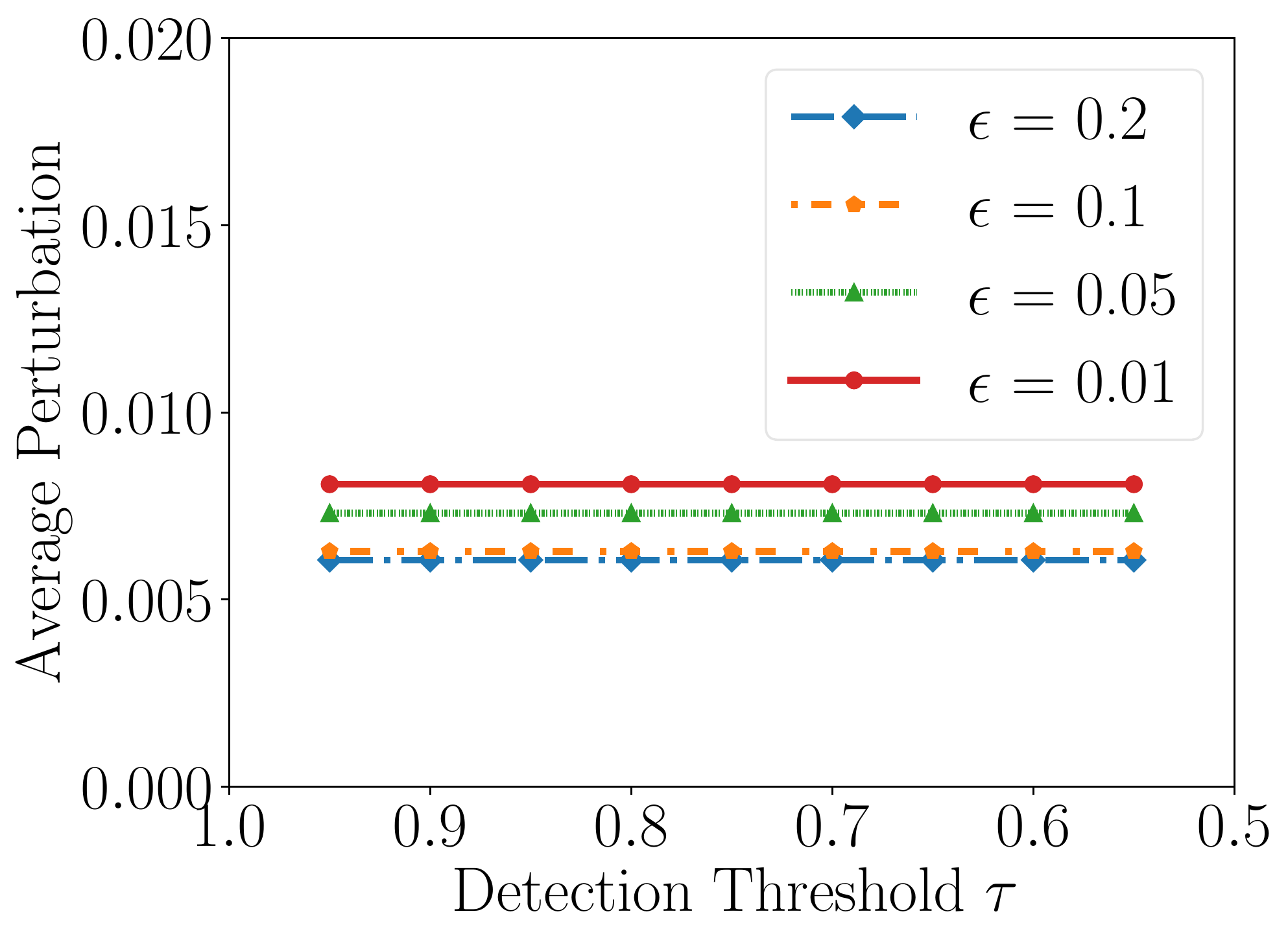}}
\caption{Comparing different $\epsilon$ values.}
\label{epsilon-comparison}
\vspace{-3mm}
\end{figure}

\begin{figure}[!t]
\centering
{\includegraphics[width=0.3\textwidth]{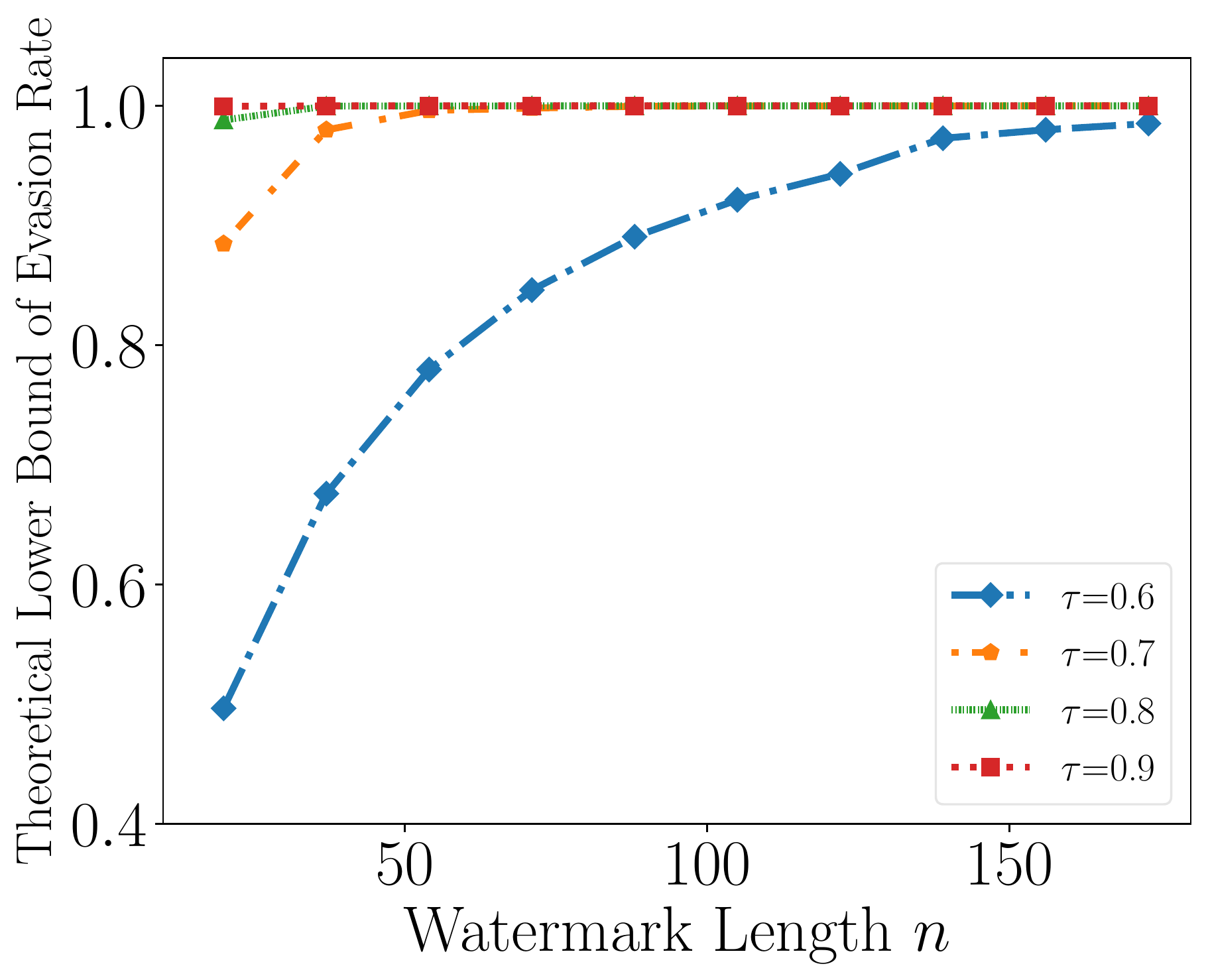}}
\caption{Impact of watermark length $n$.}
\label{n-comparison-length}
\vspace{-4mm}
\end{figure}

\myparatight{Impact of loss function} Figure~\ref{loss-function-comparison} compares different loss functions with respect to evasion rate and average perturbation of \algns-W-II. We observe that these loss functions achieve comparable results, though $\ell_2$-distance and negative cosine similarity achieve slightly smaller average perturbations. The reason is that,  in our Algorithm~\ref{WEvadeWhite}, we find the smallest perturbation that satisfies the constraint in Equation~\ref{bitconstraint} no matter what loss function is used; and in Algorithm~\ref{PGD}, we early stop as long as the constraint in Equation~\ref{bitconstraint} is satisfied. Moreover, our Theorem~\ref{theorem3} shows that the evasion rate of \algns-W-II does not depend on the loss function once the found perturbation satisfies the constraint in Equation~\ref{bitconstraint}.

\myparatight{Impact of $\epsilon$}
Figure~\ref{epsilon-comparison} compares different $\epsilon$ values with respect to evasion rate and average perturbation of \algns-W-II. We observe that $\epsilon$ achieves a trade-off between evasion rate and average perturbation. As $\epsilon$ increases, perturbation decreases because Equation~\ref{bitconstraint} is easier to be satisfied; but  evasion rate also decreases because the decoded watermark is less similar to the target watermark $w_t$.

\myparatight{Impact of watermark length $n$}
Figure~\ref{n-comparison-length} shows the theoretical lower bound of evasion rate of \algns-W-II to double-tail detector (i.e., $2P(\lfloor (\tau-\epsilon)n \rfloor)-1$) as a function of the watermark length $n$, where $\epsilon=0.01$ and $\tau$ varies from 0.6 to 0.9. We observe that the lower bound increases as $n$ increases. This is because the randomly picked target watermark $w_t$ is more likely to have a bitwise accuracy 0.5 compared to the ground-truth watermark as $n$ increases.

\begin{figure}[!t]
\centering
{
\includegraphics[width=0.23\textwidth]{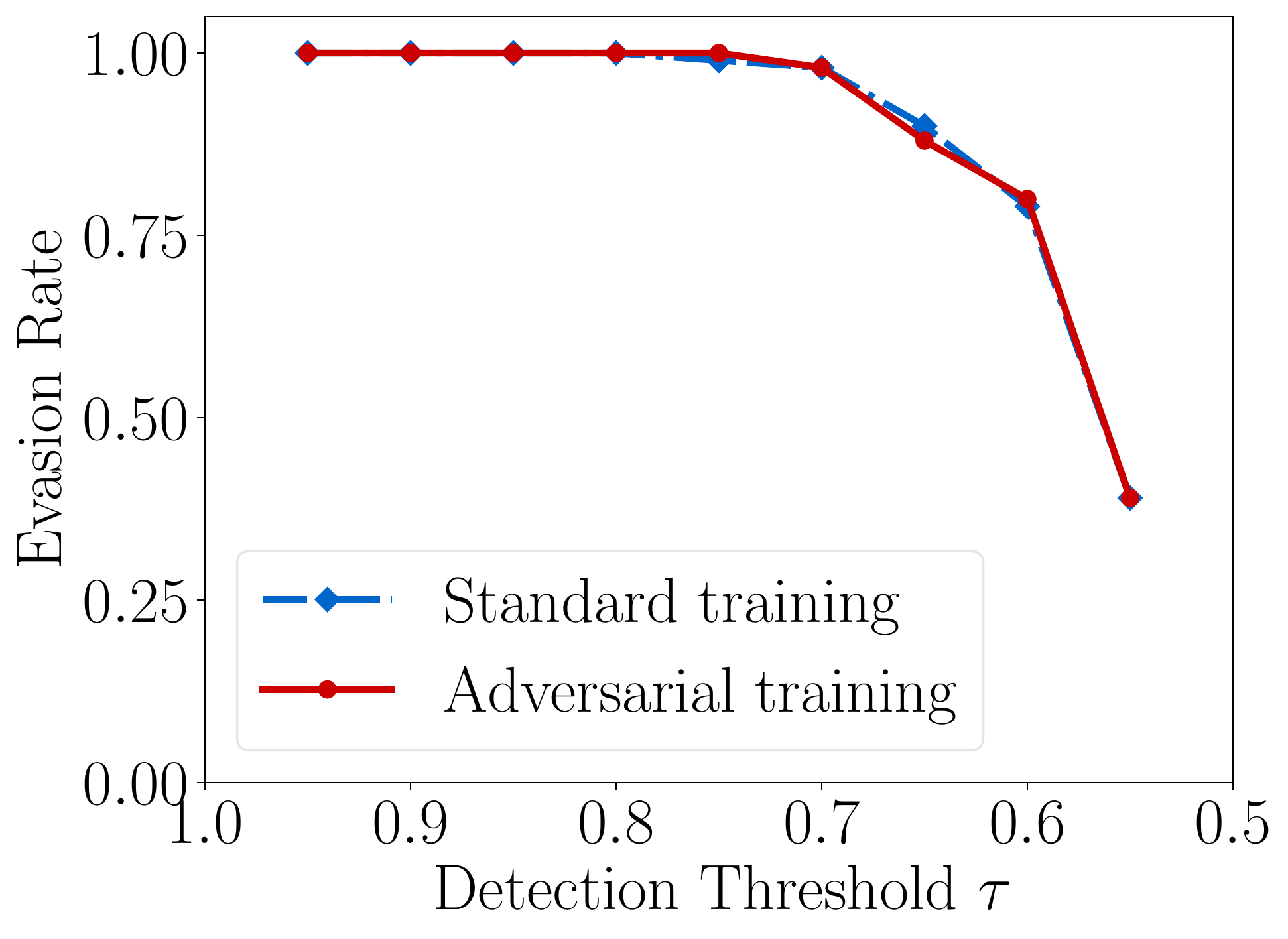}
}%
{
\includegraphics[width=0.23\textwidth]{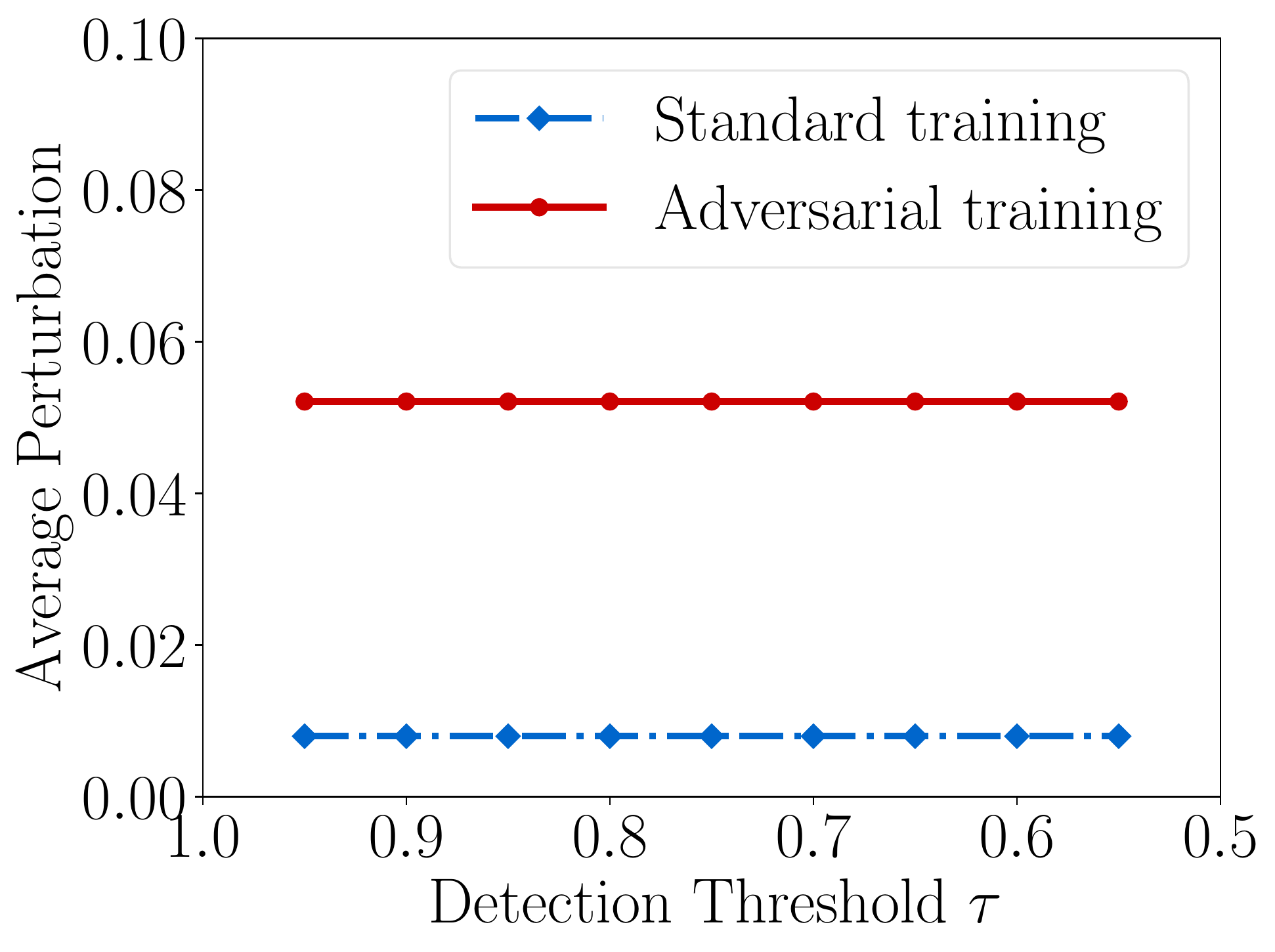}}
\caption{ Standard vs. adversarial training for \algns-W-II.}
\label{training-comparison}
\vspace{-3mm}
\end{figure}

\myparatight{Adversarial training improves robustness but is still insufficient} Figure~\ref{training-comparison} compares standard training and adversarial training with respect to the evasion rates and average perturbations of \algns-W-II.  We have three observations. First, adversarial training improves robustness of the detector. In particular, \algns-W-II achieves the same evasion rates for standard and adversarial training. This is because evasion rates of \algns-W-II do not depend on how the encoder and decoder are trained. However, \algns-W-II needs to add larger perturbations on average when adversarial training is used. Second, adversarial training is still insufficient. Specifically, the perturbations added by \algns-W-II are still small, which maintain visual quality of the images well (Figure~\ref{examples-intro} shows some example images).  Third, \algns-W-II still outperforms existing post-processing methods when adversarial training is used. In particular, Figure~\ref{adversarial-hidden-coco} in Appendix shows that  \algns-W-II still adds much smaller perturbations than existing  methods when they tune parameters to achieve similar evasion rates with \algns-W-II.

\begin{figure}[!t]
\centering
{\includegraphics[width=0.15\textwidth]{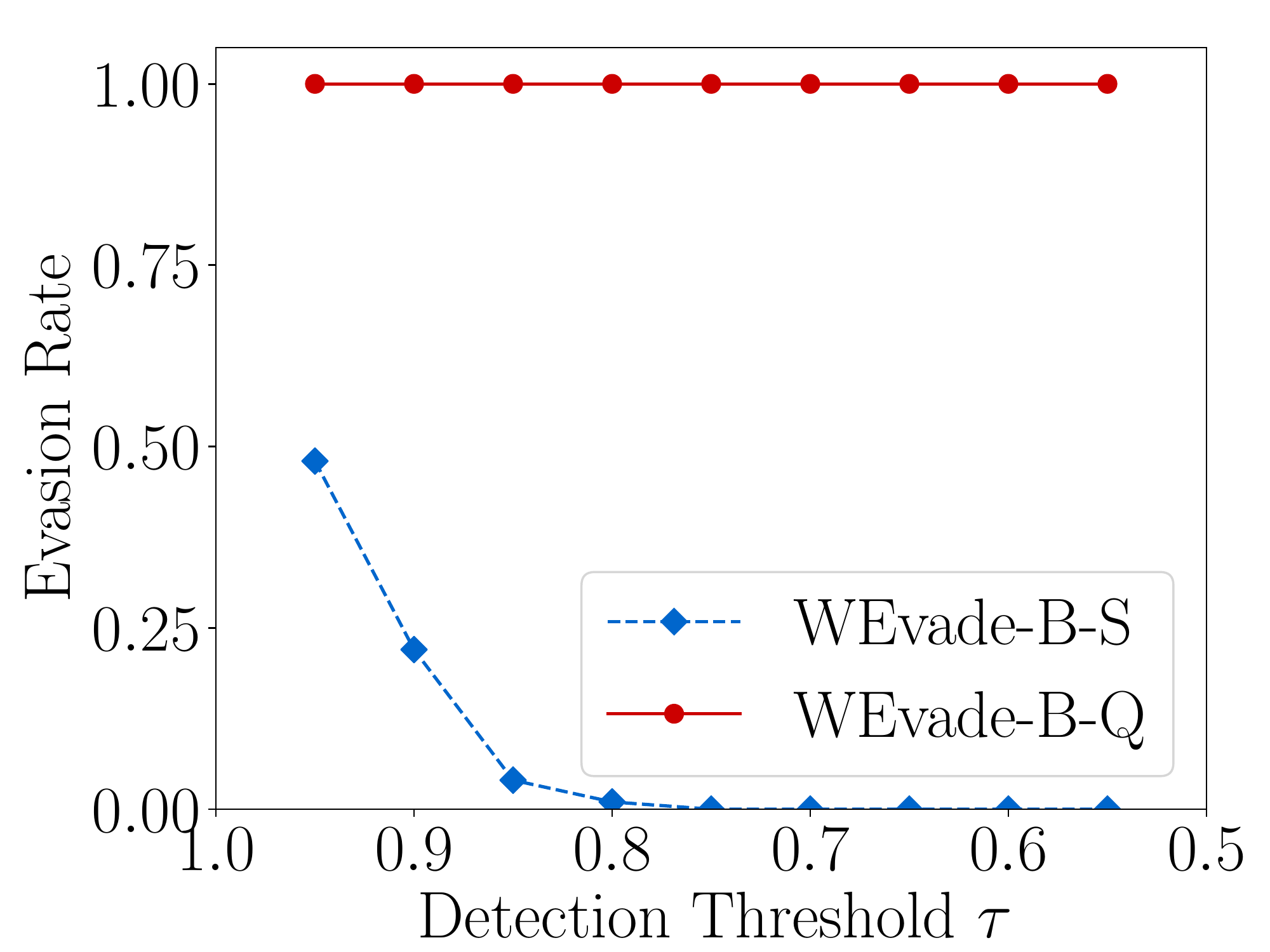}}\hspace{0.1mm}
{\includegraphics[width=0.15\textwidth]{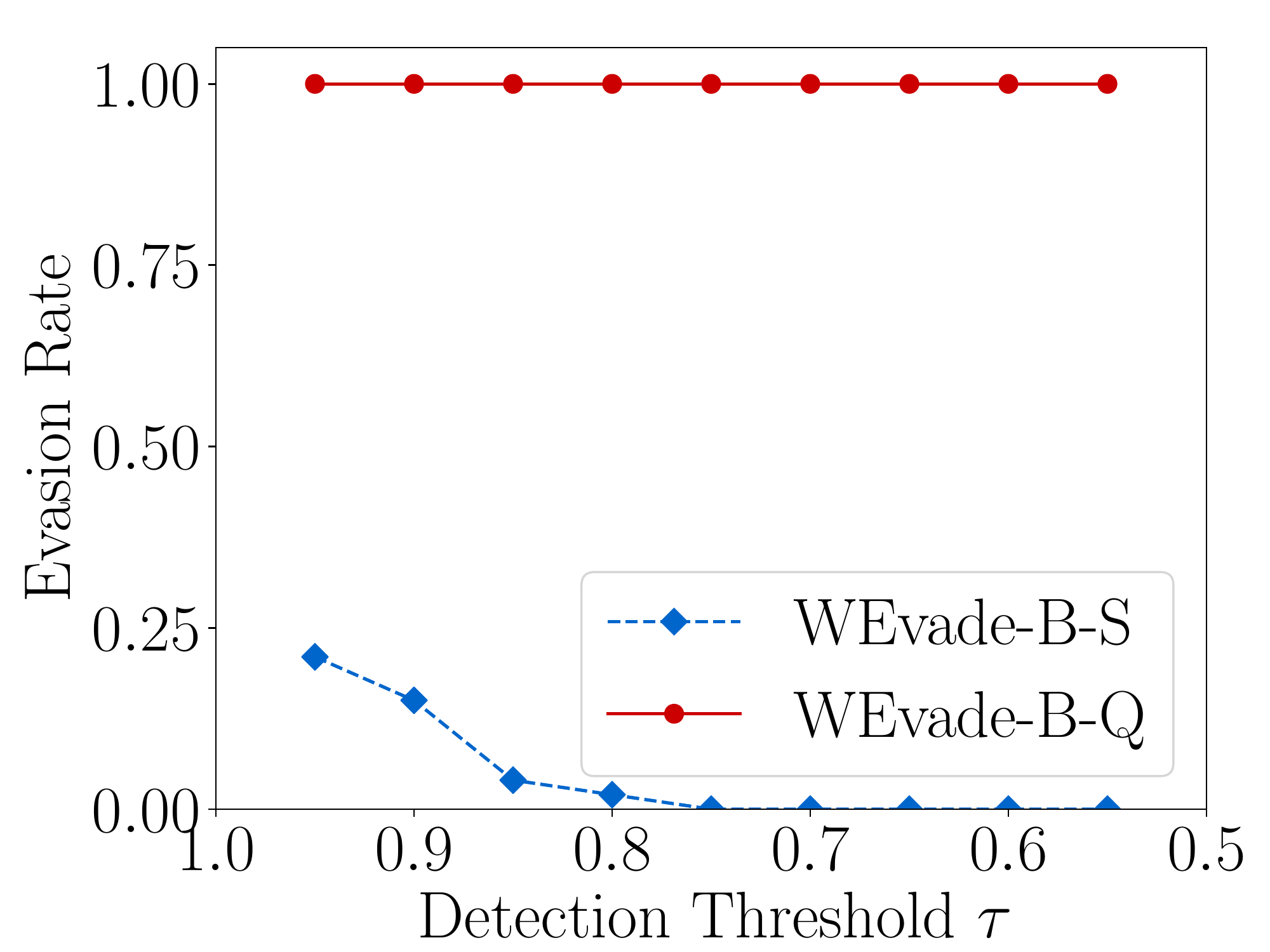}}\hspace{0.1mm}
{\includegraphics[width=0.15\textwidth]{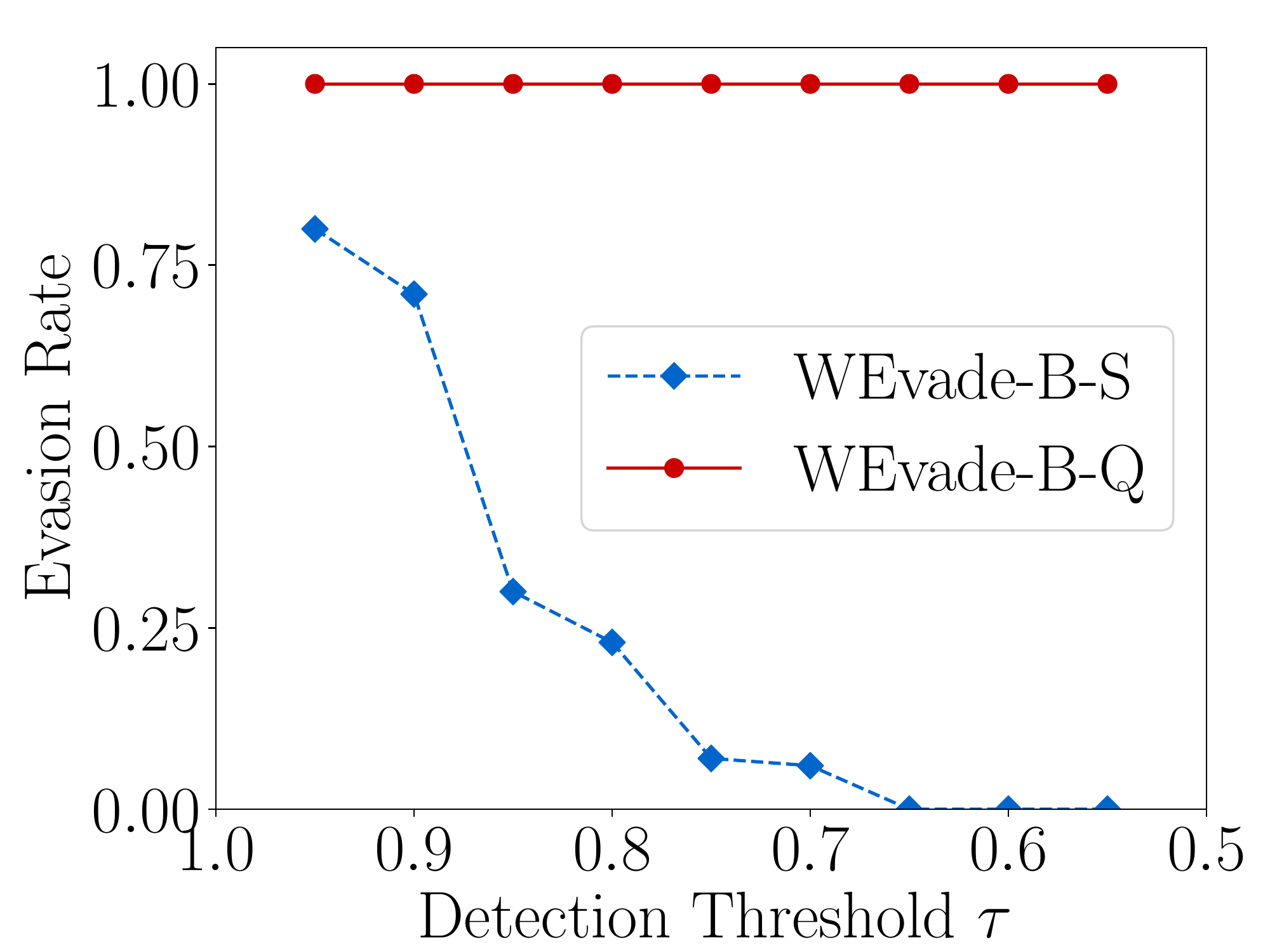}}\hspace{0.1mm}
\vspace{-2mm}

\subfloat[COCO]{\includegraphics[width=0.15\textwidth]{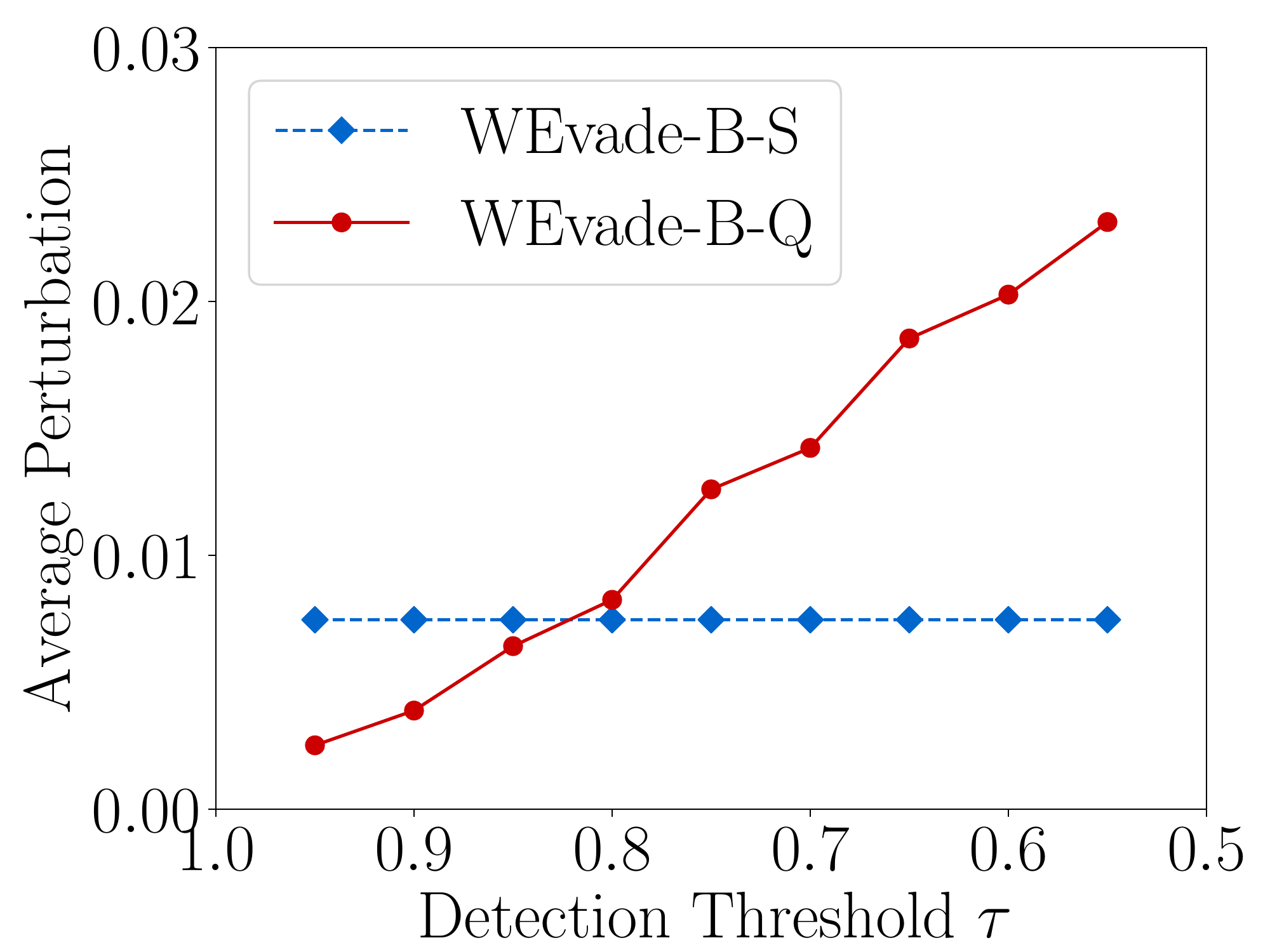}}\hspace{0.1mm}
\subfloat[ImageNet]{\includegraphics[width=0.15\textwidth]{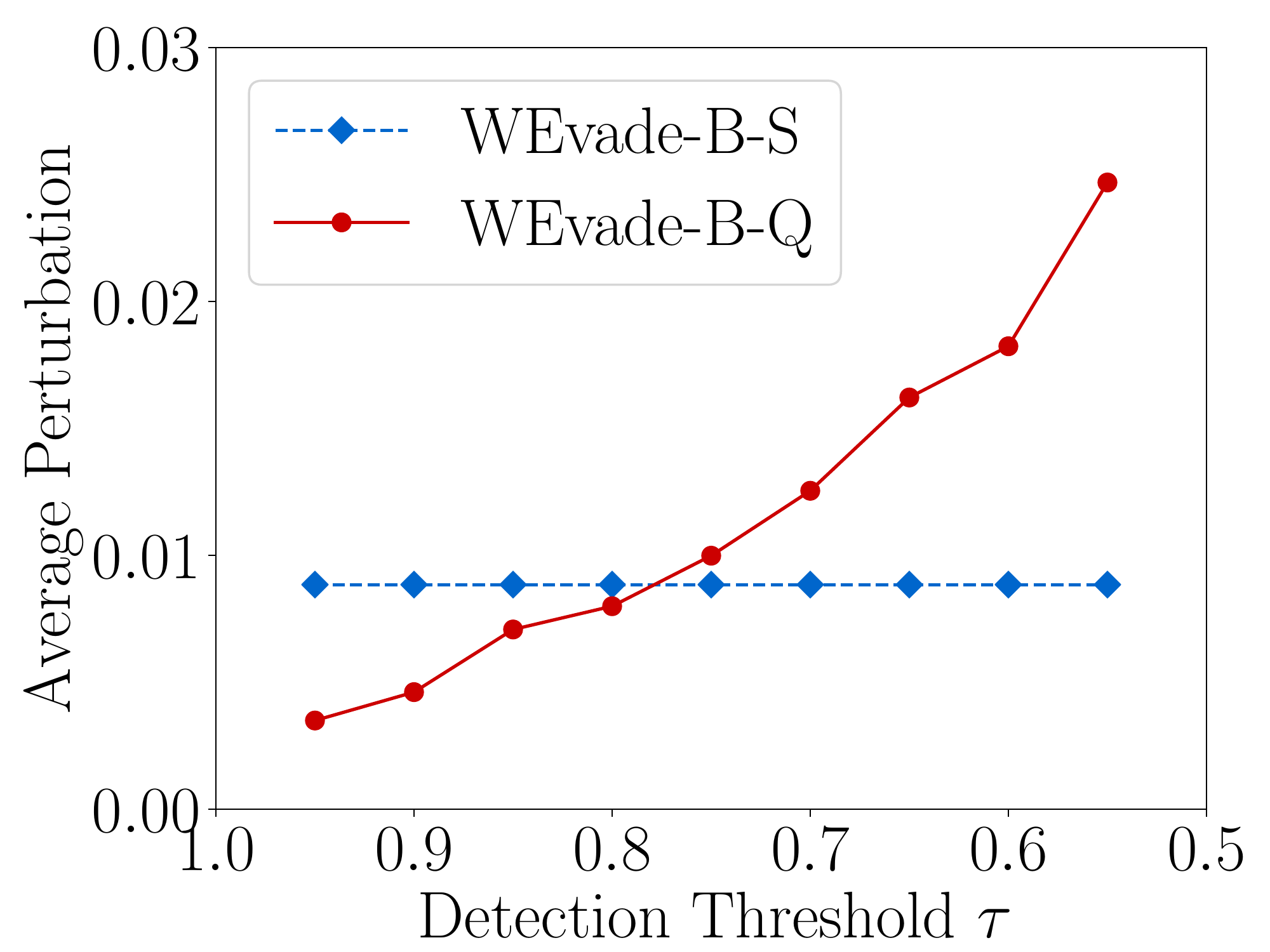}}\hspace{0.1mm}
\subfloat[CC]{\includegraphics[width=0.15\textwidth]{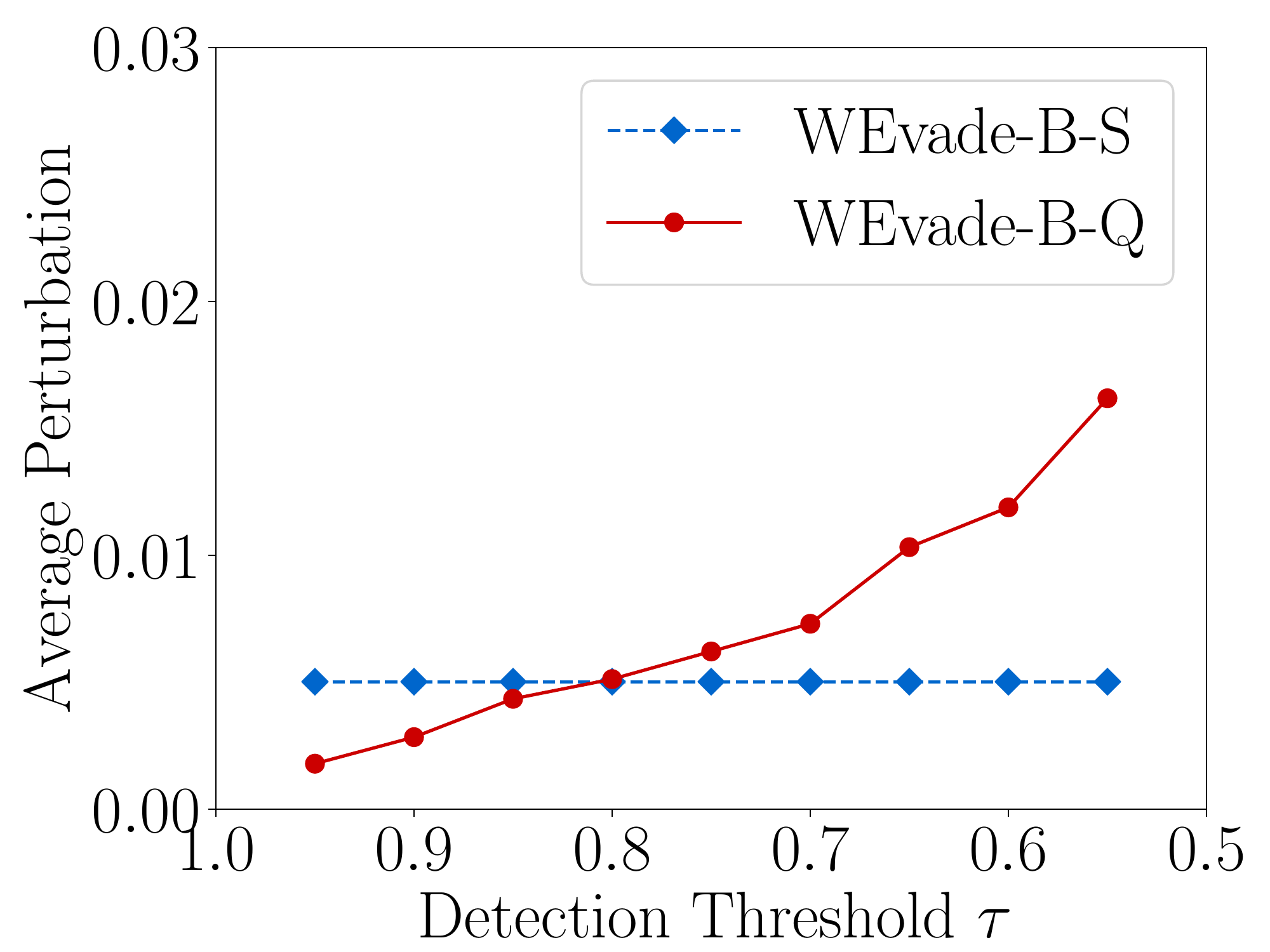}}\hspace{0.1mm}

\caption{Comparing evasion rates (\emph{first row}) and average perturbations (\emph{second row})  of \algns-B-S and \algns-B-Q in the black-box setting.  The watermarking method is HiDDeN and Figure~\ref{black-udh} in Appendix shows results for UDH. } 
\label{black-hidden}
\vspace{-3mm}
\end{figure}

\subsection{Attack Results in the Black-box Setting}
\myparatight{\algns-B-S vs. \algns-B-Q} Figure~\ref{black-hidden} shows the evasion rate and average perturbation of \algns-B-S and \algns-B-Q on the three datasets. Note that, for target detectors with different $\tau$, we apply \algns-B-Q separately to find the (different) perturbations for a watermarked image, while \algns-B-S adds $\tau$-agnostic perturbation to a watermarked image. First, \algns-B-Q always achieves evasion rate of 1 while the evasion rate of \algns-B-S decreases to 0 as the threshold $\tau$ decreases. This is because  the surrogate decoder and the target decoder output dissimilar watermarks for an image. As our Theorem~\ref{theorem-black} shows, when the surrogate decoder and the target decoder are more likely to output dissimilar watermarks, the evasion rate of \algns-B-S decreases.  Second,  \algns-B-Q adds larger perturbation as $\tau$ decreases. This is because  
the decision boundary of a detector with smaller $\tau$ is further away from the watermarked images and \algns-B-Q requires larger perturbations to move them across such boundary. Third, the perturbation of \algns-B-S does not depend on $\tau$ because it uses the white-box attack \algns-W-II to find perturbations.

\begin{figure}[!t]
\centering
\vspace{-3mm}
\subfloat[Impact of query budget max\_q]{\includegraphics[width=0.23\textwidth]{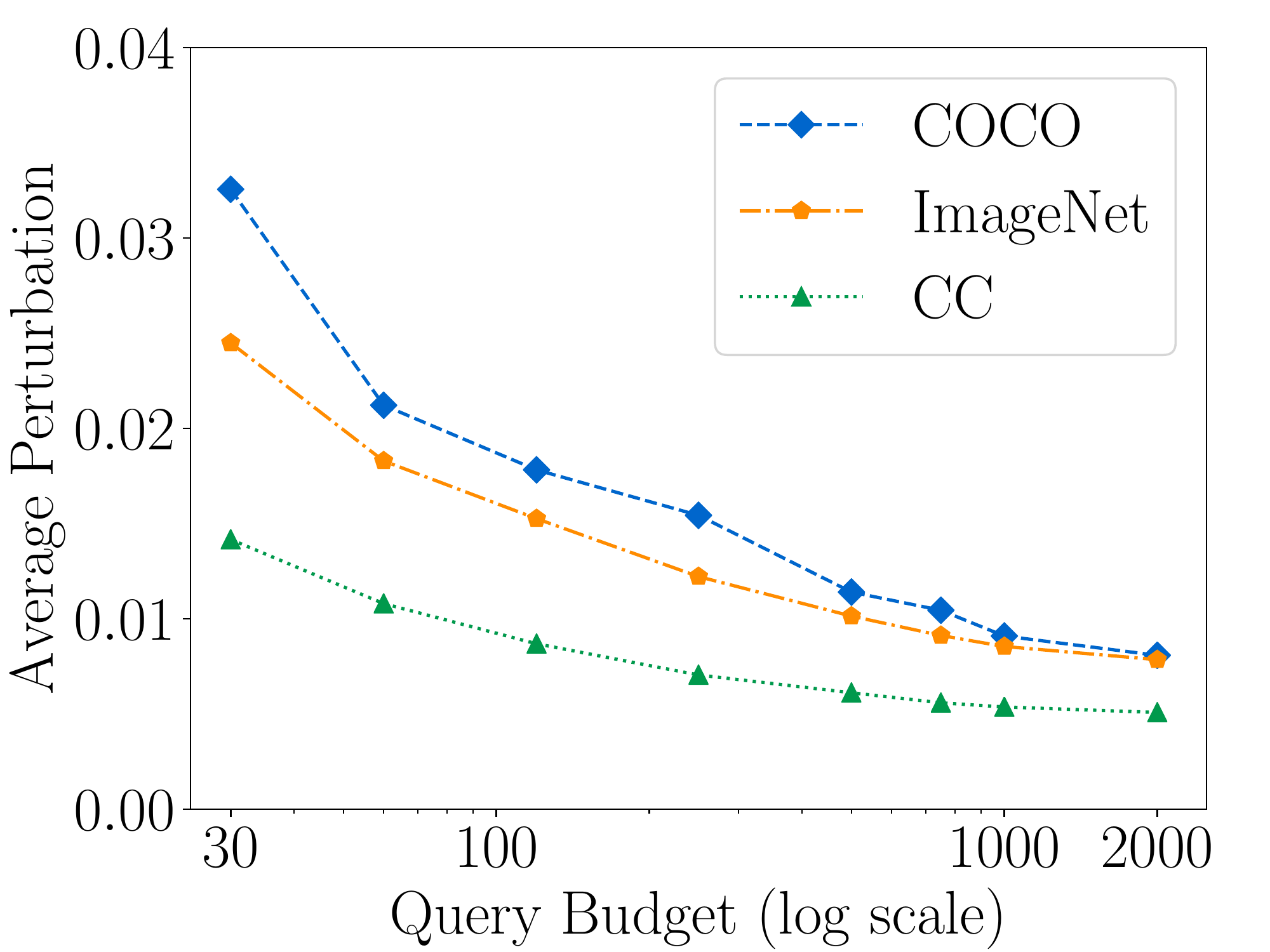} \label{black-query-hidden}}
\subfloat[Single-tail vs. double-tail detector]{\includegraphics[width=0.23\textwidth]{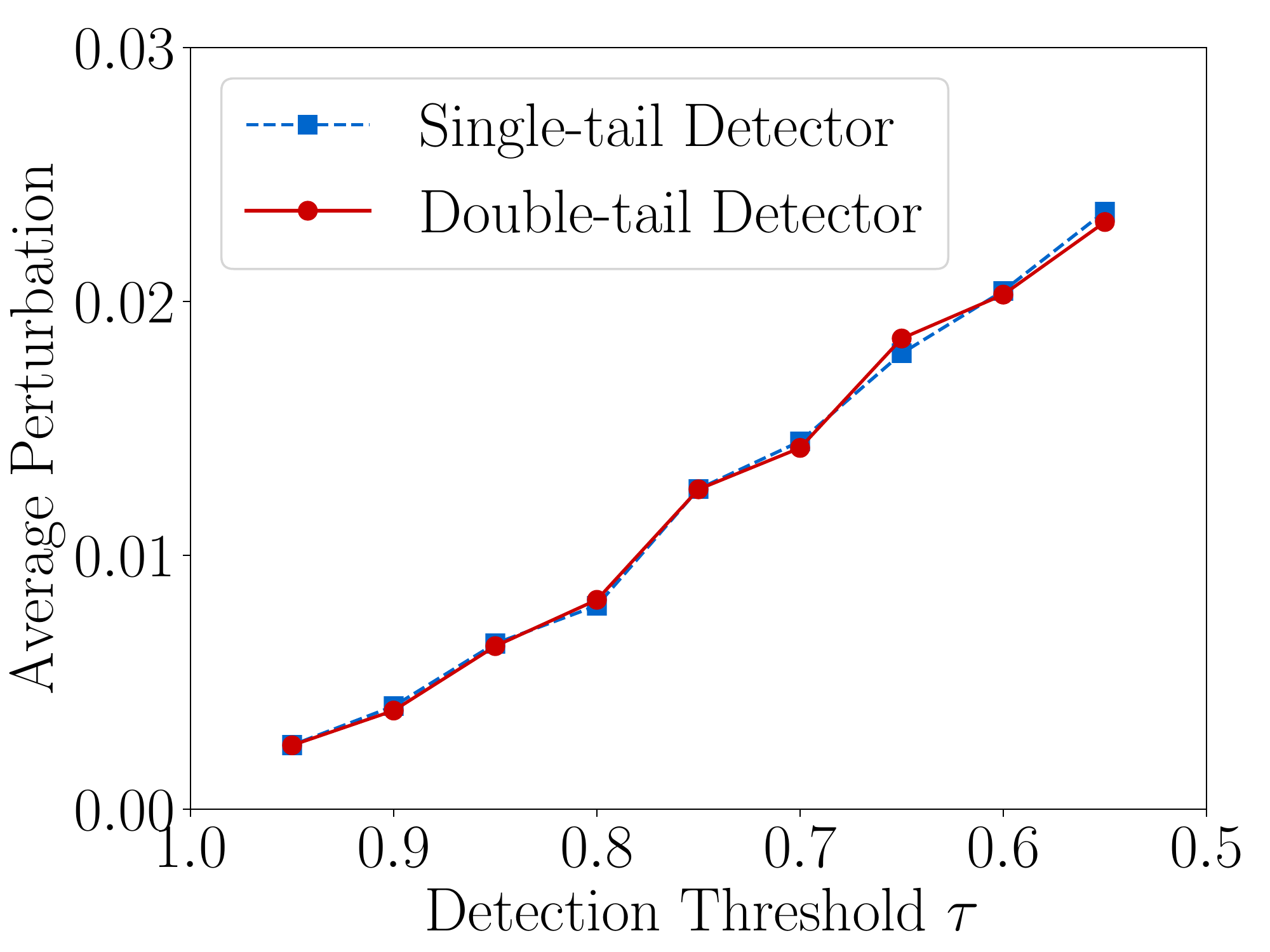}\label{black-stadard-vs-adaptive}}
\caption{(a) Average perturbation of \algns-B-Q as  query budget varies. (b) Average perturbation of \algns-B-Q to evade the single-tail detector or double-tail detector with different threshold $\tau$.}
\vspace{-3mm}
\end{figure}

\myparatight{Impact of the number of queries on \algns-B-Q}  Figure~\ref{black-query-hidden} shows the average perturbation added by \algns-B-Q when the query budget max\_q per watermarked image varies, where the threshold $\tau=\tau^*=0.83$ (corresponding to FPR=$10^{-4}$). Note that the evasion rate is always 1. We observe that the average perturbation added by \algns-B-Q decreases rapidly as the query budget increases. Moreover, when the query budget is small, the average perturbation is already small. For instance, when the query budget is 30 and dataset is COCO, the average perturbation added by WEvade-B-Q is 0.032. On the contrary, existing post-processing methods JPEG, Gaussian noise, Gaussian blur, and Brightness/Contrast respectively add average perturbations 0.211, 0.109, 0.395, and 0.439 to achieve evasion rates close to 1. We acknowledge that WEvade-B-Q requires queries for each watermarked image, so the total number of queries may be large when an attacker aims to evade detection of many watermarked images. However, we note that an attacker can perform a high-profile targeted attack by evading detection of a single or a small number of watermarked images, e.g., a fake image of Elon Musk dating GM CEO Mary Barra~\cite{fake-news}. In such scenarios, an attacker can afford a larger number of queries for the targeted watermarked images.

\myparatight{Single-tail  vs. double-tail detector} Figure~\ref{black-stadard-vs-adaptive} shows the average perturbations added by \algns-B-Q to evade the single-tail detector and double-tail detector. We observe that \algns-B-Q adds similar perturbations to evade the two detectors. The reason is that \algns-B-Q only uses the detector API without considering the internal mechanisms of the detector. Note that the evasion rates of \algns-B-Q are always 1.  

\begin{figure}[!t]
\centering
{\includegraphics[width=0.23\textwidth]{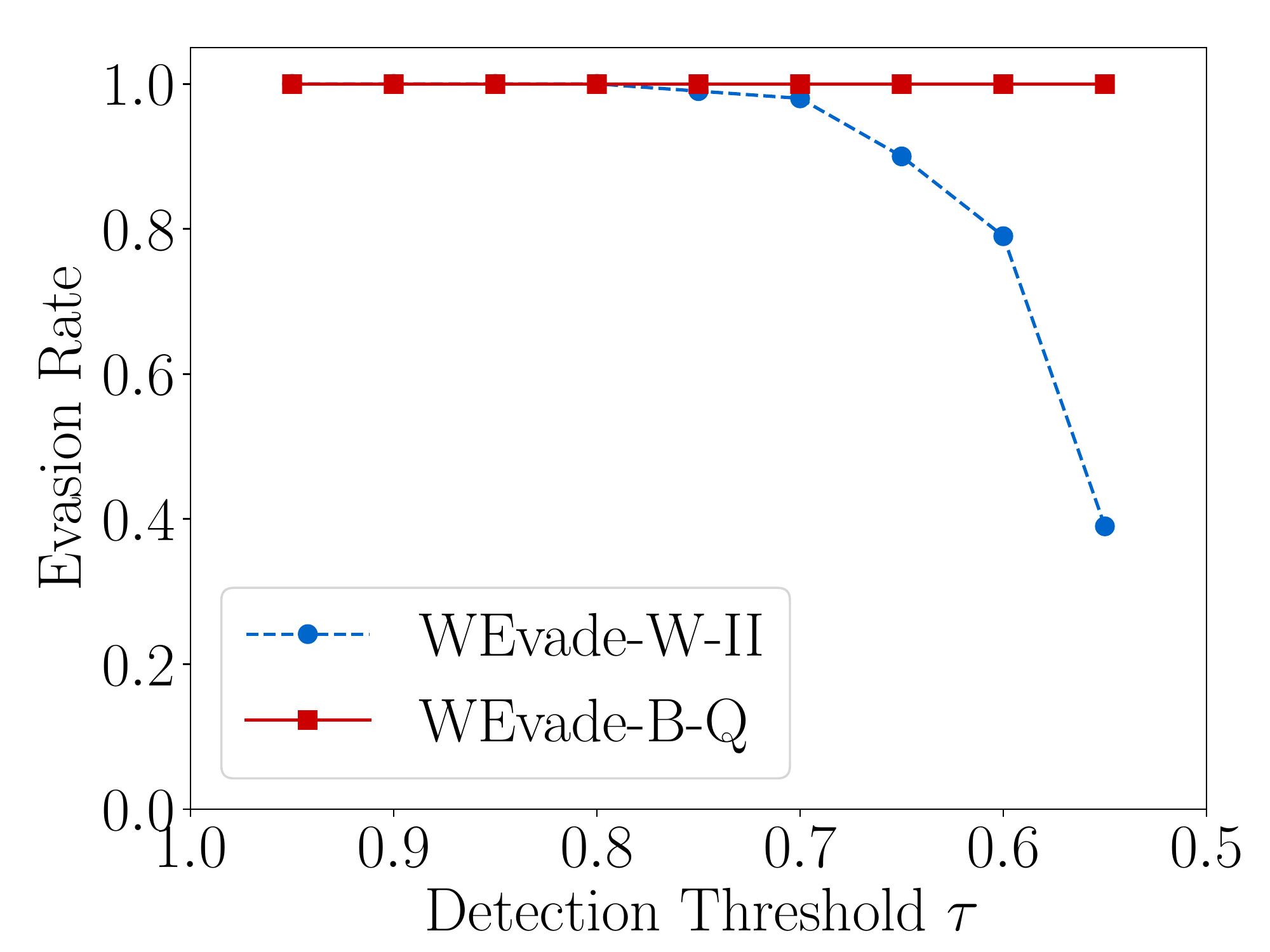}}
{\includegraphics[width=0.23\textwidth]{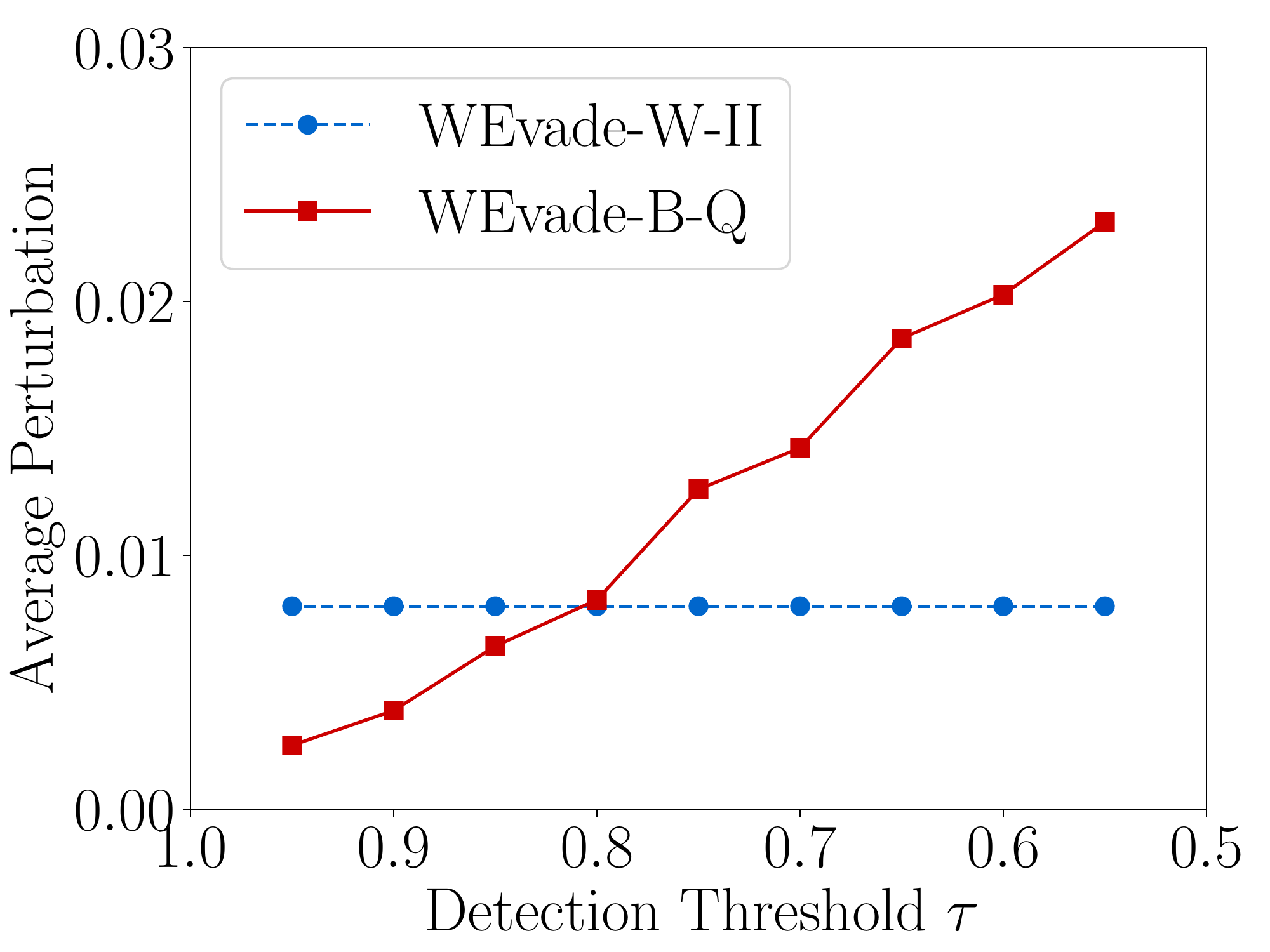}}
\caption{White-box vs. black-box.}
\label{black-white-comparison}
\vspace{-4mm}
\end{figure}

\begin{figure}[!t]
\centering
\subfloat[Standard vs. adversarial training]{\includegraphics[width=0.23\textwidth]{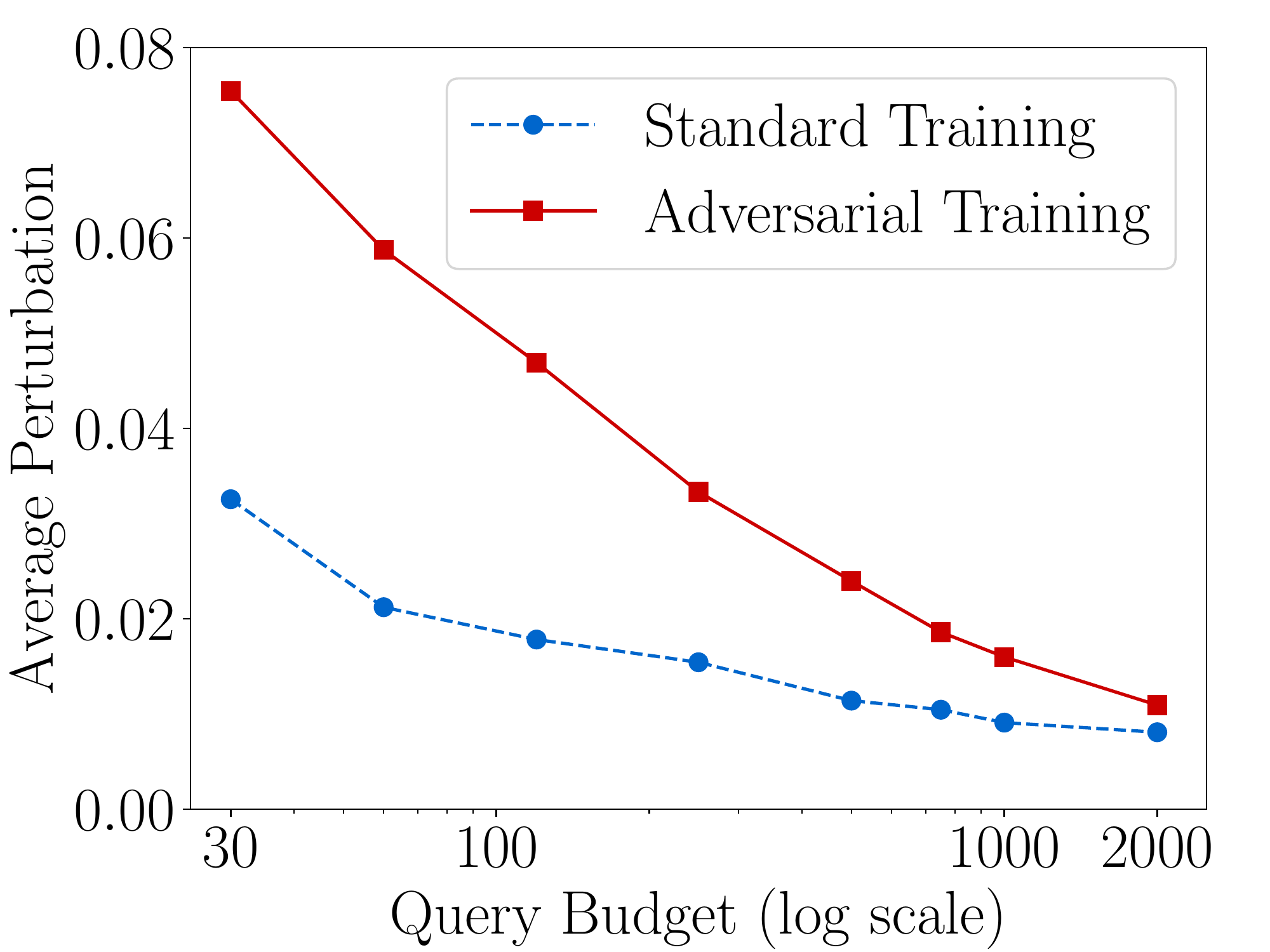} \label{at-bbx}}
\subfloat[\algns-B-Q vs. HopSkipJump]{\includegraphics[width=0.23\textwidth]{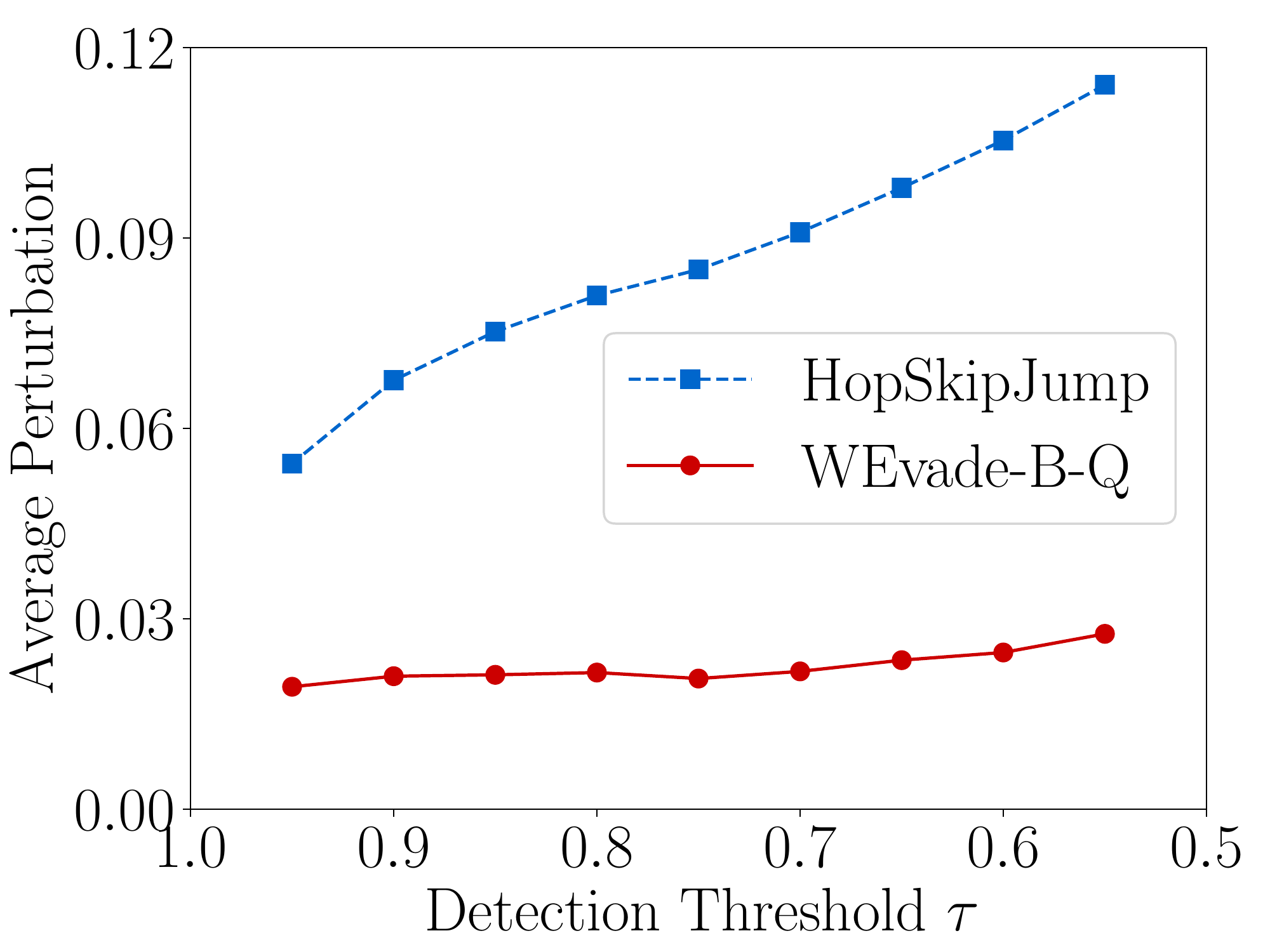} \label{es}}
\caption{(a) Average perturbation of \algns-B-Q as the query budget increases. (b)  \algns-B-Q vs. HopSkipJump.}
\vspace{-3mm}
\end{figure}

\myparatight{Black-box vs. white-box} Figure~\ref{black-white-comparison} compares the evasion rate and average perturbation of \alg in the white-box (i.e., \algns-W-II) and black-box settings (i.e., \algns-B-Q).  First, \algns-B-Q  adds smaller perturbations when $\tau$ is large (e.g., 0.9) but larger perturbations when $\tau$ is small (e.g., 0.6). This is because \algns-B-Q requires larger perturbations to move watermarked images across the decision boundary of a detector with smaller $\tau$ while \algns-W-II is agnostic to $\tau$. However, we stress that the perturbations of both \algns-B-Q and \algns-W-II are small. Second, the perturbations of \algns-B-Q are still much smaller than those of existing post-processing methods (refer to Figure~\ref{white-standard-perturbation}). Third,  \algns-B-Q achieves higher evasion rates than \algns-W-II when $\tau$ is small (e.g., 0.6). This is because \algns-B-Q guarantees evasion rate of 1.

\myparatight{Adversarial training} Figure~\ref{at-bbx} compares the average perturbations added by \algns-B-Q with different query budget max\_q for detectors obtained by standard training and adversarial training, where we set $\tau=\tau^*=0.83$ (corresponding to FPR=$10^{-4}$). Adversarial training improves robustness in the sense that an attacker needs more queries to achieve similar level of perturbation. However, we stress that adversarial training is insufficient because a moderate number of queries can still achieve small perturbations.   

\myparatight{Comparing \algns-B-Q with HopSkipJump} Figure~\ref{es} compares \algns-B-Q with HopSkipJump in terms of average perturbations, where the watermarking method is UDH and dataset is COCO. We observe that  \algns-B-Q adds much smaller perturbations than HopSkipJump. This is because \algns-B-Q uses JPEG compressed version of a watermarked image as initialization and adopts early stopping when the added perturbation increases. Figure~\ref{earlystop} in Appendix further shows that both the initialization and early stopping contribute to \algns-B-Q. We note that \algns-B-Q achieves comparable perturbations with HopSkipJump for HiDDeN. This is because HiDDeN uses 30-bit watermarks and thus the detectors have much simpler decision boundaries.

\subsection{Attacking Stable Diffusion's Detector}
\label{stablediffusion}

\begin{figure}[!t]
\centering
\vspace{-4mm}
\subfloat[JPEG]{\includegraphics[width=0.25\textwidth]{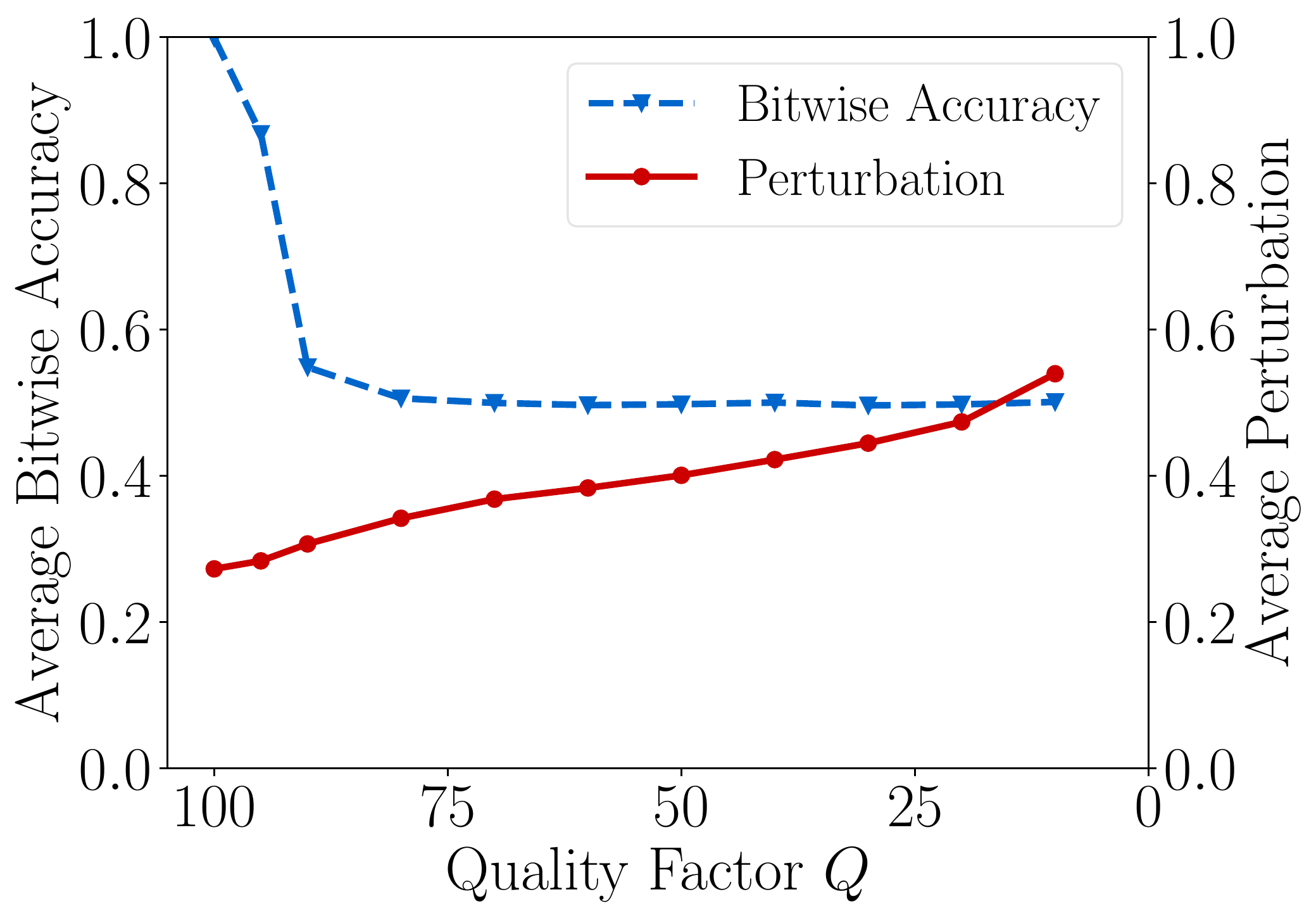}\label{SD-parameter-1}}
\subfloat[Average Perturbation]
{\includegraphics[width=0.228\textwidth]{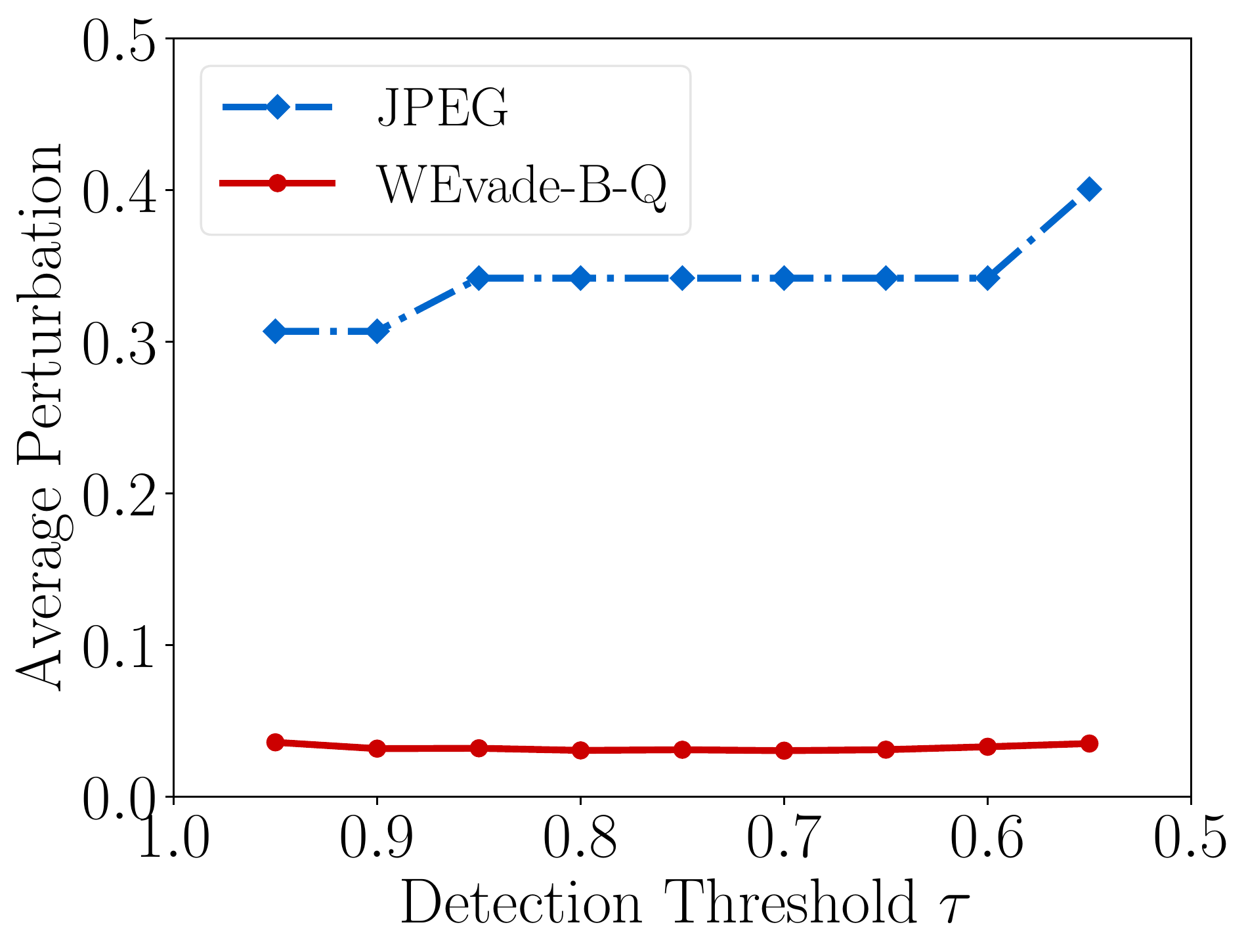}\label{SD-parameter-2}}
\caption{(a) Average bitwise accuracy and average perturbation of the Stable Diffusion watermarked images post-processed by JPEG with different quality factor $Q$. (b) Average perturbation added by JPEG compression and \algns-B-Q  to evade the double-tail detector with  different threshold $\tau$. }
\label{SD-parameter}
\vspace{-5mm}
\end{figure}

\begin{figure}[!t]
\centering
\subfloat[Watermarked]{\includegraphics[width=0.15\textwidth]{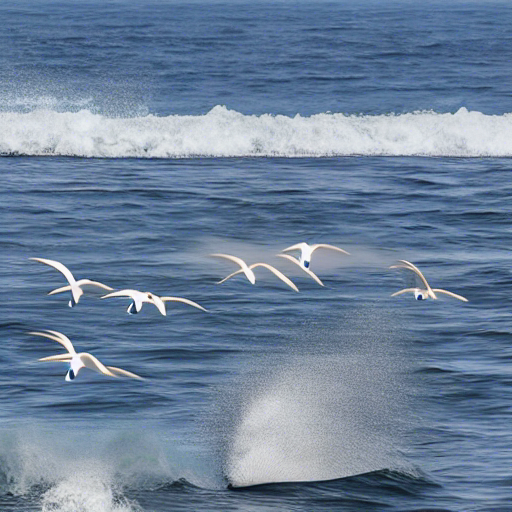}}\hspace{0.3mm}
\subfloat[JPEG]{\includegraphics[width=0.15\textwidth]{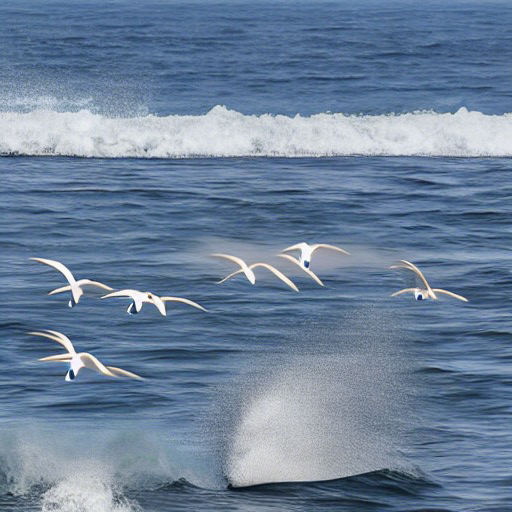}}\hspace{0.3mm}
\subfloat[\algns-B-Q]{\includegraphics[width=0.15\textwidth]{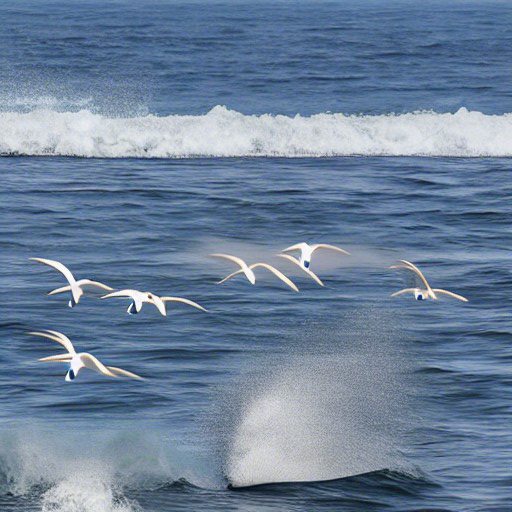}}
\caption{Illustration of a Stable Diffusion watermarked image and the versions post-processed by JPEG and \algns-B-Q to evade watermark-based detection.}
\label{examples-SD}
\vspace{-5mm}
\end{figure}

We generate 100 watermarked images using Stable Diffusion with default setting. We use \emph{sd-v1-1.ckpt} as the checkpoint. Stable Diffusion uses a watermark="StableDiffusionV1", which is represented as 136 bits. The decoder can decode the exact watermark from each of the 100 watermarked images. 
Figure~\ref{SD-parameter-1} shows the average bitwise accuracy and average perturbation of the  watermark images post-processed by JPEG with different quality factor $Q$. When $Q$ is around 80, the bitwise accuracy already reduces to be around 0.5, which means a watermark-based detector cannot distinguish JPEG compressed watermarked images  with original images. Figure~\ref{SD-parameter-2} shows the average perturbation incurred by JPEG compression and \algns-B-Q to evade the double-tail detector. Our \algns-B-Q incurs much smaller perturbations than JPEG compression. Figure~\ref{examples-SD} shows an example Stable Diffusion watermarked image, its JPEG compressed version, and the version post-processed by \algns-B-Q to evade the double-tail detector with $\tau=0.66$ (corresponding to FPR=$10^{-4}$). As we can see, both JPEG compression and \algns-B-Q can evade the Stable Diffusion's detector, which is based on a non-learning-based watermarking method, without sacrificing the image quality.

\section{Discussion and Limitations}
\begin{figure}[!t]
\centering
\vspace{-2mm}
{\includegraphics[width=0.23\textwidth]{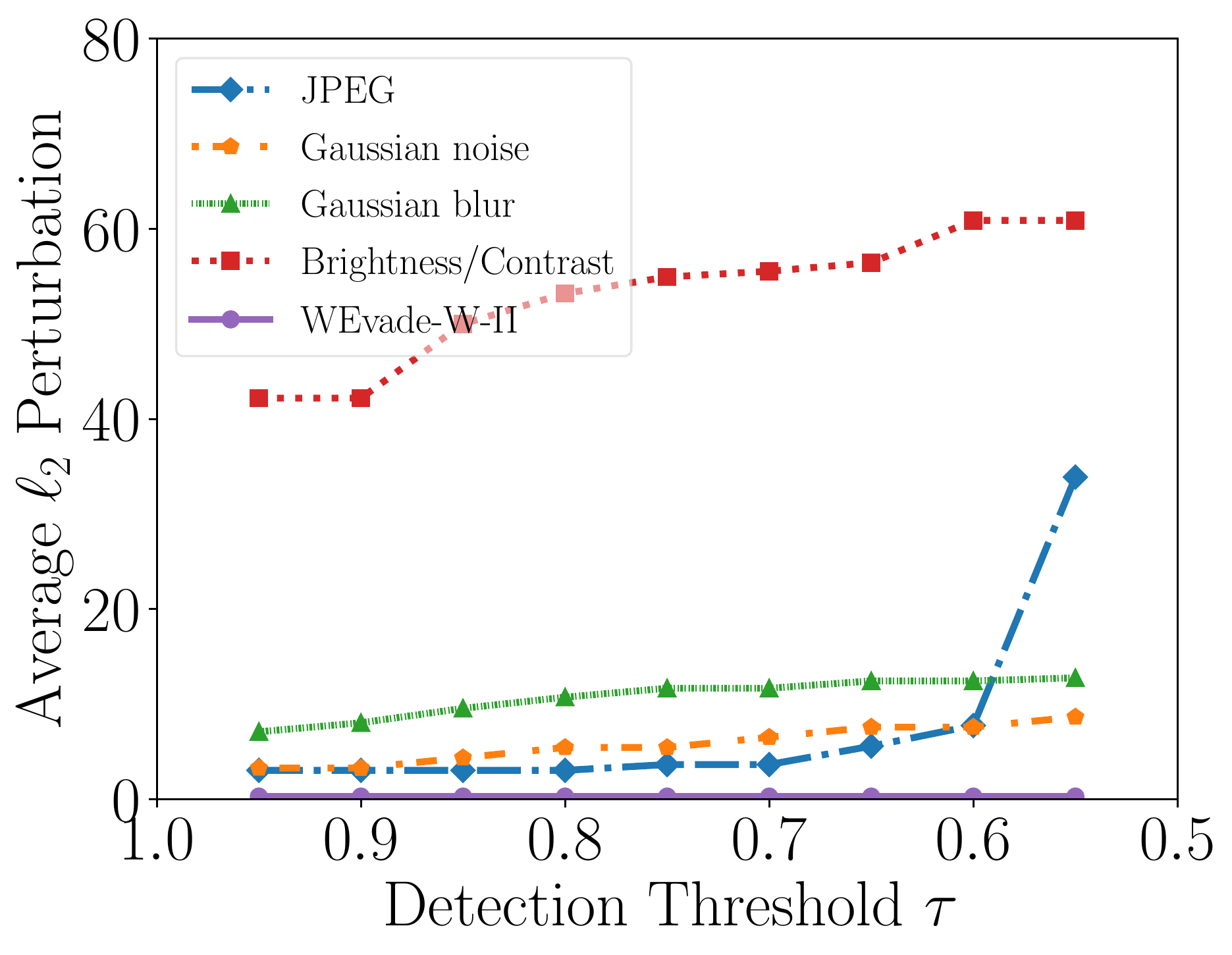}}
{\includegraphics[width=0.23\textwidth]{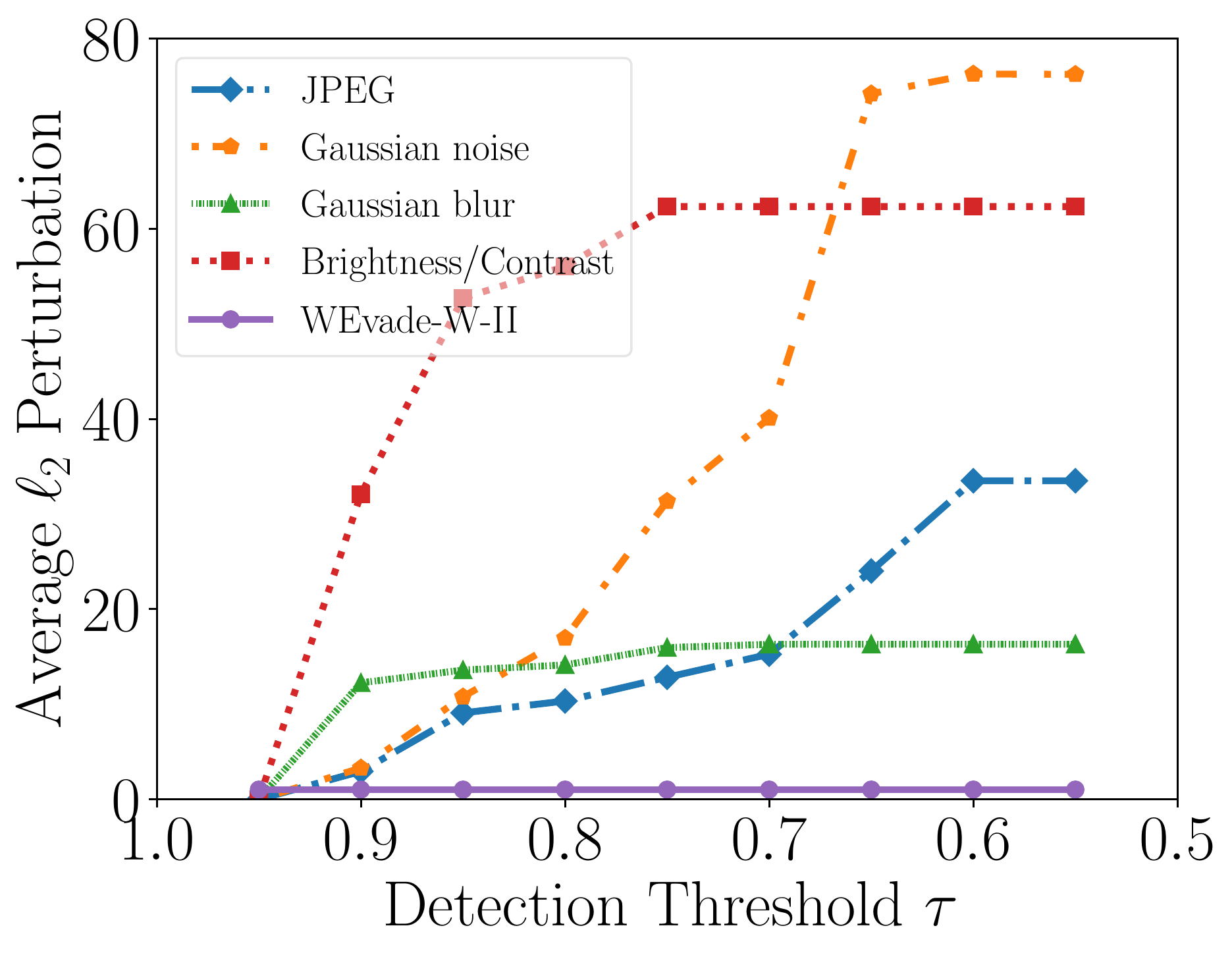}}
\vspace{-3mm}

\subfloat[Standard training]{\includegraphics[width=0.23\textwidth]{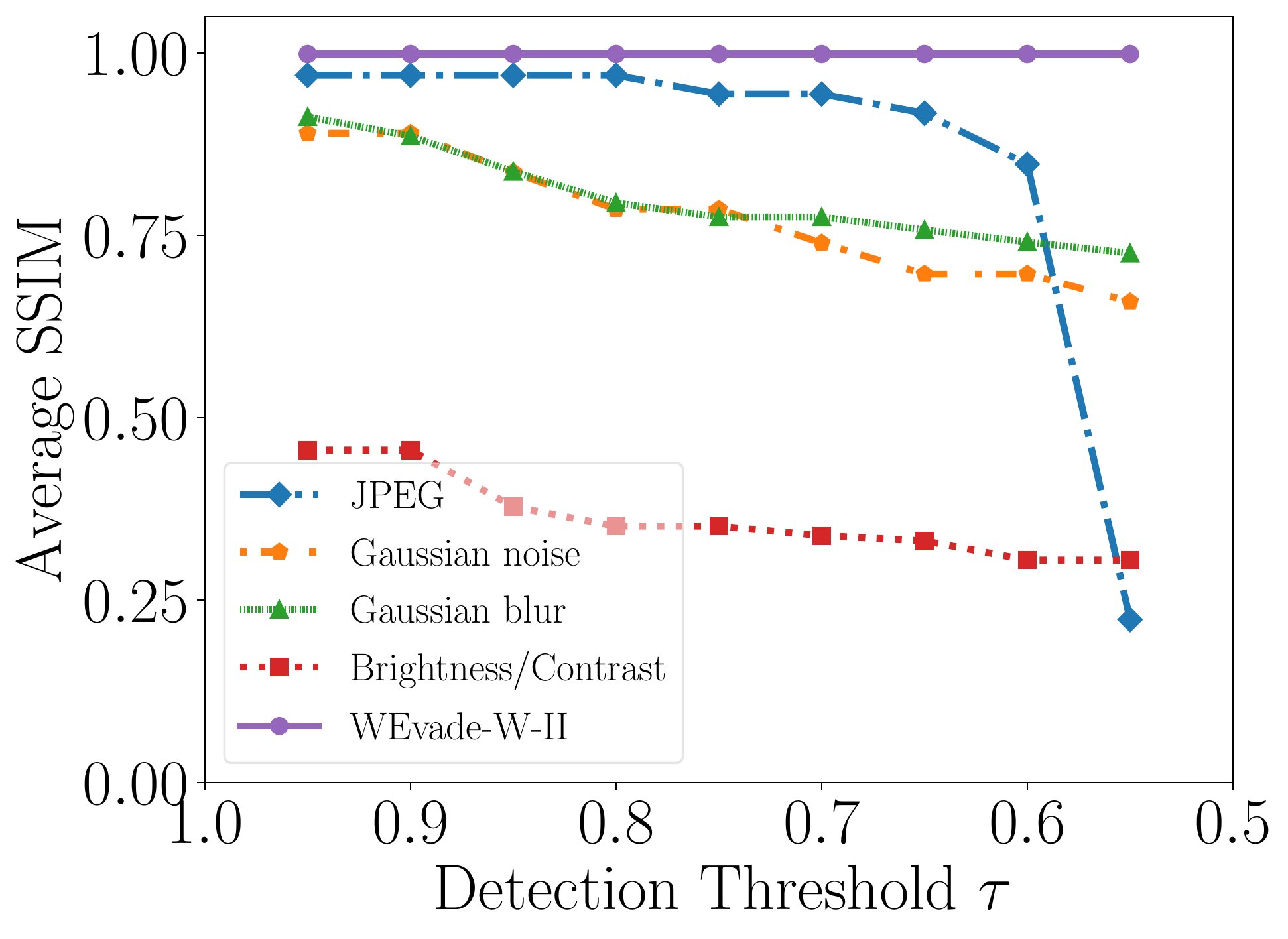}}
\subfloat[Adversarial training]{\includegraphics[width=0.23\textwidth]{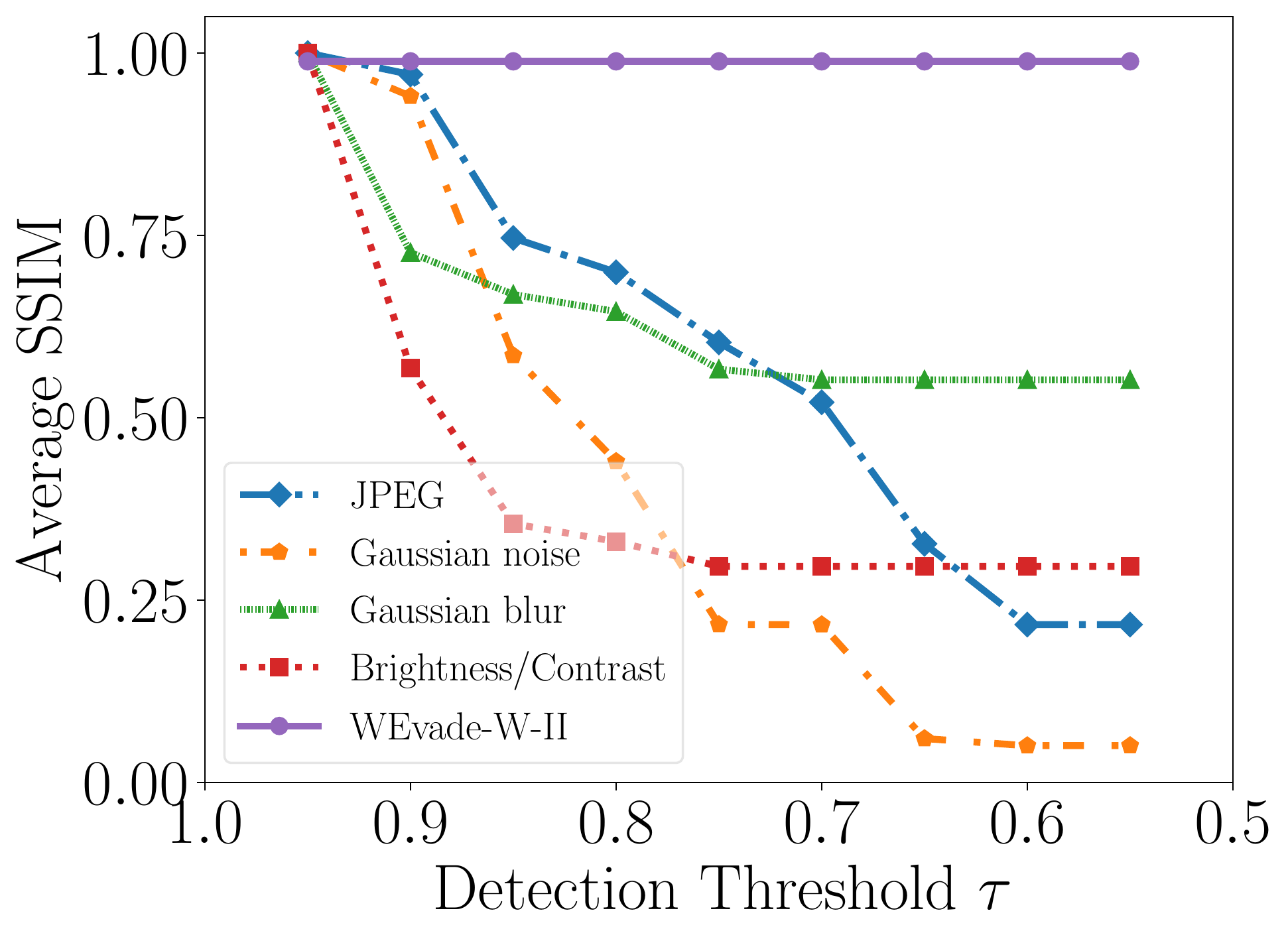}}
\caption{Average perturbation, measured by $\ell_2$-norm (first row) or SSIM (second row),  added by each post-processing method to evade the double-tail detector with different threshold $\tau$ in the white-box setting. We set the parameters of existing post-processing methods such that they achieve the same evasion rate as our \algns-W-II. The watermarking method is HiDDeN and dataset is COCO.}
\label{l2whitebox}
\vspace{-3mm}
\end{figure}

\vspace{-1mm}
\myparatight{Other metrics to quantify perturbation} Attacker's goal is to add small perturbation to evade detection while preserving visual quality of the image.  We use $\ell_\infty$-norm of the perturbation to quantify whether it preserves visual quality, which is a popular choice in adversarial examples~\cite{goodfellow2014explaining,carlini2017towards}. In particular, when  $\ell_\infty$-norm of the perturbation is small enough, the visual quality is preserved. We can also use other $\ell_p$-norms, e.g., $\ell_2$-norm, or SSIM~\cite{wang2004image} between a watermarked image and its post-processed version, to quantify the perturbation.  For instance, Figure~\ref{l2whitebox} compares the perturbations added by different post-processing methods in the white-box setting when using $\ell_2$-norm or SSIM to quantify the perturbation, while Figure~\ref{l2blackbox} in Appendix shows the results in the black-box setting, where \alg uses the default parameter settings described in Section~\ref{sec:setup}. Our results show that \alg still adds much smaller perturbations than existing methods when  $\ell_2$-norm or SSIM is used to quantify the perturbation. 
We acknowledge that $\ell_p$-norms and SSIM are approximate measures of perturbations' impact on visual quality. Previous works~\cite{sharif2018suitability} on adversarial examples showed that small $\ell_p$-norms of perturbations may not be sufficient nor necessary conditions to maintain visual quality. It is an interesting future work to explore other metrics to quantify the impact of perturbation on visual quality specifically in the generative AI domain.

\myparatight{Provably robust watermarking methods} The fundamental reason that watermarking-based detectors can be evaded by our attack is that existing watermarking methods do not have provable robustness guarantees. Specifically, an attacker can add a small perturbation to a watermarked image such that the decoder outputs a different watermark for the post-processed watermarked image. To defend against such attacks, one interesting future work is to build watermarking methods with provable robustness guarantees. In particular, a provably robust watermarking method is guaranteed to output similar watermarks for a watermarked image and its post-processed version once the added perturbation is bounded, e.g., its $\ell_\infty$-norm or $\ell_2$-norm is smaller than a threshold. For instance,  if the watermarks decoded from a watermarked image and its post-processed version are guaranteed to have bitwise accuracy of 0.85 once the $\ell_\infty$-norm of the perturbation is bounded by 0.03, then a detector with threshold $\tau=0.8$ is guaranteed to detect the post-processed version once the $\ell_\infty$-norm of the perturbation is bounded by 0.03. If the perturbation bound is large enough to be human-perceptible, an attacker has to sacrifice visual quality of the watermarked image in order to evade watermarking-based detector, leading to a dilemma for the attacker, i.e., either being detected or  perturbed images have low quality.

\myparatight{Text watermarking} In this work, we focus on AI-generated images, but our ideas of adversarial post-processing can also be generalized to text watermarking. Existing studies~\cite{krishna2023paraphrasing} showed that non-learning-based text watermarking~\cite{kirchenbauer2023watermark} is not robust to common post-processing such as paraphrasing. These common post-processing are analogous to those--such as JPEG compression, Gaussian blur, and Brightness/Contrast-- in image watermarking. Like learning-based image watermarking, we suspect learning-based text watermarking~\cite{abdelnabi2021adversarial} could be more robust to common post-processing via leveraging adversarial training, which is an interesting future work to explore. It is also an interesting future work to extend our adversarial post-processing to text watermarking. 
\section{Conclusion and Future Work}

We find that watermark-based detection of AI-generated content is vulnerable to strategic, adversarial post-processing. An attacker can add a small, human-imperceptible perturbation to an AI-generated, watermarked image to evade detection. Our results indicate that watermark-based AI-generated content detection is not as robust as previously thought.  We also find that simply extending standard adversarial examples to watermarking is insufficient since they do not take the unique characteristics of watermarking into consideration. An interesting future work is to explore watermark-based detectors with provable robustness guarantees.

\section*{Acknowledgements}

We thank the anonymous reviewers for their constructive comments. This work was supported by NSF grant No. 1937787, 1937786, 2112562, and 2125977, as well as ARO grant No. W911NF2110182.

\bibliographystyle{ACM-Reference-Format}
\bibliography{refs}


\begin{thebibliography}{47}


\ifx \showCODEN    \undefined \def \showCODEN     #1{\unskip}     \fi
\ifx \showDOI      \undefined \def \showDOI       #1{#1}\fi
\ifx \showISBNx    \undefined \def \showISBNx     #1{\unskip}     \fi
\ifx \showISBNxiii \undefined \def \showISBNxiii  #1{\unskip}     \fi
\ifx \showISSN     \undefined \def \showISSN      #1{\unskip}     \fi
\ifx \showLCCN     \undefined \def \showLCCN      #1{\unskip}     \fi
\ifx \shownote     \undefined \def \shownote      #1{#1}          \fi
\ifx \showarticletitle \undefined \def \showarticletitle #1{#1}   \fi
\ifx \showURL      \undefined \def \showURL       {\relax}        \fi
\providecommand\bibfield[2]{#2}
\providecommand\bibinfo[2]{#2}
\providecommand\natexlab[1]{#1}
\providecommand\showeprint[2][]{arXiv:#2}

\bibitem[Abdelnabi and Fritz(2021)]%
        {abdelnabi2021adversarial}
\bibfield{author}{\bibinfo{person}{Sahar Abdelnabi} {and}
  \bibinfo{person}{Mario Fritz}.} \bibinfo{year}{2021}\natexlab{}.
\newblock \showarticletitle{Adversarial watermarking transformer: Towards
  tracing text provenance with data hiding}. In \bibinfo{booktitle}{\emph{IEEE
  Symposium on Security and Privacy}}.
\newblock


\bibitem[{ARTnews}(2023)]%
        {us-copyright-office}
\bibfield{author}{\bibinfo{person}{{ARTnews}}.}
  \bibinfo{year}{2023}\natexlab{}.
\newblock \bibinfo{title}{{US Copyright Office: AI Generated Works Are Not
  Eligible for Copyright.}}
\newblock
  \bibinfo{howpublished}{\url{https://www.artnews.com/art-news/news/ai-generator-art-text-us-copyright-policy-1234661683}}.
\newblock


\bibitem[Bi et~al\mbox{.}(2007)]%
        {bi2007robust}
\bibfield{author}{\bibinfo{person}{Ning Bi}, \bibinfo{person}{Qiyu Sun},
  \bibinfo{person}{Daren Huang}, \bibinfo{person}{Zhihua Yang}, {and}
  \bibinfo{person}{Jiwu Huang}.} \bibinfo{year}{2007}\natexlab{}.
\newblock \showarticletitle{Robust image watermarking based on multiband
  wavelets and empirical mode decomposition}.
\newblock \bibinfo{journal}{\emph{IEEE Transactions on Image Processing}}
  (\bibinfo{year}{2007}).
\newblock


\bibitem[Cao and Gong(2022)]%
        {cao2022understanding}
\bibfield{author}{\bibinfo{person}{Xiaoyu Cao} {and}
  \bibinfo{person}{Neil~Zhenqiang Gong}.} \bibinfo{year}{2022}\natexlab{}.
\newblock \showarticletitle{Understanding the security of deepfake detection}.
  In \bibinfo{booktitle}{\emph{EAI International Conference on Digital
  Forensics and Cyber Crime}}.
\newblock


\bibitem[Carlini and Wagner(2017)]%
        {carlini2017towards}
\bibfield{author}{\bibinfo{person}{Nicholas Carlini} {and}
  \bibinfo{person}{David Wagner}.} \bibinfo{year}{2017}\natexlab{}.
\newblock \showarticletitle{Towards evaluating the robustness of neural
  networks}. In \bibinfo{booktitle}{\emph{IEEE Symposium on Security and
  Privacy}}.
\newblock


\bibitem[Chen et~al\mbox{.}(2020)]%
        {chen2020hopskipjumpattack}
\bibfield{author}{\bibinfo{person}{Jianbo Chen}, \bibinfo{person}{Michael~I
  Jordan}, {and} \bibinfo{person}{Martin~J Wainwright}.}
  \bibinfo{year}{2020}\natexlab{}.
\newblock \showarticletitle{Hopskipjumpattack: A query-efficient decision-based
  attack}. In \bibinfo{booktitle}{\emph{IEEE Symposium on Security and
  Privacy}}.
\newblock


\bibitem[Corvi et~al\mbox{.}(2022)]%
        {corvi2022detection}
\bibfield{author}{\bibinfo{person}{Riccardo Corvi}, \bibinfo{person}{Davide
  Cozzolino}, \bibinfo{person}{Giada Zingarini}, \bibinfo{person}{Giovanni
  Poggi}, \bibinfo{person}{Koki Nagano}, {and} \bibinfo{person}{Luisa
  Verdoliva}.} \bibinfo{year}{2022}\natexlab{}.
\newblock \showarticletitle{On the detection of synthetic images generated by
  diffusion models}.
\newblock \bibinfo{journal}{\emph{arXiv preprint arXiv:2211.00680}}
  (\bibinfo{year}{2022}).
\newblock


\bibitem[Deng et~al\mbox{.}(2009)]%
        {deng2009imagenet}
\bibfield{author}{\bibinfo{person}{Jia Deng}, \bibinfo{person}{Wei Dong},
  \bibinfo{person}{Richard Socher}, \bibinfo{person}{Li-Jia Li},
  \bibinfo{person}{Kai Li}, {and} \bibinfo{person}{Li Fei-Fei}.}
  \bibinfo{year}{2009}\natexlab{}.
\newblock \showarticletitle{Imagenet: A large-scale hierarchical image
  database}. In \bibinfo{booktitle}{\emph{IEEE/CVF Conference on Computer
  Vision and Pattern Recognition}}.
\newblock


\bibitem[Fernandez et~al\mbox{.}(2023)]%
        {fernandez2023stable}
\bibfield{author}{\bibinfo{person}{Pierre Fernandez},
  \bibinfo{person}{Guillaume Couairon}, \bibinfo{person}{Herv{\'e} J{\'e}gou},
  \bibinfo{person}{Matthijs Douze}, {and} \bibinfo{person}{Teddy Furon}.}
  \bibinfo{year}{2023}\natexlab{}.
\newblock \showarticletitle{The Stable Signature: Rooting Watermarks in Latent
  Diffusion Models}.
\newblock \bibinfo{journal}{\emph{arXiv preprint arXiv:2303.15435}}
  (\bibinfo{year}{2023}).
\newblock


\bibitem[Frank et~al\mbox{.}(2020)]%
        {frank2020leveraging}
\bibfield{author}{\bibinfo{person}{Joel Frank}, \bibinfo{person}{Thorsten
  Eisenhofer}, \bibinfo{person}{Lea Sch{\"o}nherr}, \bibinfo{person}{Asja
  Fischer}, \bibinfo{person}{Dorothea Kolossa}, {and} \bibinfo{person}{Thorsten
  Holz}.} \bibinfo{year}{2020}\natexlab{}.
\newblock \showarticletitle{Leveraging frequency analysis for deep fake image
  recognition}. In \bibinfo{booktitle}{\emph{International Conference on
  Machine Learning}}.
\newblock


\bibitem[Goodfellow et~al\mbox{.}(2020)]%
        {goodfellow2020generative}
\bibfield{author}{\bibinfo{person}{Ian Goodfellow}, \bibinfo{person}{Jean
  Pouget-Abadie}, \bibinfo{person}{Mehdi Mirza}, \bibinfo{person}{Bing Xu},
  \bibinfo{person}{David Warde-Farley}, \bibinfo{person}{Sherjil Ozair},
  \bibinfo{person}{Aaron Courville}, {and} \bibinfo{person}{Yoshua Bengio}.}
  \bibinfo{year}{2020}\natexlab{}.
\newblock \showarticletitle{Generative adversarial networks}.
\newblock \bibinfo{journal}{\emph{Commun. ACM}} (\bibinfo{year}{2020}).
\newblock


\bibitem[Goodfellow et~al\mbox{.}(2014)]%
        {goodfellow2014explaining}
\bibfield{author}{\bibinfo{person}{Ian~J Goodfellow}, \bibinfo{person}{Jonathon
  Shlens}, {and} \bibinfo{person}{Christian Szegedy}.}
  \bibinfo{year}{2014}\natexlab{}.
\newblock \showarticletitle{Explaining and harnessing adversarial examples}.
\newblock \bibinfo{journal}{\emph{arXiv preprint arXiv:1412.6572}}
  (\bibinfo{year}{2014}).
\newblock


\bibitem[{James Vincent}(2023)]%
        {Meta-leaked}
\bibfield{author}{\bibinfo{person}{{James Vincent}}.}
  \bibinfo{year}{2023}\natexlab{}.
\newblock \bibinfo{title}{{Meta’s powerful AI language model has leaked
  online.}}
\newblock
  \bibinfo{howpublished}{\url{https://www.theverge.com/2023/3/8/23629362/meta-ai-language-model-llama-leak-online-misuse}}.
\newblock


\bibitem[Kirchenbauer et~al\mbox{.}(2023)]%
        {kirchenbauer2023watermark}
\bibfield{author}{\bibinfo{person}{John Kirchenbauer}, \bibinfo{person}{Jonas
  Geiping}, \bibinfo{person}{Yuxin Wen}, \bibinfo{person}{Jonathan Katz},
  \bibinfo{person}{Ian Miers}, {and} \bibinfo{person}{Tom Goldstein}.}
  \bibinfo{year}{2023}\natexlab{}.
\newblock \showarticletitle{A watermark for large language models}.
\newblock \bibinfo{journal}{\emph{arXiv preprint arXiv:2301.10226}}
  (\bibinfo{year}{2023}).
\newblock


\bibitem[Krishna et~al\mbox{.}(2023)]%
        {krishna2023paraphrasing}
\bibfield{author}{\bibinfo{person}{Kalpesh Krishna}, \bibinfo{person}{Yixiao
  Song}, \bibinfo{person}{Marzena Karpinska}, \bibinfo{person}{John Wieting},
  {and} \bibinfo{person}{Mohit Iyyer}.} \bibinfo{year}{2023}\natexlab{}.
\newblock \showarticletitle{Paraphrasing evades detectors of ai-generated text,
  but retrieval is an effective defense}. In \bibinfo{booktitle}{\emph{Advances
  in Neural Information Processing Systems}}.
\newblock


\bibitem[Lin et~al\mbox{.}(2014)]%
        {lin2014microsoft}
\bibfield{author}{\bibinfo{person}{Tsung-Yi Lin}, \bibinfo{person}{Michael
  Maire}, \bibinfo{person}{Serge Belongie}, \bibinfo{person}{James Hays},
  \bibinfo{person}{Pietro Perona}, \bibinfo{person}{Deva Ramanan},
  \bibinfo{person}{Piotr Doll{\'a}r}, {and} \bibinfo{person}{C~Lawrence
  Zitnick}.} \bibinfo{year}{2014}\natexlab{}.
\newblock \showarticletitle{Microsoft coco: Common objects in context}. In
  \bibinfo{booktitle}{\emph{European Conference on Computer Vision}}.
\newblock


\bibitem[Luo et~al\mbox{.}(2020)]%
        {luo2020distortion}
\bibfield{author}{\bibinfo{person}{Xiyang Luo}, \bibinfo{person}{Ruohan Zhan},
  \bibinfo{person}{Huiwen Chang}, \bibinfo{person}{Feng Yang}, {and}
  \bibinfo{person}{Peyman Milanfar}.} \bibinfo{year}{2020}\natexlab{}.
\newblock \showarticletitle{Distortion agnostic deep watermarking}. In
  \bibinfo{booktitle}{\emph{IEEE/CVF Conference on Computer Vision and Pattern
  Recognition}}.
\newblock


\bibitem[Madry et~al\mbox{.}(2017)]%
        {madry2017towards}
\bibfield{author}{\bibinfo{person}{Aleksander Madry},
  \bibinfo{person}{Aleksandar Makelov}, \bibinfo{person}{Ludwig Schmidt},
  \bibinfo{person}{Dimitris Tsipras}, {and} \bibinfo{person}{Adrian Vladu}.}
  \bibinfo{year}{2017}\natexlab{}.
\newblock \showarticletitle{Towards deep learning models resistant to
  adversarial attacks}.
\newblock \bibinfo{journal}{\emph{arXiv preprint arXiv:1706.06083}}
  (\bibinfo{year}{2017}).
\newblock


\bibitem[{Makena Kelly}(2023)]%
        {developing-AI-responsibility}
\bibfield{author}{\bibinfo{person}{{Makena Kelly}}.}
  \bibinfo{year}{2023}\natexlab{}.
\newblock \bibinfo{title}{{Meta, Google, and OpenAI promise the White House
  they’ll develop AI responsibly.}}
\newblock
  \bibinfo{howpublished}{\url{https://www.theverge.com/2023/7/21/23802274/artificial-intelligence-meta-google-openai-white-house-security-safety}}.
\newblock


\bibitem[{MARKETSANDMARKETS}(2023)]%
        {generative-ai-market}
\bibfield{author}{\bibinfo{person}{{MARKETSANDMARKETS}}.}
  \bibinfo{year}{2023}\natexlab{}.
\newblock \bibinfo{title}{{Generative AI Market.}}
\newblock
  \bibinfo{howpublished}{\url{https://www.marketsandmarkets.com/Market-Reports/generative-ai-market-142870584.html}}.
\newblock


\bibitem[{Marking the Photo}(2022)]%
        {remove-dall-e}
\bibfield{author}{\bibinfo{person}{{Marking the Photo}}.}
  \bibinfo{year}{2022}\natexlab{}.
\newblock \bibinfo{title}{{How to Remove Dall-E Watermark.}}
\newblock
  \bibinfo{howpublished}{\url{https://www.youtube.com/watch?v=6EMROCxGCIA}}.
\newblock


\bibitem[Mitchell et~al\mbox{.}(2023)]%
        {mitchell2023detectgpt}
\bibfield{author}{\bibinfo{person}{Eric Mitchell}, \bibinfo{person}{Yoonho
  Lee}, \bibinfo{person}{Alexander Khazatsky}, \bibinfo{person}{Christopher~D
  Manning}, {and} \bibinfo{person}{Chelsea Finn}.}
  \bibinfo{year}{2023}\natexlab{}.
\newblock \showarticletitle{Detectgpt: Zero-shot machine-generated text
  detection using probability curvature}.
\newblock \bibinfo{journal}{\emph{arXiv preprint arXiv:2301.11305}}
  (\bibinfo{year}{2023}).
\newblock


\bibitem[{OpenAI}(2022)]%
        {openai-chatgpt}
\bibfield{author}{\bibinfo{person}{{OpenAI}}.} \bibinfo{year}{2022}\natexlab{}.
\newblock \bibinfo{title}{{Chatgpt: Optimizing language models for dialogue.}}
\newblock \bibinfo{howpublished}{\url{https://openai.com/blog/chatgpt}}.
\newblock


\bibitem[Pereira and Pun(2000)]%
        {pereira2000robust}
\bibfield{author}{\bibinfo{person}{Shelby Pereira} {and}
  \bibinfo{person}{Thierry Pun}.} \bibinfo{year}{2000}\natexlab{}.
\newblock \showarticletitle{Robust template matching for affine resistant image
  watermarks}.
\newblock \bibinfo{journal}{\emph{IEEE Transactions on Image Processing}}
  (\bibinfo{year}{2000}).
\newblock


\bibitem[{Qingquan Wang and buley}(2020)]%
        {invisible-watermark}
\bibfield{author}{\bibinfo{person}{{Qingquan Wang and buley}}.}
  \bibinfo{year}{2020}\natexlab{}.
\newblock \bibinfo{title}{{Invisible watermark.}}
\newblock
  \bibinfo{howpublished}{\url{https://github.com/ShieldMnt/invisible-watermark}}.
\newblock


\bibitem[Ramesh et~al\mbox{.}(2021)]%
        {ramesh2021zero}
\bibfield{author}{\bibinfo{person}{Aditya Ramesh}, \bibinfo{person}{Mikhail
  Pavlov}, \bibinfo{person}{Gabriel Goh}, \bibinfo{person}{Scott Gray},
  \bibinfo{person}{Chelsea Voss}, \bibinfo{person}{Alec Radford},
  \bibinfo{person}{Mark Chen}, {and} \bibinfo{person}{Ilya Sutskever}.}
  \bibinfo{year}{2021}\natexlab{}.
\newblock \showarticletitle{Zero-shot text-to-image generation}. In
  \bibinfo{booktitle}{\emph{International Conference on Machine Learning}}.
\newblock


\bibitem[{Robin Rombach}(2022)]%
        {stable-diffusion-watermark-decoder}
\bibfield{author}{\bibinfo{person}{{Robin Rombach}}.}
  \bibinfo{year}{2022}\natexlab{}.
\newblock \bibinfo{title}{{Stable Diffusion watermark decoder.}}
\newblock
  \bibinfo{howpublished}{\url{https://github.com/CompVis/stable-diffusion/blob/main/scripts/tests/test_watermark.py}}.
\newblock


\bibitem[Rombach et~al\mbox{.}(2022)]%
        {rombach2022high}
\bibfield{author}{\bibinfo{person}{Robin Rombach}, \bibinfo{person}{Andreas
  Blattmann}, \bibinfo{person}{Dominik Lorenz}, \bibinfo{person}{Patrick
  Esser}, {and} \bibinfo{person}{Bj{\"o}rn Ommer}.}
  \bibinfo{year}{2022}\natexlab{}.
\newblock \showarticletitle{High-resolution image synthesis with latent
  diffusion models}. In \bibinfo{booktitle}{\emph{IEEE/CVF Conference on
  Computer Vision and Pattern Recognition}}.
\newblock


\bibitem[Sadasivan et~al\mbox{.}(2023)]%
        {sadasivan2023can}
\bibfield{author}{\bibinfo{person}{Vinu~Sankar Sadasivan},
  \bibinfo{person}{Aounon Kumar}, \bibinfo{person}{Sriram Balasubramanian},
  \bibinfo{person}{Wenxiao Wang}, {and} \bibinfo{person}{Soheil Feizi}.}
  \bibinfo{year}{2023}\natexlab{}.
\newblock \showarticletitle{Can AI-Generated Text be Reliably Detected?}
\newblock \bibinfo{journal}{\emph{arXiv preprint arXiv:2303.11156}}
  (\bibinfo{year}{2023}).
\newblock


\bibitem[Sha et~al\mbox{.}(2022)]%
        {sha2022fake}
\bibfield{author}{\bibinfo{person}{Zeyang Sha}, \bibinfo{person}{Zheng Li},
  \bibinfo{person}{Ning Yu}, {and} \bibinfo{person}{Yang Zhang}.}
  \bibinfo{year}{2022}\natexlab{}.
\newblock \showarticletitle{DE-FAKE: Detection and Attribution of Fake Images
  Generated by Text-to-Image Diffusion Models}.
\newblock \bibinfo{journal}{\emph{arXiv preprint arXiv:2210.06998}}
  (\bibinfo{year}{2022}).
\newblock


\bibitem[Sharif et~al\mbox{.}(2018)]%
        {sharif2018suitability}
\bibfield{author}{\bibinfo{person}{Mahmood Sharif}, \bibinfo{person}{Lujo
  Bauer}, {and} \bibinfo{person}{Michael~K Reiter}.}
  \bibinfo{year}{2018}\natexlab{}.
\newblock \showarticletitle{On the suitability of lp-norms for creating and
  preventing adversarial examples}. In \bibinfo{booktitle}{\emph{IEEE/CVF
  Conference on Computer Vision and Pattern Recognition Workshops}}.
\newblock


\bibitem[Sharma et~al\mbox{.}(2018)]%
        {sharma2018conceptual}
\bibfield{author}{\bibinfo{person}{Piyush Sharma}, \bibinfo{person}{Nan Ding},
  \bibinfo{person}{Sebastian Goodman}, {and} \bibinfo{person}{Radu Soricut}.}
  \bibinfo{year}{2018}\natexlab{}.
\newblock \showarticletitle{Conceptual captions: A cleaned, hypernymed, image
  alt-text dataset for automatic image captioning}. In
  \bibinfo{booktitle}{\emph{Annual Meeting of the Association for Computational
  Linguistics}}.
\newblock


\bibitem[{Shivdeep Dhaliwal}(2023)]%
        {fake-news}
\bibfield{author}{\bibinfo{person}{{Shivdeep Dhaliwal}}.}
  \bibinfo{year}{2023}\natexlab{}.
\newblock \bibinfo{title}{{Elon Musk isn't dating GM's Mary Barra: he has this
  to say though on the photos.}}
\newblock
  \bibinfo{howpublished}{\url{https://www.benzinga.com/news/23/03/31505898/elon-musk-isnt-dating-gms-mary-barra-he-has-this-to-say-though-on-the-photos}}.
\newblock


\bibitem[Szegedy et~al\mbox{.}(2013)]%
        {szegedy2013intriguing}
\bibfield{author}{\bibinfo{person}{Christian Szegedy},
  \bibinfo{person}{Wojciech Zaremba}, \bibinfo{person}{Ilya Sutskever},
  \bibinfo{person}{Joan Bruna}, \bibinfo{person}{Dumitru Erhan},
  \bibinfo{person}{Ian Goodfellow}, {and} \bibinfo{person}{Rob Fergus}.}
  \bibinfo{year}{2013}\natexlab{}.
\newblock \showarticletitle{Intriguing properties of neural networks}.
\newblock \bibinfo{journal}{\emph{arXiv preprint arXiv:1312.6199}}
  (\bibinfo{year}{2013}).
\newblock


\bibitem[Tancik et~al\mbox{.}(2020)]%
        {tancik2020stegastamp}
\bibfield{author}{\bibinfo{person}{Matthew Tancik}, \bibinfo{person}{Ben
  Mildenhall}, {and} \bibinfo{person}{Ren Ng}.}
  \bibinfo{year}{2020}\natexlab{}.
\newblock \showarticletitle{Stegastamp: Invisible hyperlinks in physical
  photographs}. In \bibinfo{booktitle}{\emph{IEEE/CVF Conference on Computer
  Vision and Pattern Recognition}}.
\newblock


\bibitem[Wang et~al\mbox{.}(2019)]%
        {wang2019detecting}
\bibfield{author}{\bibinfo{person}{Sheng-Yu Wang}, \bibinfo{person}{Oliver
  Wang}, \bibinfo{person}{Andrew Owens}, \bibinfo{person}{Richard Zhang}, {and}
  \bibinfo{person}{Alexei~A Efros}.} \bibinfo{year}{2019}\natexlab{}.
\newblock \showarticletitle{Detecting photoshopped faces by scripting
  photoshop}. In \bibinfo{booktitle}{\emph{IEEE/CVF International Conference on
  Computer Vision}}.
\newblock


\bibitem[Wang et~al\mbox{.}(2004)]%
        {wang2004image}
\bibfield{author}{\bibinfo{person}{Zhou Wang}, \bibinfo{person}{Alan~C Bovik},
  \bibinfo{person}{Hamid~R Sheikh}, {and} \bibinfo{person}{Eero~P Simoncelli}.}
  \bibinfo{year}{2004}\natexlab{}.
\newblock \showarticletitle{Image quality assessment: from error visibility to
  structural similarity}.
\newblock \bibinfo{journal}{\emph{IEEE Transactions on Image Processing}}
  (\bibinfo{year}{2004}).
\newblock


\bibitem[Wen and Aydore(2019)]%
        {wen2019romark}
\bibfield{author}{\bibinfo{person}{Bingyang Wen} {and} \bibinfo{person}{Sergul
  Aydore}.} \bibinfo{year}{2019}\natexlab{}.
\newblock \showarticletitle{Romark: A robust watermarking system using
  adversarial training}.
\newblock \bibinfo{journal}{\emph{arXiv preprint arXiv:1910.01221}}
  (\bibinfo{year}{2019}).
\newblock


\bibitem[Wen et~al\mbox{.}(2023)]%
        {wen2023tree}
\bibfield{author}{\bibinfo{person}{Yuxin Wen}, \bibinfo{person}{John
  Kirchenbauer}, \bibinfo{person}{Jonas Geiping}, {and} \bibinfo{person}{Tom
  Goldstein}.} \bibinfo{year}{2023}\natexlab{}.
\newblock \showarticletitle{Tree-Ring Watermarks: Fingerprints for Diffusion
  Images that are Invisible and Robust}.
\newblock \bibinfo{journal}{\emph{arXiv preprint arXiv:2305.20030}}
  (\bibinfo{year}{2023}).
\newblock


\bibitem[Yang et~al\mbox{.}(2021)]%
        {yang2021faceguard}
\bibfield{author}{\bibinfo{person}{Yuankun Yang}, \bibinfo{person}{Chenyue
  Liang}, \bibinfo{person}{Hongyu He}, \bibinfo{person}{Xiaoyu Cao}, {and}
  \bibinfo{person}{Neil~Zhenqiang Gong}.} \bibinfo{year}{2021}\natexlab{}.
\newblock \showarticletitle{Faceguard: Proactive deepfake detection}.
\newblock \bibinfo{journal}{\emph{arXiv preprint arXiv:2109.05673}}
  (\bibinfo{year}{2021}).
\newblock


\bibitem[Yu et~al\mbox{.}(2019)]%
        {yu2019attributing}
\bibfield{author}{\bibinfo{person}{Ning Yu}, \bibinfo{person}{Larry~S Davis},
  {and} \bibinfo{person}{Mario Fritz}.} \bibinfo{year}{2019}\natexlab{}.
\newblock \showarticletitle{Attributing fake images to gans: Learning and
  analyzing gan fingerprints}. In \bibinfo{booktitle}{\emph{IEEE/CVF
  International Conference on Computer Vision}}.
\newblock


\bibitem[Yu et~al\mbox{.}(2021)]%
        {yu2021artificial}
\bibfield{author}{\bibinfo{person}{Ning Yu}, \bibinfo{person}{Vladislav
  Skripniuk}, \bibinfo{person}{Sahar Abdelnabi}, {and} \bibinfo{person}{Mario
  Fritz}.} \bibinfo{year}{2021}\natexlab{}.
\newblock \showarticletitle{Artificial fingerprinting for generative models:
  Rooting deepfake attribution in training data}. In
  \bibinfo{booktitle}{\emph{IEEE/CVF International Conference on Computer
  Vision}}.
\newblock


\bibitem[Zhang et~al\mbox{.}(2020a)]%
        {zhang2020udh}
\bibfield{author}{\bibinfo{person}{Chaoning Zhang}, \bibinfo{person}{Philipp
  Benz}, \bibinfo{person}{Adil Karjauv}, \bibinfo{person}{Geng Sun}, {and}
  \bibinfo{person}{In~So Kweon}.} \bibinfo{year}{2020}\natexlab{a}.
\newblock \showarticletitle{Udh: Universal deep hiding for steganography,
  watermarking, and light field messaging}.
\newblock \bibinfo{journal}{\emph{Advances in Neural Information Processing
  Systems}} (\bibinfo{year}{2020}).
\newblock


\bibitem[Zhang et~al\mbox{.}(2020b)]%
        {zhang2020towards}
\bibfield{author}{\bibinfo{person}{Chaoning Zhang}, \bibinfo{person}{Adil
  Karjauv}, \bibinfo{person}{Philipp Benz}, {and} \bibinfo{person}{In~So
  Kweon}.} \bibinfo{year}{2020}\natexlab{b}.
\newblock \showarticletitle{Towards robust data hiding against (jpeg)
  compression: A pseudo-differentiable deep learning approach}.
\newblock \bibinfo{journal}{\emph{arXiv preprint arXiv:2101.00973}}
  (\bibinfo{year}{2020}).
\newblock


\bibitem[Zhang et~al\mbox{.}(2019)]%
        {zhang2019steganogan}
\bibfield{author}{\bibinfo{person}{Kevin~Alex Zhang}, \bibinfo{person}{Alfredo
  Cuesta-Infante}, \bibinfo{person}{Lei Xu}, {and} \bibinfo{person}{Kalyan
  Veeramachaneni}.} \bibinfo{year}{2019}\natexlab{}.
\newblock \showarticletitle{SteganoGAN: High capacity image steganography with
  GANs}.
\newblock \bibinfo{journal}{\emph{arXiv preprint arXiv:1901.03892}}
  (\bibinfo{year}{2019}).
\newblock


\bibitem[Zhao et~al\mbox{.}(2021)]%
        {zhao2021multi}
\bibfield{author}{\bibinfo{person}{Hanqing Zhao}, \bibinfo{person}{Wenbo Zhou},
  \bibinfo{person}{Dongdong Chen}, \bibinfo{person}{Tianyi Wei},
  \bibinfo{person}{Weiming Zhang}, {and} \bibinfo{person}{Nenghai Yu}.}
  \bibinfo{year}{2021}\natexlab{}.
\newblock \showarticletitle{Multi-attentional deepfake detection}. In
  \bibinfo{booktitle}{\emph{IEEE/CVF Conference on Computer Vision and Pattern
  Recognition}}.
\newblock


\bibitem[Zhu et~al\mbox{.}(2018)]%
        {zhu2018hidden}
\bibfield{author}{\bibinfo{person}{Jiren Zhu}, \bibinfo{person}{Russell
  Kaplan}, \bibinfo{person}{Justin Johnson}, {and} \bibinfo{person}{Li
  Fei-Fei}.} \bibinfo{year}{2018}\natexlab{}.
\newblock \showarticletitle{Hidden: Hiding data with deep networks}. In
  \bibinfo{booktitle}{\emph{European Conference on Computer Vision}}.
\newblock


\end{thebibliography}

\appendix

\begin{figure}[!t]
\centering
\subfloat[FPR]{\includegraphics[width=0.24\textwidth]{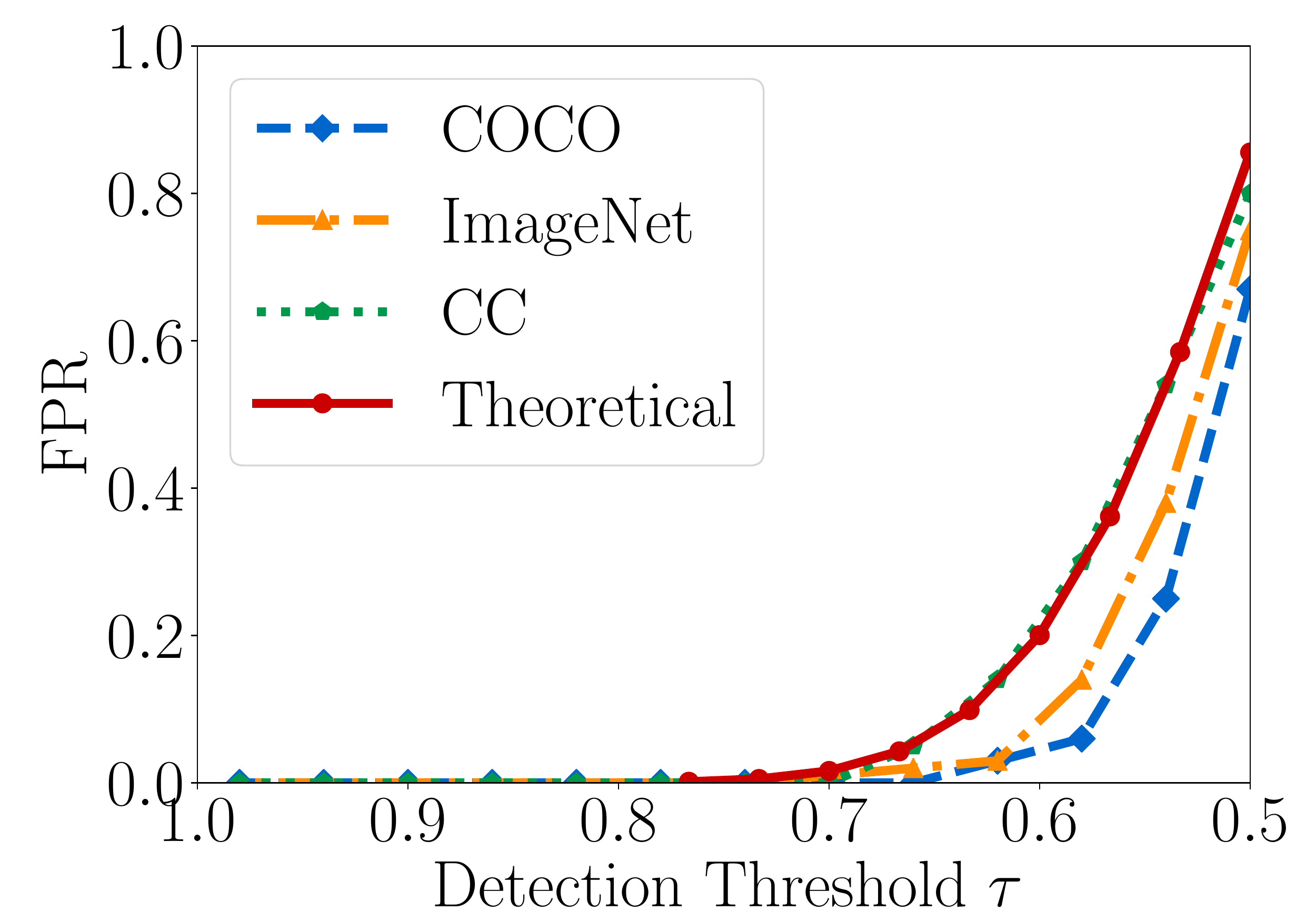}}
\subfloat[FNR]{\includegraphics[width=0.24\textwidth]{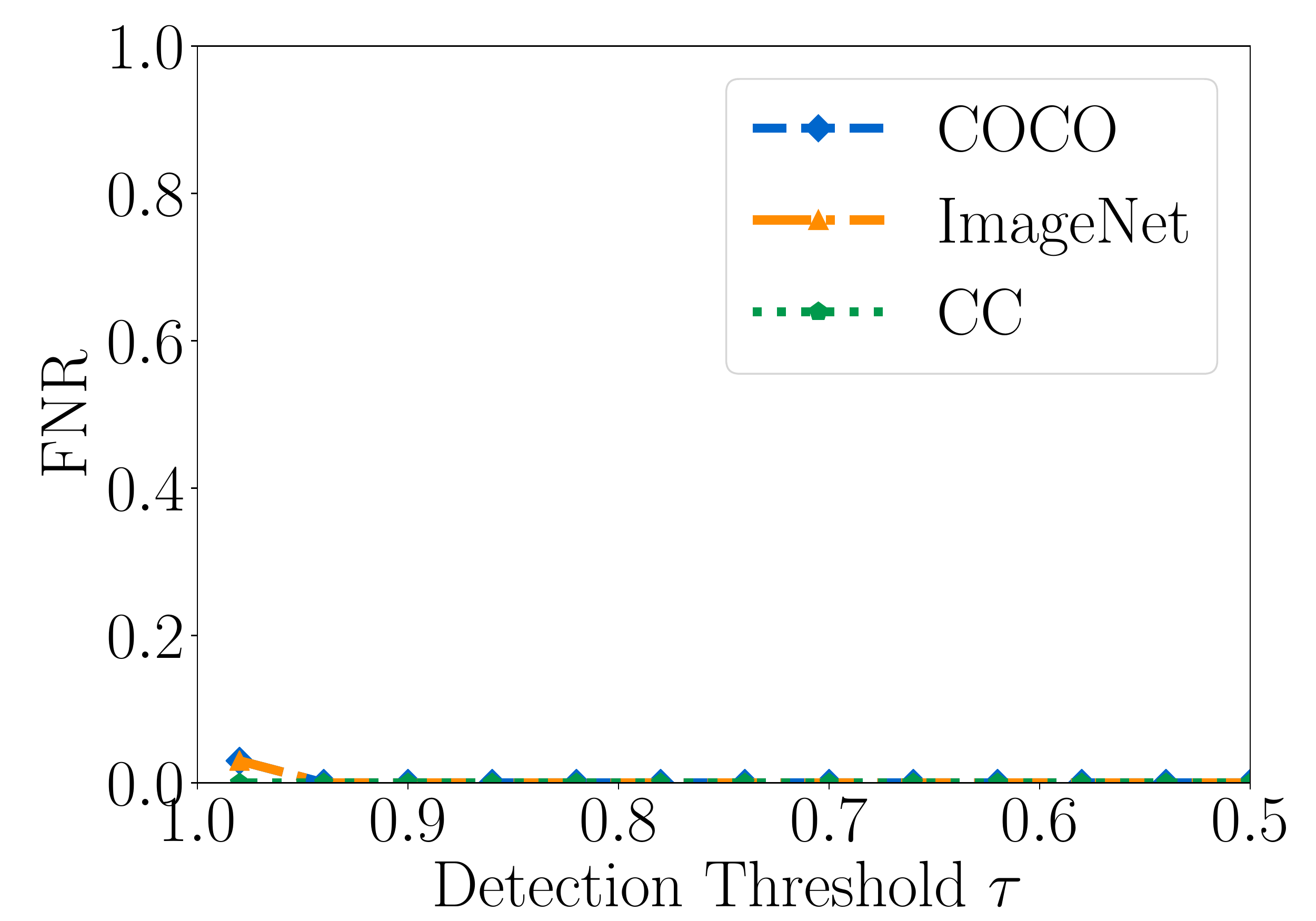}}
\caption{ FPR and  FNR of the double-tail detector based on HiDDeN as the threshold $\tau$ varies  when there are no attacks to post-process the watermarked images. }
\label{no-attack-detector-hidden}
\end{figure}

\begin{figure}[!t]
\centering
\subfloat[HiDDeN, FPR]{\includegraphics[width=0.24\textwidth]{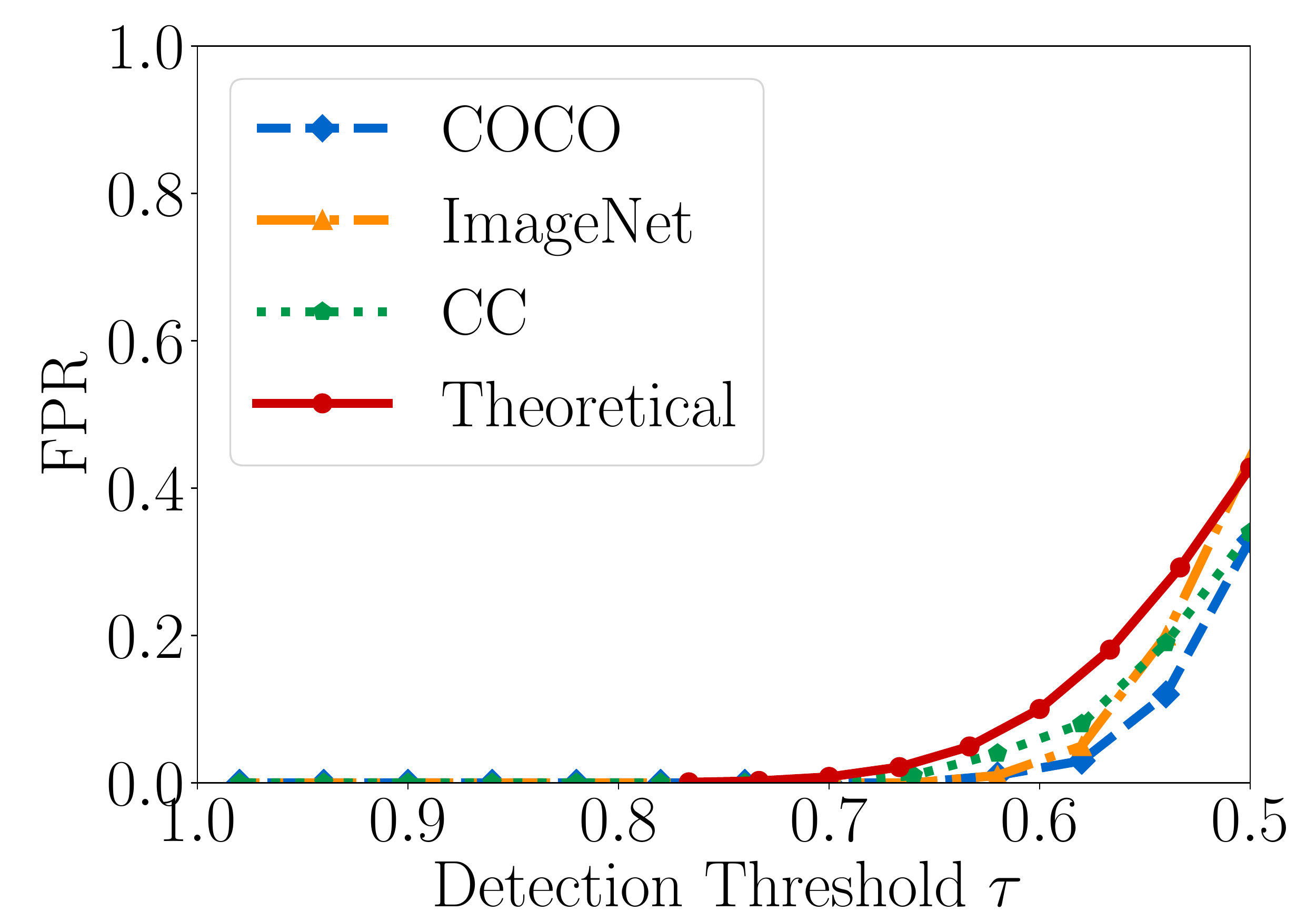}}
\subfloat[HiDDeN, FNR]{\includegraphics[width=0.24\textwidth]{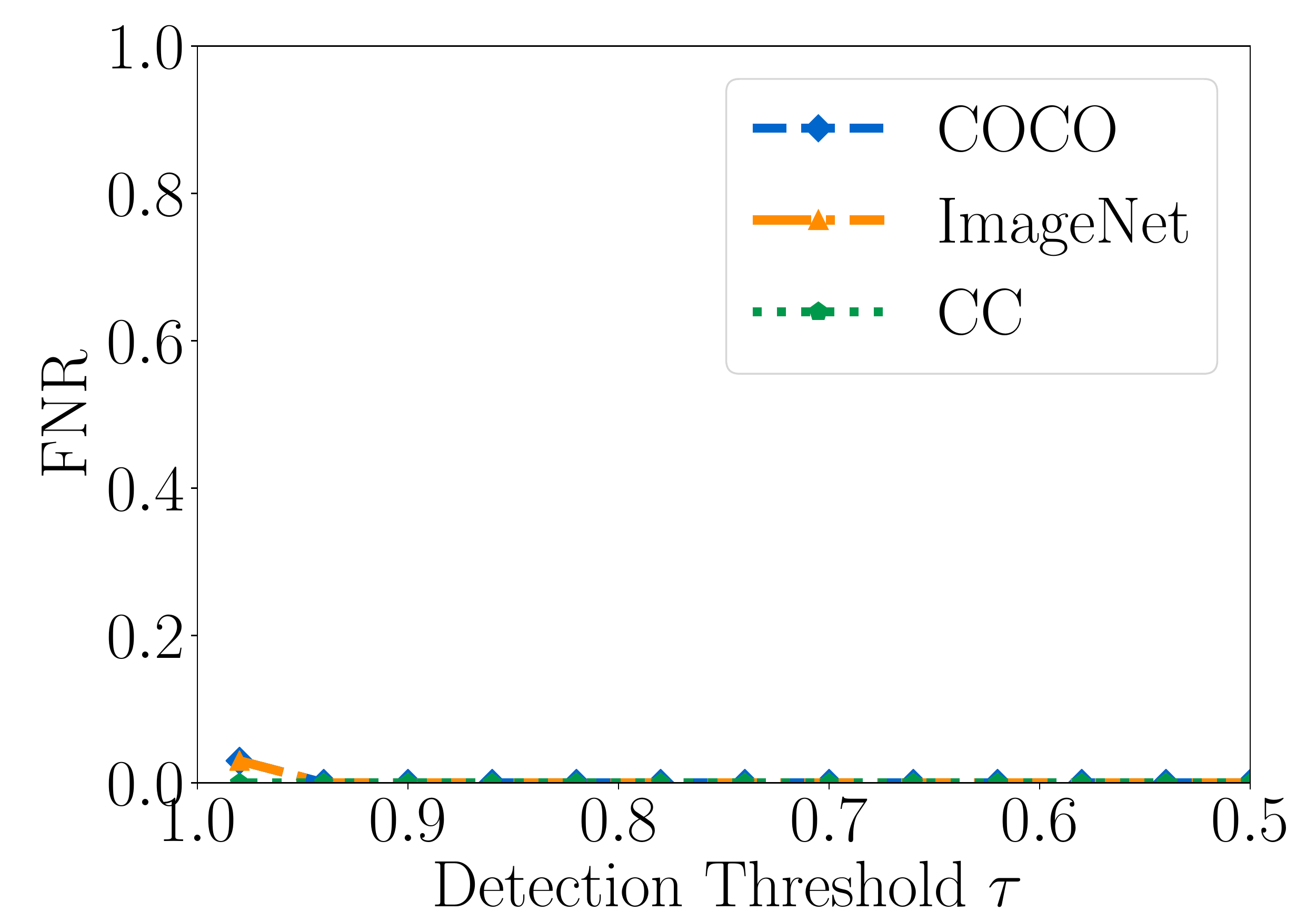}}

\subfloat[UDH, FPR]{\includegraphics[width=0.24\textwidth]{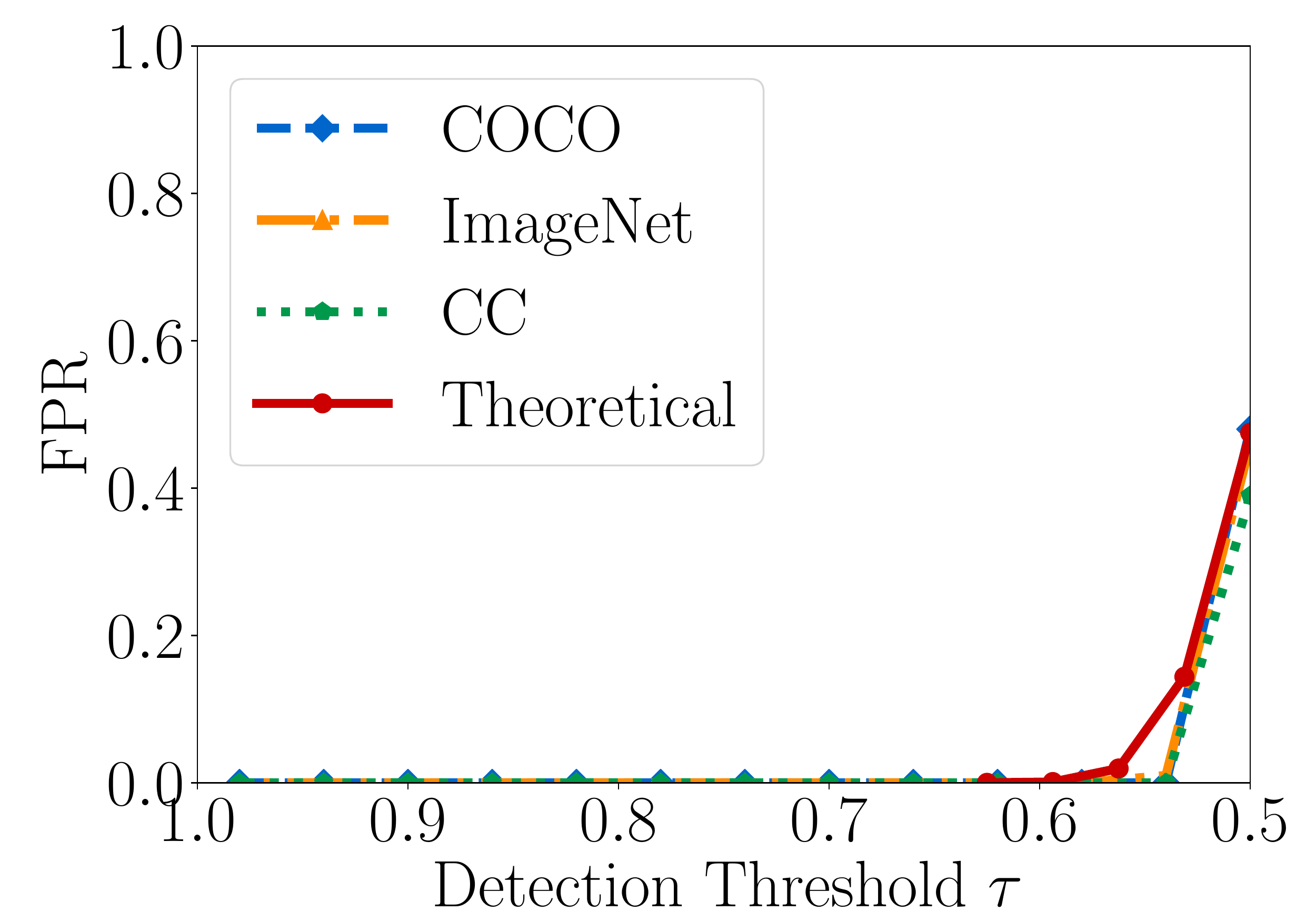}}
\subfloat[UDH, FNR]{\includegraphics[width=0.24\textwidth]{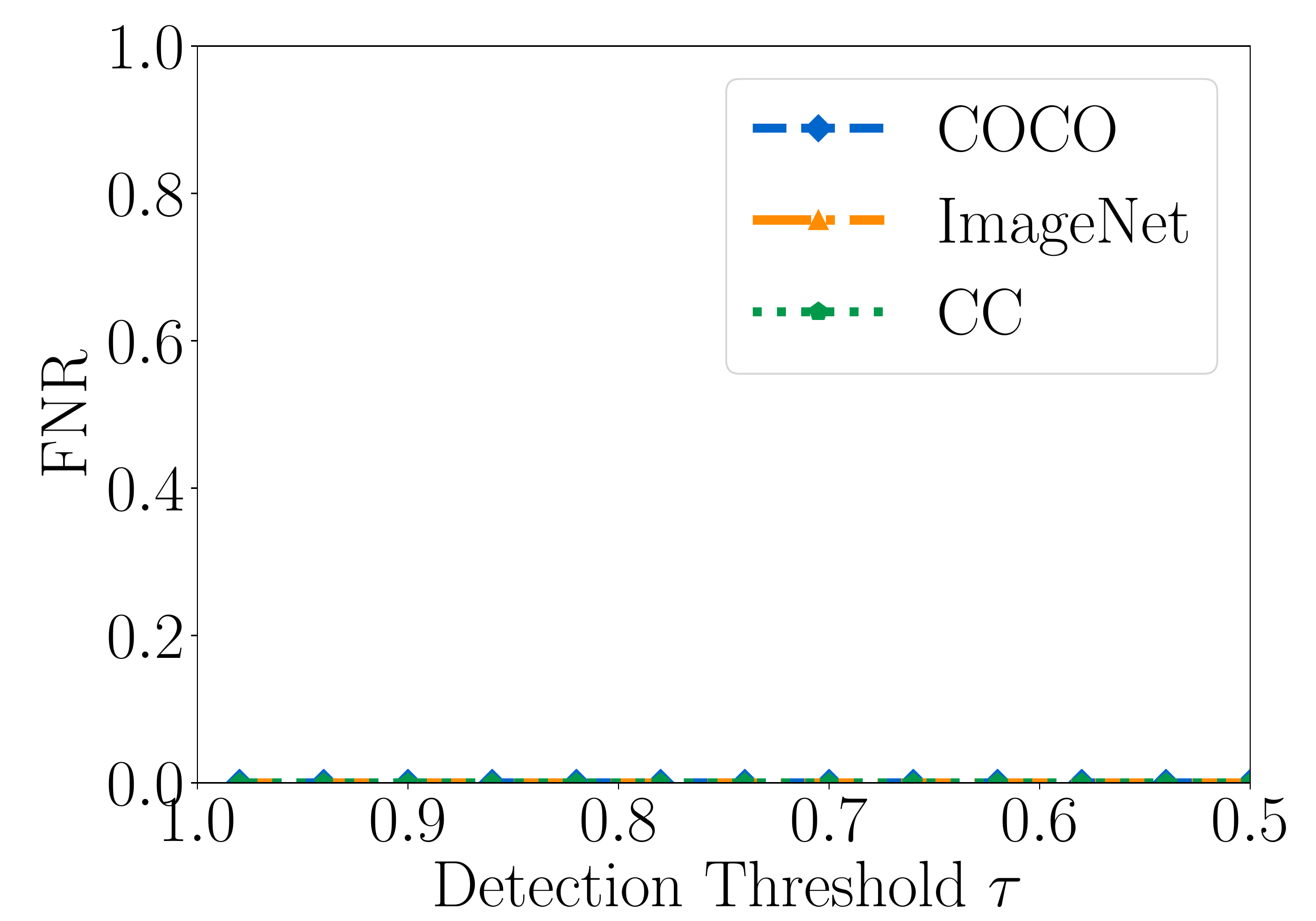}}
\caption{FPR and FNR of the single-tail detector as the threshold $\tau$ varies  when there are no attacks to post-process the watermarked images. }
\label{no-attack-standard-detector}
\end{figure}

\begin{figure}[!t]
\centering
{\includegraphics[width=0.227\textwidth]{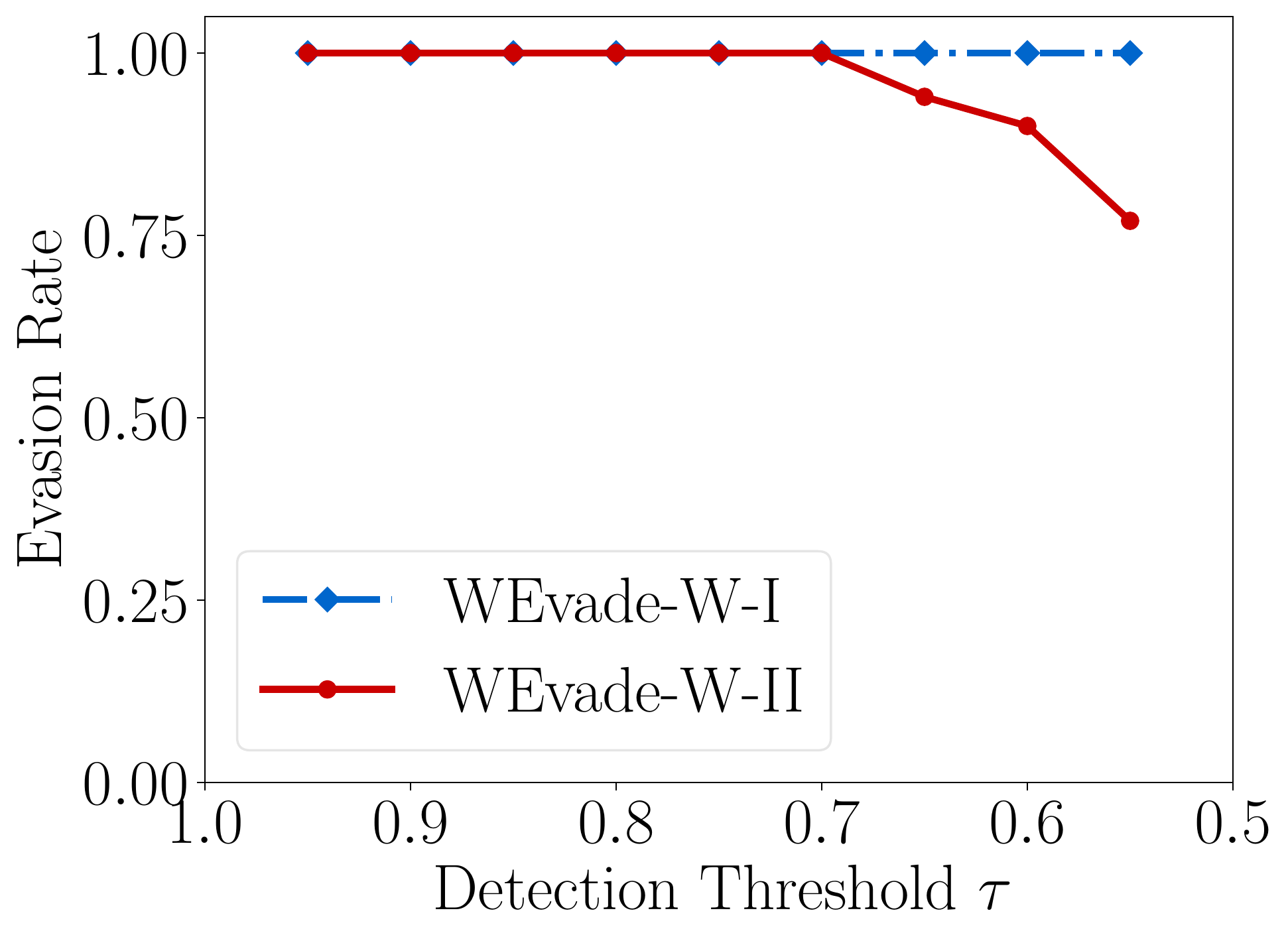}}\hspace{0.2mm}
{\includegraphics[width=0.233\textwidth]{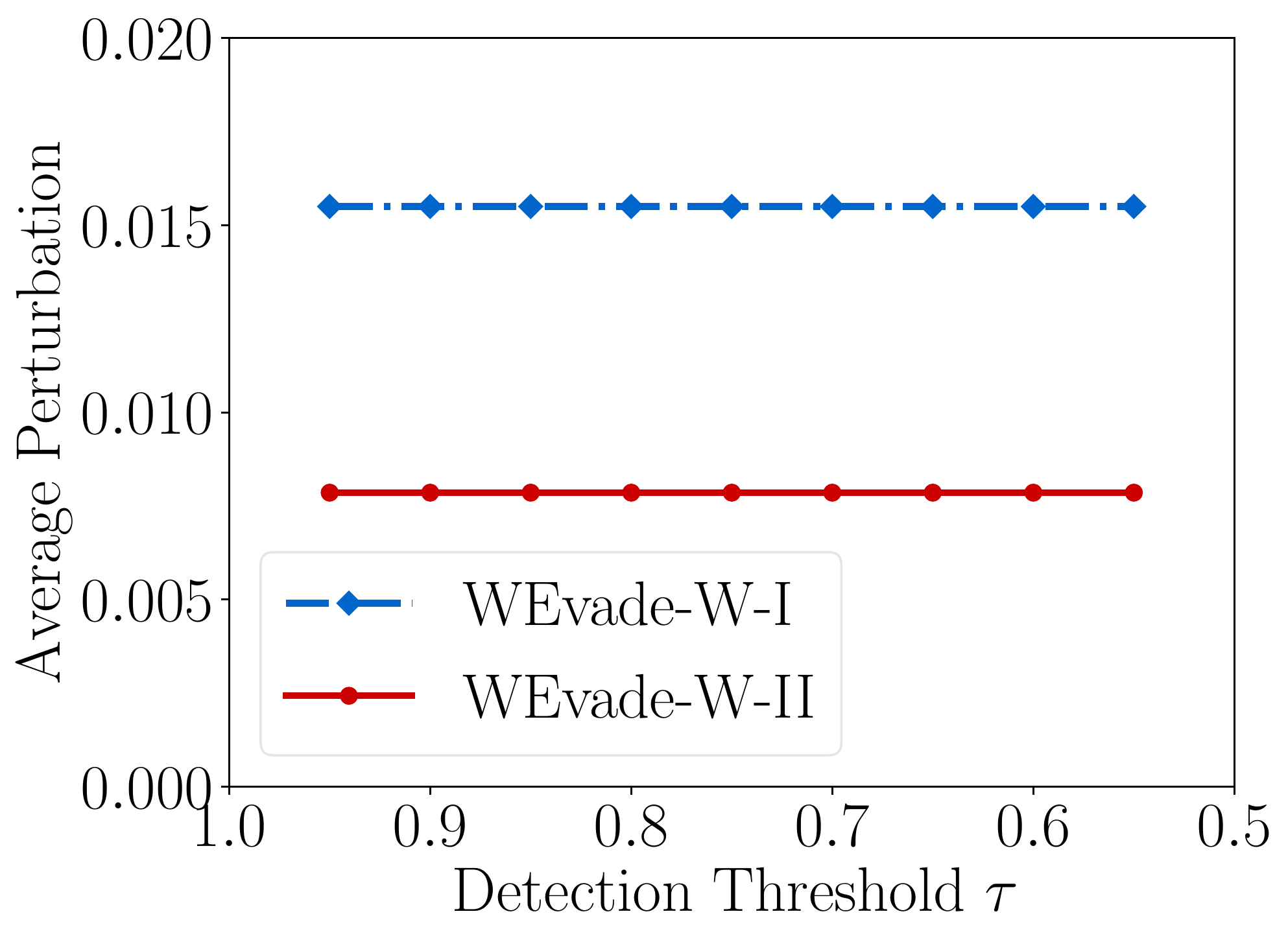}}

\subfloat[Evasion rate]{\hspace{-1mm}
\includegraphics[width=0.227\textwidth]{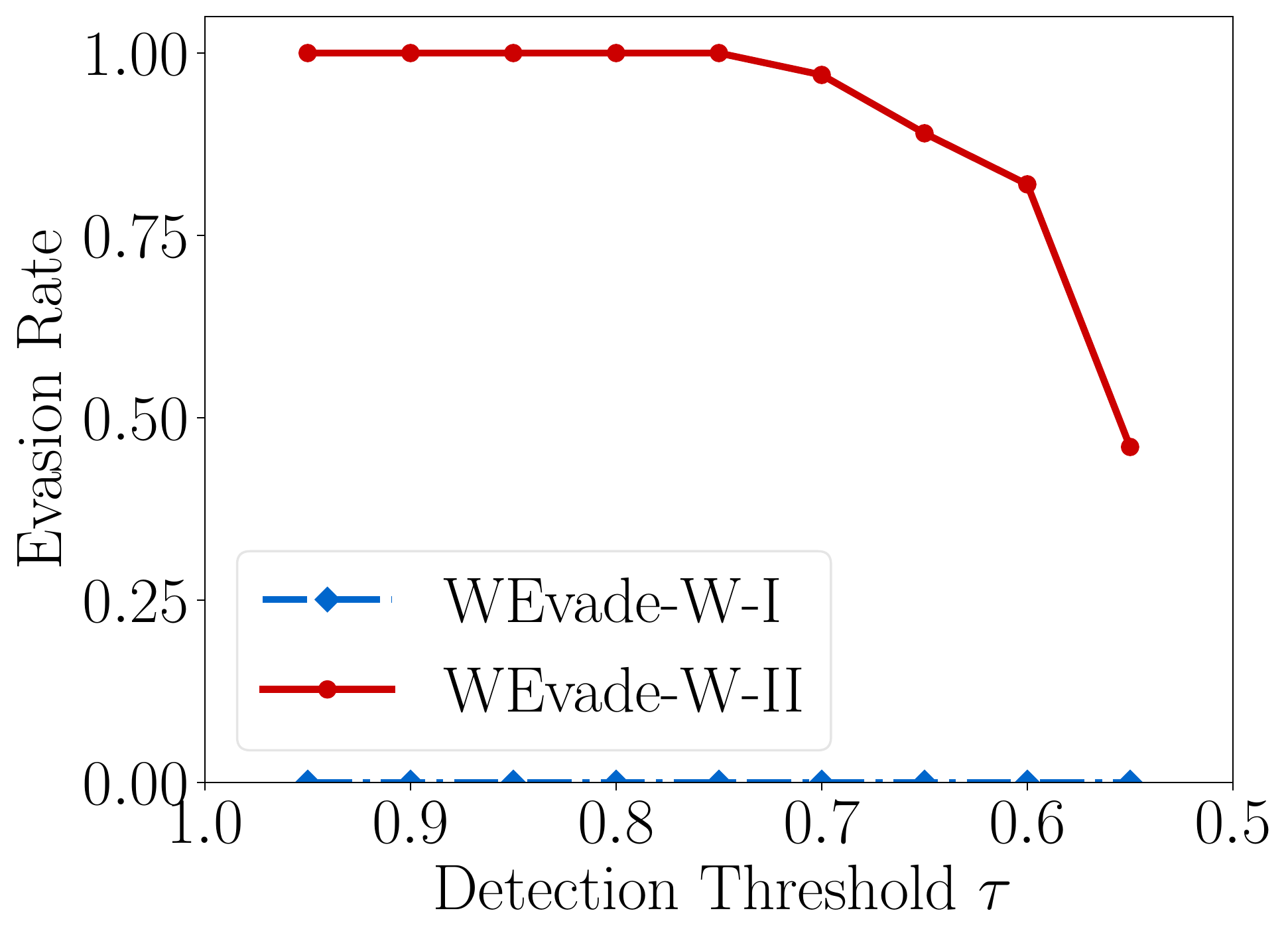}}
\subfloat[Average perturbation]{
\includegraphics[width=0.233\textwidth]{figures/WEvade_W_I_II/adaptive_pert.pdf}}
\caption{Comparing \algns-W-I with \algns-W-II against the single-tail (\emph{first row}) and double-tail (\emph{second row}) detector.}
\label{variant-comparison}
\end{figure}

\begin{figure*}[!t]
\centering
{
\includegraphics[width=0.24\textwidth]{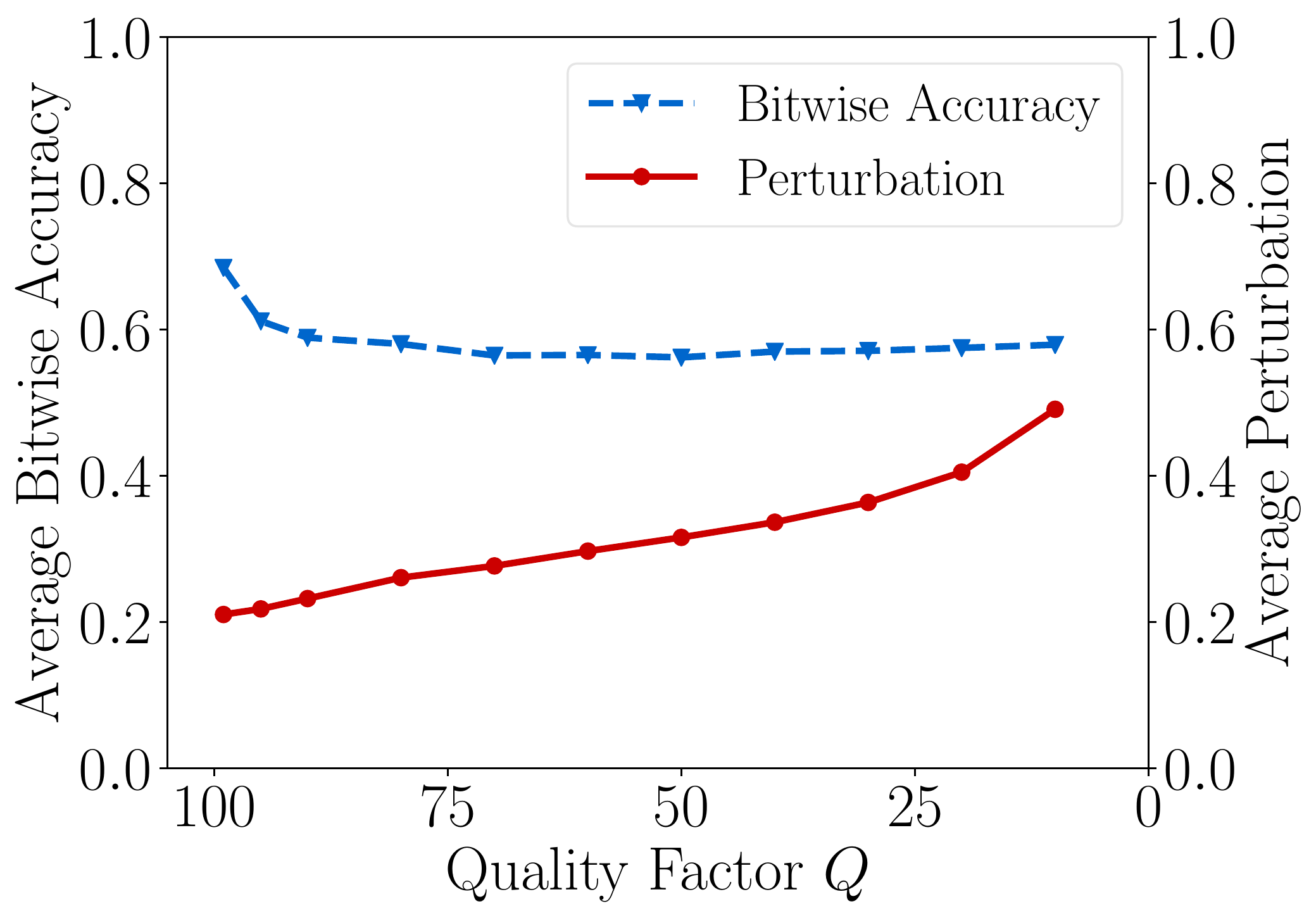}
}%
{
\includegraphics[width=0.24\textwidth]{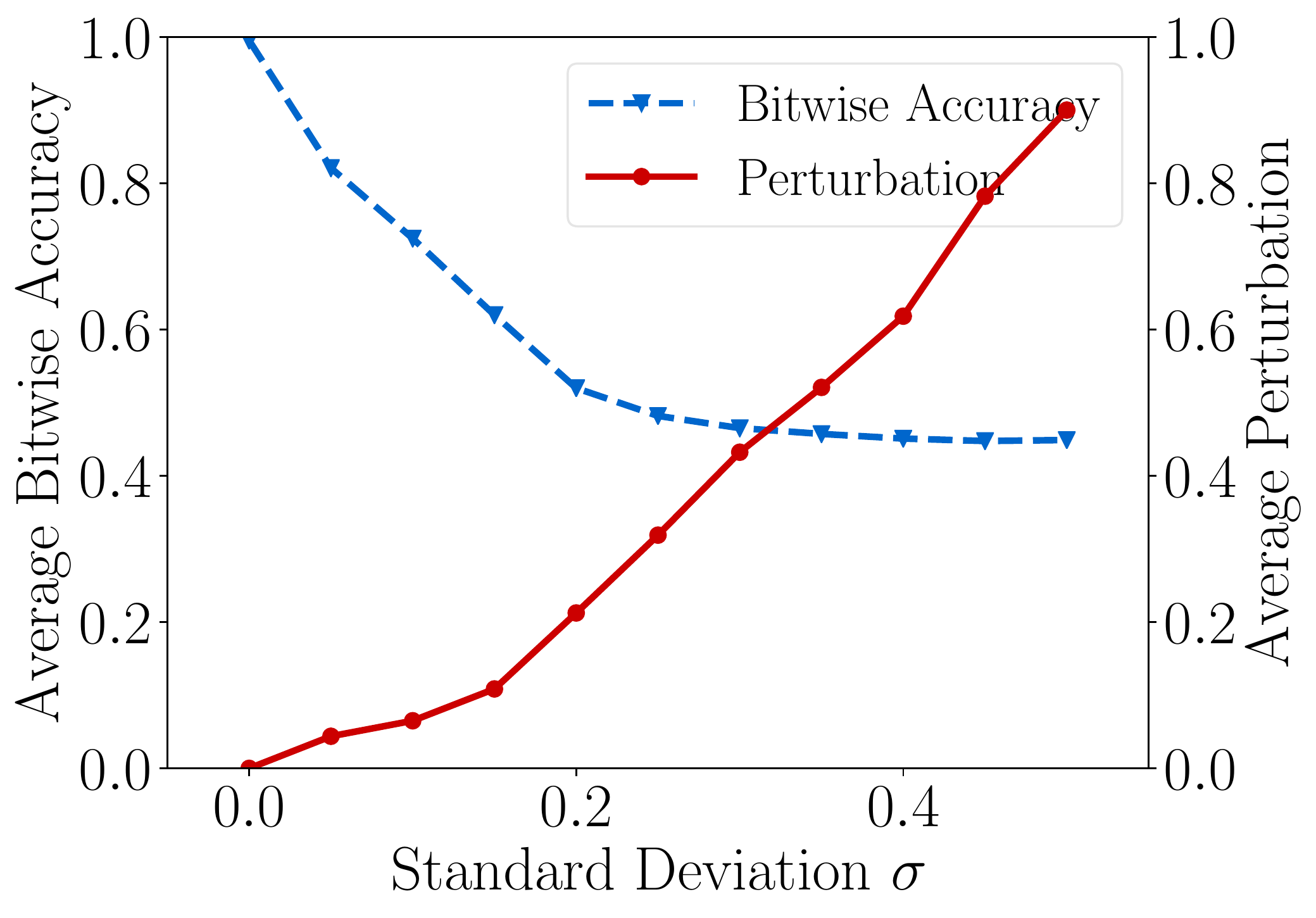}
}%
{
\includegraphics[width=0.24\textwidth]{figures/HiDDeN/COCO_Gaussian_Blur.pdf}
}%
{
\includegraphics[width=0.24\textwidth]{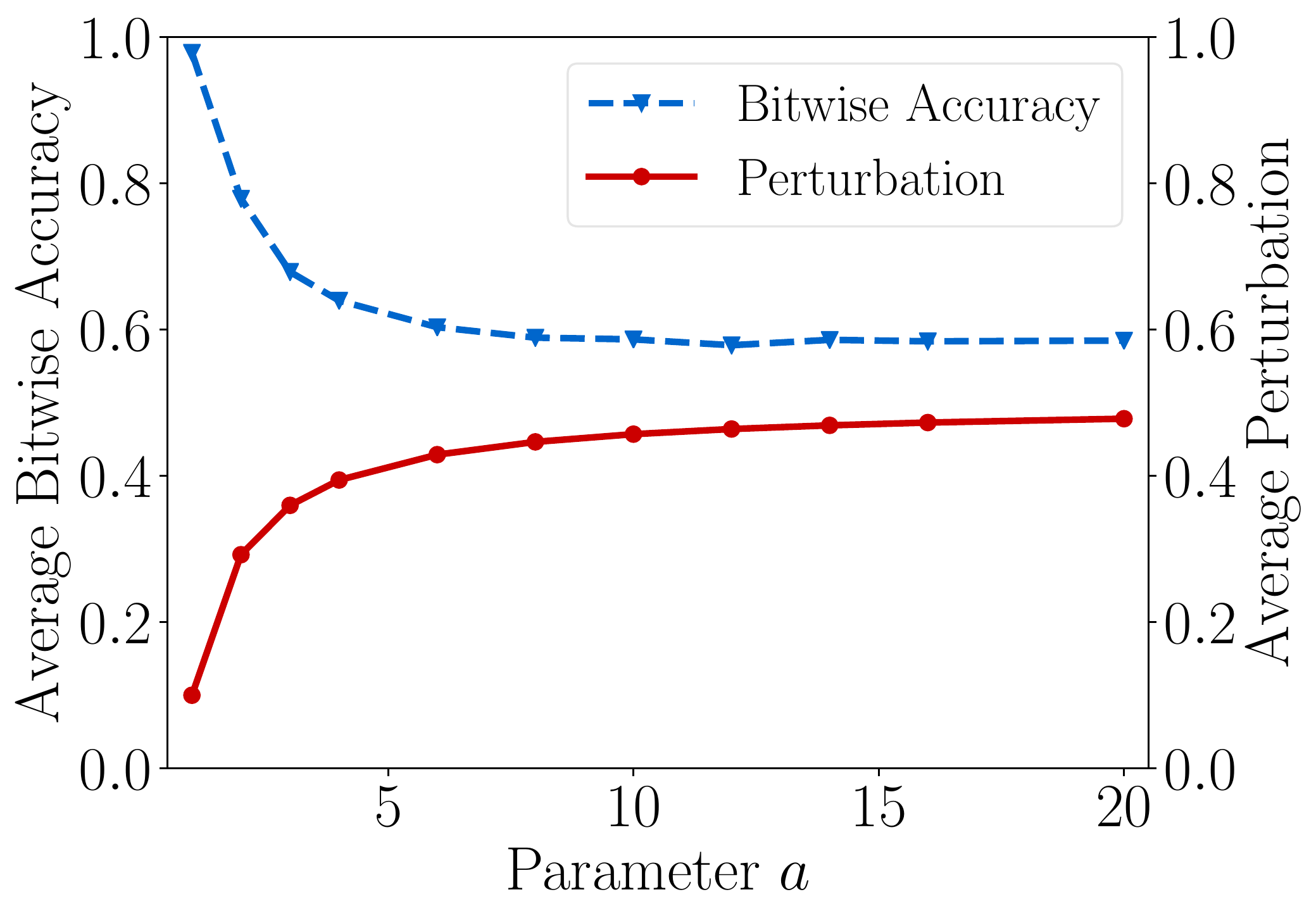}
}%

{
\includegraphics[width=0.24\textwidth]{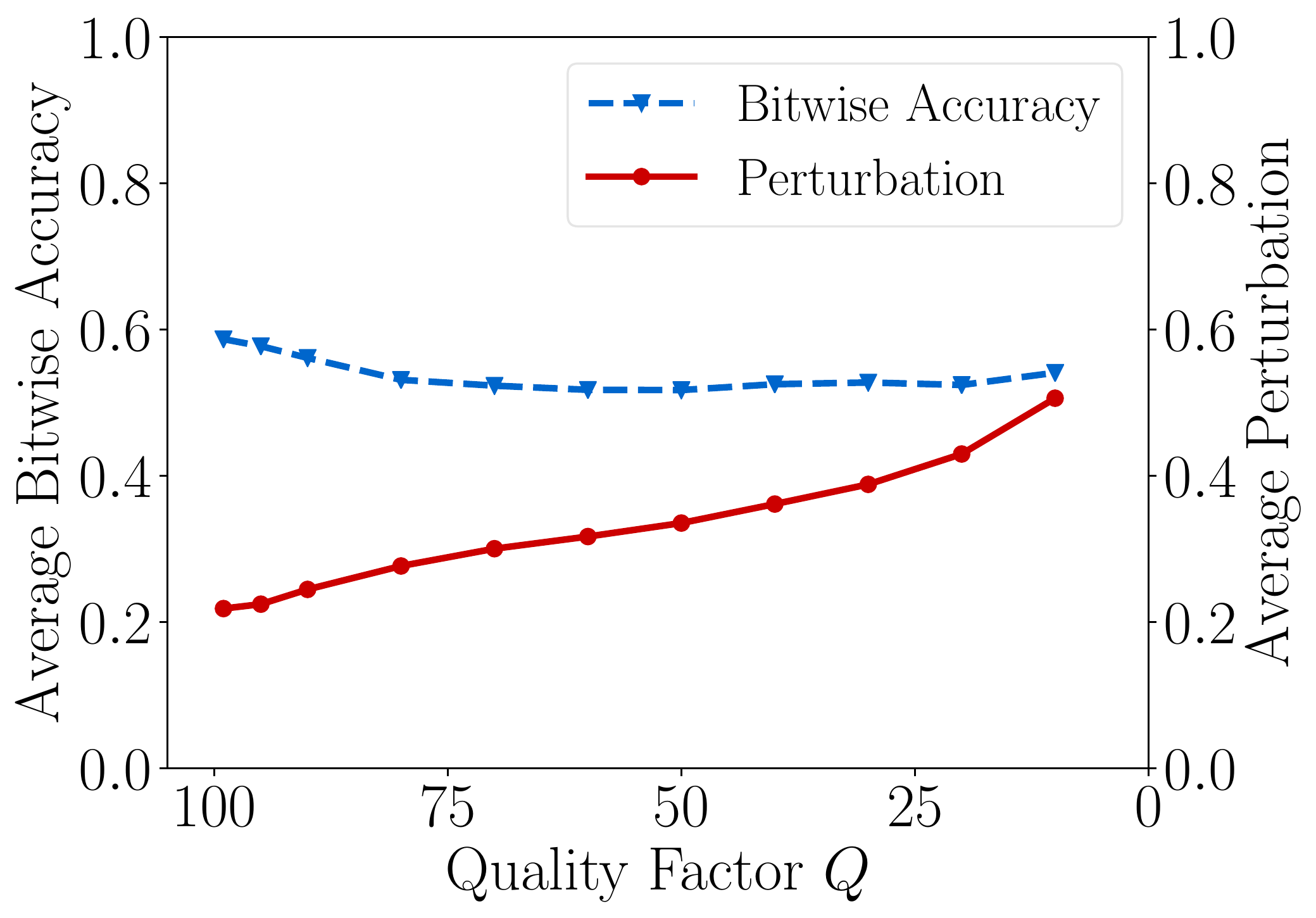}
}%
{
\includegraphics[width=0.24\textwidth]{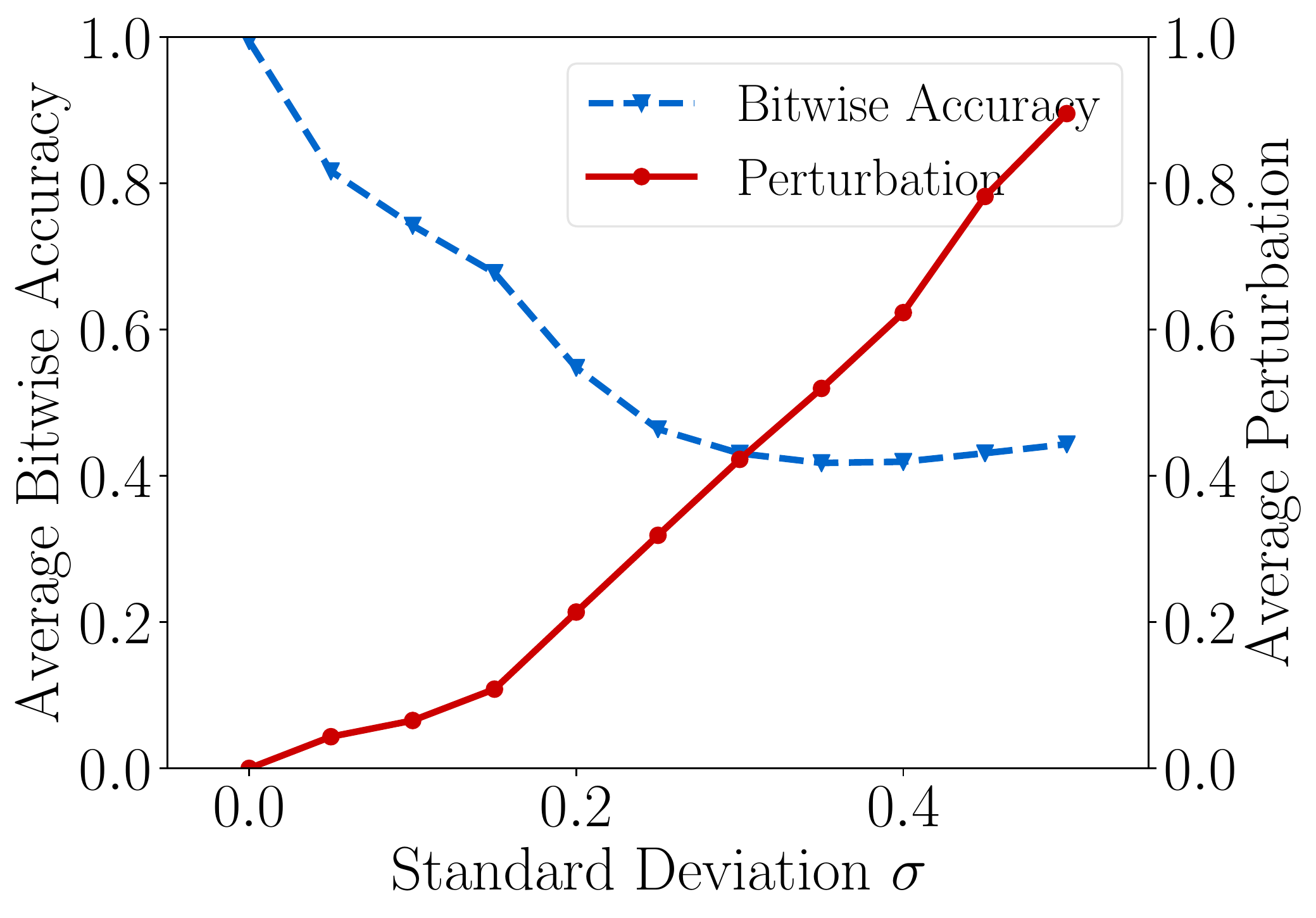}
}%
{
\includegraphics[width=0.24\textwidth]{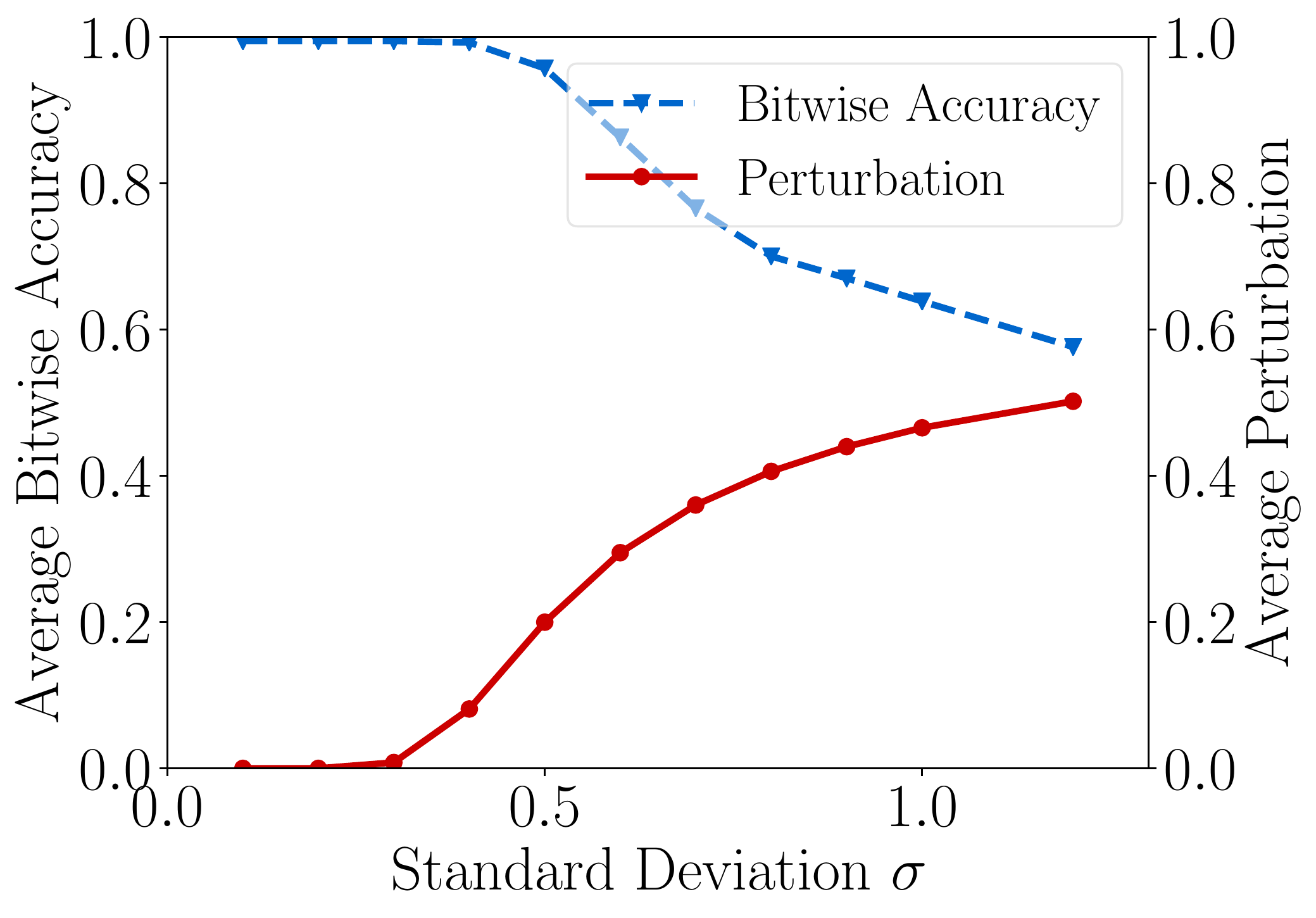}
}%
{
\includegraphics[width=0.24\textwidth]{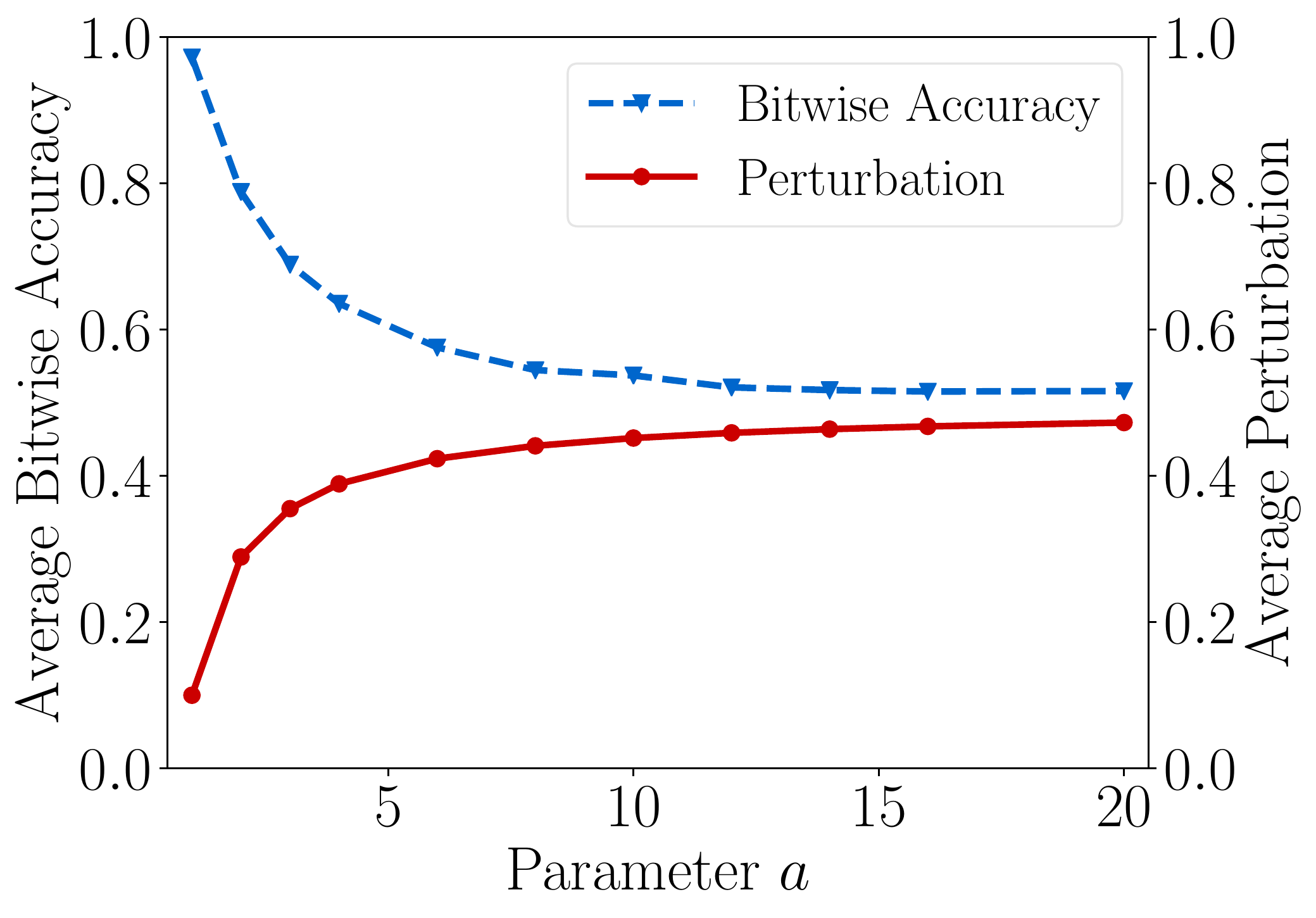}
}%

\subfloat[JPEG]{
\includegraphics[width=0.24\textwidth]{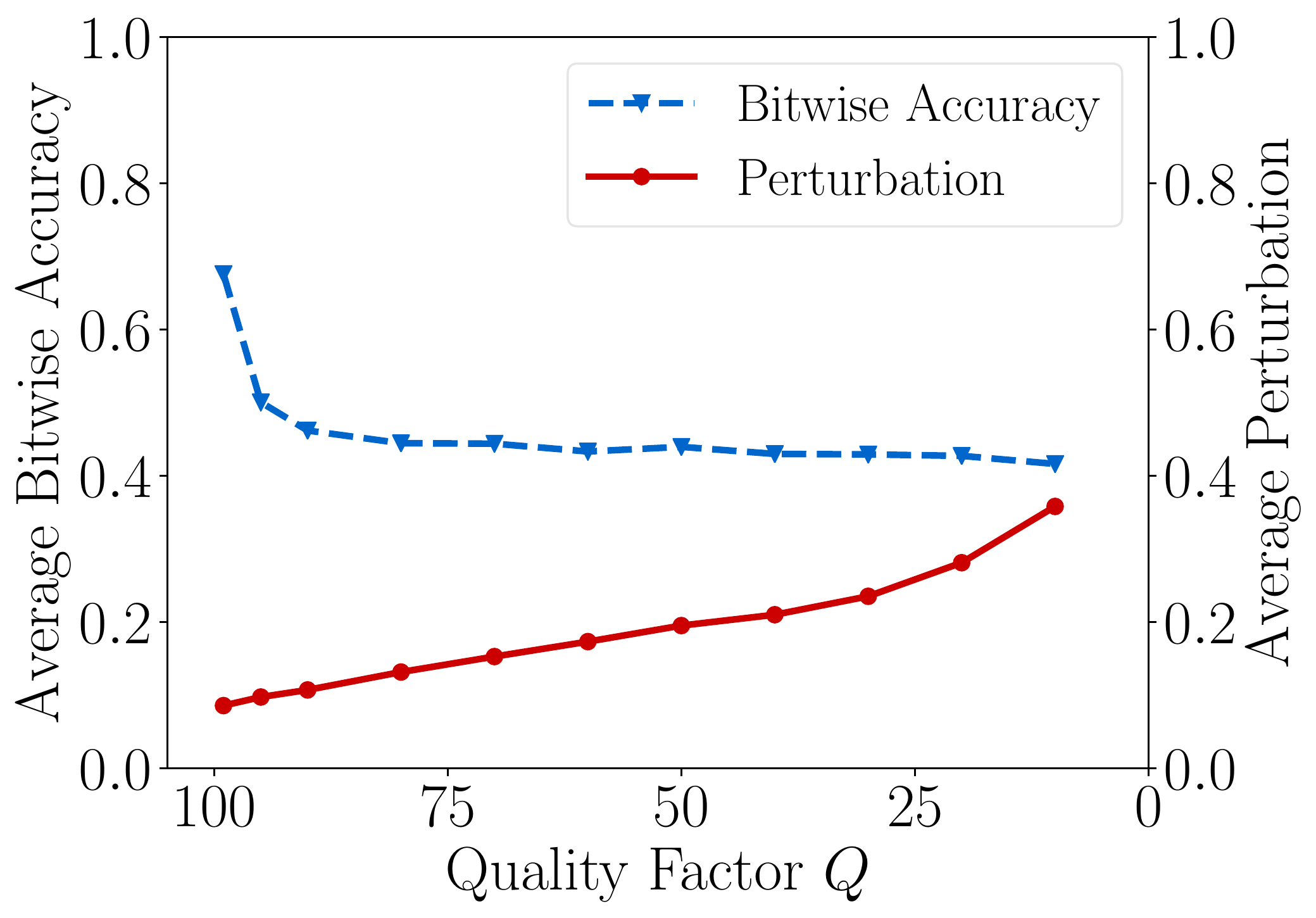}
}%
\subfloat[Gaussian noise]{
\includegraphics[width=0.24\textwidth]{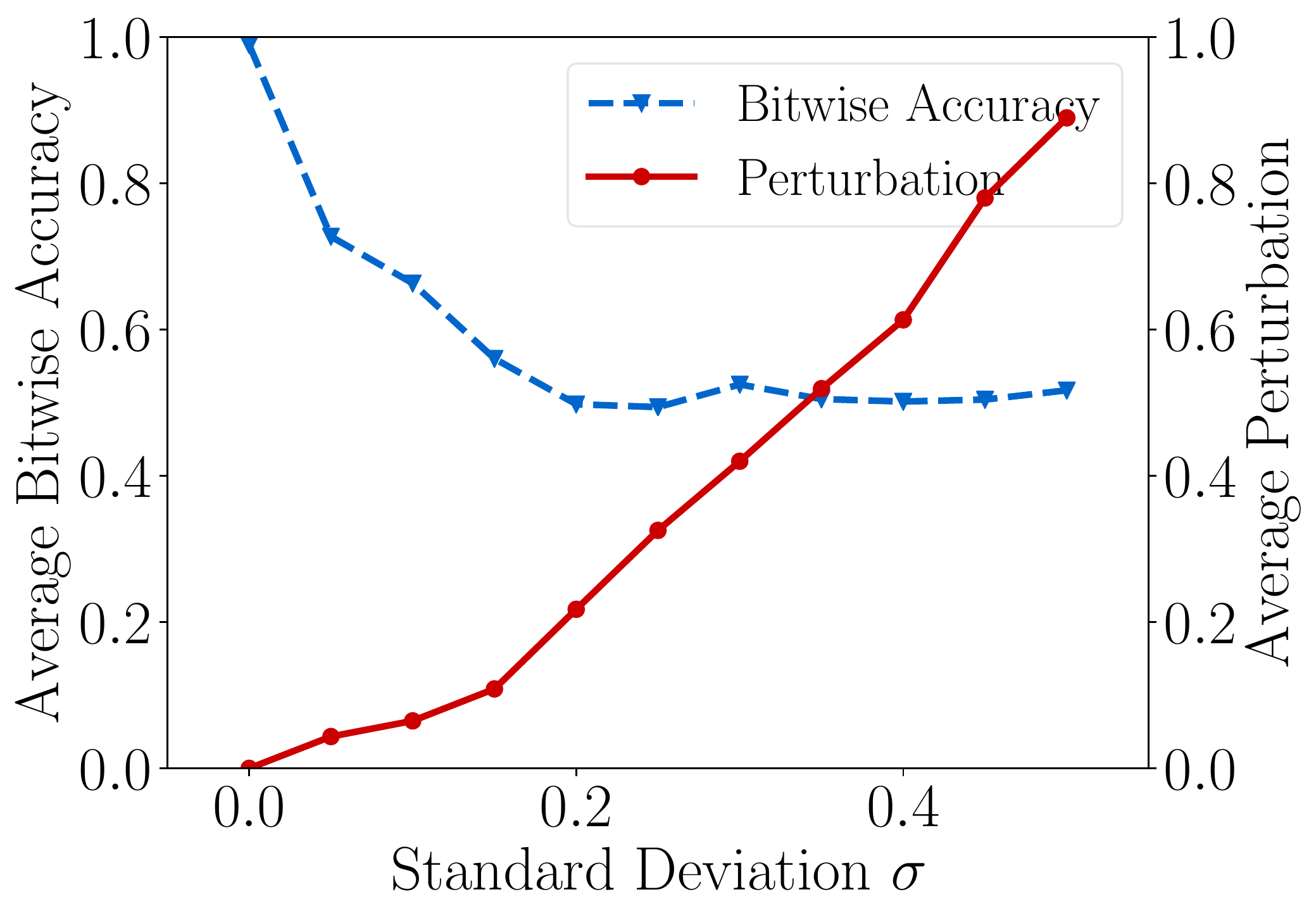}
}%
\subfloat[Gaussian blur]{
\includegraphics[width=0.24\textwidth]{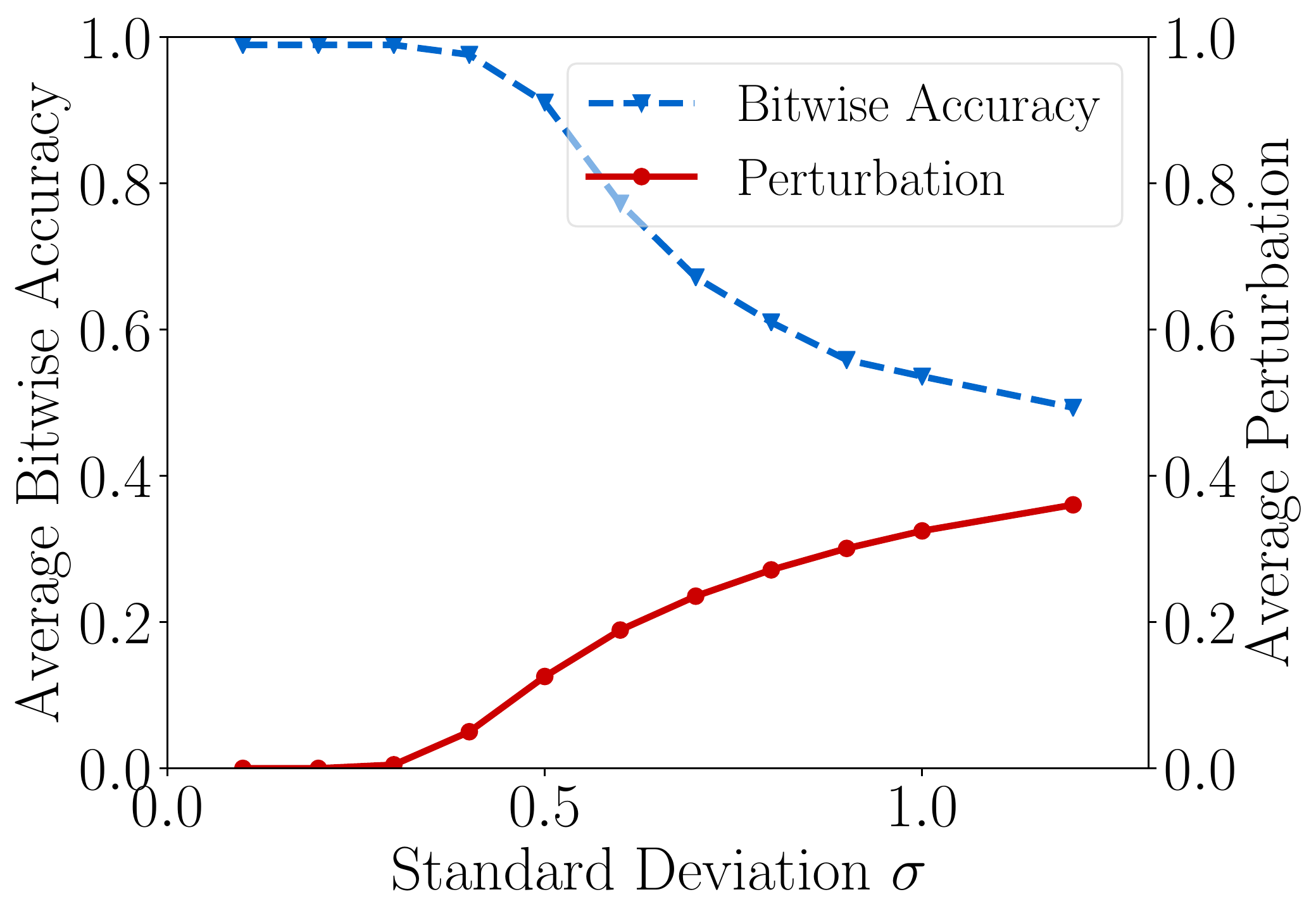}
}%
\subfloat[Brightness/Contrast]{
\includegraphics[width=0.24\textwidth]{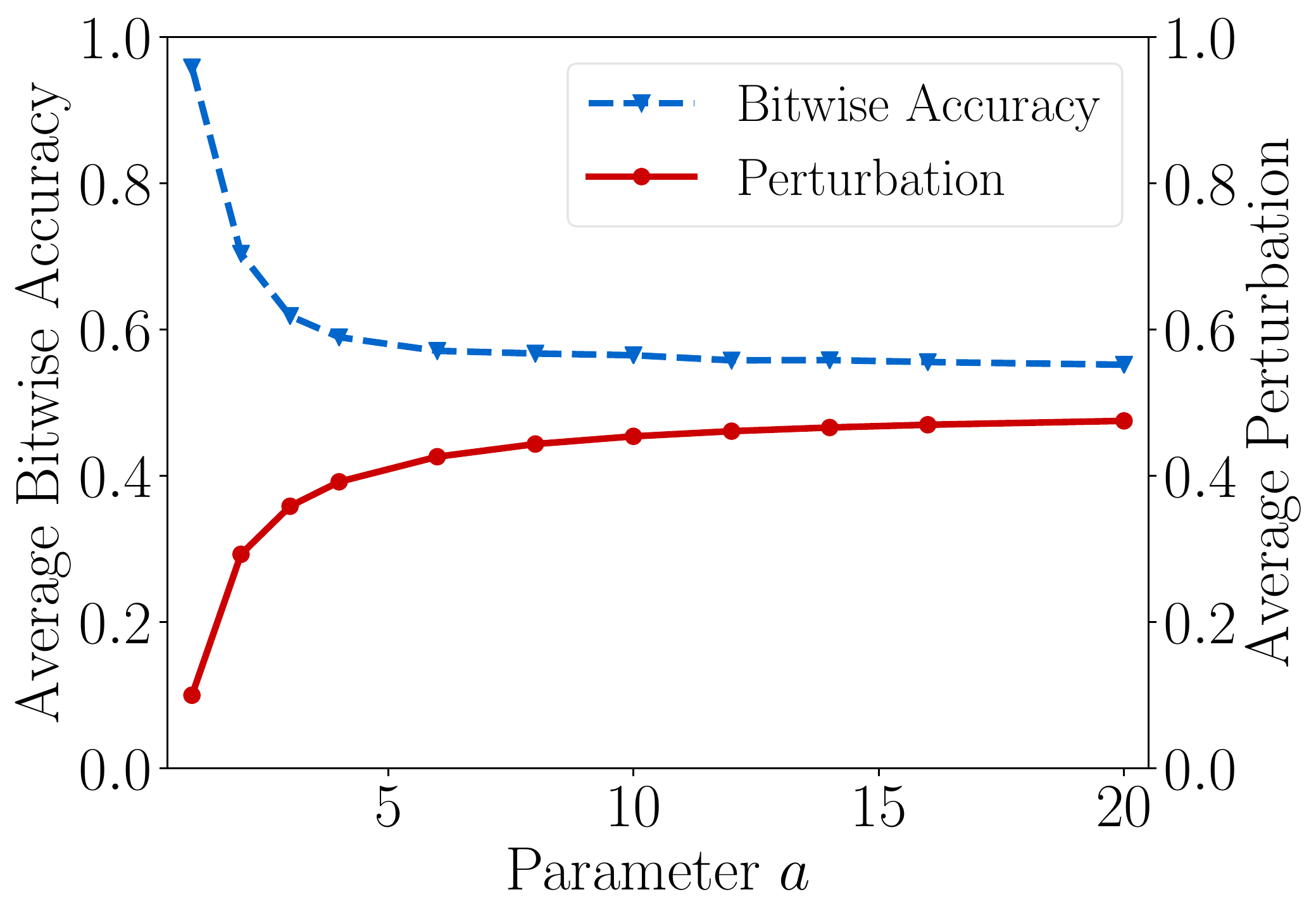}
}%
\caption{Average bitwise accuracy and average perturbation of the post-processed watermarked images  when an existing post-processing method uses different parameter values.  The watermarking method is HiDDeN. The datasets are COCO (first row), ImageNet (second row), and CC (third row).}
\label{hidden-imagenet-cc-parameter}
\end{figure*}

\begin{figure*}[!t]
\centering
{
\includegraphics[width=0.24\textwidth]{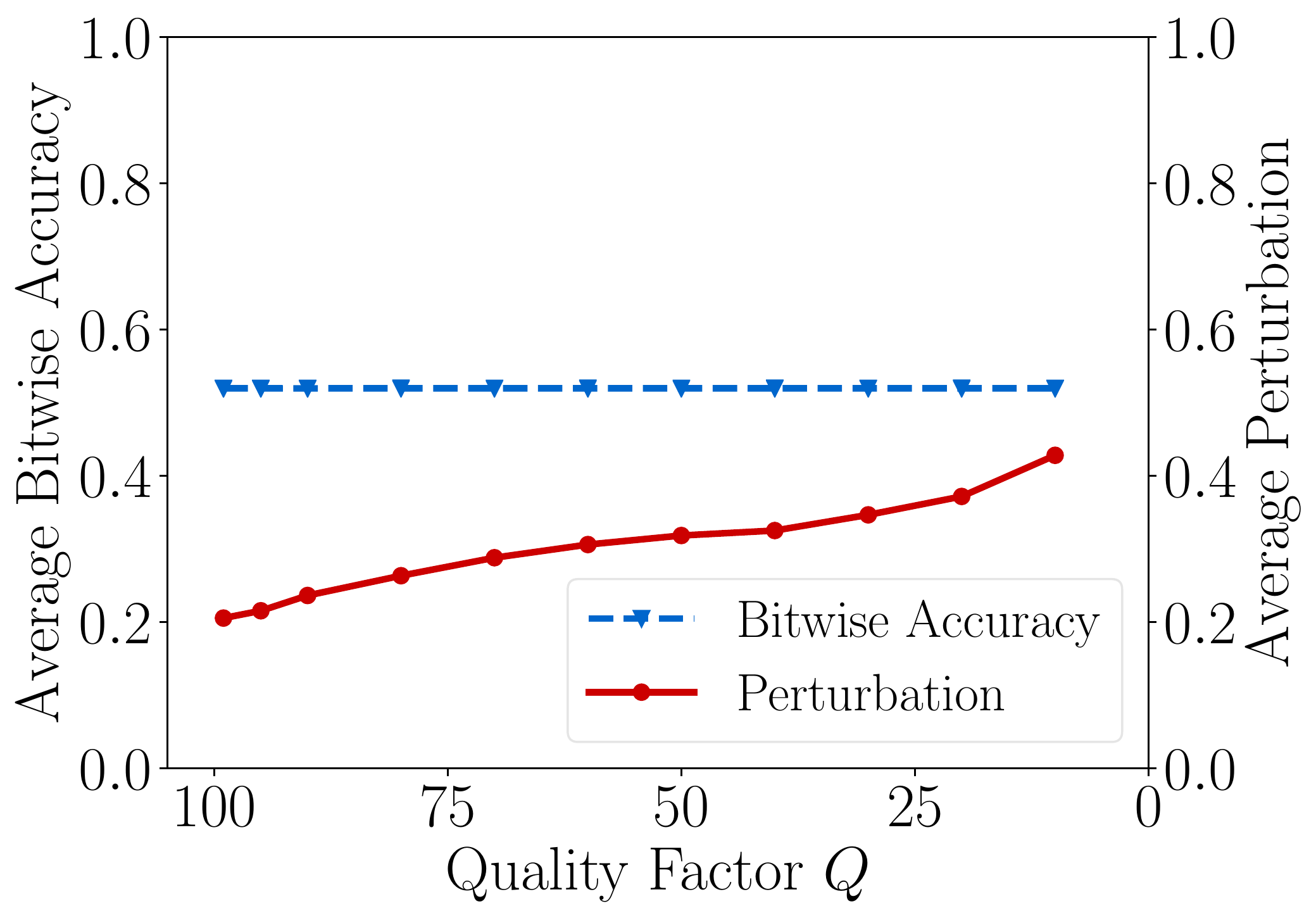}
}%
{
\includegraphics[width=0.24\textwidth]{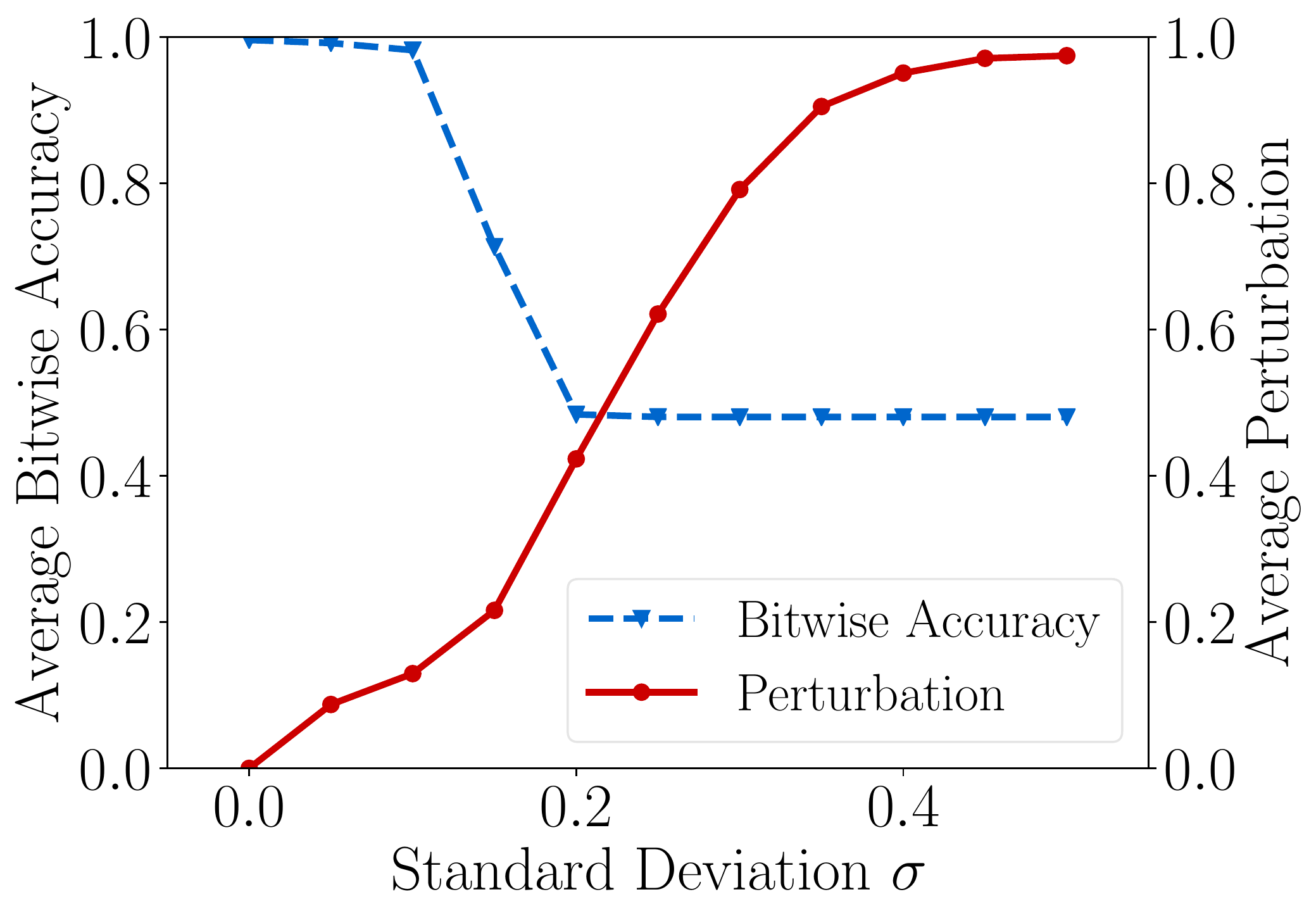}
}%
{
\includegraphics[width=0.24\textwidth]{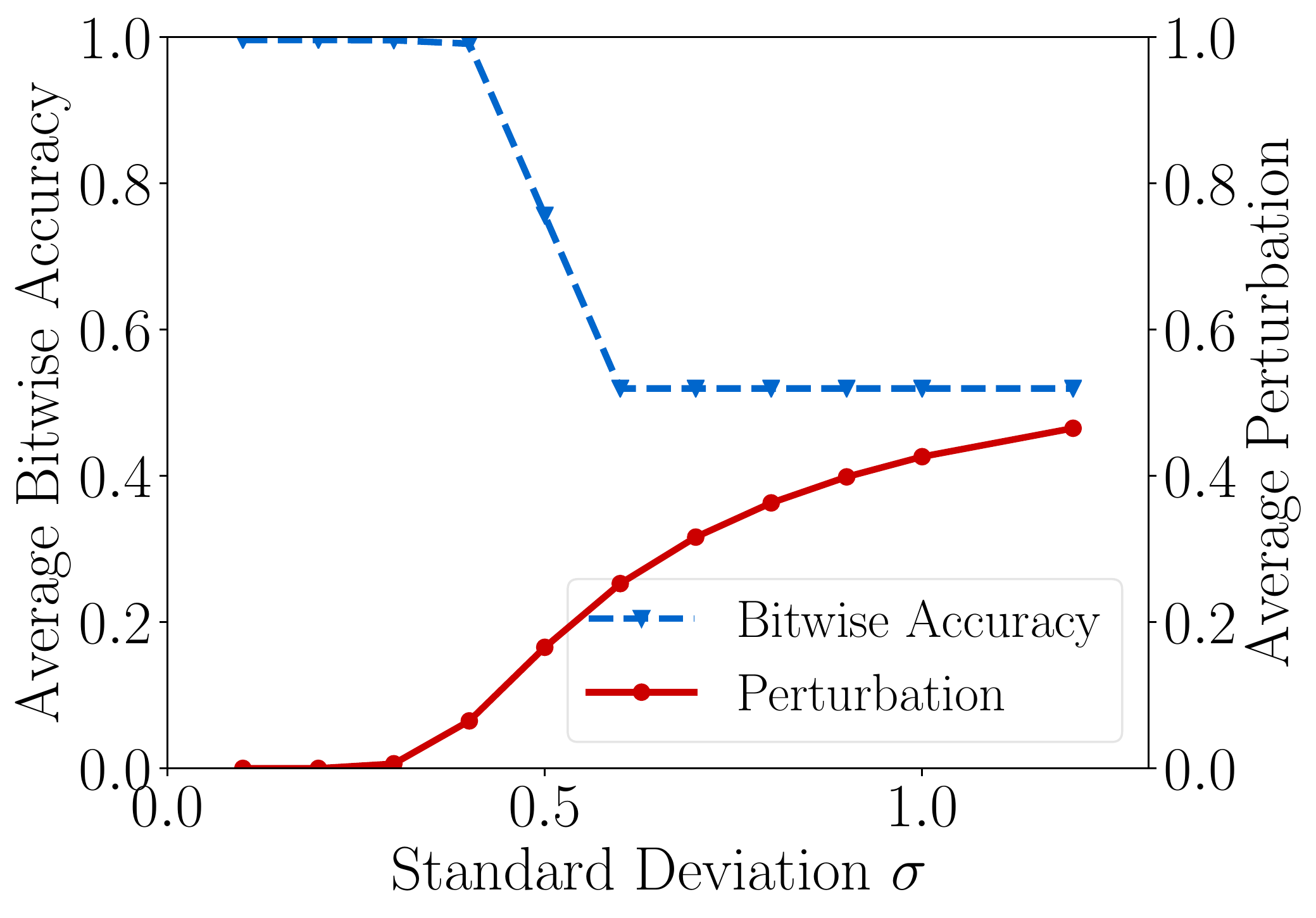}
}%
{
\includegraphics[width=0.24\textwidth]{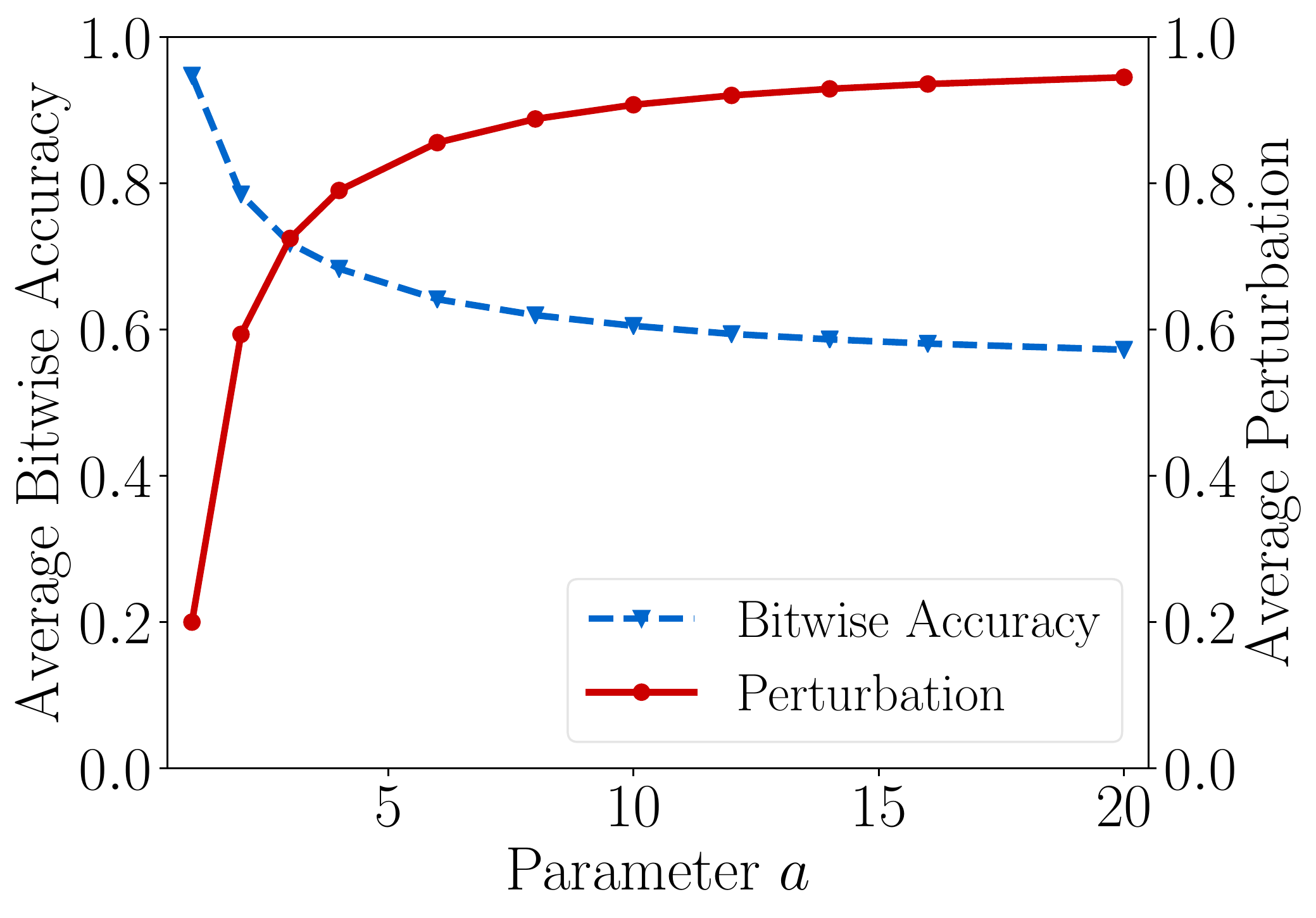}
}%

{
\includegraphics[width=0.24\textwidth]{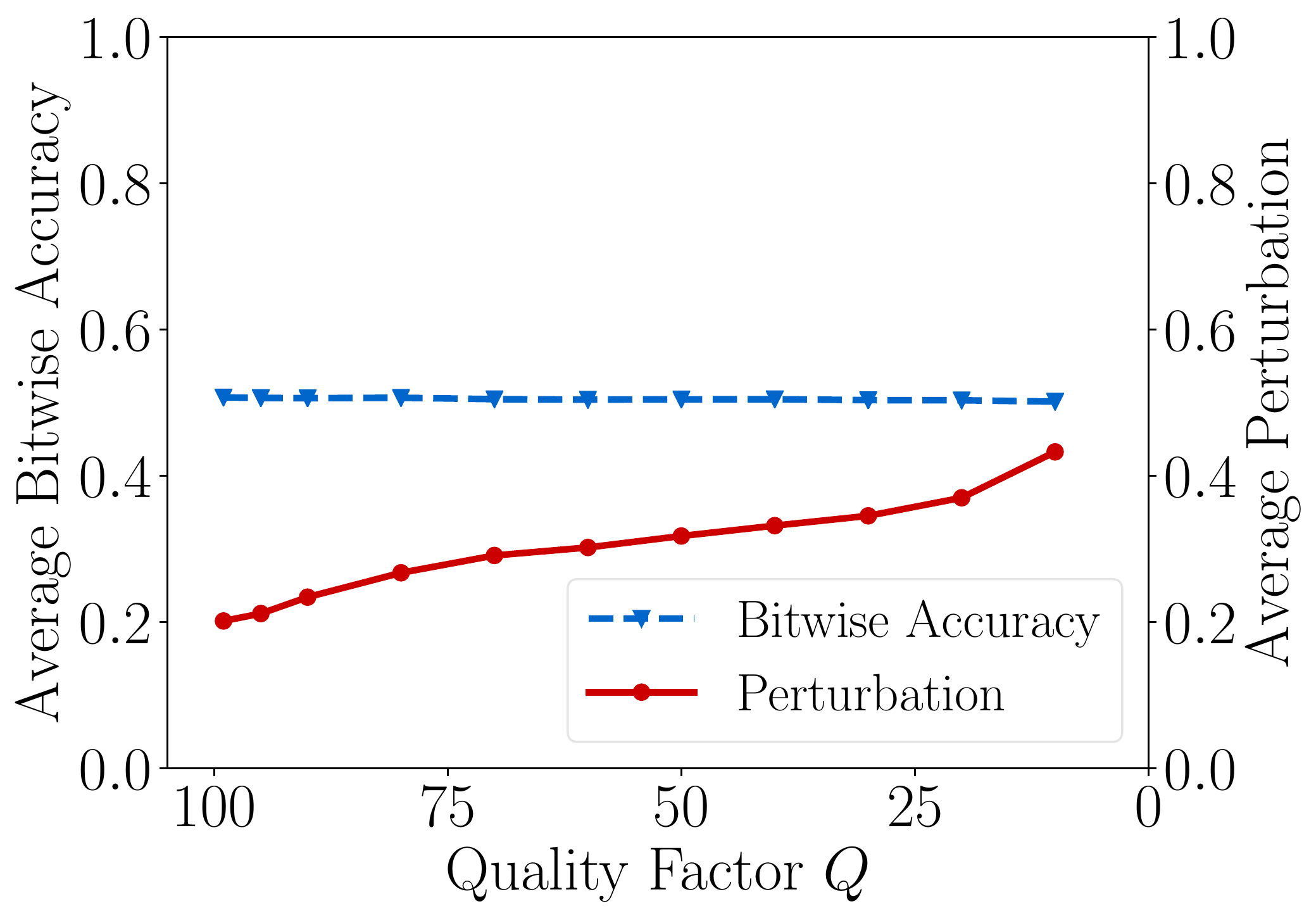}
}%
{
\includegraphics[width=0.24\textwidth]{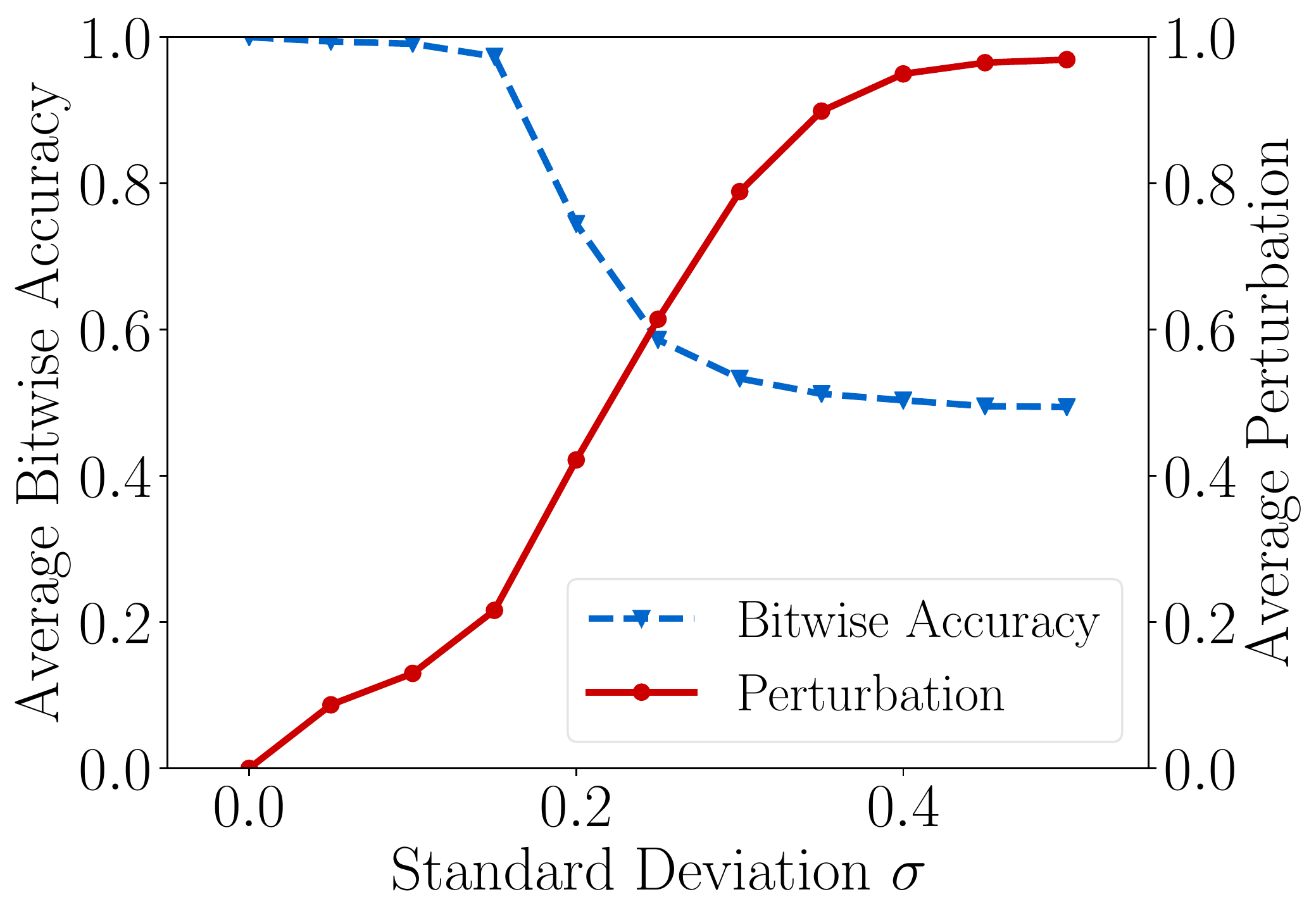}
}%
{
\includegraphics[width=0.24\textwidth]{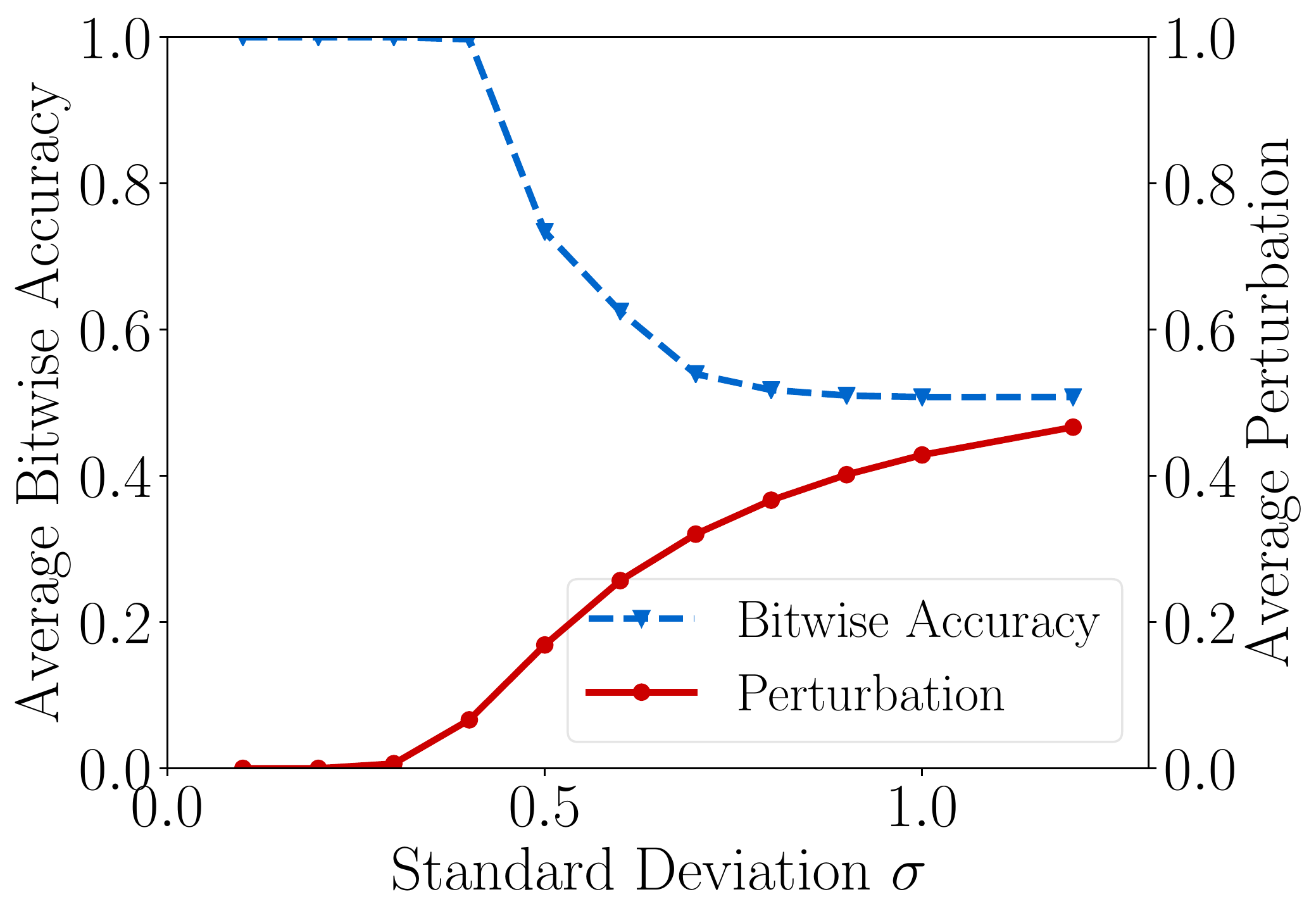}
}%
{
\includegraphics[width=0.24\textwidth]{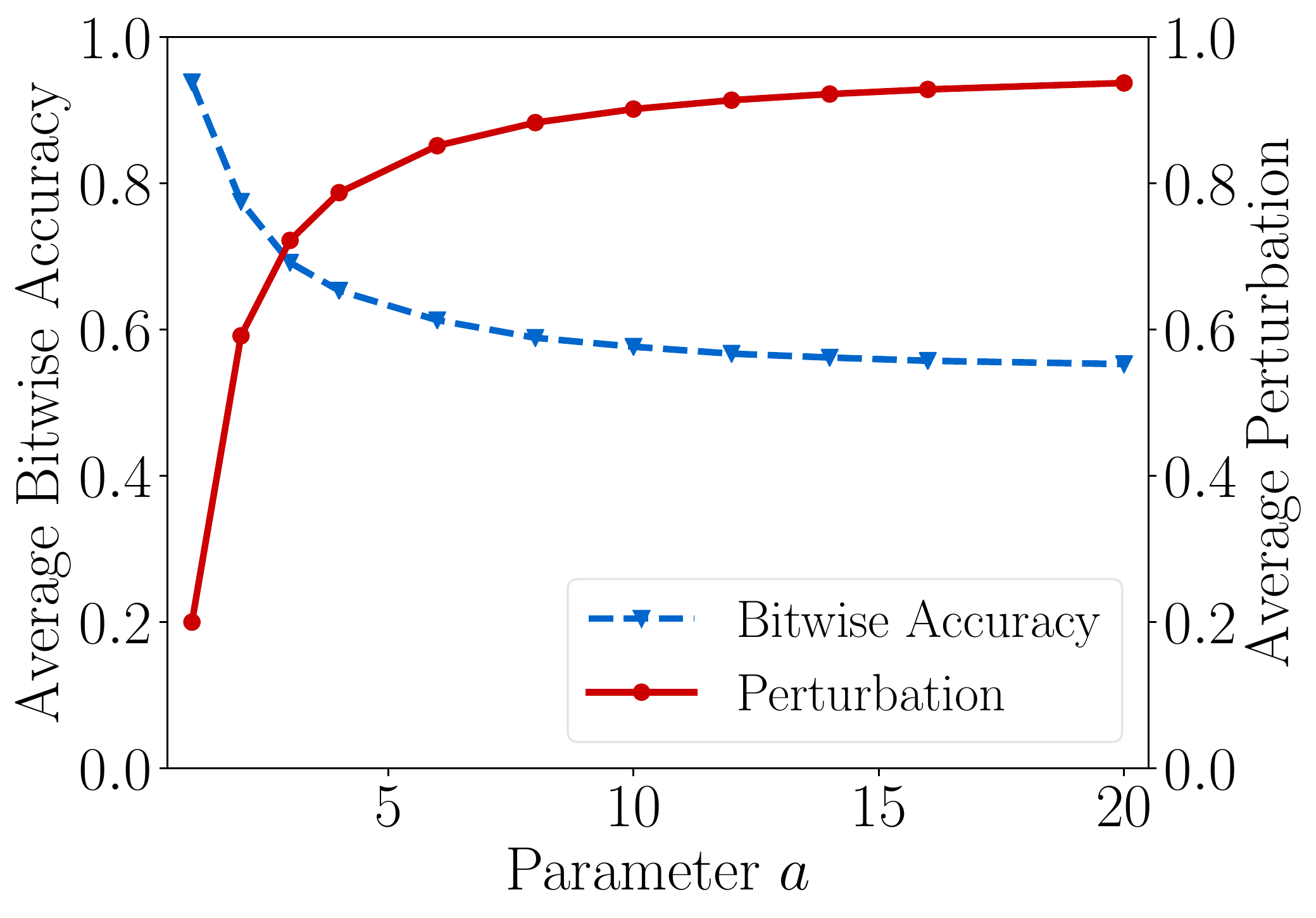}
}%

\subfloat[JPEG]{
\includegraphics[width=0.24\textwidth]{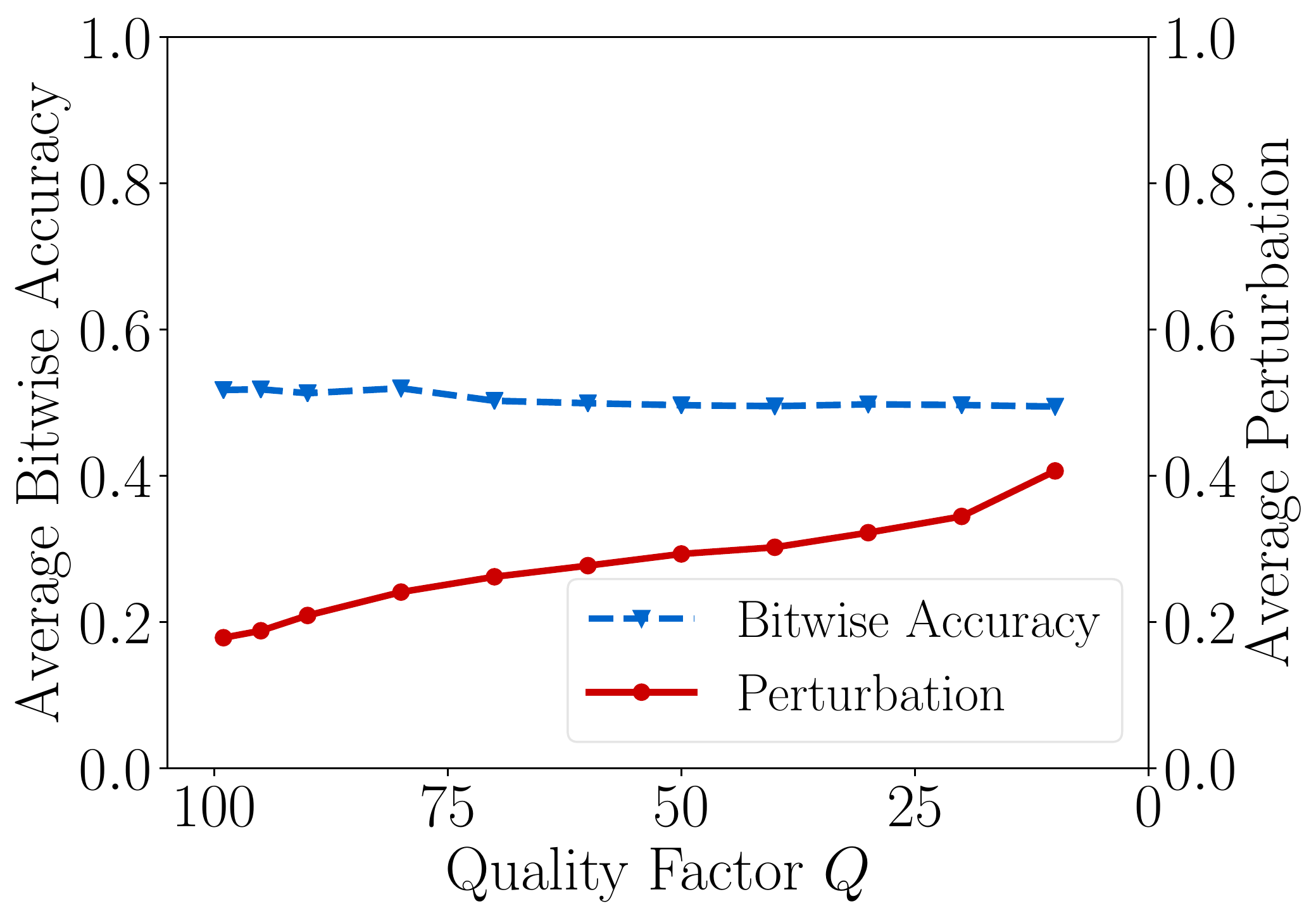}
}%
\subfloat[Gaussian noise]{
\includegraphics[width=0.24\textwidth]{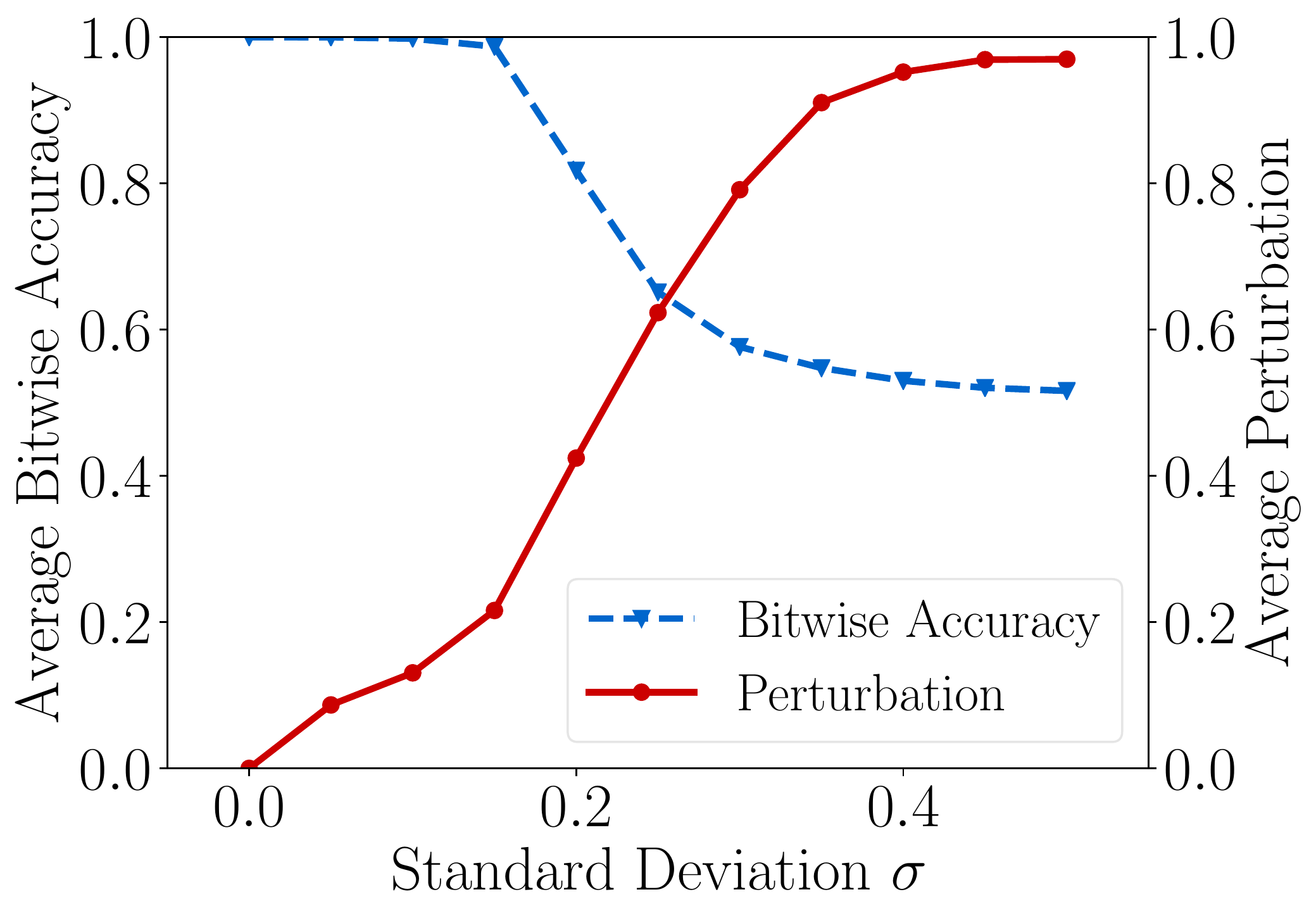}
}%
\subfloat[Gaussian blur]{
\includegraphics[width=0.24\textwidth]{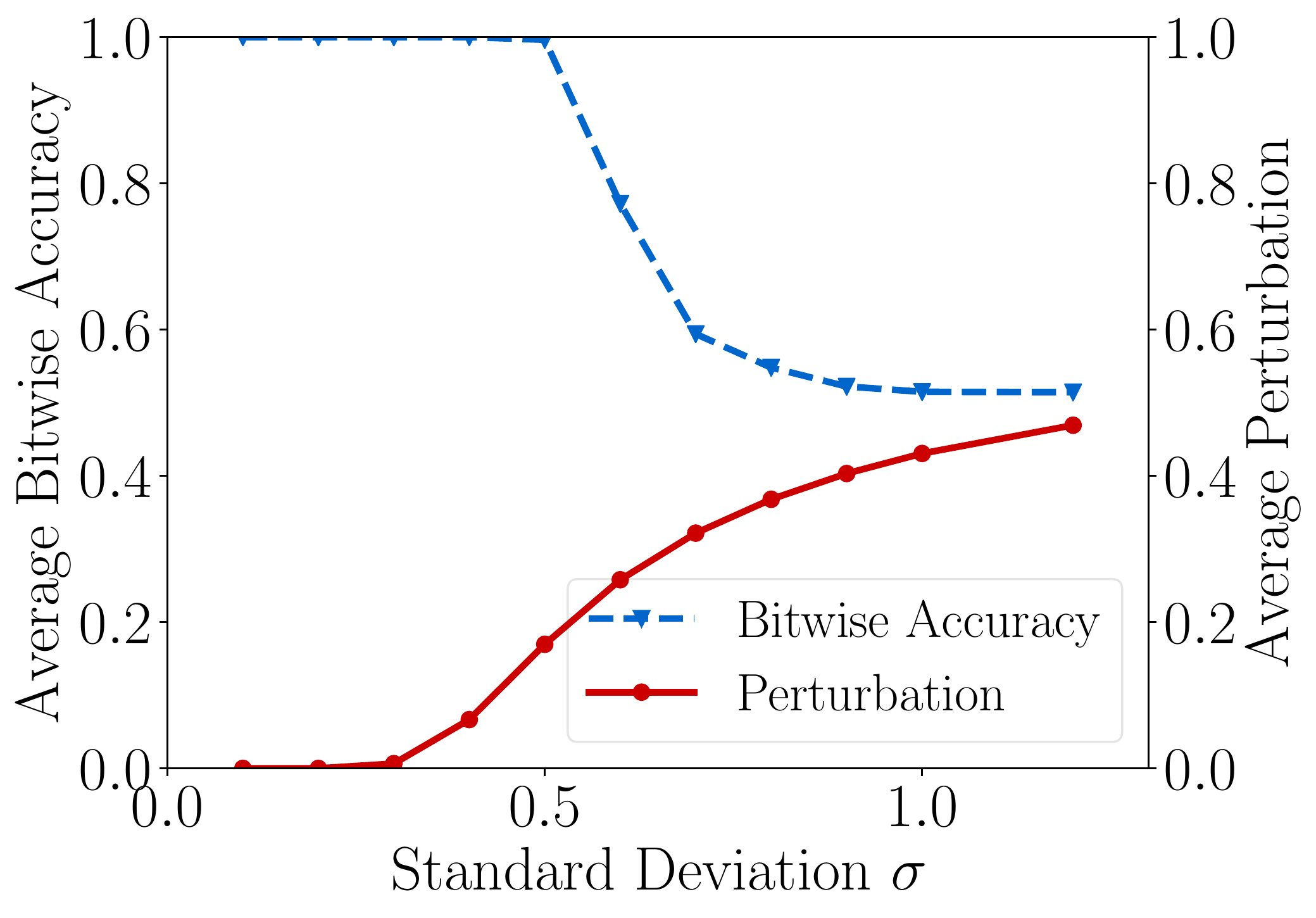}
}%
\subfloat[Brightness/Contrast]{
\includegraphics[width=0.24\textwidth]{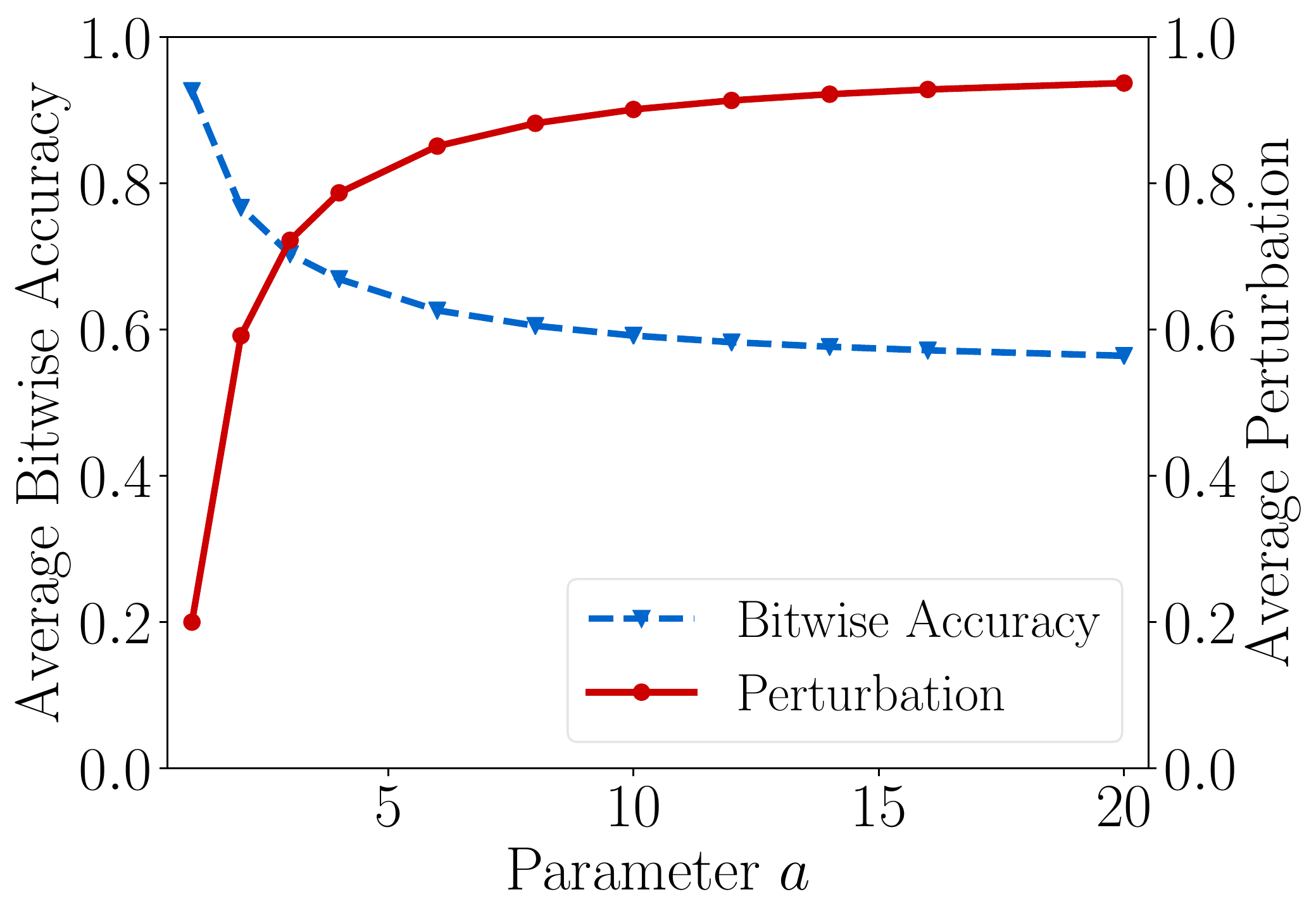}
}%
\caption{Average bitwise accuracy and average perturbation of the post-processed watermarked images  when an existing post-processing method uses different parameter values.  The watermarking method is UDH. The datasets are COCO (first row),  ImageNet (second row), and CC (third row). }
\label{udh-standard-parameter}
\end{figure*}

\begin{figure*}[!t]
\centering
\subfloat[COCO]{\includegraphics[width=0.33\textwidth]{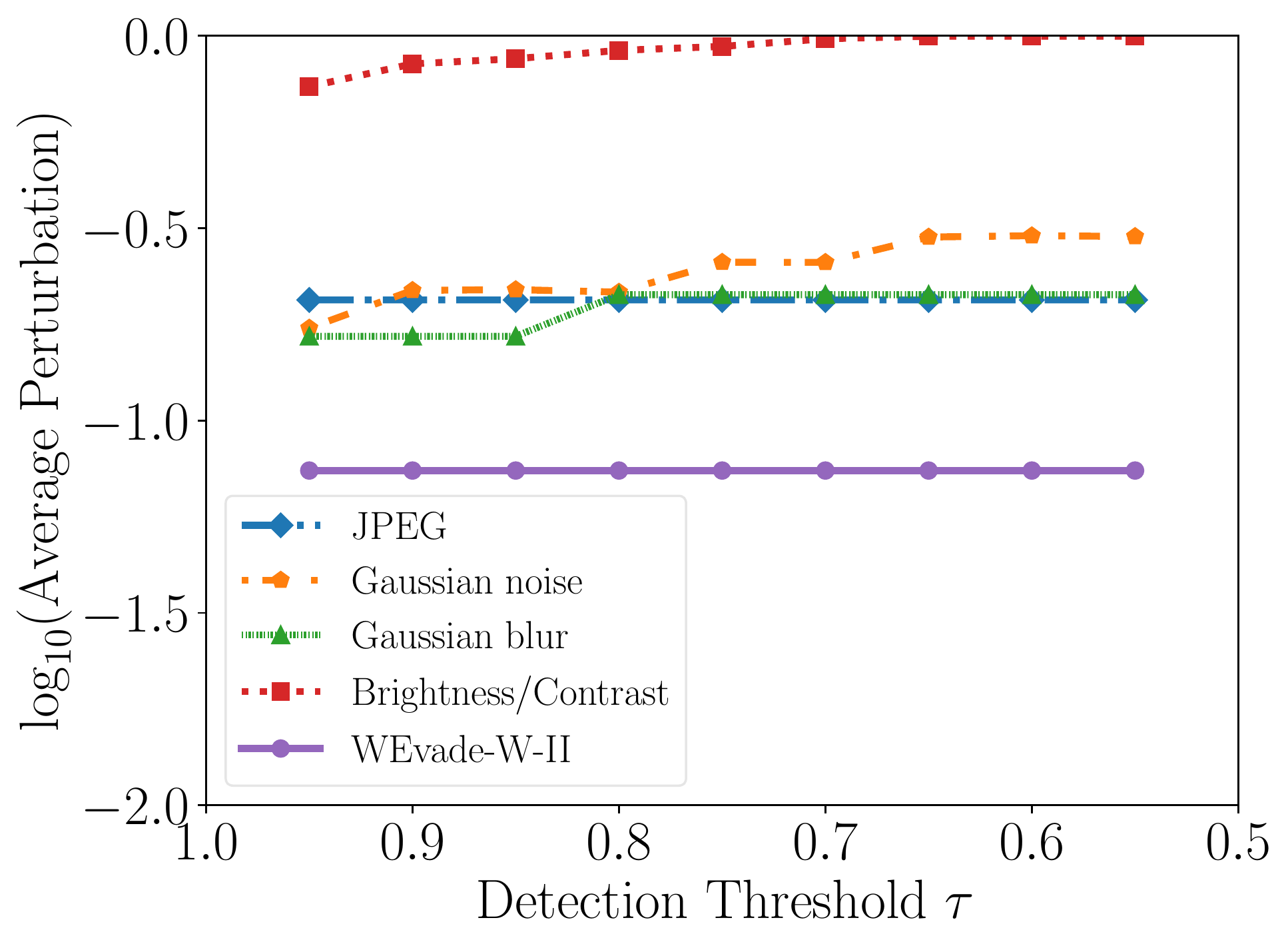}}
\subfloat[ImageNet]{\includegraphics[width=0.33\textwidth]{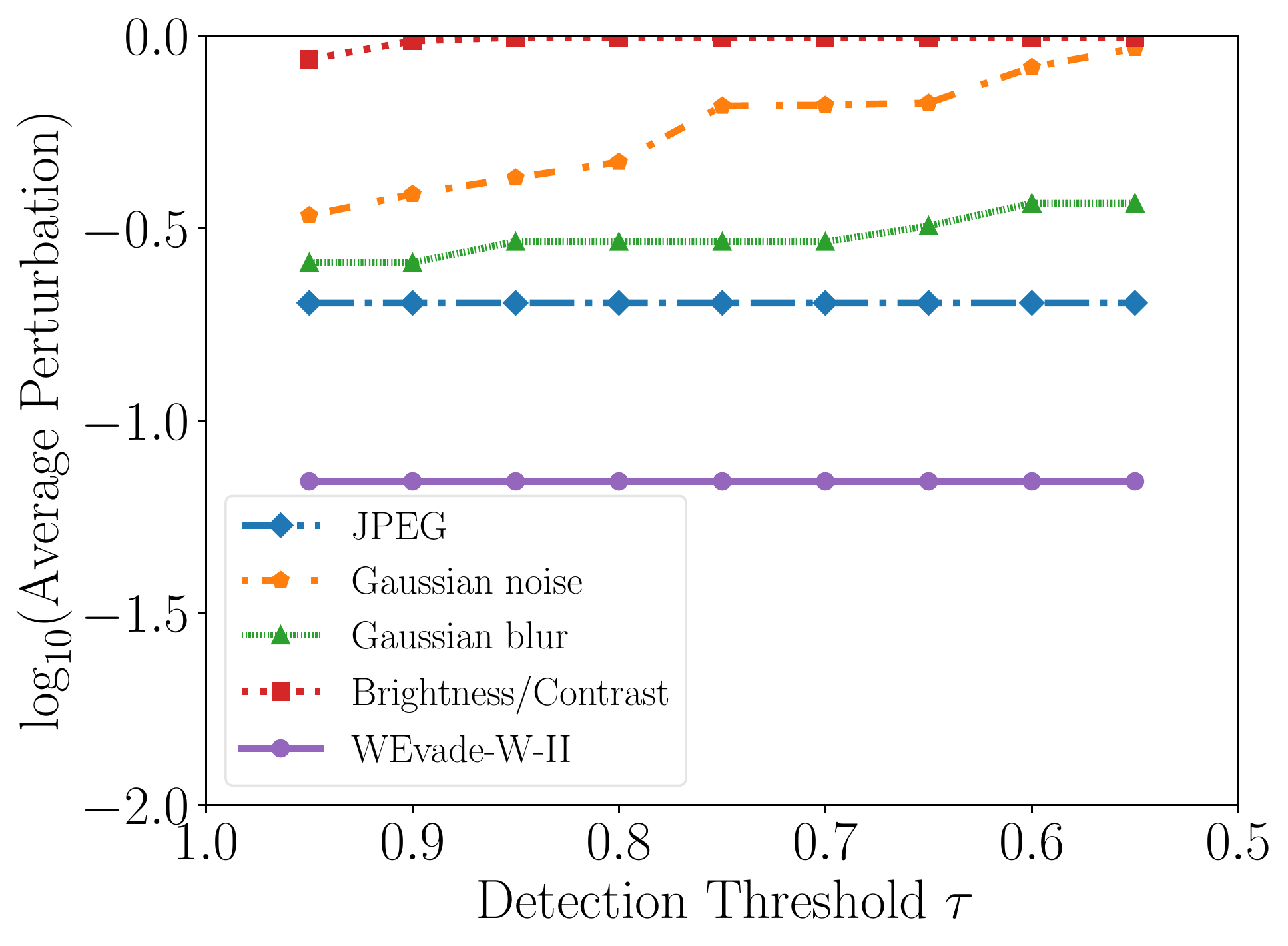}}
\subfloat[CC]{\includegraphics[width=0.33\textwidth]{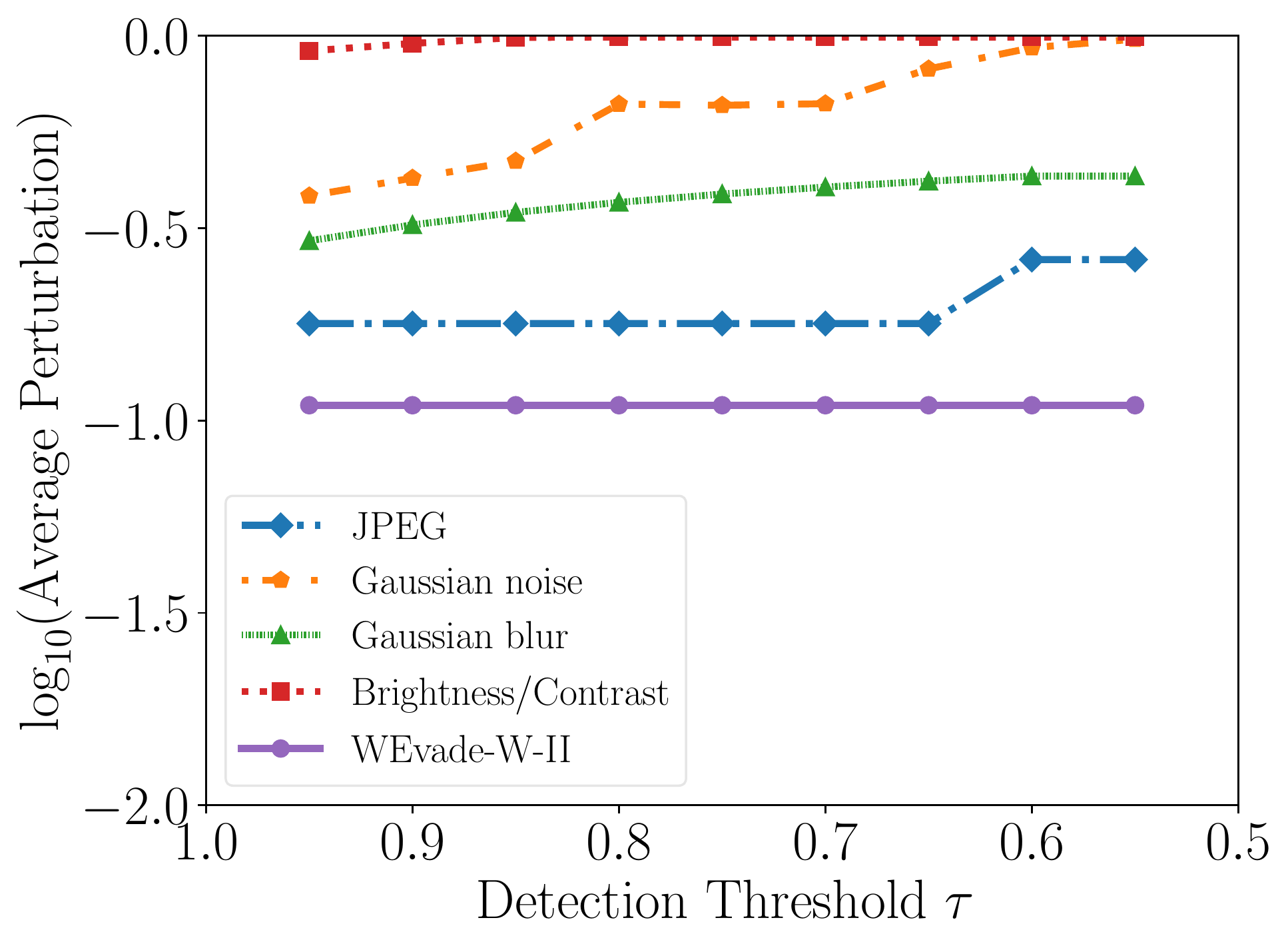}}
\caption{Average perturbation added by each post-processing method to evade the double-tail detector with different threshold $\tau$ in the white-box setting. We set the parameters of existing post-processing methods such that they achieve the same evasion rate as our \algns-W-II. The watermarking method is UDH.}
\label{white-standard-perturbation-udh}
\end{figure*}

\begin{figure}[!t]
\centering
{\includegraphics[width=0.35\textwidth]{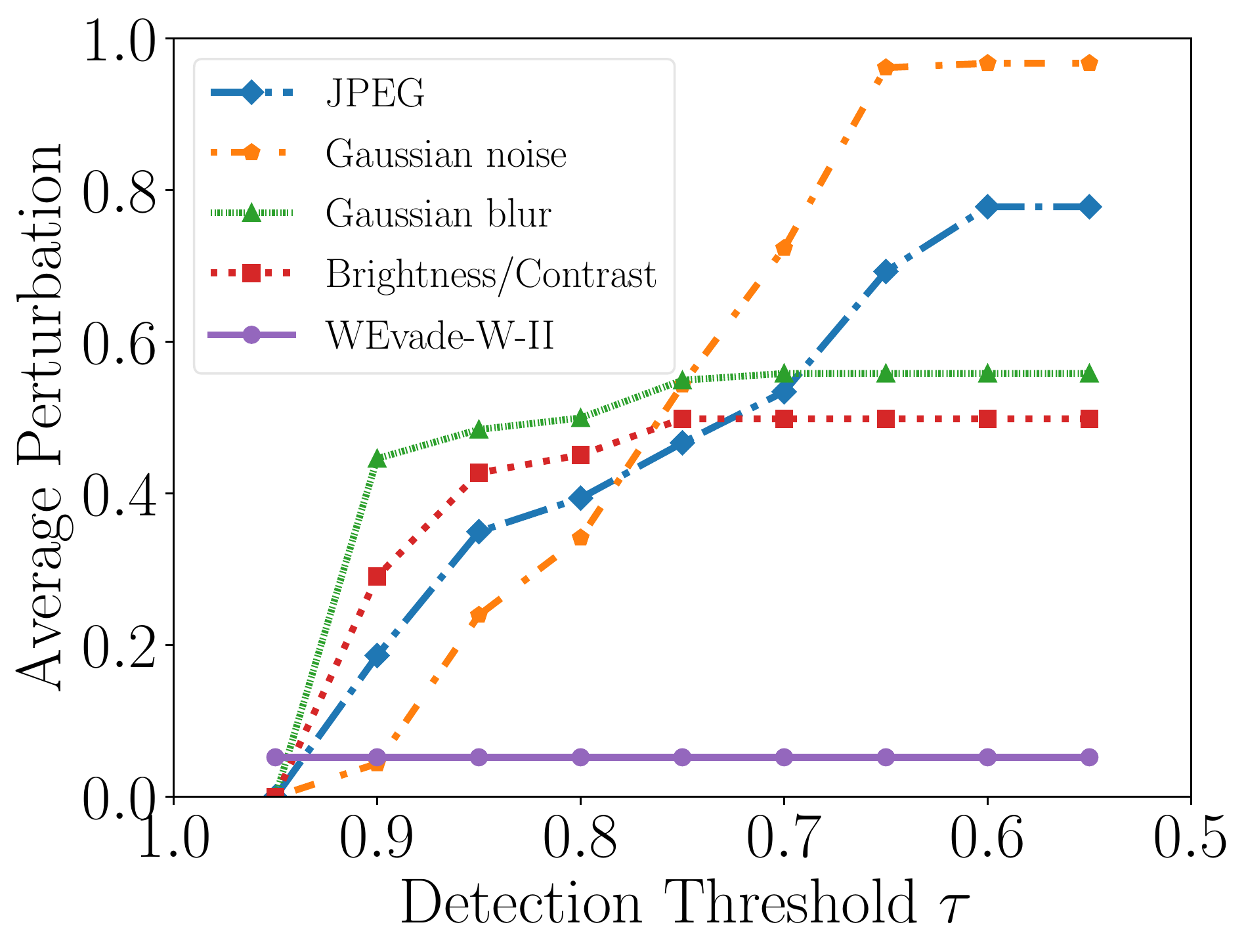}}
\caption{Average perturbation added by each post-processing method to evade the double-tail detector with different threshold $\tau$ for the COCO dataset. We set the parameters of existing post-processing methods such that they achieve the same evasion rate as \algns-W-II. The watermarking method is HiDDeN and adversarial training is used. After adversarial training, the average bitwise accuracy is around 0.87. When $\tau$ is 0.95, empirical FNR is 99.6\%, and thus existing post-processing methods do not add perturbations to a large fraction of watermarked images based on how we evaluate them, leading to 0 perturbations. However, they need much larger perturbations  when $\tau$ is smaller than 0.9.}
\label{adversarial-hidden-coco}
\end{figure}

\begin{figure}[!t]
\centering
{\includegraphics[width=0.15\textwidth]{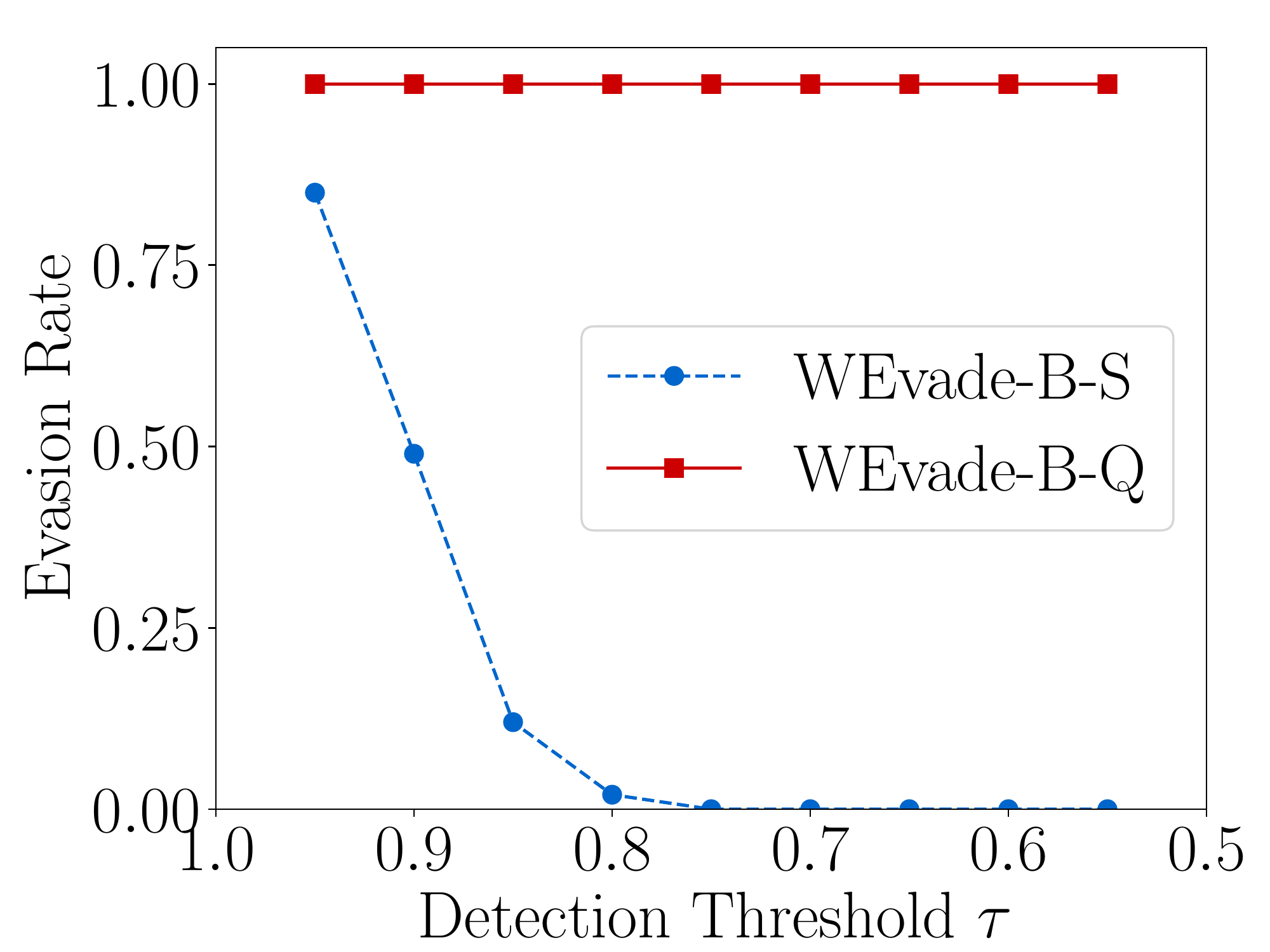}}\hspace{0.2mm}
{\includegraphics[width=0.15\textwidth]{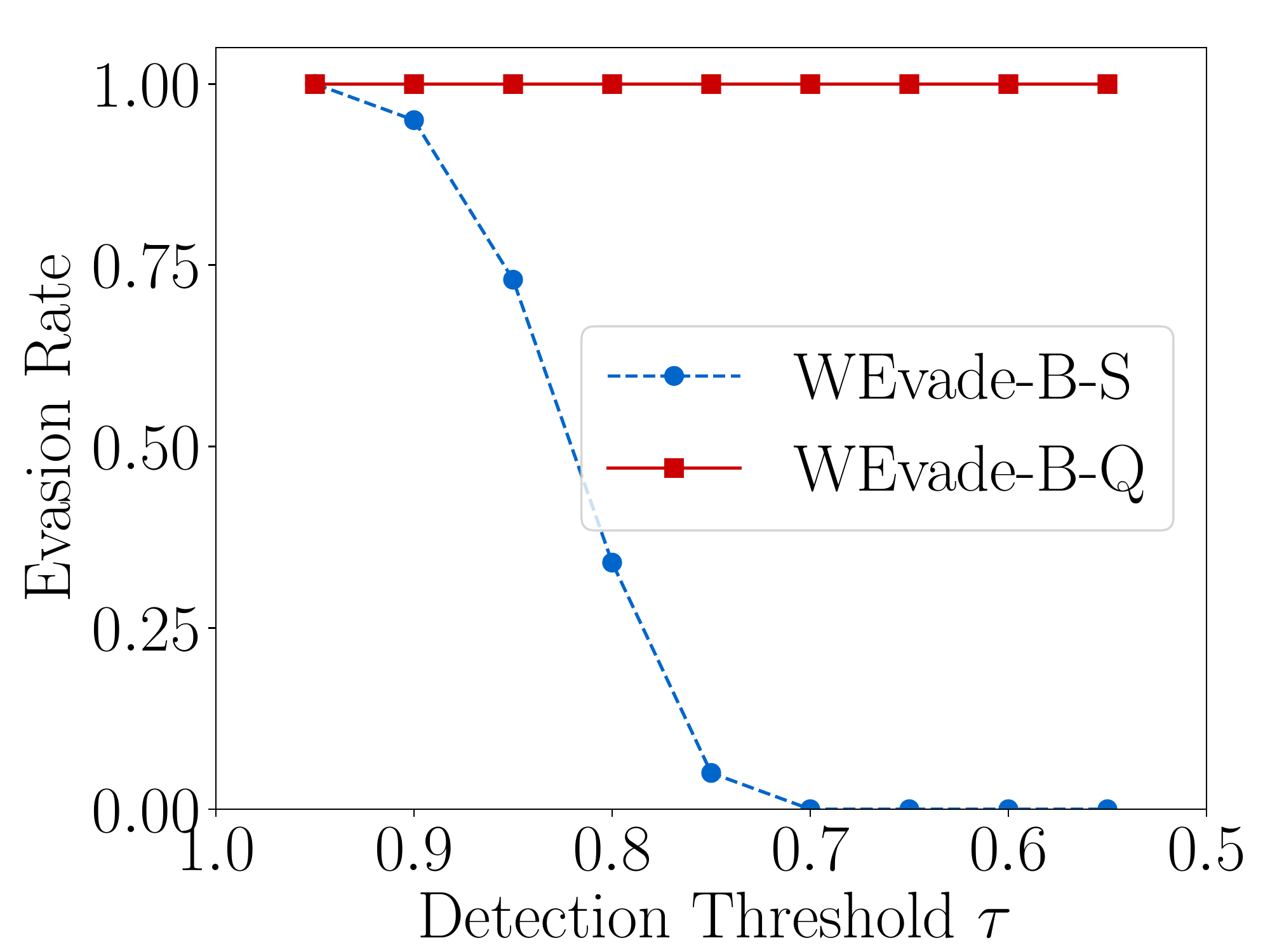}}\hspace{0.2mm}
{\includegraphics[width=0.15\textwidth]{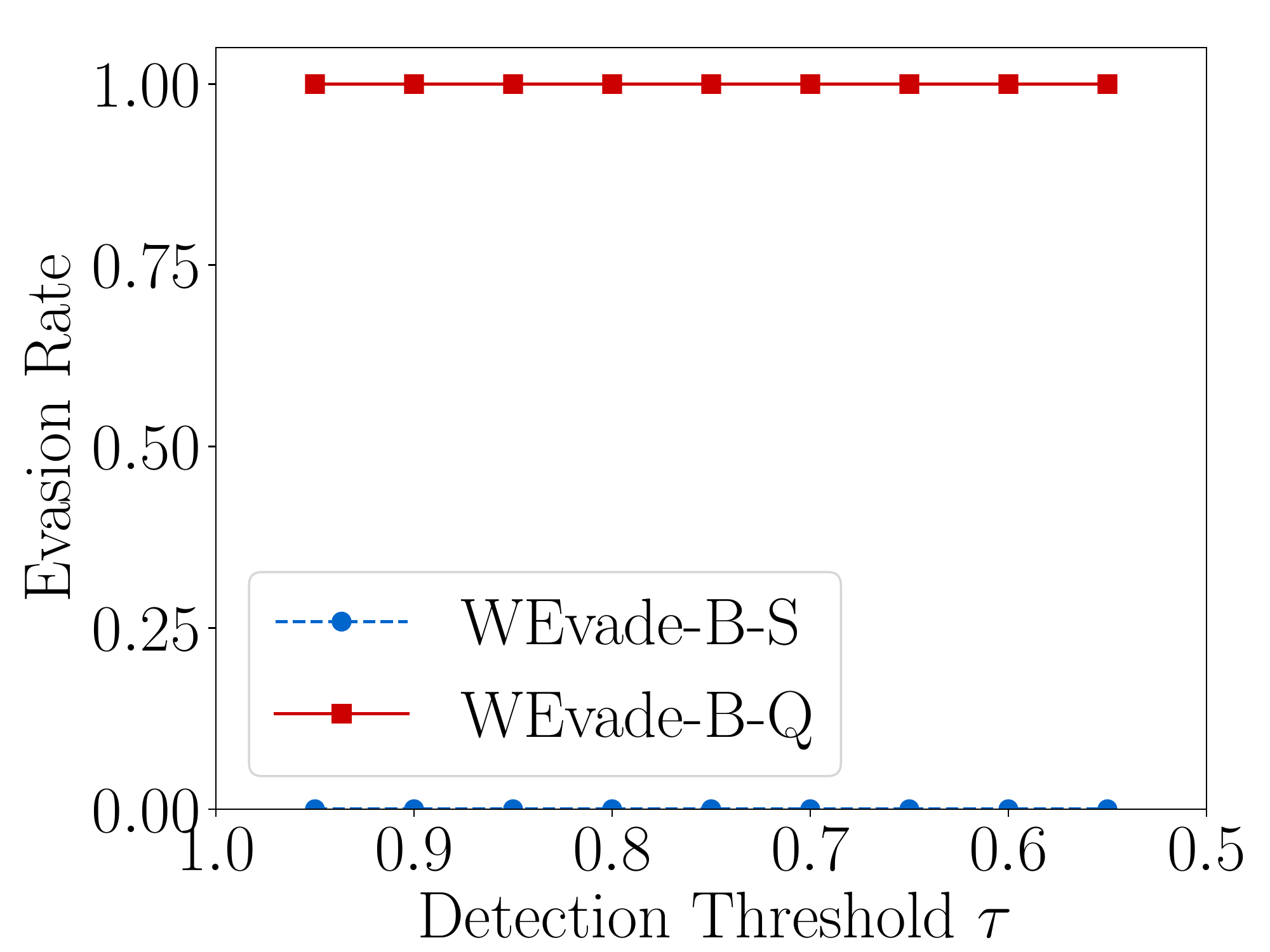}}
\subfloat[COCO]{\includegraphics[width=0.15\textwidth]{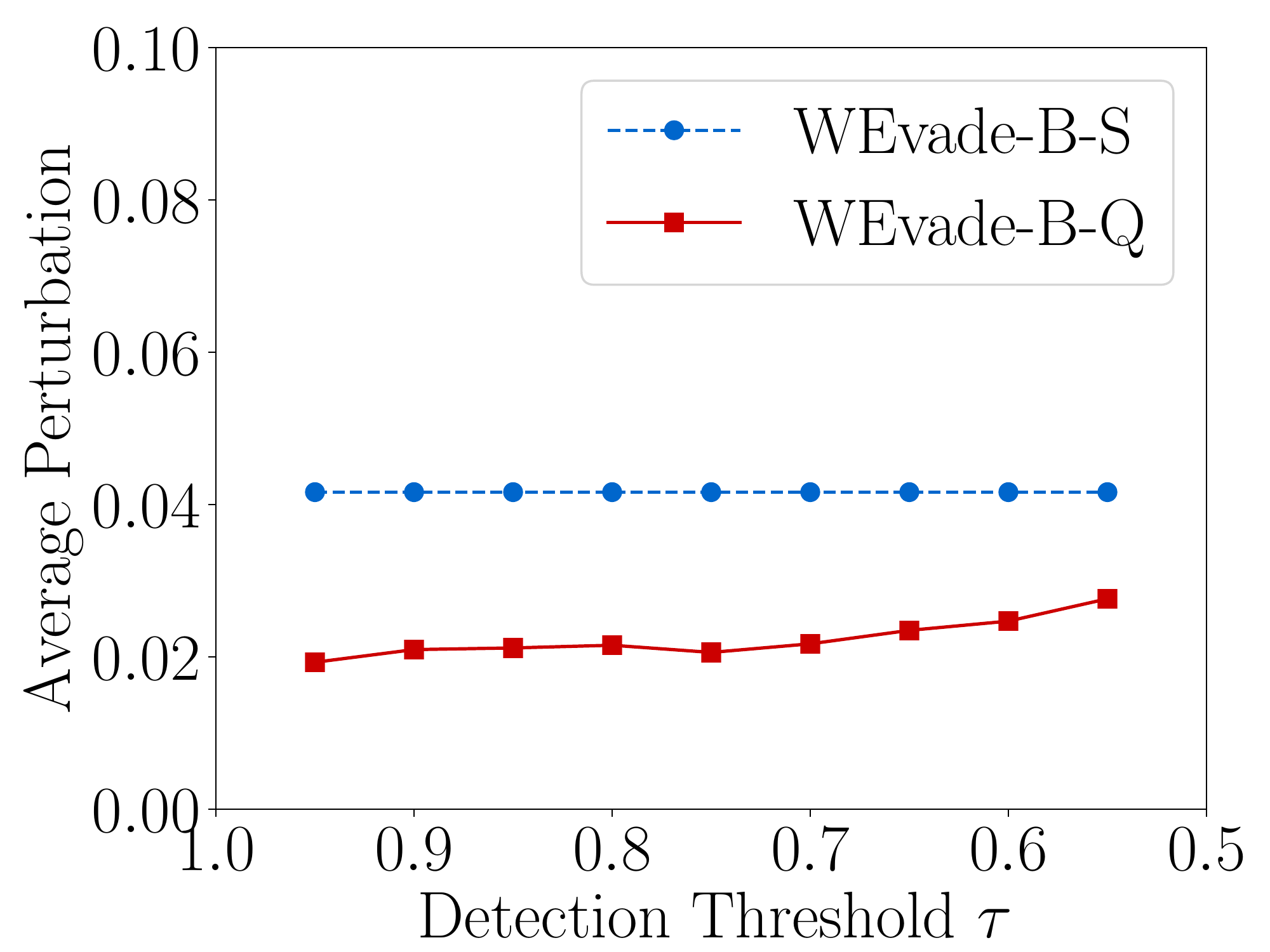}}\hspace{0.2mm}
\subfloat[ImageNet]{\includegraphics[width=0.15\textwidth]{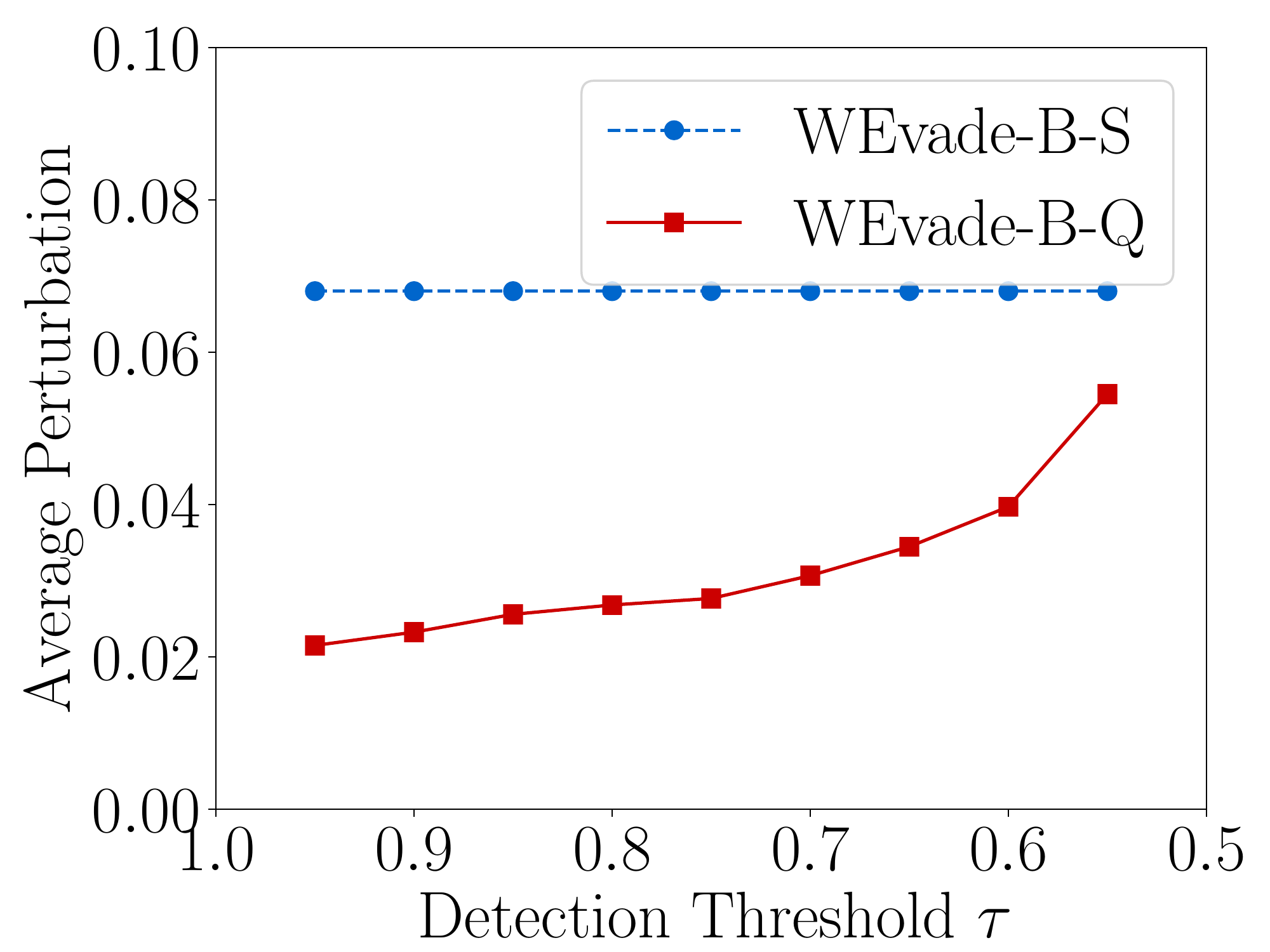}}\hspace{0.2mm}
\subfloat[CC]{\includegraphics[width=0.15\textwidth]{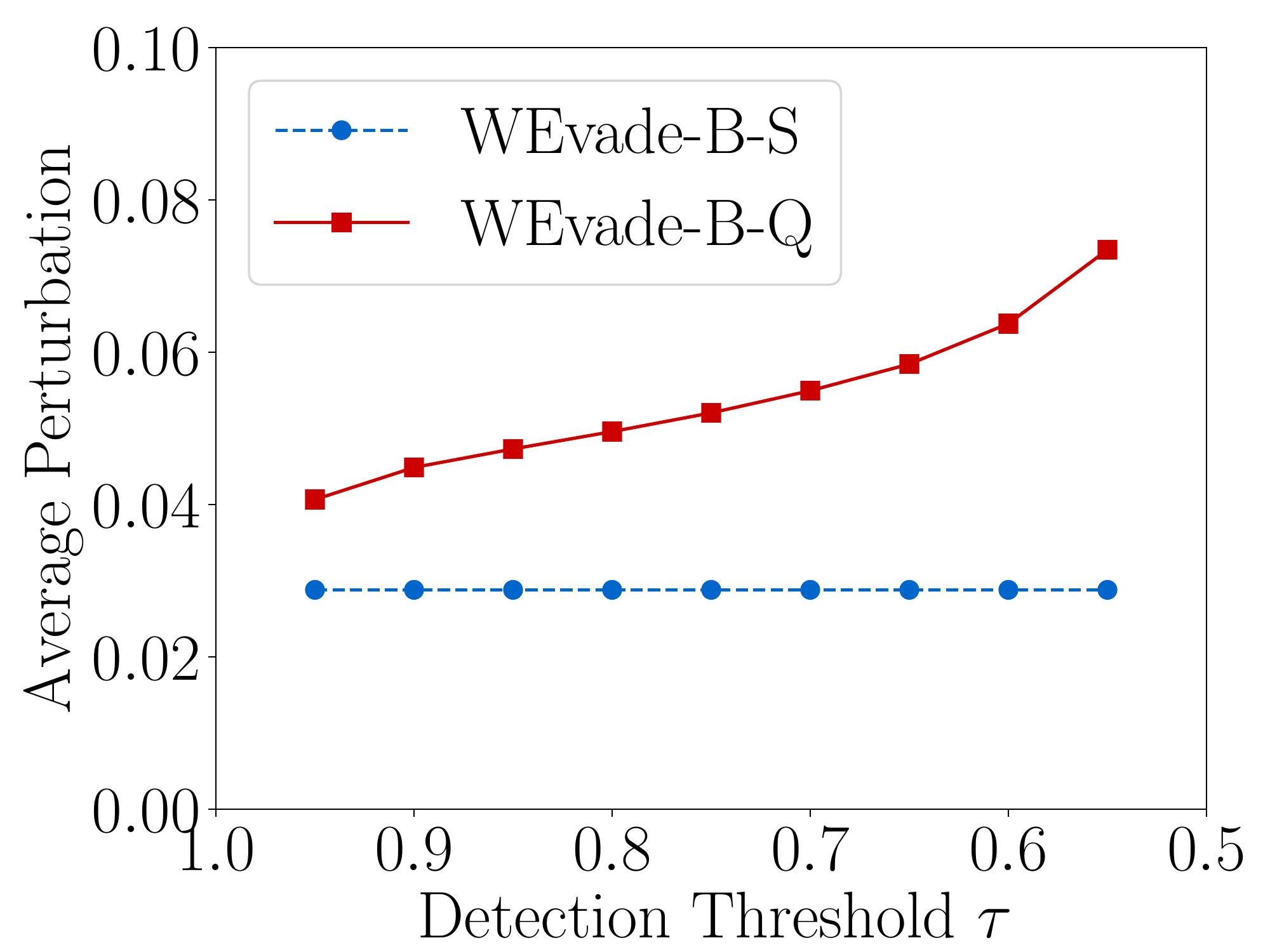}}
\caption{Comparing evasion rates (\emph{first row}) and average perturbations (\emph{second row})  of \algns-B-S and \algns-B-Q in the black-box setting. Watermarking method is UDH.} 
\label{black-udh}
\end{figure}

\begin{figure}[!t]
\centering
\subfloat[Initialization]{\includegraphics[width=0.23\textwidth]{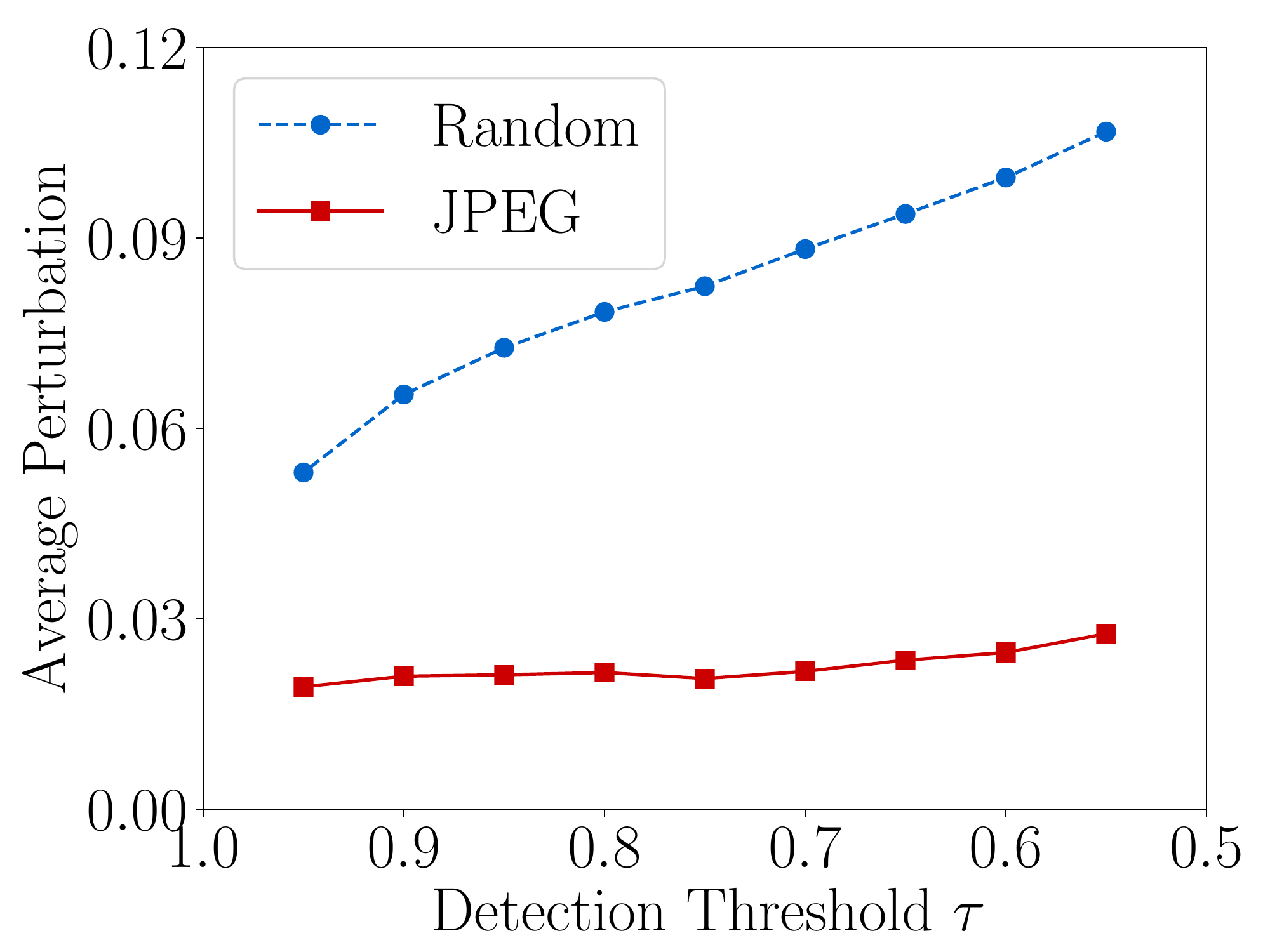} \label{initialization}}
\subfloat[Early stopping]{\includegraphics[width=0.23\textwidth]{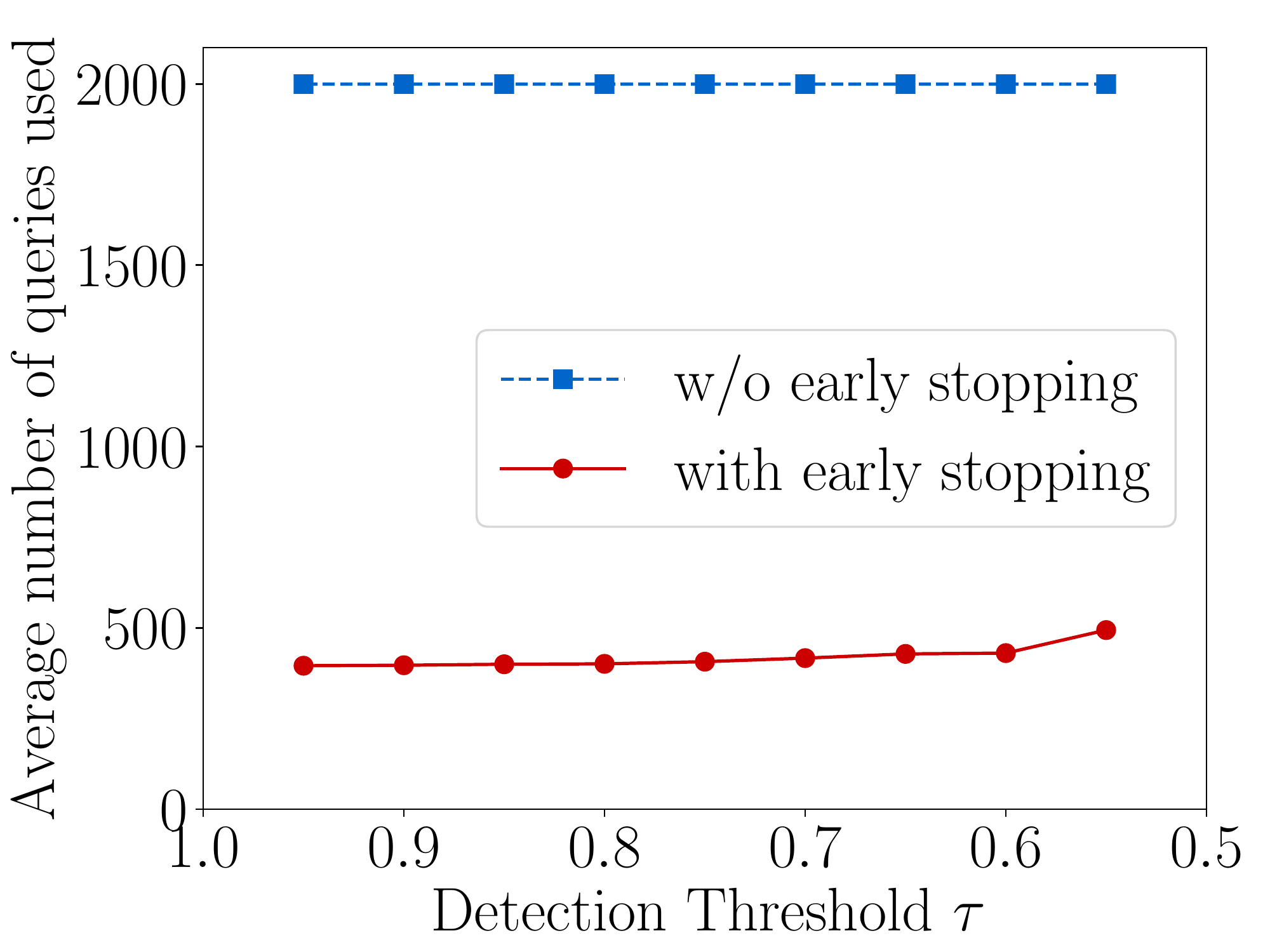} \label{es-q}}
\vspace{0.1cm}
\caption{Impact of (a) initialization and (b) early stopping on our \algns-B-Q for UDH and COCO dataset.}
\label{earlystop}
\end{figure}

\begin{figure}[!t]
\centering
{\includegraphics[width=0.23\textwidth]{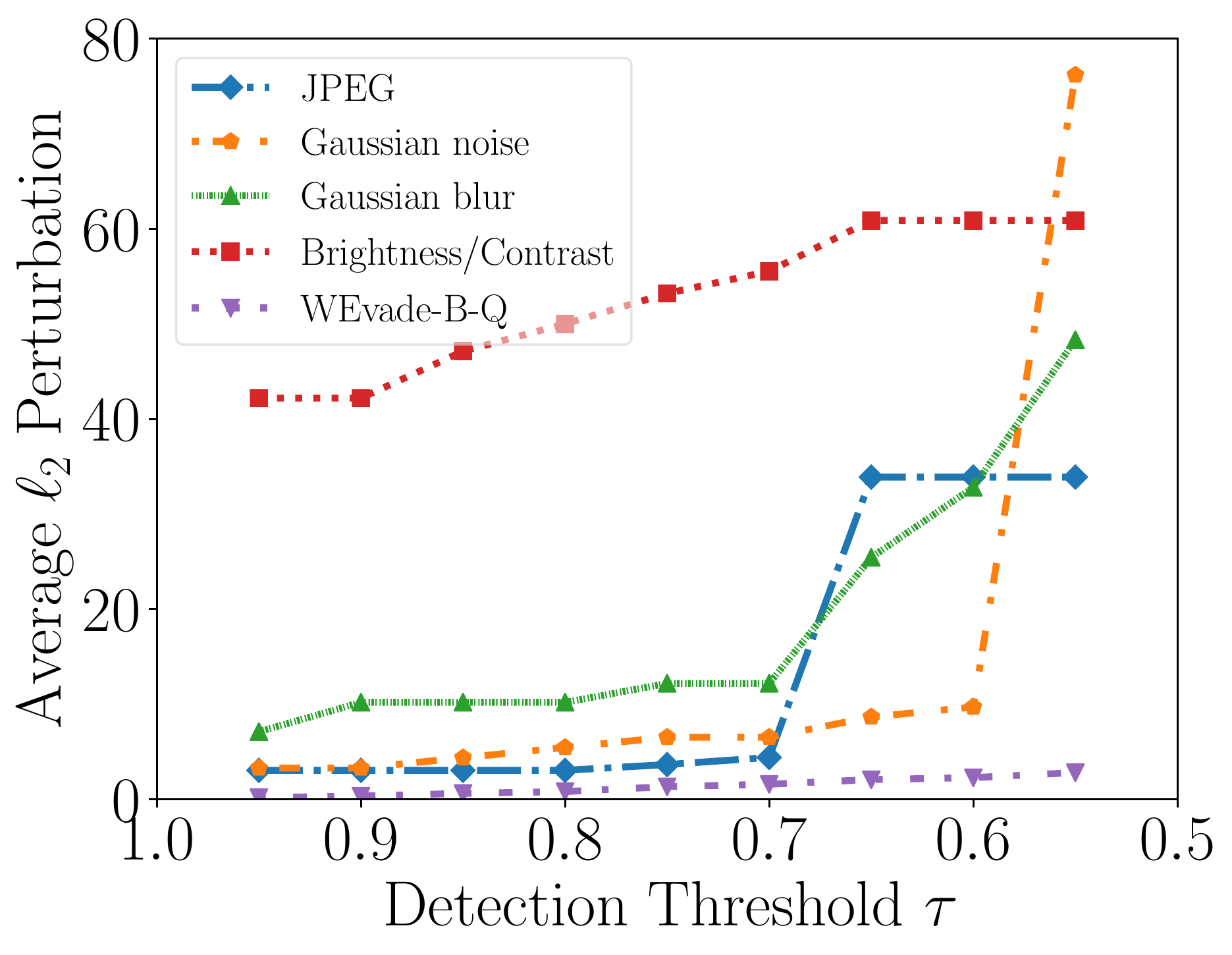}}
{\includegraphics[width=0.23\textwidth]{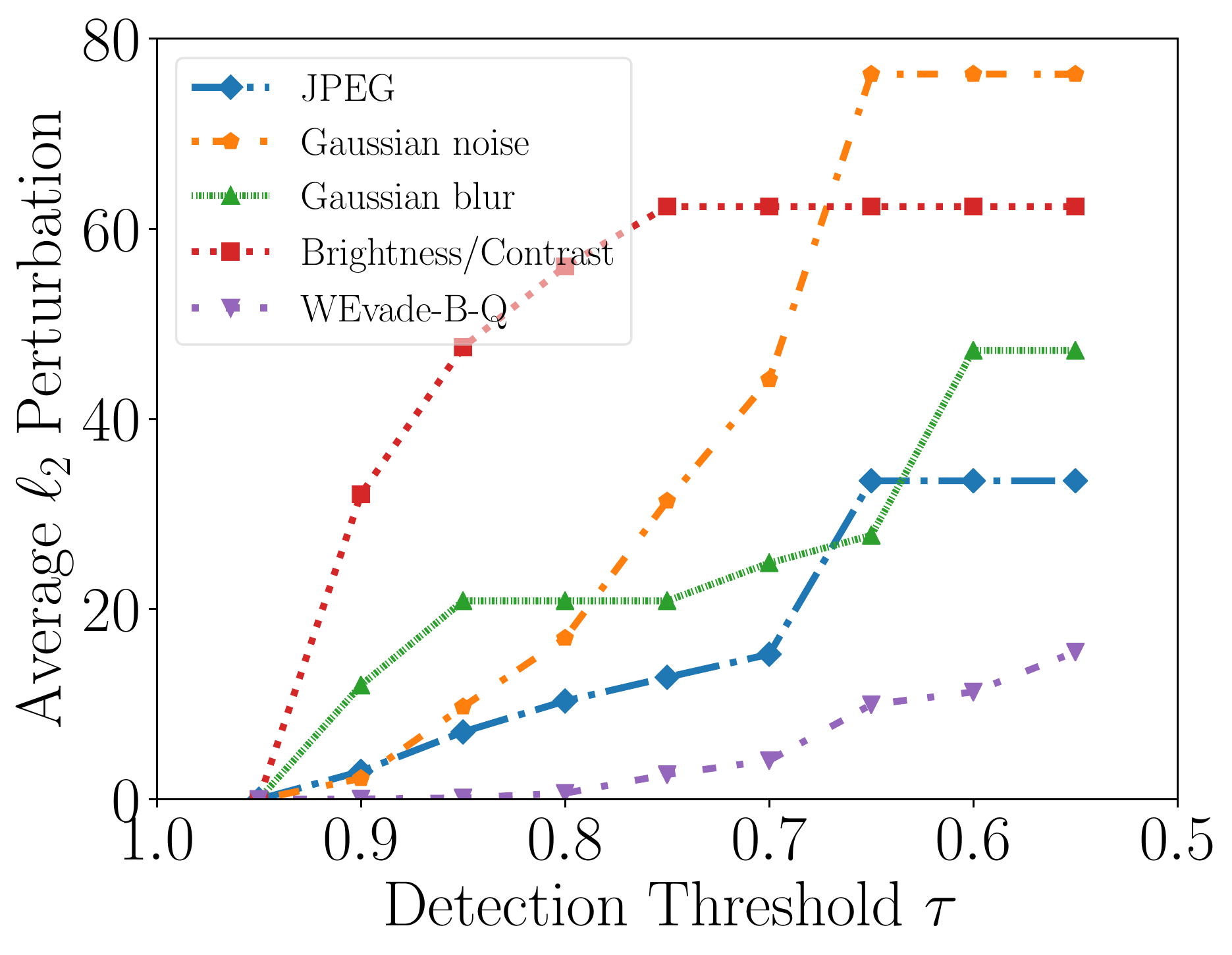}}

\subfloat[Standard training]{\includegraphics[width=0.23\textwidth]{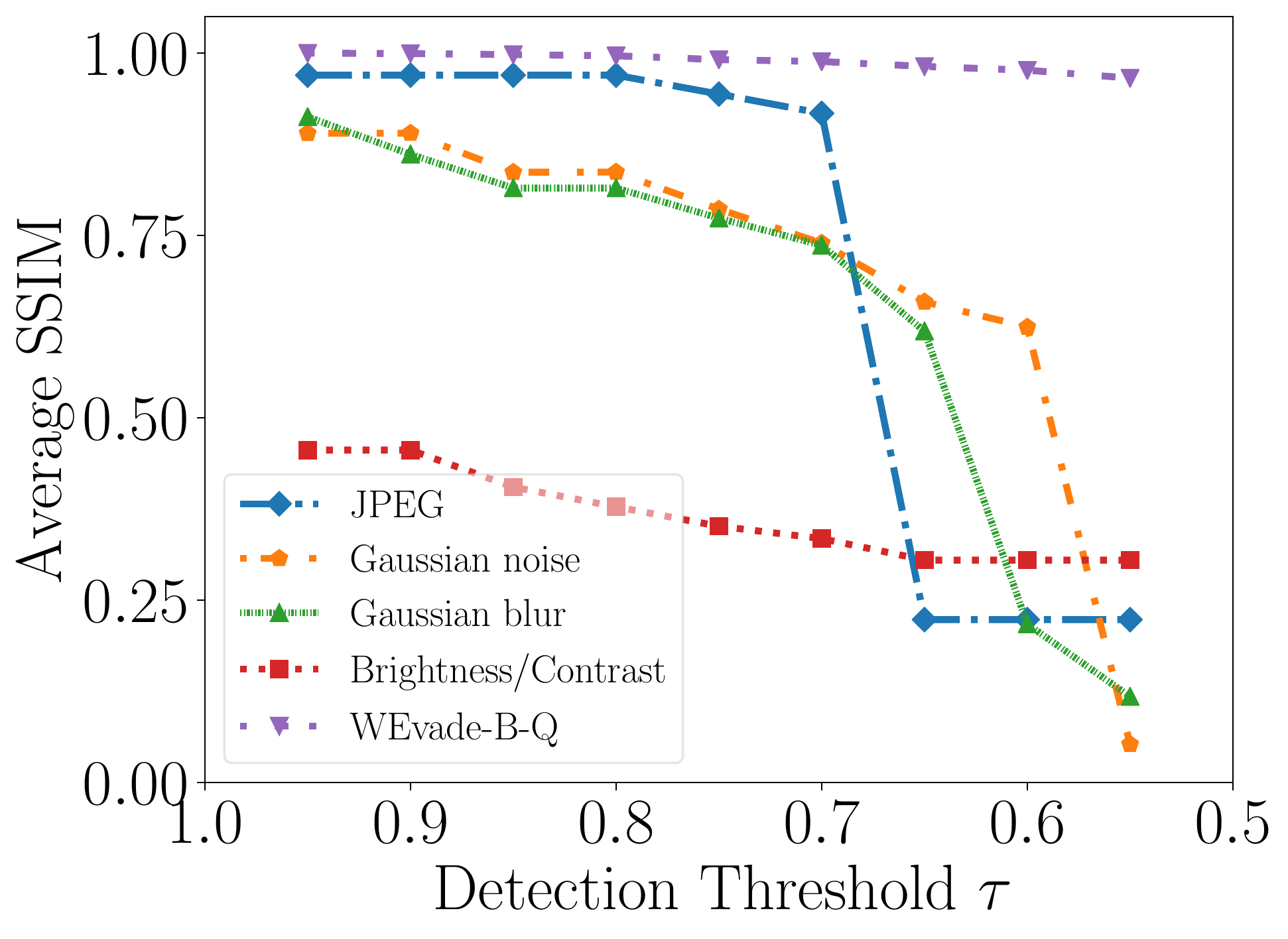}}
\subfloat[Adversarial training]{\includegraphics[width=0.23\textwidth]{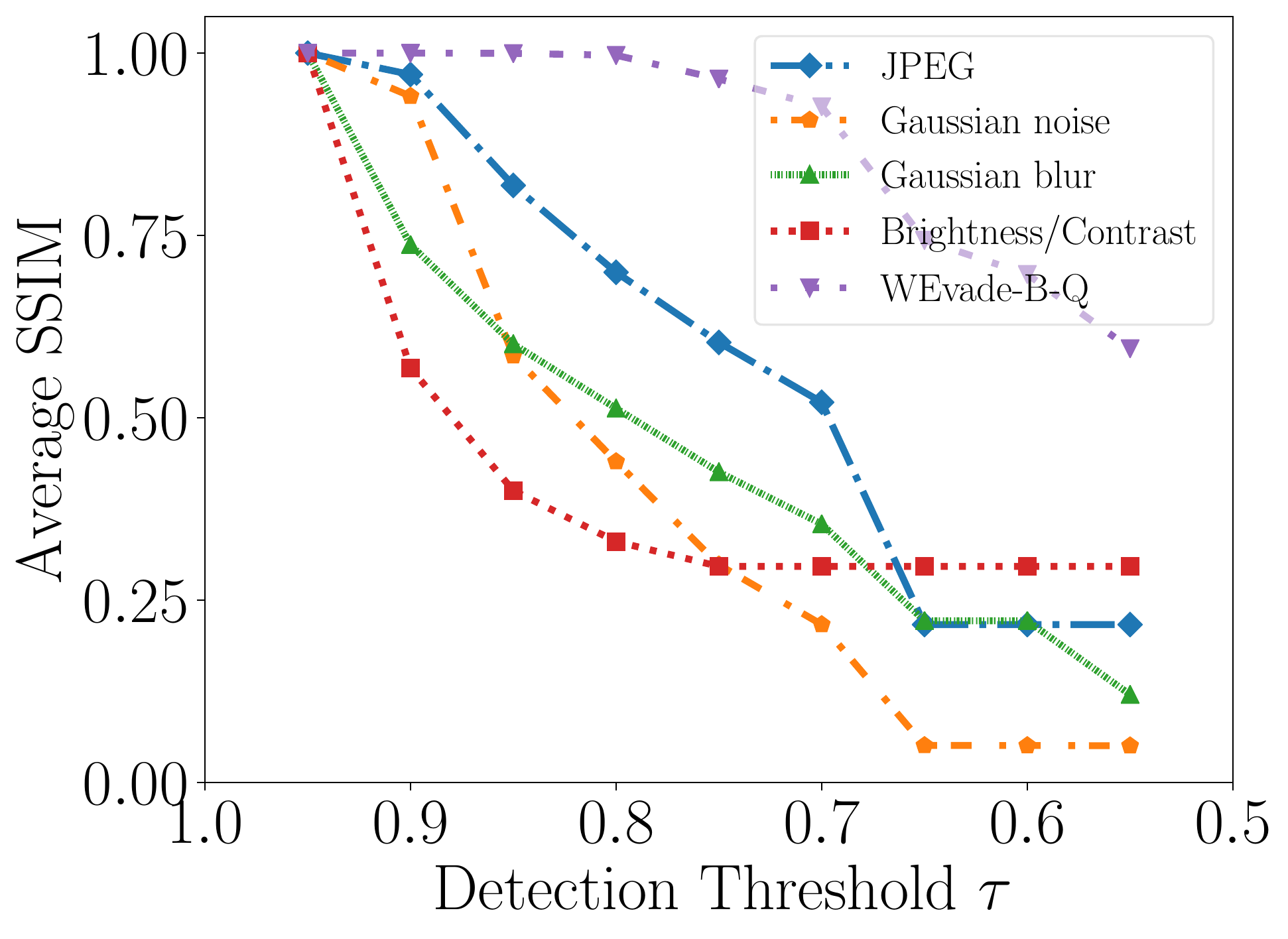}}
\caption{Average perturbation, measured by $\ell_2$-norm (first row) or SSIM (second row),  added by each post-processing method to evade the double-tail detector with different $\tau$ in the black-box setting. \algns-B-Q always achieves evasion rate 1, and  we set the parameters of existing post-processing methods such that they achieve evasion rates as close to 1 as possible. The watermarking method is HiDDeN and dataset is COCO. When generating these perturbations, we change the $\ell_\infty$-norm to $\ell_2$-norm at Line~\ref{lpnormcheck} in Algorithm~\ref{blackboxquery}.}
\label{l2blackbox}
\end{figure}

\begin{figure}[!t]
\centering
{\includegraphics[width=0.25\textwidth]{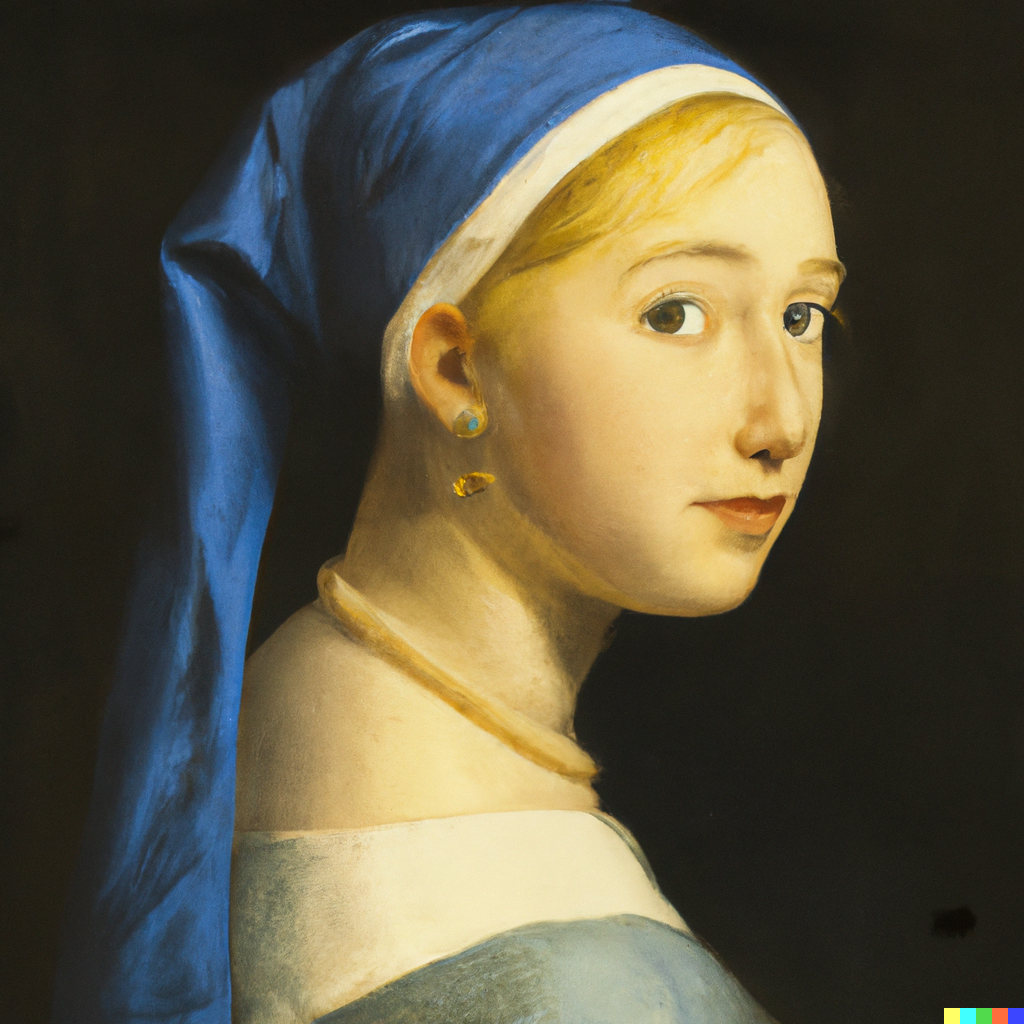}}
\vspace{0.2cm}
\caption{DALL-E generated image with a visible watermark at the bottom right corner.}
\label{dall-e}
\end{figure}

\begin{algorithm}[!t] 
\renewcommand{\algorithmicrequire}{\textbf{Input:}} 
\renewcommand{\algorithmicensure}{\textbf{Output:}}
    \caption{\algns-W-I and \algns-W-II} 
    \label{WEvadeWhite} 
    \begin{algorithmic}[1]
        \REQUIRE    Watermarked image $I_w$ and target watermark $w_t$
        \ENSURE Post-processed watermarked image $I_{pw}$  \\
        \STATE $r_b \leftarrow 2$
        \STATE $r_a \leftarrow 0$
        \WHILE{$r_b-r_a > 0.001$}
            \STATE $r \leftarrow (r_a + r_b)/2$
            \STATE $\delta' \leftarrow$ FindPerturbation ($I_w$, $w_t$, $r$)
            \IF{((\algns-W-I $\&$ Equation~\ref{adv-re-constraint} is satisfied) \\
            or (\algns-W-II $\&$ Equation~\ref{bitconstraint} is satisfied))}
                \STATE $r_b \leftarrow r$
                \STATE $\delta \leftarrow \delta'$
            \ELSE
                \STATE $r_a \leftarrow r$
            \ENDIF
        \ENDWHILE
        \STATE return $I_w+\delta$
    \end{algorithmic}
\end{algorithm}

\begin{algorithm}[!t] 
\renewcommand{\algorithmicrequire}{\textbf{Input:}} 
\renewcommand{\algorithmicensure}{\textbf{Output:}}
    \caption{FindPerturbation ($I_w$, $w_t$, $r$)} 
    \label{PGD} 
    \begin{algorithmic}[1]
        \REQUIRE   Decoder $D$, objective function $l$, learning rate $\alpha$, and maximum number of iterations \emph{max\_iter}.
        \ENSURE Perturbation $\delta$  \\

        \STATE $\delta \leftarrow 0$
        \FOR {$k$ = 1 to \emph{max\_iter} }
            \STATE $g \leftarrow \nabla_\delta l(D(I_w+\delta), w_t)$
            \STATE $\delta \leftarrow \delta - \alpha \cdot g$

            \STATE//Projection to satisfy the perturbation bound
            \IF {$\|\delta\|_{\infty} > r$}
                \STATE $\delta \leftarrow  \delta \cdot \frac{r}{\|\delta\|_{\infty}}$
            \ENDIF
            
            \STATE//Early stopping
            \IF{((\algns-W-I $\&$ Equation~\ref{adv-re-constraint} is satisfied) \\
            or (\algns-W-II $\&$ Equation~\ref{bitconstraint} is satisfied))}
                \STATE return $\delta$
            \ENDIF
        \ENDFOR
        \STATE return $\delta$
    \end{algorithmic}
\end{algorithm}

\begin{algorithm}[!t] 
\renewcommand{\algorithmicrequire}{\textbf{Input:}} 
\renewcommand{\algorithmicensure}{\textbf{Output:}}
    \caption{\algns-B-Q}
    \label{blackboxquery}
    \begin{algorithmic}[1]
        \REQUIRE API of the target detector, a watermarked image $I_w$, query budget max\_q, and early stop threshold $ES$. 
        \ENSURE Post-processed image $I_{pw}$  \\
        
        \STATE $q \leftarrow 0$
        \STATE //Initializing $I_{pw}$
        \label{algo_initialization}
        \FOR {$Q\in $  [99, 90, 70, 50, 30, 10, 1]}
            \STATE $q \leftarrow q + 1$
            \IF {$API$(JPEG($I_w$, $Q$))=="non-AI-generated"}
                \STATE $I_{pw} \leftarrow$ JPEG($I_w$, $Q$)
                \STATE \textbf{break}
            \ENDIF
        \ENDFOR

        \STATE //Iteratively move $I_{pw}$ towards $I_{w}$
        \STATE $\delta_{min} \leftarrow I_{pw}-I_w$
        \STATE $es \leftarrow 0$
        \WHILE {$q \leq$ \emph{max\_q} and $es \leq ES$}
        \STATE $I_{pw},  q' \leftarrow$ HopSkipJump($I_{pw}$)
        \STATE $q \leftarrow q + q'$
        \IF {$\|I_{pw}-I_w\|_{\infty} < \|\delta_{min}\|_{\infty}$}
        \label{lpnormcheck}
            \STATE $\delta_{min} \leftarrow I_{pw}-I_w$
            \STATE $es \leftarrow 0$
        \ELSE
            \STATE $es \leftarrow es + 1$
        \ENDIF
        \ENDWHILE
        \STATE return $I_{w}+\delta_{min}$
    \end{algorithmic}
\end{algorithm}

\section{Proof of Theorem~\ref{theorem1}}
\label{proof1}
For the standard detector, $I_w$ is correctly detected and thus we have $BA(D(I_{w}), w) > \tau >0.5$. Therefore, we have:
\begin{align}
    & BA(D(I_{pw}), w) \nonumber \\
    & = BA(\neg D(I_{w}), w) = 1-BA(D(I_{w}), w) \nonumber\\
    & < 1 - \tau < \tau \nonumber.
\end{align}

For the adaptive detector, $I_w$ is correctly detected and thus we have $BA(D(I_{w}), w) > \tau$ or $BA(D(I_{w}), w) < 1- \tau$, where $\tau>0.5$. Since $BA(D(I_{pw}), w)=1-BA(D(I_{w}), w)$, we have $BA(D(I_{pw}), w) > \tau$ or $BA(D(I_{pw}), w) < 1- \tau$.

\section{Proof of Theorem~\ref{theorem2}}
\label{proof2}
We denote $D(I_{pw})$ = $w_{I_{pw}}$. According to Equation~\ref{bitconstraint}, we have:
\begin{align}
    & BA(w_{I_{pw}}, w_t) = 1 - \frac{|w_{I_{pw}} - w_t|_1}{n} \geq 1 - \epsilon, \nonumber \\
    & \Longrightarrow |w_{I_{pw}} - w_t|_1 \leq \epsilon n \nonumber,
\end{align}
where $|\cdot|_1$ is $\ell_1$ distance between two binary vectors. Then, according to the triangle inequality, we have:
\begin{align}
    & |w_t-w|_1 = |w_t-w_{I_{pw}}+w_{I_{pw}}-w|_1 \nonumber \\
    & \leq |w_t-w_{I_{pw}}|_1 + |w_{I_{pw}}-w|_1 \nonumber \\
    & \leq \epsilon n + |w_{I_{pw}}-w|_1 \nonumber.
\end{align}
Therefore, we have:
\begin{align}
    & \text{Pr}(BA(D(I_{pw}), w) \leq \tau) \nonumber \\
    & = \text{Pr}(BA(w_{I_{pw}}, w) \leq \tau) \nonumber \\
    & = \text{Pr}(1-\frac{|w_{I_{pw}}-w|_1}{n} \leq \tau) \nonumber \\
    & = \text{Pr}(|w_{I_{pw}}-w|_1 \geq (1-\tau)n) \nonumber \\
    & \geq \text{Pr}(|w_t-w|_1-\epsilon n \geq (1-\tau)n) \nonumber \\
    & = \text{Pr}(|w_t-w|_1 \geq (1-\tau+\epsilon)n) \nonumber,
\end{align}
Since $w_t$ is picked uniformly at random, we know  $|w_t-w|_1$ follows a \emph{binomial distribution}, i.e., $|w_t-w|_1 \sim B(n,0.5)$. Thus, we have: 
\begin{align}
    & \text{Pr}(BA(D(I_{pw}), w) \leq \tau) \nonumber \\
    & \geq \text{Pr}(|w_t-w|_1 \geq \lceil (1-\tau+\epsilon)n \rceil) \nonumber \\
    & = P(\lfloor (\tau-\epsilon)n \rfloor) \nonumber,
\end{align}
where $P(t)=\text{Pr}(m \leq t)$ is the cumulative distribution function of the binomial distribution $m \sim B(n,0.5)$.

\section{Proof of Theorem~\ref{theorem3}}
\label{proof3}
According to Equation~\ref{bitconstraint}, we have: 
\begin{align}
    & BA(w_{I_{pw}}, w_t) = 1 - \frac{|w_{I_{pw}} - w_t|_1}{n} \leq 1 - \epsilon, \nonumber \\
    & \Longrightarrow |w_{I_{pw}} - w_t|_1 \leq \epsilon n \nonumber.
\end{align}
Then, according to the triangle inequality, we have: 
\begin{align}
    & |w_{I_{pw}}-w|_1 = |w_{I_{pw}}-w_t+w_t-w|_1 \nonumber \\
    & \leq |w_{I_{pw}}-w_t|_1 + |w_t-w|_1  \leq \epsilon n + |w_t-w|_1 \nonumber.
\end{align}
Similarly, we have:
\begin{align}
    & |w_t-w|_1 = |w_t-w_{I_{pw}}+w_{I_{pw}}-w|_1 \nonumber \\
    & \leq |w_t-w_{I_{pw}}|_1 + |w_{I_{pw}}-w|_1  \leq \epsilon n + |w_{I_{pw}}-w|_1 \nonumber.
\end{align}
Therefore, we have:
\begin{align}
    & |w_t-w|_1 - \epsilon n \leq |w_{I_{pw}}-w|_1 \leq |w_t-w|_1 + \epsilon n \nonumber.
\end{align}
Thus, we have: 
\begin{align}
    & \text{Pr}(1-\tau \leq BA(D(I_{pw}), w) \leq \tau) \nonumber \\
    & = \text{Pr}(1-\tau \leq BA(w_{I_{pw}}, w) \leq \tau) \nonumber \\
    & = \text{Pr}(1-\tau \leq 1-\frac{|w_{I_{pw}}-w|_1}{n} \leq \tau) \nonumber \\
    & = \text{Pr}((1-\tau)n \leq |w_{I_{pw}}-w|_1 \leq \tau n) \nonumber \\
    & = 1 - \text{Pr}((1-\tau)n \textgreater |w_{I_{pw}}-w|_1) - \text{Pr} (|w_{I_{pw}}-w|_1 \textgreater \tau n) \nonumber \\
    & \geq 1 - \text{Pr}((1-\tau)n \textgreater |w_t-w|_1 - \epsilon n) - \text{Pr} (|w_t-w|_1 + \epsilon n \textgreater \tau n) \nonumber \\
    & = 1 - \text{Pr}((1-\tau+\epsilon)n \textgreater |w_t-w|_1) - \text{Pr} (|w_t-w|_1 \textgreater (\tau-\epsilon)n) \nonumber \\
    & = 1 - 2\text{Pr} (|w_t-w|_1 \textgreater (\tau-\epsilon)n) \nonumber \\
    & = 1 - 2(1 - \text{Pr} (|w_t-w|_1 \leq (\tau-\epsilon)n)) \nonumber \\
    & = 2\text{Pr} (|w_t-w|_1 \leq (\tau-\epsilon)n) - 1 \nonumber.
\end{align}
Since $|w_t-w|_1 \sim B(n,0.5)$, we have:
\begin{align}
    & \text{Pr}(1-\tau \leq BA(D(I_{pw}), w) \leq \tau) \nonumber \\
    & \geq 2\text{Pr} (|w_t-w|_1 \leq (\tau-\epsilon)n) - 1  \nonumber\\
    & = 2\text{Pr} (|w_t-w|_1 \leq \lfloor (\tau-\epsilon)n \rfloor) - 1  \nonumber\\
    & = 2P(\lfloor (\tau-\epsilon)n \rfloor) - 1 \nonumber.
\end{align}

\section{Proof of Theorem~\ref{theorem-black}}
\label{proof-black}
For single-tail detector, we denote $D^{\prime}(I_{pw})=w^{\prime}_{I_{pw}}$. According to Equation~\ref{bitconstraint}, we have:
\begin{align}
    & BA(w^{\prime}_{I_{pw}}, w_t) = 1 - \frac{|w^{\prime}_{I_{pw}} - w_t|_1}{n} \leq 1 - \epsilon, \nonumber \\
    & \Longrightarrow |w^{\prime}_{I_{pw}} - w_t|_1 \leq \epsilon n \nonumber.
\end{align}
Then, according to the triangle inequality, we have: 
\begin{align}
    & |w^{\prime}_{I_{pw}}-w|_1 = |w^{\prime}_{I_{pw}}-w_t+w_t-w|_1 \nonumber \\
    & \leq |w^{\prime}_{I_{pw}}-w_t|_1 + |w_t-w|_1  \leq \epsilon n + |w_t-w|_1 \nonumber.
\end{align}
Similarly, we have:
\begin{align}
    & |w_t-w|_1 = |w_t-w^{\prime}_{I_{pw}}+w^{\prime}_{I_{pw}}-w|_1 \nonumber \\
    & \leq |w_t-w^{\prime}_{I_{pw}}|_1 + |w^{\prime}_{I_{pw}}-w|_1  \leq \epsilon n + |w^{\prime}_{I_{pw}}-w|_1 \nonumber.
\end{align}
Therefore, we have:
\begin{align}
    & |w_t-w|_1 - \epsilon n \leq |w^{\prime}_{I_{pw}}-w|_1 \leq |w_t-w|_1 + \epsilon n \nonumber.
\end{align}
Moreover, according to Definition~\ref{definition4}, we have:
\begin{align}
    & \text{Pr}(BA(w^{\prime}_{I_{pw}}, w_{I_{pw}}) \geq \beta) \nonumber \\
    & = \text{Pr}(1 - \frac{|w^{\prime}_{I_{pw}} - w_{I_{pw}}|_1}{n} \geq \beta) \geq \gamma, \nonumber \\
    & \Longrightarrow \text{Pr}(|w^{\prime}_{I_{pw}} - w_{I_{pw}}|_1 \leq (1-\beta)n) \geq \gamma \nonumber.
\end{align}
Thus, we have:
\begin{align}
    & \text{Pr}(BA(D(I_{pw}), w) \leq \tau) \nonumber \\
    & = \text{Pr}(BA(w_{I_{pw}}, w) \leq \tau) \nonumber \\
    & = \text{Pr}(1-\frac{|w_{I_{pw}}-w|_1}{n} \leq \tau) \nonumber \\
    & = \text{Pr}(|w_{I_{pw}}-w|_1 \geq (1-\tau)n) \nonumber.
\end{align}
Then, according to the triangle inequality, we have:
\begin{align}
    & |w^{\prime}_{I_{pw}}-w|_1 = |w^{\prime}_{I_{pw}}-w_{I_{pw}}+w_{I_{pw}}-w|_1 \nonumber \\
    & \leq |w^{\prime}_{I_{pw}}-w_{I_{pw}}|_1 + |w_{I_{pw}}-w|_1, \nonumber \\
    & \Longrightarrow |w_{I_{pw}}-w|_1 \geq |w^{\prime}_{I_{pw}} - w|_1 - |w^{\prime}_{I_{pw}} - w_{I_{pw}}|_1 \nonumber.
\end{align}
Similarly, we have:
\begin{align}
    & |w_{I_{pw}}-w|_1 = |w_{I_{pw}}-w^{\prime}_{I_{pw}}+w^{\prime}_{I_{pw}}-w|_1 \nonumber \\
    & \leq |w^{\prime}_{I_{pw}}-w|_1 + |w_{I_{pw}}-w^{\prime}_{I_{pw}}|_1 \nonumber.
\end{align}
Thus, we have:
\begin{align}
    & \text{Pr}(|w_{I_{pw}}-w|_1 \geq (1-\tau)n) \nonumber \\
    & \geq \text{Pr}(|w^{\prime}_{I_{pw}} - w|_1 - |w^{\prime}_{I_{pw}} - w_{I_{pw}}|_1 \geq (1-\tau)n) \nonumber \\
    & \geq \text{Pr}(|w^{\prime}_{I_{pw}} - w|_1 - (1-\beta)n) \geq (1-\tau)n) \nonumber \\
    & \quad \quad  \cdot \text{Pr} (|w^{\prime}_{I_{pw}} - w_{I_{pw}}|_1 \leq (1-\beta)n) \nonumber \\
    & \geq \text{Pr}(|w^{\prime}_{I_{pw}} - w|_1 \geq (2-\tau-\beta)n) \cdot \gamma \nonumber \\
    & \geq \gamma \text{Pr}(|w_t - w|_1 -\epsilon n \geq (2-\tau-\beta)n) \nonumber \\
    & = \gamma \text{Pr}(|w_t - w|_1 \geq (2-\tau-\beta+\epsilon)n) \nonumber.
\end{align}
Since $|w_t-w|_1 \sim B(n,0.5)$, we have:
\begin{align}
    & \text{Pr}(BA(D(I_{pw}), w) \leq \tau) \nonumber \\
    & \geq \gamma \text{Pr}(|w_t-w|_1 \geq \lceil (2-\tau-\beta+\epsilon)n \rceil) \nonumber \\
    & = \gamma (1-P(\lceil (2-\tau-\beta+\epsilon) \rceil)) \nonumber \\
    & = \gamma P(\lfloor (\tau+\beta-\epsilon-1)n \rfloor) \nonumber.
\end{align}
For double-tail detector, we have:
\begin{align}
    & \text{Pr}(1-\tau \leq BA(D(I_{pw}), w) \leq \tau) \nonumber \\
    & = \text{Pr}(1-\tau \leq BA(w_{I_{pw}}, w) \leq \tau) \nonumber \\
    & = \text{Pr}(1-\tau \leq 1-\frac{|w_{I_{pw}}-w|_1}{n} \leq \tau) \nonumber \\
    & = \text{Pr}((1-\tau)n \leq |w_{I_{pw}}-w|_1 \leq \tau n) \nonumber \\
    & = 1 - \text{Pr}((1-\tau)n \textgreater |w_{I_{pw}}-w|_1) - \text{Pr} (|w_{I_{pw}}-w|_1 \textgreater \tau n) \nonumber \\
    & \geq 1 - \text{Pr}(|w^{\prime}_{I_{pw}} - w|_1 - |w^{\prime}_{I_{pw}} - w_{I_{pw}}|_1 \textless (1-\tau)n) \nonumber \\
    & \quad \quad - \text{Pr}(|w^{\prime}_{I_{pw}}-w|_1 + |w_{I_{pw}}-w^{\prime}_{I_{pw}}|_1 \textgreater \tau n) \nonumber \\
    & \geq \text{Pr}(|w^{\prime}_{I_{pw}} - w|_1 - |w^{\prime}_{I_{pw}} - w_{I_{pw}}|_1 \geq (1-\tau)n) \nonumber \\
    & \quad \quad  + \text{Pr}(|w^{\prime}_{I_{pw}}-w|_1 + |w_{I_{pw}}-w^{\prime}_{I_{pw}}|_1 \leq \tau n) - 1 \nonumber \\
    & \geq \gamma P(\lfloor (\tau+\beta-\epsilon-1)n \rfloor) + \text{Pr}(|w^{\prime}_{I_{pw}}-w|_1 + (1-\beta)n \leq \tau n) \nonumber \\
    & \quad \quad \cdot \text{Pr}(|w^{\prime}_{I_{pw}} - w_{I_{pw}}|_1 \leq (1-\beta)n) - 1 \nonumber \\
    & \geq \gamma P(\lfloor (\tau+\beta-\epsilon-1)n \rfloor) \nonumber \\
    & \quad \quad + \text{Pr}(|w_t - w|_1 + \epsilon n \leq (\tau+\beta-1) n) \cdot \gamma - 1 \nonumber \\
    & \geq \gamma P(\lfloor (\tau+\beta-\epsilon-1)n \rfloor) \nonumber \\
    & \quad \quad + \gamma \text{Pr}(|w_t - w|_1  \leq (\tau+\beta-\epsilon-1) n) - 1 \nonumber \\
    & \geq 2 \gamma P(\lfloor (\tau+\beta-\epsilon-1)n \rfloor) - 1 \nonumber.
\end{align}

\end{document}